\documentclass[10pt,twocolumn,letterpaper]{article}

\usepackage{iccv}
\usepackage{times}
\usepackage{epsfig}
\usepackage{graphicx}
\usepackage{amsmath}
\usepackage{amssymb}
\usepackage{booktabs}
\usepackage{caption}
\usepackage{subfig}
\usepackage[dvipsnames]{xcolor}
\usepackage[export]{adjustbox}
\newcommand{\ra}[1]{\renewcommand{\arraystretch}{#1}}
\usepackage{color}
\usepackage{rotating}
\usepackage{bm}

\makeatletter
\let\@fnsymbol\@arabic
\makeatother

\usepackage[pagebackref=true,breaklinks=true,letterpaper=true,colorlinks,bookmarks=false]{hyperref}

\iccvfinalcopy 

\def\iccvPaperID{835} 

\newif\ifkeepComments
\keepCommentstrue

\newcommand{\spaceparagraph}{\vspace{-.35cm}}
\newcommand{\gotoline}{\vspace{.05cm}}

\begin{document}

\title{Weakly-supervised learning of visual relations}

\author{Julia Peyre\footnotemark[1] \textsuperscript{,}\footnotemark[2]
 \qquad Ivan Laptev\footnotemark[1] \textsuperscript{,}\footnotemark[2] \qquad Cordelia Schmid\footnotemark[2] \textsuperscript{,}\footnotemark[4] \qquad Josef Sivic\footnotemark[1] \textsuperscript{,}\footnotemark[2] \textsuperscript{,}\footnotemark[3] \\
}

\maketitle

\footnotetext[1]{D\'epartement d'informatique de l'ENS, ´ Ecole normale sup\'erieure, ´
CNRS, PSL Research University, 75005 Paris, France.}
\footnotetext[2]{INRIA}
\footnotetext[3]{Czech Institute of Informatics, Robotics and Cybernetics at the Czech Technical University in Prague.}
\footnotetext[4]{Univ. Grenoble Alpes, Inria, CNRS, Grenoble INP, LJK, 38000 Grenoble, France.}

\begin{abstract}

This paper introduces a novel approach for modeling visual relations
        between pairs of objects.
 We call relation a triplet of the form $(subject,
        predicate, object)$ where the predicate is typically a
        preposition (eg. 'under', 'in front of') or a verb ('hold',
        'ride') that links a pair of objects $(subject,
        object)$. Learning such relations is challenging as the
        objects have different spatial configurations and appearances
        depending on the relation in which they occur. Another major
        challenge comes from the difficulty to get annotations,
        especially at box-level, for all possible triplets, which
        makes both learning and evaluation difficult. The
        contributions of this paper are threefold. First, we design strong yet flexible visual features that 		encode the appearance and spatial configuration for pairs of objects. Second, we propose a
        weakly-supervised discriminative clustering model to learn
        relations from image-level labels only. Third we introduce a
        new challenging dataset of unusual relations (UnRel) together
        with an exhaustive annotation, that enables accurate
        evaluation of visual relation retrieval. We show
        experimentally that our model results in state-of-the-art
        results on the visual relationship dataset~\cite{Lu16}
        significantly improving performance on previously
        unseen relations (zero-shot learning), and confirm this
        observation on our newly introduced UnRel dataset.    
   
\end{abstract}

\section{Introduction}

\begin{figure}[t]
\centering
	\begin{minipage}[t]{0.23\textwidth}
    	\centering
    	\includegraphics[trim={1cm 2cm 2.5cm 1cm},clip,width=\linewidth]{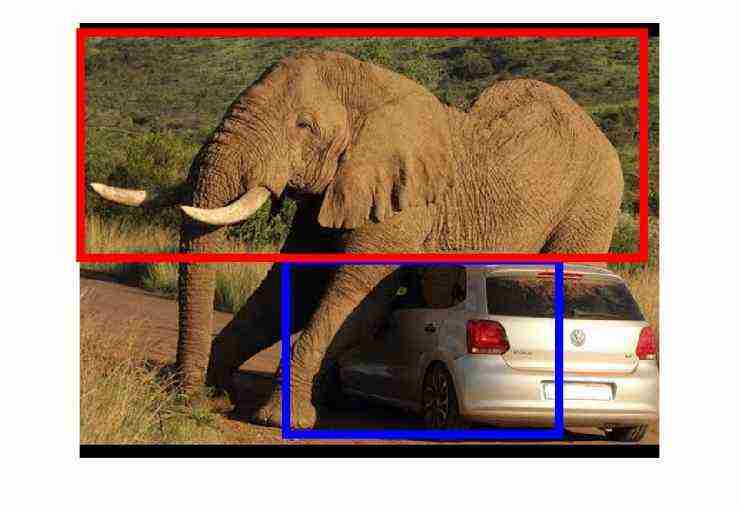}\\
       	\color{blue}{car} \textcolor{Green}{under} \color{red}{elephant}
  		\vspace{1.5ex}
    \end{minipage}
    \begin{minipage}[t]{0.23\textwidth}
    	\centering
    	\includegraphics[trim={8.3cm 2.2cm 2.3cm 1cm},clip,width=\linewidth]{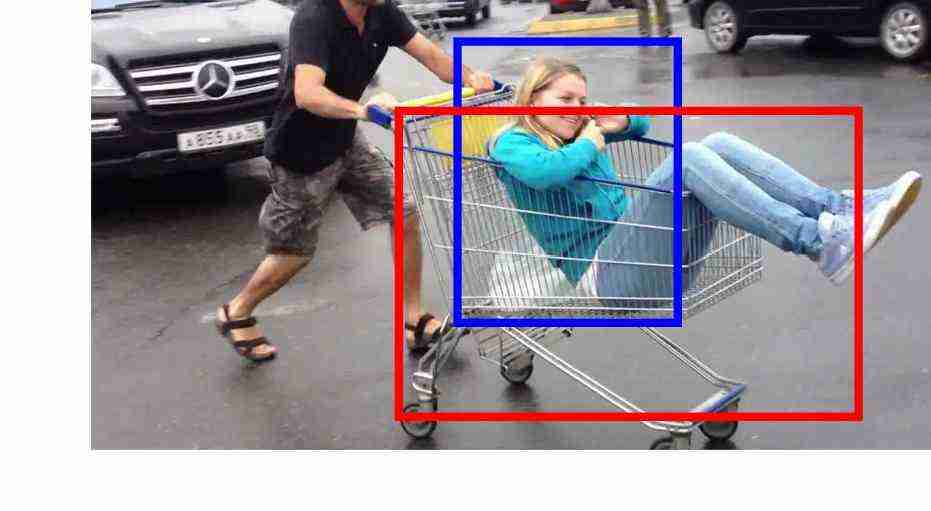}\\ 
    		\color{blue}{person} \textcolor{Green}{in} \color{red}{cart}
    		\vspace{1.5ex}
    \end{minipage}
    \begin{minipage}[t]{0.23\textwidth}
    	\centering
    	\includegraphics[trim={1.5cm 2.5cm 2.5cm 1.3cm},clip,width=\linewidth]{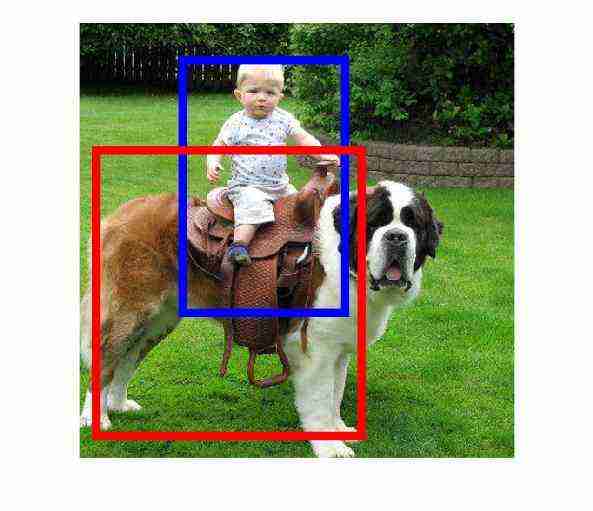}\\ 
 		\color{blue}{person} \textcolor{Green}{ride} \color{red}{dog}		
    \end{minipage}
    \begin{minipage}[t]{0.23\textwidth}
    	\centering
    	\includegraphics[trim={6.8cm 2.5cm 3cm 1.9cm},clip,width=\linewidth]{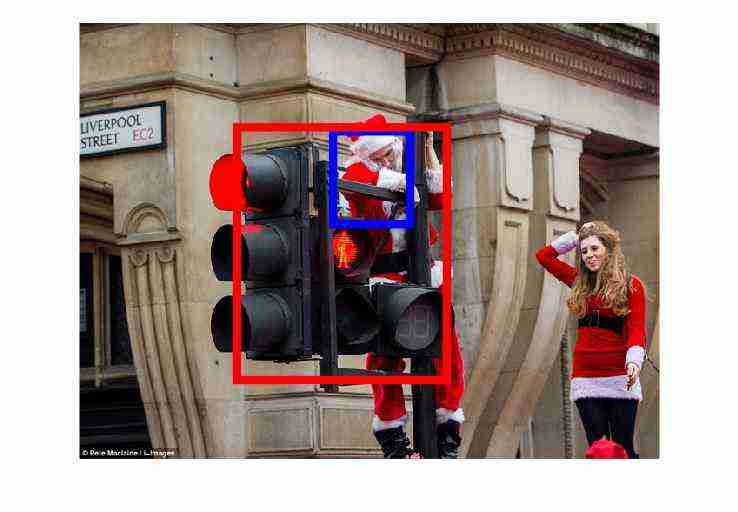}\\
  		\color{blue}{person} \textcolor{Green}{on top of} \color{red}{traffic light}    	
    \end{minipage}
    \setlength\abovecaptionskip{0.2cm}
    \caption{Examples of top retrieved pairs of boxes in UnRel dataset for unusual queries (indicated below each image) with our weakly-supervised model described in \ref{model}.}
    \vspace{-.4cm}
    \label{teaser_figure}
\end{figure}

While a great progress has been made on the detection and localization of individual objects~\cite{ren15,zagoruyko2016multipath},
it is now time to move one step forward towards understanding complete scenes.
For example, if we want to localize ``a  person 
sitting on a chair under an umbrella'', we not only need to detect the objects
involved : ``person", ``chair", ``umbrella",  but also need to find the correspondence of the semantic relations ``sitting on" and
``under" with the correct pairs of objects in the image. Thus, an
important challenge is automatic understanding of how entities in an image interact with each other.

\gotoline
\vspace{4pt}
This task presents two main challenges. 
First, the appearance of objects can change significantly due to interactions with other objects (person cycling, person driving).
This visual complexity can be tackled by learning ``visual phrases"~\cite{Sadeghi2011} capturing the pair
of objects in a relation as one entity, as opposed to first detecting individual entities in an image and then modeling their relations. 
This approach, however, does not scale to the large number of relations as the number of such visual phrases grows combinatorially, requiring large amounts of training data.
To address this challenge, we need a method that can share knowledge among similar relations.
Intuitively, it seems possible to generalize frequent relations to unseen triplets like those depicted in Figure \ref{teaser_figure} : for example having seen ``person ride horse" at training could help recognizing ``person ride dog" at test time.

\gotoline
The second main challenge comes from the difficulty to provide exhaustive annotations on the object level for relations that are by their nature non mutually-exclusive (i.e.\ ``on the left of" is also ``next to"). A complete labeling of $R$ relations for all pairs of $N$ objects in an image would indeed require $\mathcal{O}(N^{2}R)$ annotations {\em for each image}. Such difficulty makes both learning and evaluation very challenging. For learning, it would be desirable to learn relations from image-level annotations only. For evaluation, current large-scale datasets \cite{Krishna2016,Lu16} do not allow retrieval evaluation due to large amount of missing annotations. 

\spaceparagraph
\paragraph{Contributions.}
The contributions of this work are three-fold. First, to address the combinatorial challenge, we develop a method that can handle a large number of relations by sharing parameters among them. 
For example, we learn  a single ``on" classifier that can recognize
both ``person on bike" and ``dog on bike", even when ``dog on bike" has not been seen in training.  The main innovation is a
new model of an object relation that represents a pair of boxes by
explicitly incorporating their spatial configuration as well as the
appearance of individual objects. 
Our model relies on a multimodal representation of object configurations for each relation to handle the variability of relations. Second, to
address the challenge of missing training data, we develop a model that, given
pre-trained object detectors, is able to learn classifiers for object
relations from image-level supervision only. It is, thus, sufficient 
to provide an image-level annotation, such as  ``person on bike", without
annotating the objects involved in the relation. 
Finally, to address the issue of missing annotations in test data, we introduce a new
dataset of unusual relations (UnRel), with exhaustive annotation for a
set of unusual triplet queries, that enables to evaluate retrieval on
rare triplets and validate the generalization capabilities the learned
model.  

\section{Related Work}

\paragraph{Alignment of images with language.}
Learning correspondences between fragments of sentences and image regions has been addressed by the visual-semantic alignment which has been used for applications in image retrieval and caption generation \cite{chang2015text,Karpathy2014,Karpathy2014a}. With the appearance of new datasets providing box-level natural language annotations \cite{Kazemzadeh2014,Krishna2016,Mao2016,Plummer2015}, recent works have also investigated caption generation at the level of image regions for the tasks of natural language object retrieval \cite{Hu2015,Mao2016,Rohrbach2015} or dense captioning \cite{Johnson2015}. Our approach is similar in the sense that we aim at aligning a language triplet with a pair of boxes in the image. Typically, existing approaches do not explicitly represent relations between noun phrases in a sentence to improve visual-semantic alignment. We believe that understanding these relations is the next step towards image understanding with potential applications in tasks such as Visual Question Answering \cite{Andreas2016}.

\spaceparagraph
\paragraph{Learning triplets.} 
Triplet learning has been addressed in various tasks such as mining typical relations (knowledge extraction)~\cite{chen2013neil,sadeghi2015viske,yatskar2016stating,Zhu2014}, reasoning~\cite{jenatton2012,Movshovitz-attias,Socher2013b}, object detection~\cite{Gupta08,Sadeghi2011}, image retrieval~\cite{Johnson15a} or fact retrieval \cite{elhoseiny2015sherlock}. In this work, we address the task of relationship detection in images. This task was studied for the special case of human-object interactions \cite{Delaitre11,Desai2010,Gupta2009,Prest12,ramanathan15,Yao2010,Yao2010a,Yao11}, where the triplet is in the form $(person, action, object)$. Contrary to these approaches, we do not restrict the $subject$ to be a person and we cover a broader class of predicates that includes prepositions and comparatives. Moreover, most of the previous work in human-object interaction was tested on small datasets only and does not explicitly address the combinatorial challenge in modeling relations \cite{Sadeghi2011}. Recently,~\cite{Lu16} tried to generalize this setup to non-human subjects and scale to a larger vocabulary of objects and relations by developing a language model sharing knowledge among relations for visual relation detection. In our work we address this combinatorial challenge by developing a new visual representation that generalizes better to unseen triplets without the need for a strong language model. This visual representation shares the spirit of~\cite{galleguillos2008object,Johnson15a,li2012automatic} and we show it can handle multimodal relations and generalizes well to unseen triplets. Our model also handles a weakly-supervised set-up when only image-level annotations for object relations are available. It can thus learn from complex scenes with many objects participating in different relations, whereas previous work either uses full supervision or typically assumes only one object relation per image, for example, in images returned by a web search engine. 
Finally, we also address the problem to evaluate accurately due to missing annotations also pointed out in~\cite{elhoseiny2015sherlock,Lu16}. We introduce a new dataset of unusual relations exhaustively labeled for a set of triplet queries, the UnRel dataset. This dataset enables the evaluation of relation retrieval and localization. 
Our dataset is related to the ``Out of context" dataset of~\cite{Choi2012} which also presents objects in unusual configurations. However, the dataset of \cite{Choi2012} is not annotated with relations and does not match the vocabulary of objects in \cite{Lu16}, which prevents direct comparisons with existing methods that use data from~\cite{Lu16} for training.

\spaceparagraph
\paragraph{Weak supervision.} Most of the work on weakly-supervised learning for visual recognition has focused on learning objects \cite{ Bilen16,fangCVPR15,Oquab15}. Here, we want to tackle the task of weakly-supervised detection of relations. This task is more complex as we need to detect the individual objects that satisfy the specific relation. We assume that pre-trained detectors for individual objects are available and learn relations among objects with image-level labels. Our work uses a discriminative clustering objective \cite{bach2008diffrac}, that has been successful in several computer vision tasks \cite{Bojanowski2014,joulin2014}, but has not been so far, to the best of our knowledge, used for modeling relations.


\spaceparagraph
\paragraph{Zero-shot learning.} Zero-shot learning has been mostly explored for object classification~\cite{Frome2013,Lazaridou2014a,Socher2013a,Xian16} and recently for the task of describing images with novel objects~\cite{Hendricks2015,Venugopalan2016}. In our work, we address zero-shot learning of relations in the form of triplets $(subject,predicate,object)$, where each term has already been seen independently during training, but not in that specific combination. We develop a model to detect and localize such zero-shot relations.

\section{Representing and learning visual relations}

We want to represent triplets $t = (s, r, o)$ where $s$ is the
subject, $o$ the object and $r$ is the predicate. $s$ and $o$ are
nouns and can be objects like ``person", ``horse", ``car" or regions such as  ``sky",
``street", ``mountain". The predicate $r$ is a term that links the subject and the object in a sentence and can be a preposition (``in front of", ``under"), a verb (``ride", ``hold") or a comparative adjective (``taller than"). To detect and localize such triplets in test images, we assume that the candidate object detections for $s$ and $o$ are given by a detector trained with full supervision. Here we use the object detector \cite{girshick15fastrcnn} trained on the Visual Relationship Detection training set \cite{Lu16}. In \ref{visual_representation}, we will explain our representation of a triplet $t = (s, r, o)$  and show in \ref{model} how we can learn to detect triplets in images given weak image-level supervision for relations.

\subsection{Visual representation of relations}
\label{visual_representation}

\begin{figure}[t]
	\begin{center}
   	\includegraphics[trim={0 100 0 50},clip,width=\linewidth]{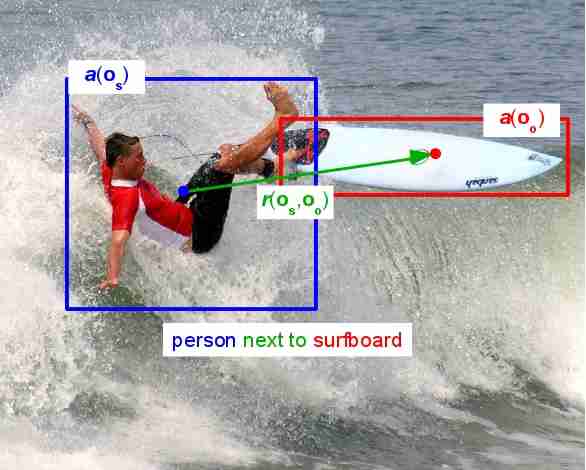}
	\end{center}
	\setlength\abovecaptionskip{-5pt}
   	\caption{Our visual representation is the composition of appearance features for each object $[\bm{a}(\bm{o_s}), \bm{a}(\bm{o_o})]$ and their spatial configuration $\bm{r}(\bm{o_s},\bm{o_o})$ represented by the green arrow.}
   	\vspace{-.4cm}
	\label{fig:visual_representation}
\end{figure}

A triplet $t=(s,r,o)$ such as ``person next to surfboard" in Figure \ref{fig:visual_representation} visually corresponds to a pair of objects $(s,o)$ in a certain configuration. We represent such pairs by the spatial configuration between object bounding boxes $(\bm{o_s},\bm{o_o})$ and the individual appearance of each object. 

\begin{figure*}[t]
\centering
    \begin{minipage}[b]{\textwidth}
       \includegraphics[trim={6 6 6 6},clip,width=0.16\linewidth]{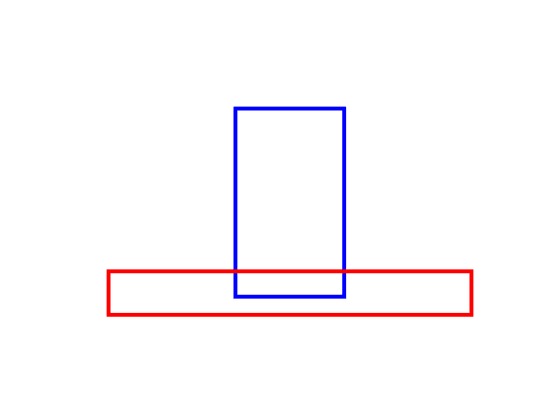}
       \includegraphics[trim={1.5cm 2.5cm 1.5cm 1.5cm},clip,width=0.16\linewidth]{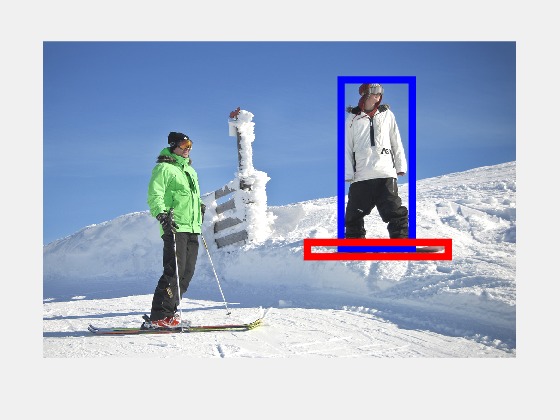}
       \includegraphics[trim={1.5cm 2.5cm 1.5cm 1.5cm},clip,width=0.16\linewidth]{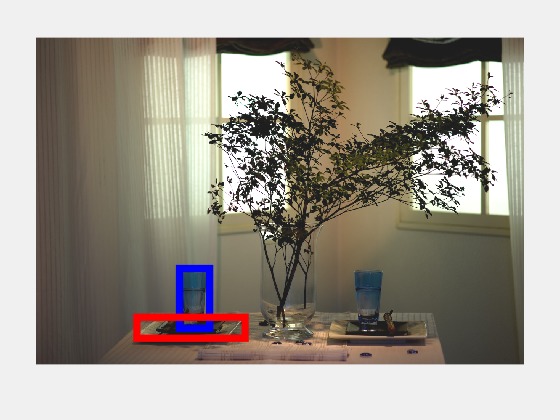}
       \includegraphics[trim={1.5cm 2.5cm 1.5cm 1.5cm},clip,width=0.16\linewidth]{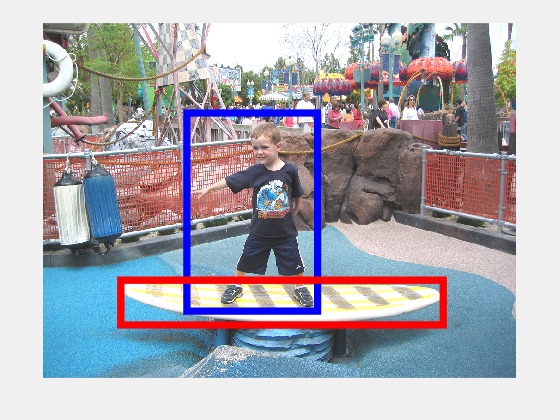}
       \includegraphics[trim={1.5cm 3.5cm 4.6cm 2.5cm},clip,width=0.16\linewidth]{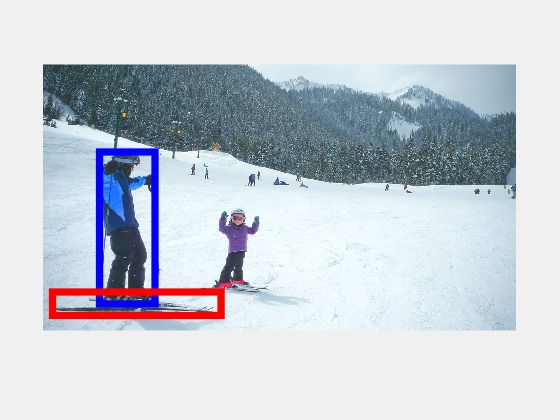}
       \includegraphics[trim={1.5cm 2.5cm 1.5cm 1.5cm},clip,width=0.16\linewidth]{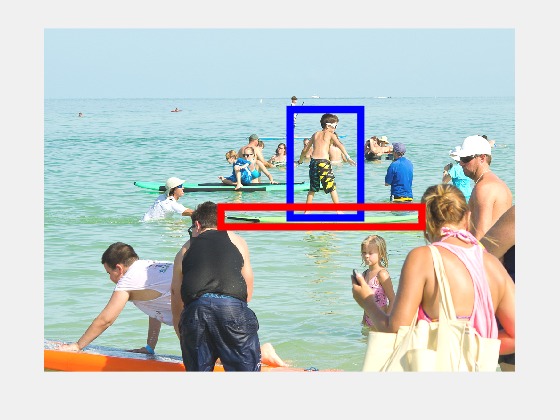}
       \vspace{-2ex}
	\end{minipage}   
	    
    \begin{minipage}[b]{\textwidth}   
       \includegraphics[trim={0 0 0 0},clip,width=0.16\linewidth]{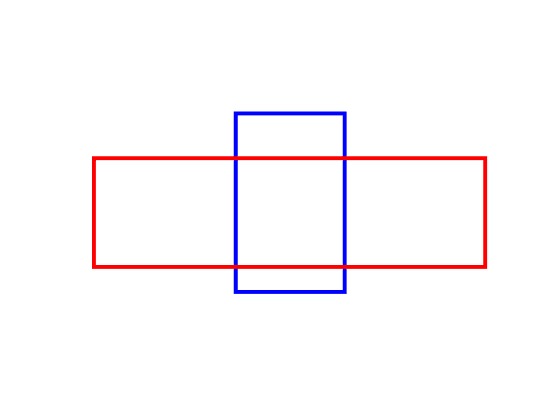}
       \includegraphics[trim={1.5cm 2.5cm 1.5cm 1.5cm},clip,width=0.16\linewidth]{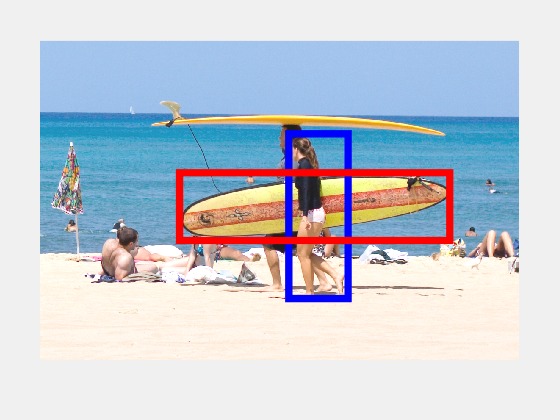}
       \includegraphics[trim={1.5cm 2.5cm 1.5cm 1.5cm},clip,width=0.16\linewidth]{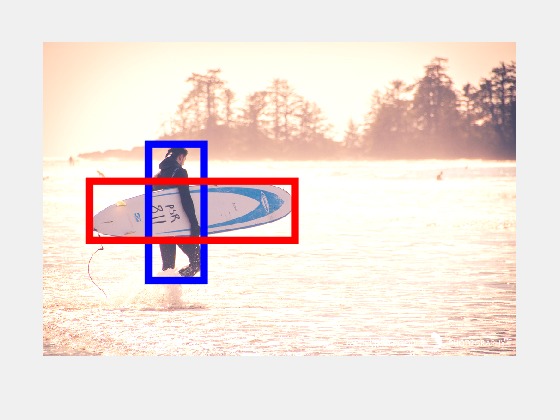}
       \includegraphics[trim={1.5cm 3cm 1.5cm 1cm},clip,width=0.16\linewidth]{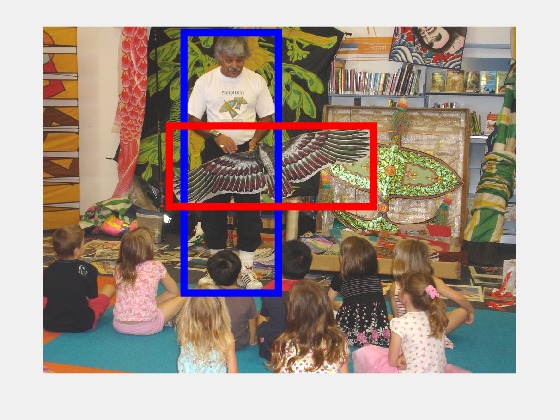}
       \includegraphics[trim={1.5cm 2.5cm 1.5cm 1.5cm},clip,width=0.16\linewidth]{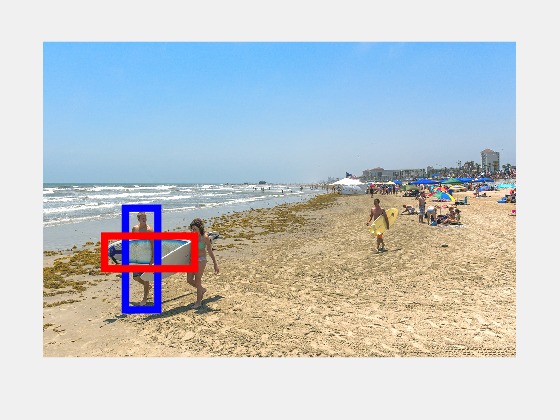}
       \includegraphics[trim={2.2cm 2.5cm 1.5cm 2cm},clip,width=0.16\linewidth]{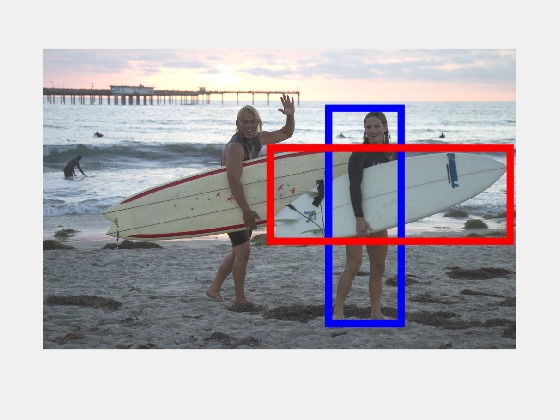}
	\vspace{-1ex}
    \end{minipage}

    \begin{minipage}[b]{\textwidth}
       \includegraphics[trim={1cm 0 0 0},clip,width=0.16\linewidth]{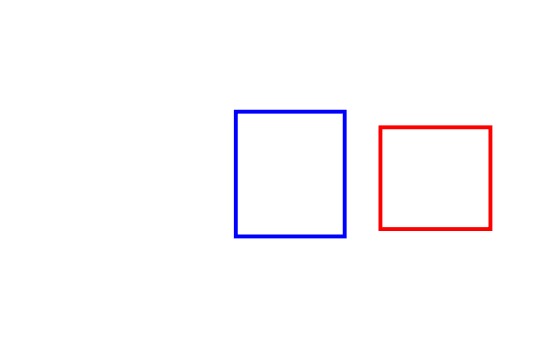}
       \includegraphics[trim={1.5cm 2.5cm 1.5cm 1.5cm},clip,width=0.16\linewidth]{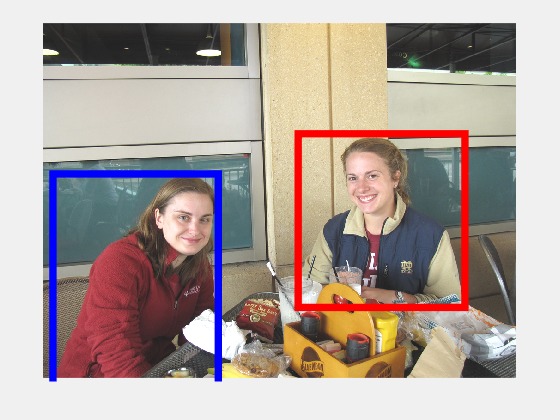}
       \includegraphics[trim={1.5cm 2.5cm 1.5cm 1.5cm},clip,width=0.16\linewidth]{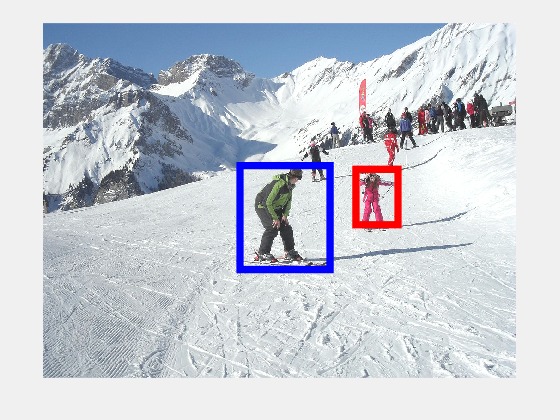}
       \includegraphics[trim={1.5cm 2.5cm 1.5cm 1.5cm},clip,width=0.16\linewidth]{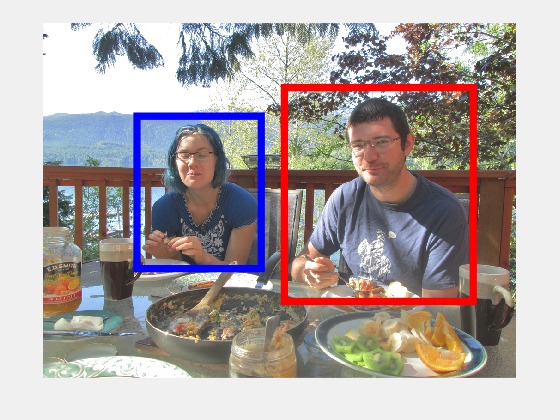}
       \includegraphics[trim={1.5cm 2.5cm 1.5cm 1.5cm},clip,width=0.16\linewidth]{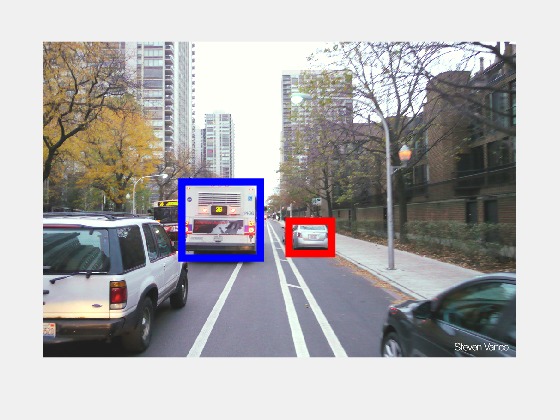}
       \includegraphics[trim={1.5cm 2.5cm 1.5cm 1.5cm},clip,width=0.16\linewidth]{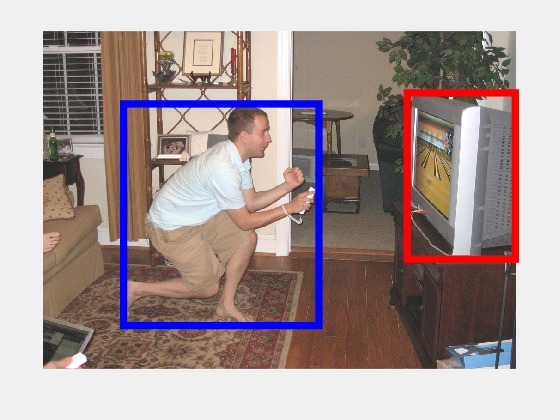}
       \vspace{-0.5ex}
    \end{minipage}   
    \setlength\abovecaptionskip{0.3cm}
    \caption{
      Examples for different GMM components of our spatial configuration model (one
      per row). In the first column we show the spatial configuration
      corresponding to the mean of the pairs of boxes per
      component. Note that our representation can capture subtle
      differences between spatial configurations, see e.g., row 1 and 2.} 
      \vspace{-.4cm}
    \label{GMM_components}
\end{figure*}

\spaceparagraph
\paragraph{Representing spatial configurations of objects.} Given two boxes $\bm{o_s} = [x_{s}, y_{s}, w_s, h_s]$, $\bm{o_o} = [x_{o}, y_{o}, w_o, h_o]$, where $(x, y)$ are the coordinates of the center of the box, and $(w,h)$ are the width and height of the box, we encode the spatial configuration with a 6-dimensional vector:
\vspace{-2pt}
\begin{align}
\label{eq:spatial}
\begin{split}
\bm{r}(\bm{o_s},\bm{o_o})& = [\underbrace{\frac{x_{o}-x_{s}}{\sqrt{w_s h_s}}}_{r_1}, \underbrace{\frac{y_{o}-y_{s}}{\sqrt{w_s h_s}}}_{r_2}, \underbrace{\sqrt{\frac{w_o h_o}{w_s h_s}}}_{r_3}, \\
&\underbrace{\frac{\bm{o_s} \cap \bm{o_o}}{\bm{o_s} \cup \bm{o_o}}}_{r_4}, \underbrace{\frac{w_s}{h_s}}_{r_5}, \underbrace{\frac{w_o}{h_o}}_{r_6}]
\end{split}
\end{align}
 
\noindent where $r_1$ and $r_2$ represent the renormalized translation between the two boxes, $r_3$ is the ratio of box sizes, $r_4$ is the overlap between boxes, and $r_5$, $r_6$ encode the aspect ratio of each box respectively. Directly adopting this feature as our representation might not be well suited for some spatial relations like ``next to" which are multimodal. Indeed, ``$s$ next to $o$" can either correspond to the spatial configuration ``$s$ left of $o$" or ``$s$ right of $o$". Instead, we propose to discretize the feature vector (\ref{eq:spatial}) into $k$ bins. For this, we assume that the spatial configurations $\bm{r}(\bm{o_s},\bm{o_o})$ are generated by a mixture of $k$ Gaussians and we fit the parameters of the Gaussian Mixture Model to the training pairs of boxes. We take the scores representing the probability of assignment to each of the $k$ clusters as our spatial representation. In our experiments, we use $k=400$, thus the spatial representation is a 400-dimensional vector. In Figure \ref{GMM_components}, we show examples of pairs of boxes for the most populated components of the trained GMM. We can observe that our spatial representation can capture subtle differences between configurations of boxes, see row 1 and row 2 of Figure~\ref{GMM_components}, where ``person on board" and ``person carry board" are in different clusters.
  
\spaceparagraph
\paragraph{Representing appearance of objects.} 
Our appearance features are given by the fc7 output of a Fast-RCNN~\cite{girshick15fastrcnn} trained to detect individual objects. In our experiments, we use Fast-RCNN with VGG16 pre-trained on ImageNet. As the extracted features have high dimensionality, we perform PCA on the L2-normalized features to reduce the number of dimensions from 4096 to 300. We concatenate the appearance features of the subject and object and apply L2-normalization again. 

\gotoline
\vspace{4pt}
Our final visual feature is a concatenation of the spatial configuration $\bm{r}(\bm{o_s},\bm{o_o})$ and the appearance of objects $[\bm{a}(\bm{o_s}), \bm{a}(\bm{o_o})]$. In our experiments, it has a dimensionality of $d=1000$. 
In the fully supervised setup, where each relation annotation is associated with a pair of object boxes in the image, we use ridge regression to train a multi-way relation classifier to assign a relation to a given 
visual feature. Training is performed jointly on all relation examples of the training set.

\gotoline
\vspace{4pt}
In the next section, we describe how we learn relation classifiers with only weak, image-level, annotations.

\subsection{Weakly-supervised learning of relations}
\label{model}

\begin{figure}[t]
	\begin{center}
   	\includegraphics[trim={0cm 0cm 0cm 0.2cm},clip,width=\linewidth]{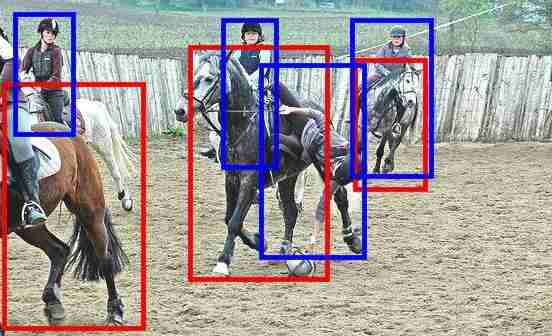}
	\end{center}
	\setlength\abovecaptionskip{-5pt}
   	\caption{Image from the COCO dataset~\cite{Lin2014a} associated with caption : \textit{``A \textcolor{blue}{person} \textcolor{Green}{falling off} the side of a \textcolor{red}{horse} as it rides"}. The boxes correspond to the possible candidates for object category $person$ (blue) and $horse$ (red). Our model has to disambiguate the right pair for the relation ``falling off" among 12 candidate pairs.}
   	\vspace{-.4cm}
\label{fig:coco_weak}
\end{figure}

Equipped with pre-trained detectors for individual objects, our goal here is to learn to detect and localize relations between objects, given image-level supervision only. For example, for a relation ``person falling off horse" we are given (multiple) object detections for ``person" and ``horse", but do not know which objects participate in the relation, as illustrated in Figure \ref{fig:coco_weak}. Our model is based on a weakly-supervised discriminative clustering objective~\cite{bach2008diffrac}, where we introduce latent variables to model which pairs of objects participate in the relation. We train a classifier for each predicate $r$ and incorporate weak annotations in the form of constraints on latent variables. Note that the relation classifiers are shared across object categories (eg. the relations ``person on horse" and ``cat on table" share the same classifier ``on") and can thus be used to predict unseen triplets.

\spaceparagraph
\paragraph{Discriminative clustering of relations.}
Our goal is to learn a set of classifiers $W = [\mathbf{w_1},...,\mathbf{w_R}] \in \mathbb{R}^{d \times R}$ where each classifier $\mathbf{w_r}$ predicts the likelihood of a pair of objects $(s,o)$ to belong to the $r^{th}$ predicate in a vocabulary of $R$ predicates. If strong supervision was provided for each pair of objects, we could learn $W$ by solving a ridge regression~:
\vspace{-0.5cm}
\begin{align}
\label{eq:ridgereg}
\min_{W \in \mathbb{R}^{d \times R}} \frac{1}{N} \|Z - XW \|^2_F + \lambda \|W\|^2_F
\end{align}
where $Z \in \{0,1\}^{N \times R}$ are the ground truth labels for each of the $N$ pairs of objects across all training images, and $X = [\mathbf{x_1}, ..., \mathbf{x_N}]^T$ is a $N \times d$ matrix where each row $\mathbf{x_k}$ is the visual feature corresponding to the $k^{th}$ pair of objects. However, in a weakly-supervised setup the entire matrix $Z$ is unknown. Building on DIFFRAC \cite{bach2008diffrac}, our approach is to optimize the cost : 
\vspace{-0.2cm}
\begin{align}
\label{DIFFRAC}
\min_{Z \in \mathcal{Z}} \min_{W \in \mathbb{R}^{d \times R}} \frac{1}{N} \|Z - XW \|^2_F + \lambda \|W\|^2_F
\end{align}
\noindent which treats $Z$ as a latent assignment matrix to be learnt together with the classifiers $W \in \mathbb{R}^{d \times R}$. Minimizing the first term encourages the predictions made by $W$ to match the latent assignments $Z$, while the second term is a L2-regularization on the classifiers $W$. This framework enables to incorporate weak annotations by constraining the space of valid assignment matrices $Z \in \mathcal{Z}$. The valid matrices $Z \in \{0,1\}^{N \times R}$ satisfy the multiclass constraint $Z \textbf{1}_R = \textbf{1}_N$ which assigns a pair of objects to one and only one predicate $r$. We describe in the next paragraph how to incorporate the weak annotations as constraints. 

\spaceparagraph
\paragraph{Weak annotations as constraints.} For an image, we are given weak annotations in the form of triplets $t = (s,r,o) \in \mathcal{T}$. Having such triplet $(s,r,o)$ indicates that at least one of the pairs of objects with object categories $(s,o)$ is in relation $r$. Let us call $\mathcal{N}_t$ the subset of pairs of objects in the image that correspond to object categories $(s,o)$. The rows of $Z$ that are in subset $\mathcal{N}_t$ should satisfy the constraint : 
\vspace{-0.2cm}
\begin{align}
\label{eq:multi_instance}
\sum_{n \in \mathcal{N}_t} Z_{nr} \ge 1
\end{align}
\vspace{-0.4cm}

This constraint ensures that at least one of the pair of objects in the subset $\mathcal{N}_t$ is assigned to predicate $r$. For instance, in case of the image in Figure~\ref{fig:coco_weak} that contains 12 candidate pairs of objects that potentially match the triplet $t=(person,falling~off,horse)$, the constraint (\ref{eq:multi_instance}) imposes that at least one of them is in relation $falling ~off$.

\spaceparagraph
\paragraph{Triplet score.} 
At test time, we can compute a score for a pair of boxes $(\bm{o}_s,\bm{o}_o)$ relative to a triplet $t=(s,r,o)$ as
\begin{align}
\label{eq:score}
\begin{split}
S((\bm{o}_s,\bm{o}_o) ~|~ t) = v_{rel}((\bm{o}_s,\bm{o}_o) ~|~ r) + \alpha_{sub} v_{sub}(\bm{o}_s ~|~ s) \\ + \alpha_{obj} v_{obj}(\bm{o}_o ~|~ o) + \alpha_{lang} \ell((s,o) ~|~ r),
\end{split}
\end{align}
\vspace{-0.4cm}

\noindent where $v_{rel}((\bm{o}_s,\bm{o}_o) | r) = \mathbf{x}_{(\bm{o}_s,\bm{o}_o)} \mathbf{w_r}$
is the score returned by the classifier $\mathbf{w}_r$ for predicate $r$ (learnt by optimizing (\ref{DIFFRAC})) for the visual representation $\mathbf{x}_{(\bm{o}_s,\bm{o}_o)}$ of the pair of boxes. $v_{sub}(\bm{o}_s | s)$ and $v_{obj}(\bm{o}_o | o)$ are the object class likelihoods returned by the object detector. $\ell((s,o) | r)$ is a score of a language model that we can optionally combine with our visual model. 

\spaceparagraph
\paragraph{Optimization.} We optimize the cost in (\ref{DIFFRAC}) on pairs of objects in the training set using a variant of the Frank-Wolfe algorithm~\cite{miech17,Osokin16}. The hyperparameters $(\alpha_{sub}, \alpha_{obj}, \alpha_{lang})$ are optimized on an held-out fully-annotated validation set which has no overlap with our training and test sets. In our experiments we use the validation split of \cite{Johnson2015} of the Visual Genome dataset \cite{Krishna2016}. Unless otherwise specified, the candidate pairs, both at training and test time, are the outputs of the object detector \cite{girshick15fastrcnn} that we fine-tuned on the Visual Relationship Detection dataset \cite{Lu16}. For each image, we keep the object candidates whose confidence scores is above 0.3 among the top 100 detections. Non-maximum suppression with threshold 0.3 is applied to handle multiple detections. This results in an average of 18 object detections per image, i.e.\ around 300 pairs of boxes.

\section{Experiments}

In this section, we evaluate the performance of our model on two
datasets for different evaluation setups. First, we evaluate our new
visual representation for relations on the Visual Relationship
Detection dataset \cite{Lu16}. We show results with our
weakly-supervised model learned from image-level supervision and
present large improvements over the state of the art for detecting
unseen triplets (zero-shot detection). Second, we evaluate our model for the task of unusual
triplets retrieval and localization on our new UnRel dataset.

\subsection{Recall on Visual Relationship Detection dataset}

\begin{table*}\centering
{\small
\ra{1}
\begin{tabular}{@{}rrrrrcrrcrr@{}}\toprule
&& & \multicolumn{2}{c}{Predicate Det.} & \phantom{abc} & \multicolumn{2}{c}{Phrase Det.} & \phantom{abc}& \multicolumn{2}{c}{Relationship Det.}\\
&&								& All & Unseen && All & Unseen && All & Unseen \\\midrule
&& \textbf{Full sup.}\\
a. && Visual Phrases \cite{Sadeghi2011} &  0.9 	& -		&& 0.04 & -		&&  -    & - \\
b. && Visual \cite{Lu16} 				&  7.1	& 3.5	&& 2.2 	& 1.0	&&  1.6	 & 0.7 \\
c. && Language (likelihood) \cite{Lu16} 			& 18.2  &  5.1	&& 0.08	& 0.00	&& 0.08  & 0.00	\\
d. && Visual + Language \cite{Lu16} 	&  47.9 & 8.5	&& 16.2 & 3.4	&&  13.9 & 3.1 	\\
e. && Language (full) \cite{Lu16} 			&  48.4 & 12.9 	&& 15.8	& 4.6	&&  13.9 & 4.3	\\
\rule{0pt}{3ex}  
f. && Ours [S] 						&  42.2 & 22.2	&& 13.8	& 7.4	&& 	12.4 & 7.0	\\
g. && Ours [A] 						&  46.3 & 16.1	&& 14.9	& 5.6	&&	12.9 & 5.0	\\
h. && Ours [S+A] 					&  50.4 & \textbf{23.6} && 16.7  &\textbf{7.4}	&& 14.9 &  \textbf{7.1}	\\
i. && Ours [S+A] + Language \cite{Lu16} 
										& \textbf{52.6} & 21.6 && \textbf{17.9} & 6.8 && \textbf{15.8} & 6.4 \\
\rule{0pt}{3ex}  
 && \textbf{Weak sup.}\\
j. && Ours [S+A]					& 46.8 	& 19.0	&& 	16.0 & 6.9	&& 14.1 & 6.7	\\
k. && Ours [S+A] - Noisy 		& 46.4  	& 17.6	&& 	15.1& 6.0	&& 13.4 & 5.6	\\
\bottomrule
\end{tabular}
\setlength\abovecaptionskip{5pt}
\caption{Results on Visual Relationship Detection dataset \cite{Lu16} for R@50. See appendix for results with R@100.} 
\vspace{-.4cm}
\label{tab:results_vrd}
}
\end{table*}

\paragraph{Dataset.} 
We evaluate our method on the Visual Relationship Detection dataset~\cite{Lu16} following the original experimental setup. This dataset contains 4000 training and 1000 test images with ground truth annotations for relations between pairs of objects. Due to the specific train/test split provided by~\cite{Lu16}, 10\% of test triplets are not seen at training and allow for evaluation of zero-shot learning. Some of these triplets are rare in the linguistic and visual world (e.g. ``laptop on stove"), but most of them are only infrequent in the training set or have not been annotated (e.g. ``van on the left of car"). Around 30K triplets are annotated in the training set, with an average of 7.5 relations per image. The dataset contains 100 objects and 70 predicates, i.e.\ $100\times 100 \times 70$ possible triplets. However there are only 6672 different annotated triplets. 

\spaceparagraph
\paragraph{Evaluation set-up.} Following \cite{Lu16}, we compute
\verb+recall@x+ which corresponds to the proportion of ground truth
pairs among the \verb+x+ top scored candidate pairs in each image. We use
the scores returned by (\ref{eq:score}) to sort the candidate pairs of
boxes. The evaluation is reported for three setups. In
\textbf{predicate detection}, candidate pairs of boxes are ground
truth boxes, and the evaluation only focuses on the quality of the
predicate classifier. In the other two more realistic setups, the
subject and object confidence scores are provided by an object
detector and we also check whether the candidate boxes intersect the
ground truth boxes : either both subject and object boxes should match
(\textbf{relationship detection}), or the union of them should match
(\textbf{phrase detection}). For a fair comparison with~\cite{Lu16},
we report results using the same set of R-CNN~\cite{girshick2014rcnn}
object detections as them. The localization is evaluated with IoU = 0.5.

\spaceparagraph
\paragraph{Benefits of our visual representation.}
We first evaluate the quality of our visual representation in a fully
supervised setup where the ground truth spatial localization for each
relation is known, i.e.\ we know which objects in the image are
involved in each relation at training time. For this, we solve the
multi-label ridge regression in (\ref{eq:ridgereg}). Training with one-vs-rest SVMs gives similar results. We compare three types of features described in Section~\ref{visual_representation} in Table~\ref{tab:results_vrd}:
[S] the spatial representation (f.), [A] the appearance representation (g.) and [S+A]
the concatenation of the two (h.). We compare with the Visual Phrases model~\cite{Sadeghi2011} and several variants of~\cite{Lu16}
\footnote{When running the evaluation code of \cite{Lu16}, we found slighlty better performance than what is reported in their paper. See appendix for more details.}
: Visual model alone (b.), Language (likelihood of a relationship) (c.), combined Visual+Language model (d.). In row (e.) we also report the performance of the full language model of \cite{Lu16}, that scores the candidate pairs of boxes based on their predicted object categories, that we computed using the model and word embeddings provided by the authors.
Because their language model is orthogonal to our visual model, we can combine them together (i.).  
The results are presented on the complete test set (column All) and on the zero-shot learning split (column Unseen).
Table~\ref{tab:results_vrd} shows that our combined visual features [S+A] improve over the visual features of \cite{Lu16} by  $40\%$ on the task of predicate detection and more than $10\%$ on the hardest task of relationship detection. Furthermore, our purely visual features without any use of language (h.) reach comparable performance to the combined Visual+Language features of~\cite{Lu16} and reach state-of-the-art performance (i.) when combined with the language scores of~\cite{Lu16}. The good performance of our spatial features [S] alone (f.) confirms the observation we made in Figure \ref{GMM_components} that our spatial clusters group pairs of objects in similar relations. That could partly explain why the visual model of~\cite{Lu16} has low performance. Their model learns a classifier only based on the appearance of the union of the two object boxes and lacks information about their spatial configuration.

\spaceparagraph
\paragraph{Weak supervision.}
We evaluate our weakly-supervised classifiers $W$ learned on image-level labels as described in Section \ref{model}. We use the ground truth annotations of the Visual Relationship Detection dataset as image-level labels. We report the results using our combined spatial and appearance features (j.) in Table \ref{tab:results_vrd}. We see that when switching to weak supervision the recall@50 only drops from $50.4\%$ to $46.8\%$ for predicate detection and has limited influence on the other tasks. This is an interesting result as it suggests that, given pre-trained object detectors, weak image-level annotations are enough to learn good classifiers for relations. To investigate this further we have also tried to learn relation classifiers directly from noisy image-level labels without inferring at training time which objects participate in which relation.
For each triplet $t=(s,r,o)$ in an image containing candidate pairs of boxes $(\bm{o_s}, \bm{o_o})$ we randomly select one of the pairs as being in relation $r$ and discard the other object pairs. This is equivalent to training in a fully-supervised setup but with noisy labels. 
The performance obtained by this classifier (k.) is  below our weakly-supervised learning set-up but is surprisingly high. We believe that this is related to a particular bias present in the Visual Relationship Detection dataset~\cite{Lu16}, which contains many images with only two prominent objects involved in a specific relation (more than half of the triplets fall into this category). To underline the ability of the weakly-supervised model to disambiguate the correct bounding boxes, we evaluate in a more difficult setup where we replace the candidate test pairs of \cite{Lu16} by all candidate pairs formed by objects of confidence scores above 0.3. This multiplies by 5 the number of candidate pairs, resulting in an increased level of ambiguity. In this more challenging setup, our approach obtains a recall@50 for Phrase Detection (resp. Relationship Detection) of 17.9\% (resp. 12.0\%) compared to the "Ours [S+A] Noisy" baseline which drops to 15.3\% (resp. 10.1\%).

\begin{figure*}[t]
\centering
	\hspace{0.005\textwidth}
	\begin{minipage}[b]{0.60\textwidth}
    	\centering
    	\emph{correctly recognized relations}\\
    	\vspace{0.4ex}
	\end{minipage}
	\begin{minipage}[b]{0.19\textwidth}
    	\centering
    	\emph{missing ground truth}\\
    	\vspace{0.4ex}
	\end{minipage}
	\begin{minipage}[b]{0.19\textwidth}
    \centering
    \emph{incorrectly recognized}\\
    	\vspace{0.4ex}
	\end{minipage}
	
	\begin{minipage}[b]{0.005\textwidth}
    	\centering
    	\begin{turn}{90}
    	seen triplets
    	\end{turn}
    	\vspace{3.5ex}
    \end{minipage}
    \hspace{0.01\textwidth}
    \begin{minipage}[b]{0.18\textwidth}
    	\centering
       	\includegraphics[trim={0cm 4cm 0cm 2cm},clip,width=0.95\linewidth,cfbox={green 2pt 2pt}]{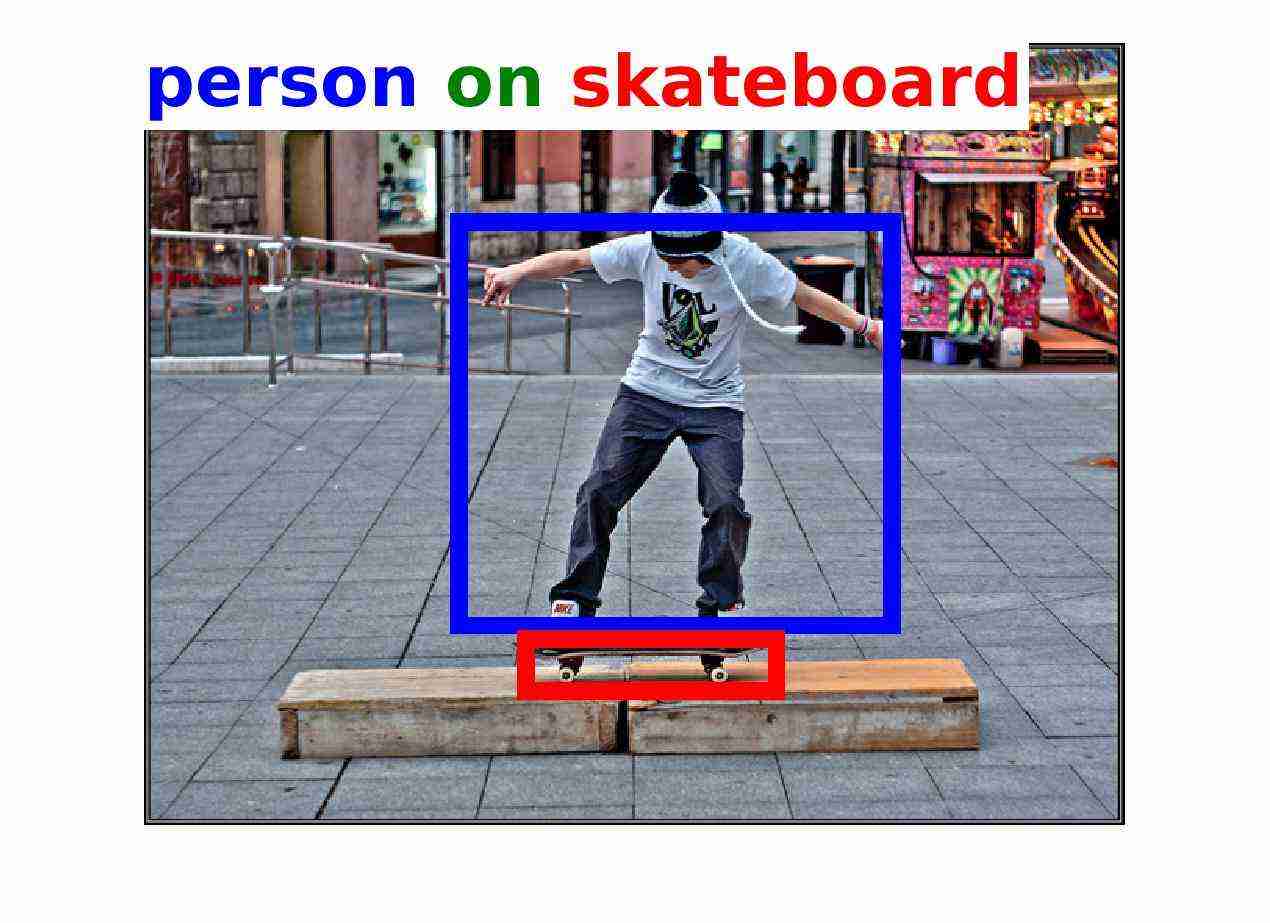}\\
       	\small{
       	\vspace{0.2ex}
       	GT: on, has\\
       	\cite{Lu16}: ride
       	}
       	\vspace{0.3ex}
    \end{minipage}
    \hspace{0.005\textwidth}
    \begin{minipage}[b]{0.18\textwidth}
       \centering
       \includegraphics[trim={0cm 3.7cm 0cm 1cm},clip,width=0.95\linewidth,cfbox={green 2pt 2pt}]{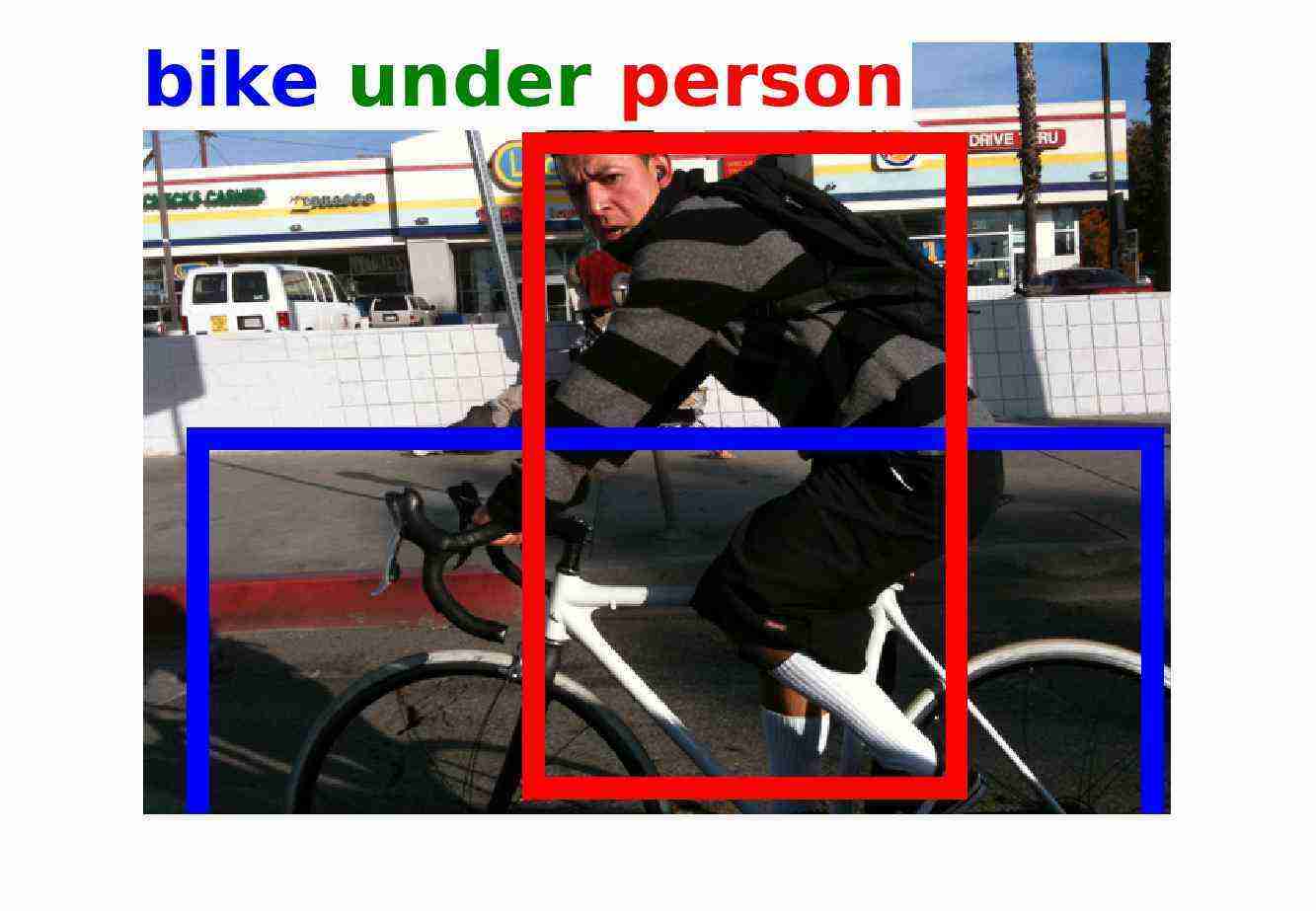}\\
       \vspace{0.2ex}
       \small{
       GT: under\\
       \cite{Lu16}: on
       }
       \vspace{0.3ex}
    \end{minipage}
    \hspace{0.005\textwidth}
    \begin{minipage}[b]{0.18\textwidth}
    	\centering
       	\includegraphics[trim={2cm 3.2cm 2cm 1cm},clip,width=0.95\linewidth,cfbox={green 2pt 2pt}]{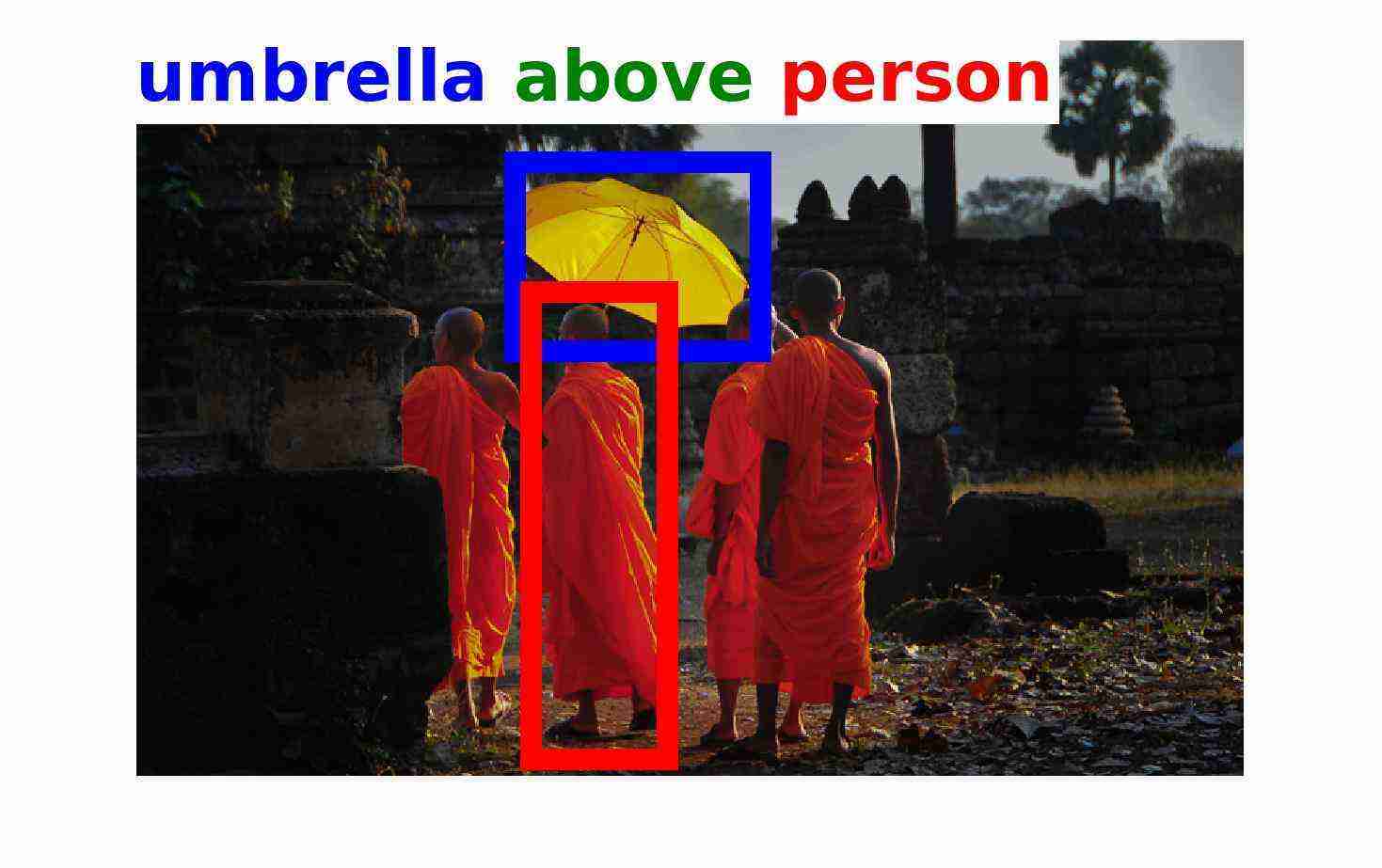}\\
       	\vspace{0.2ex}
       	\small{
       	GT: above\\
       	\cite{Lu16}: cover
       	}
       	\vspace{0.3ex}
    \end{minipage}
    \hspace{0.005\textwidth}
    \begin{minipage}[b]{0.18\textwidth}
    	\centering
       	\includegraphics[trim={1.9cm 3cm 1.8cm 1cm},clip,width=0.95\linewidth,cfbox={yellow 2pt 2pt}]{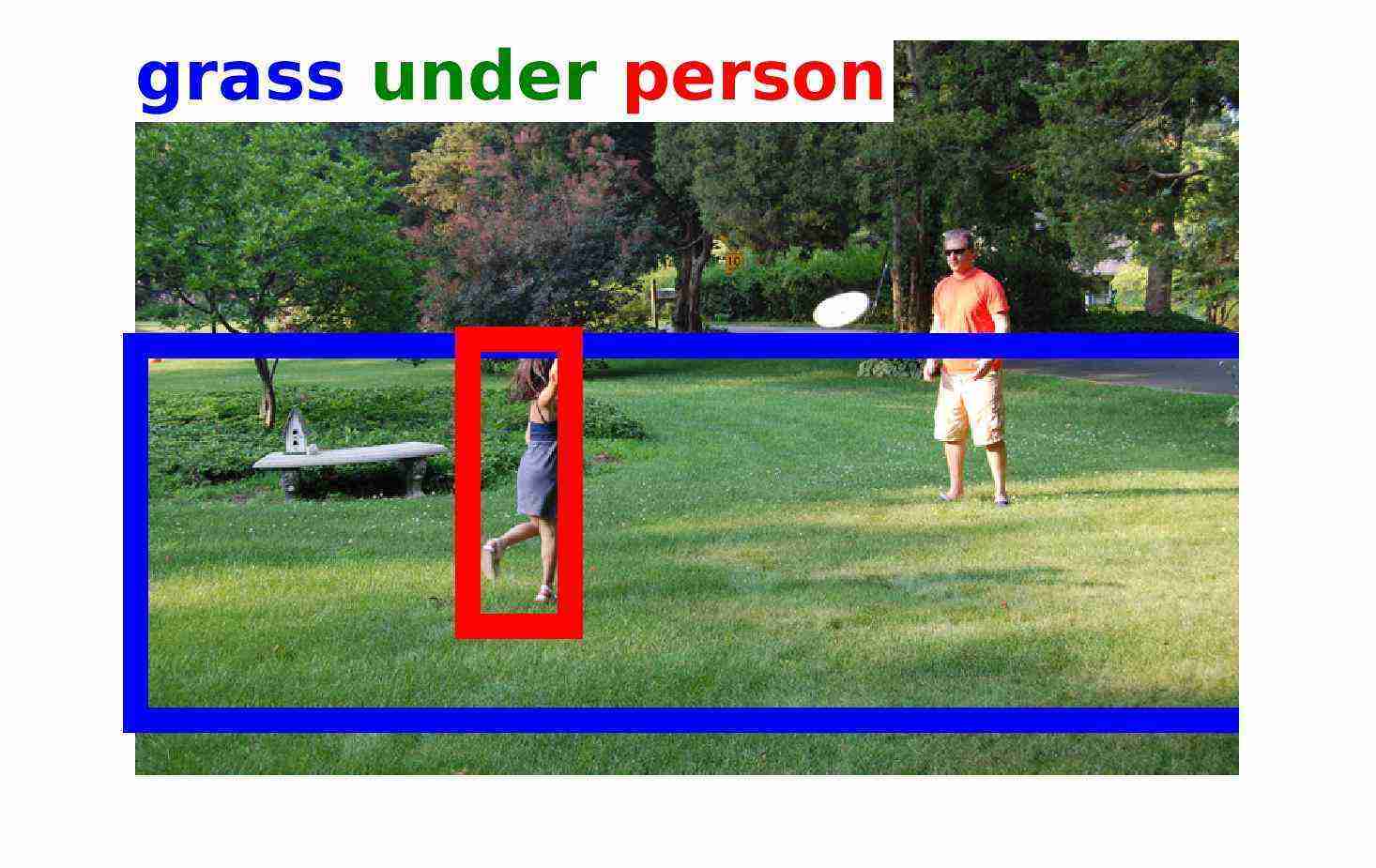}\\
       	\vspace{0.2ex}
       	\small{
       	GT: none\\
       	\cite{Lu16}: behind
       	}
       	\vspace{0.3ex}
    \end{minipage}
    \hspace{0.005\textwidth}  
    \begin{minipage}[b]{0.18\textwidth}
    	\centering
       	\includegraphics[trim={1.5cm 3cm 1.7cm 1cm},clip,width=0.95\linewidth,cfbox={red 2pt 2pt}]{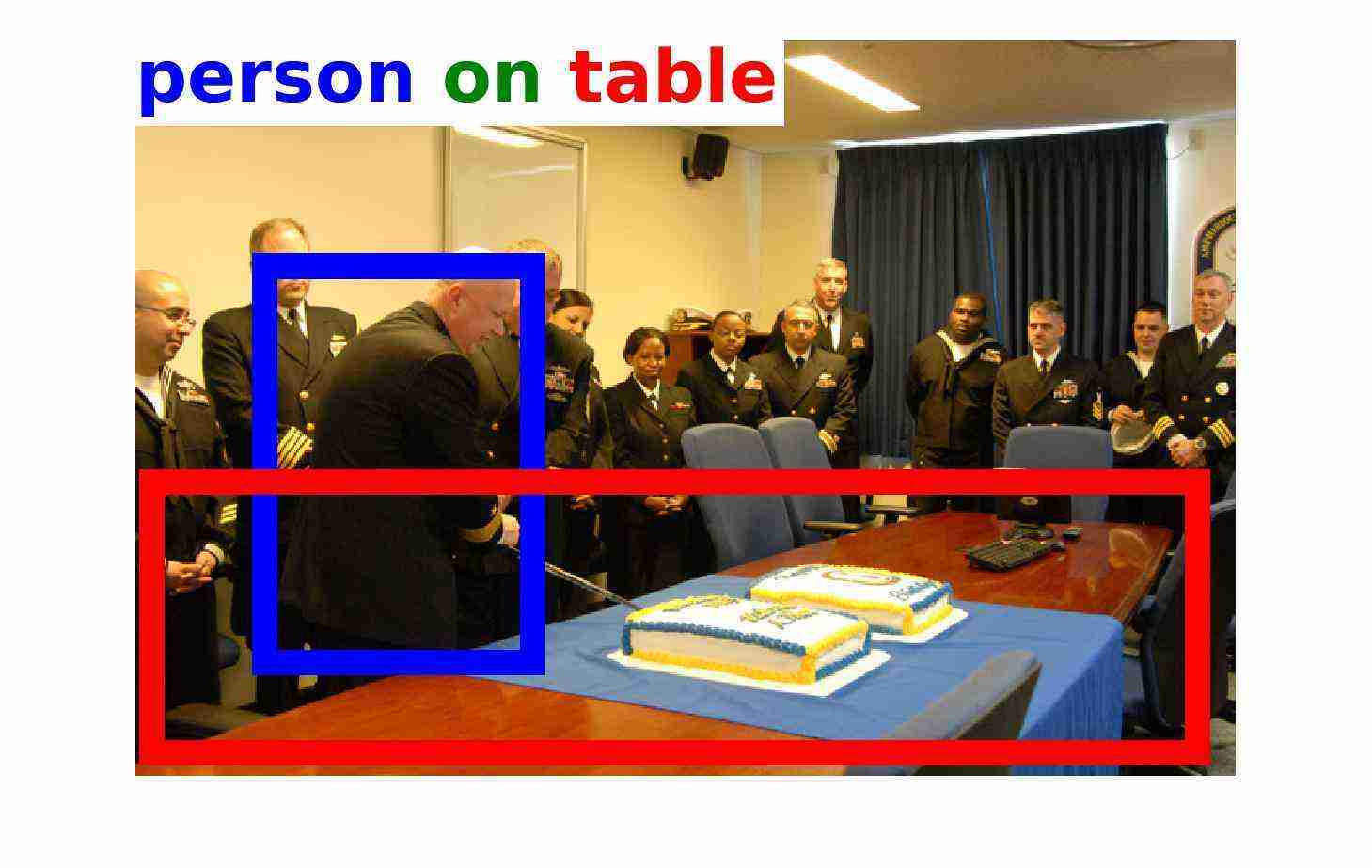}\\
       	\vspace{0.2ex}
       	\small{
       	GT: next to, behind\\
       	\cite{Lu16}: on
       	}
      	\vspace{0.3ex}
    \end{minipage} 
    
	\begin{minipage}[b]{0.005\textwidth}
    	\centering
    	\begin{turn}{90}
    	seen triplets
    	\end{turn}
    	\vspace{3.5ex}
    \end{minipage}
    \hspace{0.01\textwidth}
	\begin{minipage}[b]{0.18\textwidth}
    	\centering
       	\includegraphics[trim={0cm 3.5cm 1cm 0cm},clip,width=0.95\linewidth,cfbox={green 2pt 2pt}]{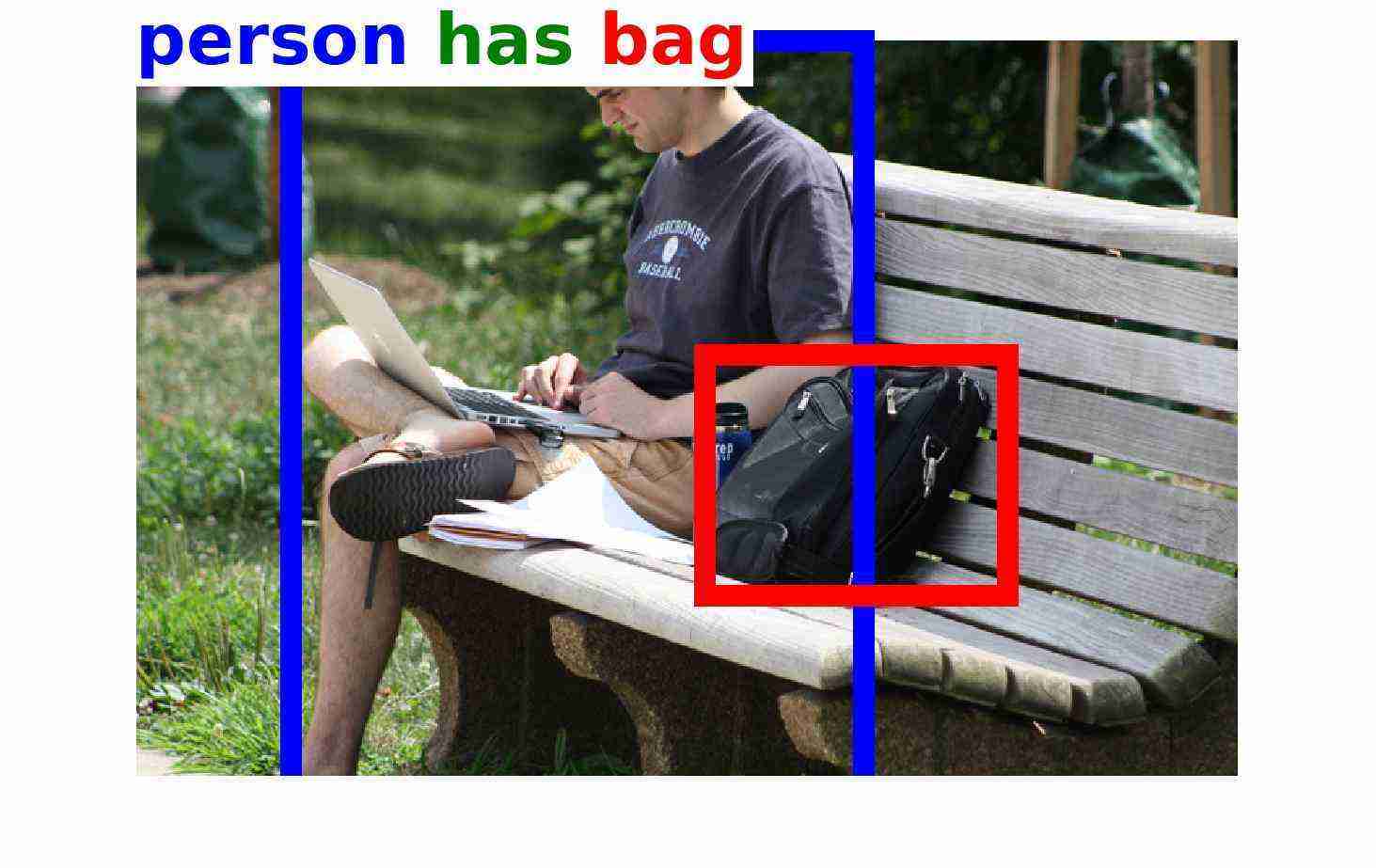}\\
       	\vspace{0.2ex}
       	\small{
       	GT: has\\
       	\cite{Lu16}: hold
       	}
       	\vspace{0.3ex}
    \end{minipage}
    \hspace{0.005\textwidth}      
	\begin{minipage}[b]{0.18\textwidth}
    	\centering
       	\includegraphics[trim={2cm 4cm 1.8cm 1cm},clip,width=0.95\linewidth,cfbox={green 2pt 2pt}]{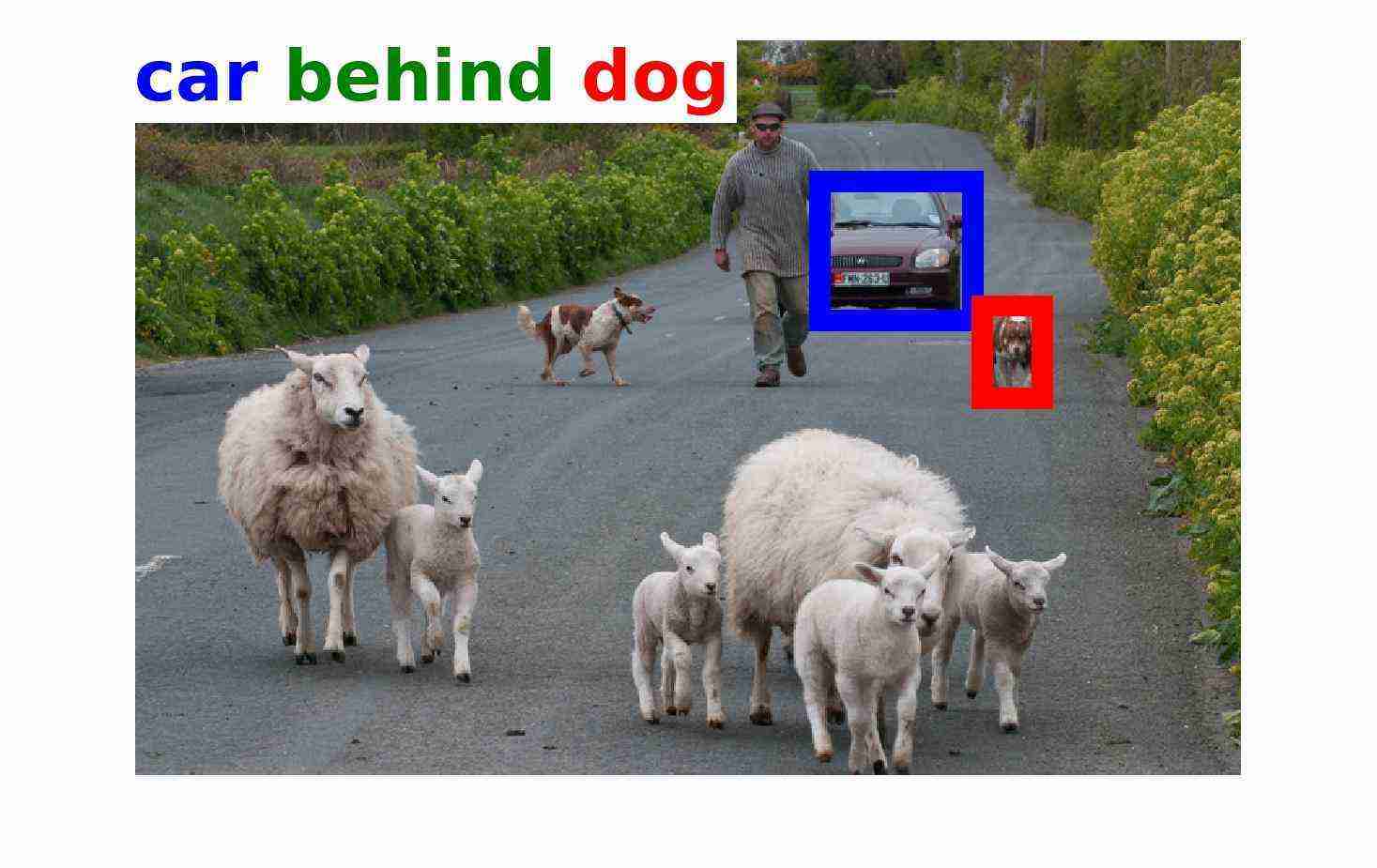}\\
       	\vspace{0.2ex}
       	\small{
       	GT: behind\\
       	\cite{Lu16}: behind
       	}
       	\vspace{0.3ex}
    \end{minipage}
    \hspace{0.005\textwidth}
	\begin{minipage}[b]{0.18\textwidth}
    	\centering
       	\includegraphics[trim={1cm 3.5cm 1.5cm 1cm},clip,width=0.95\linewidth,cfbox={green 2pt 2pt}]{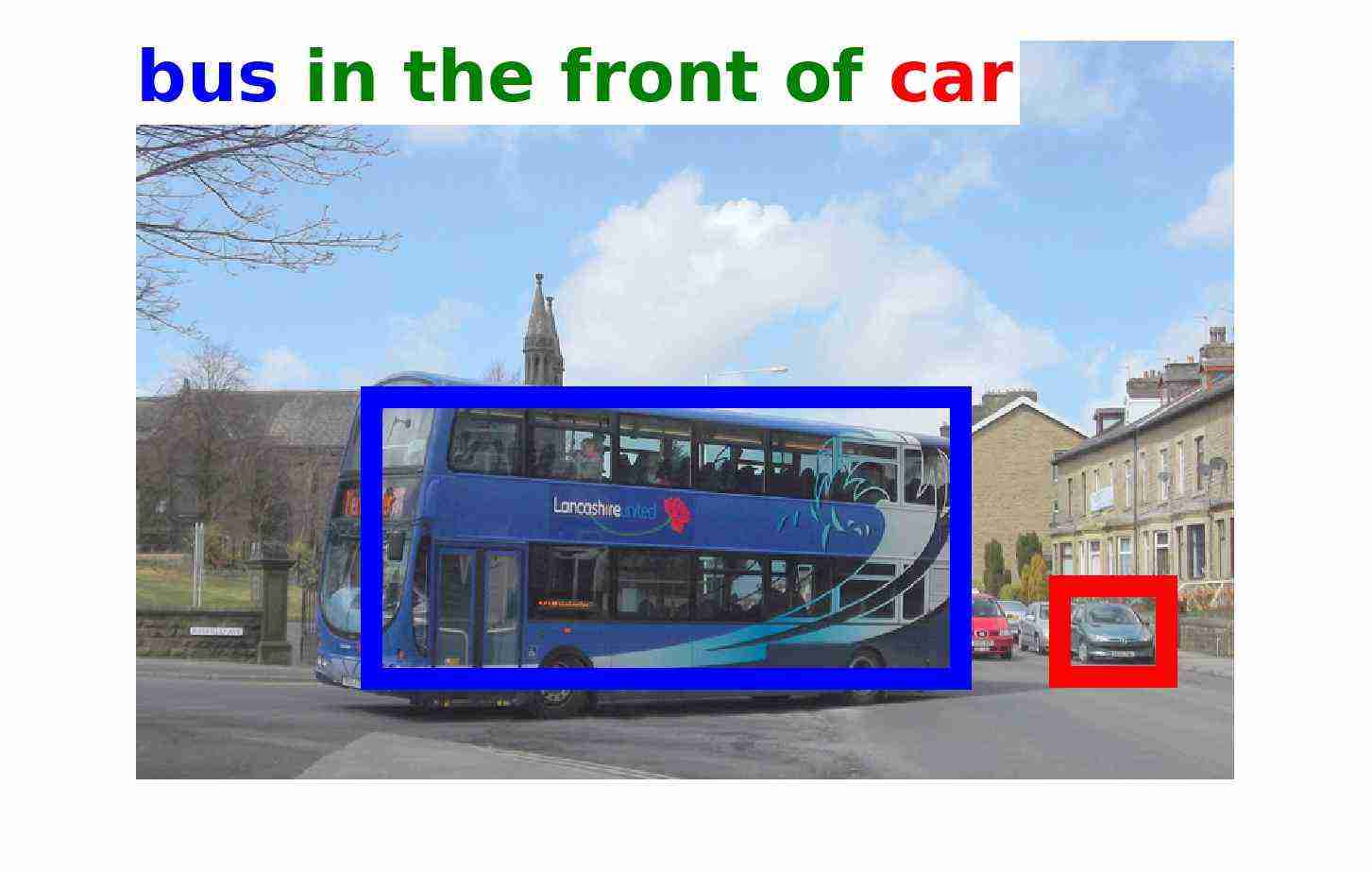}\\
       	\vspace{0.2ex}
       	\small{
       	GT: in the front of\\
       	\cite{Lu16}: behind
       	} 
       	\vspace{0.3ex}
    \end{minipage}
    \hspace{0.005\textwidth}
    \begin{minipage}[b]{0.18\textwidth}
    	\centering
      	\includegraphics[trim={2cm 4cm 2cm 1cm},clip,width=0.95\linewidth,cfbox={yellow 2pt 2pt}]{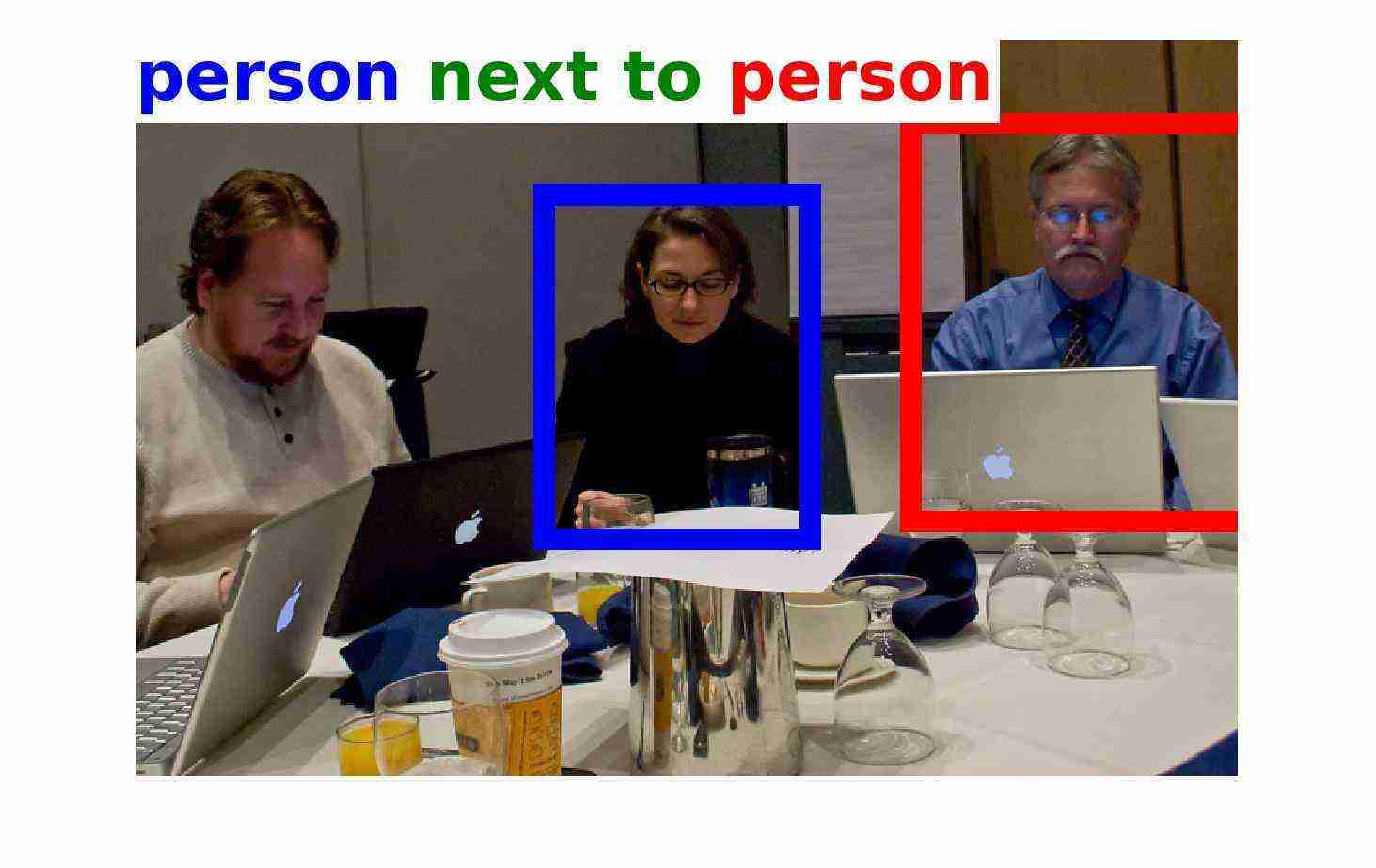}\\
      	\vspace{0.2ex}
      	\small{
       	GT: on the right of\\
       	\cite{Lu16}: next to
       	}
       	\vspace{0.3ex}
    \end{minipage}
    \hspace{0.005\textwidth}
    \begin{minipage}[b]{0.18\textwidth}
    	\centering
       	\includegraphics[trim={1cm 4cm 1cm 0cm},clip,width=0.95\linewidth,cfbox={red 2pt 2pt}]{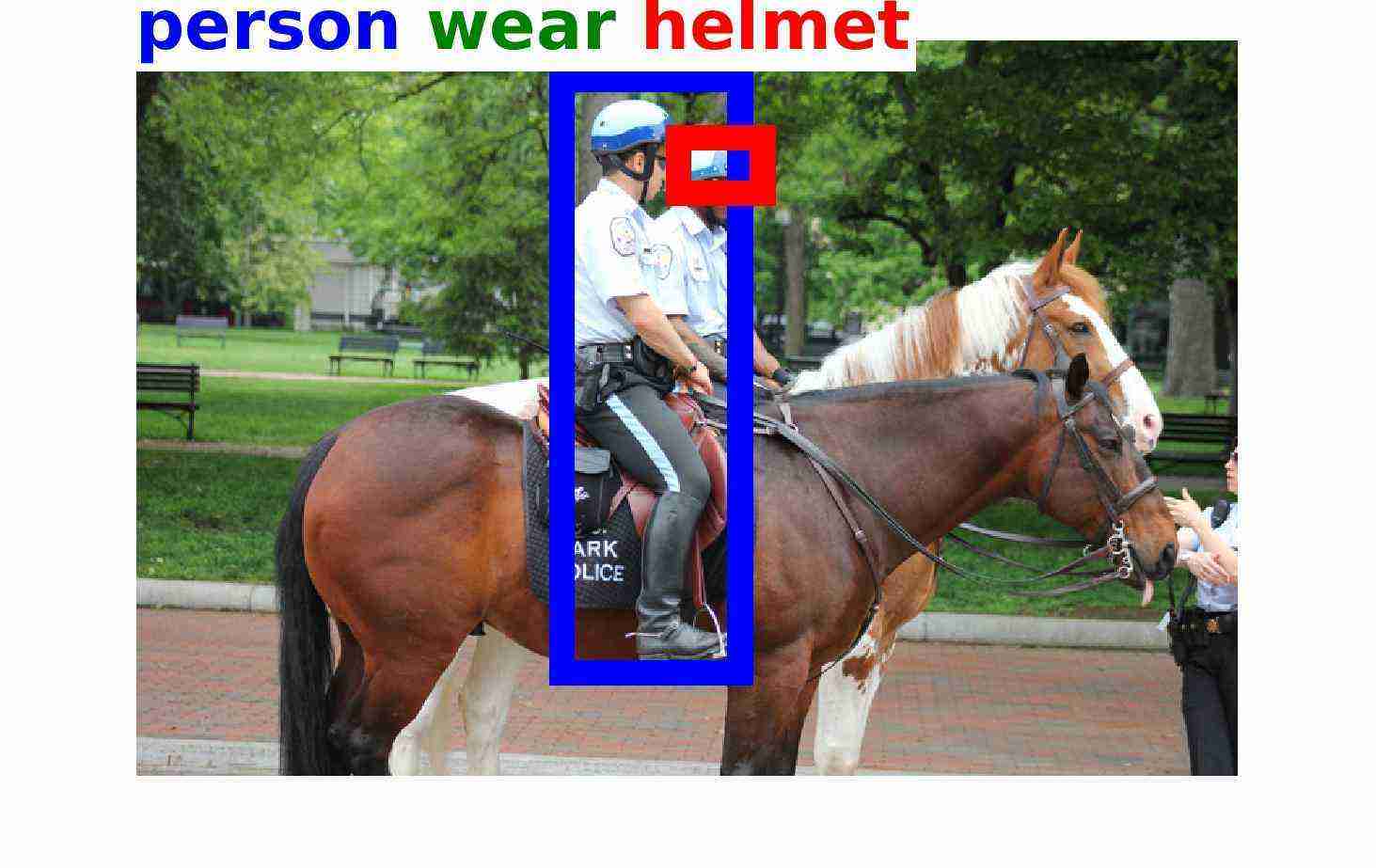} \\
       	\vspace{0.2ex}
       	\small{
       	GT: none\\
       	\cite{Lu16}: has
       	}
       	\vspace{0.3ex}
    \end{minipage}
     	
	\begin{minipage}[b]{0.005\textwidth}
    	\centering
    	\begin{turn}{90}
    	unseen triplets
    	\end{turn}
    	\vspace{1.5ex}
   	\end{minipage}
    \hspace{0.01\textwidth}
    \begin{minipage}[b]{0.18\textwidth}
    	\centering
       	\includegraphics[trim={0cm 4cm 0cm 1.2cm},clip,width=0.95\linewidth,cfbox={green 2pt 2pt}]{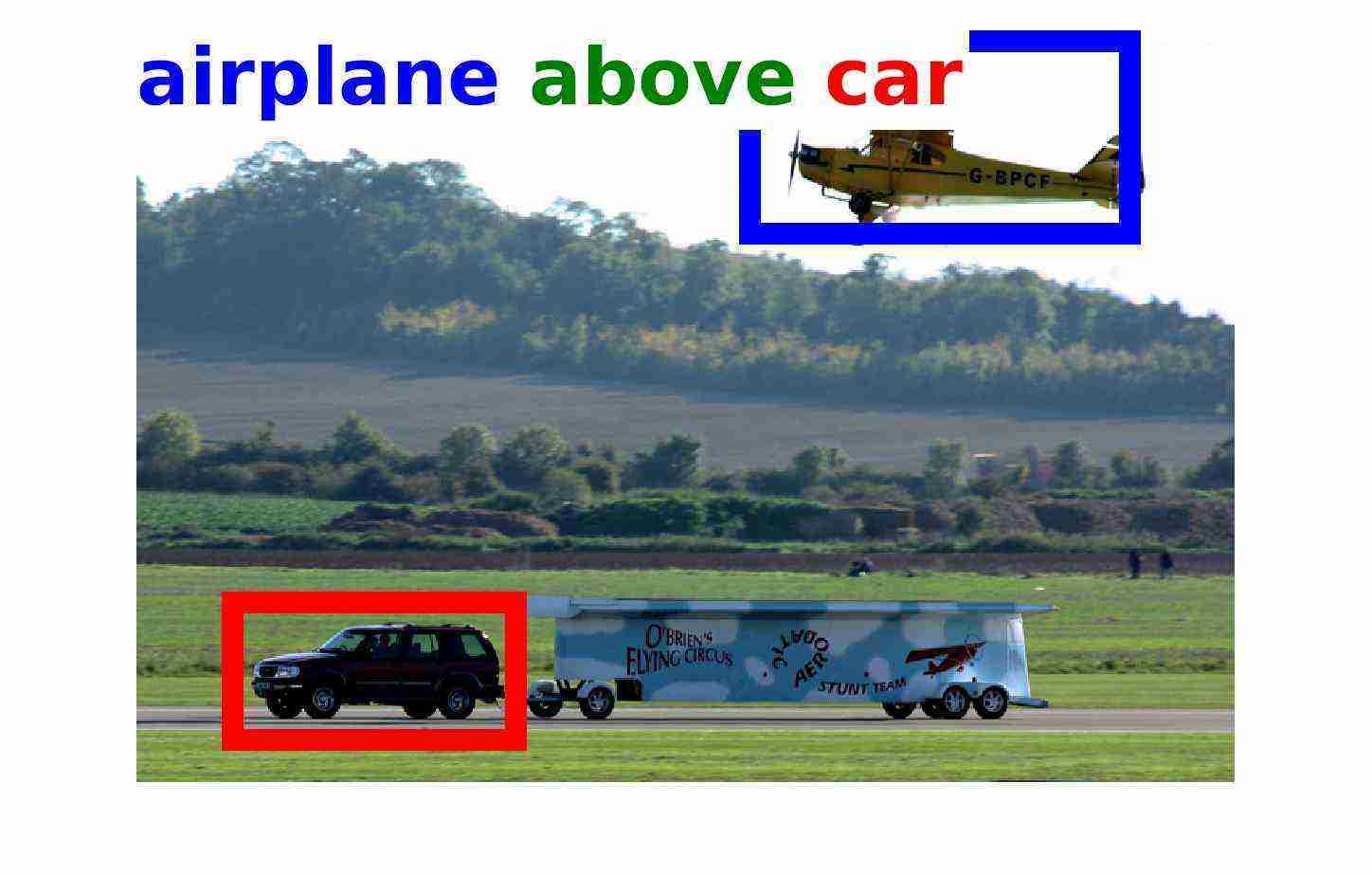}\\
       	\vspace{0.2ex}
       	\small{
       	GT: above\\
       	\cite{Lu16}: in the front of
       	}
       	\vspace{0.3ex}
    \end{minipage}
    \hspace{0.005\textwidth}
    \begin{minipage}[b]{0.18\textwidth}
    	\centering
       	\includegraphics[trim={0cm 3.5cm 0cm 1.5cm},clip,width=0.95\linewidth,cfbox={green 2pt 2pt}]{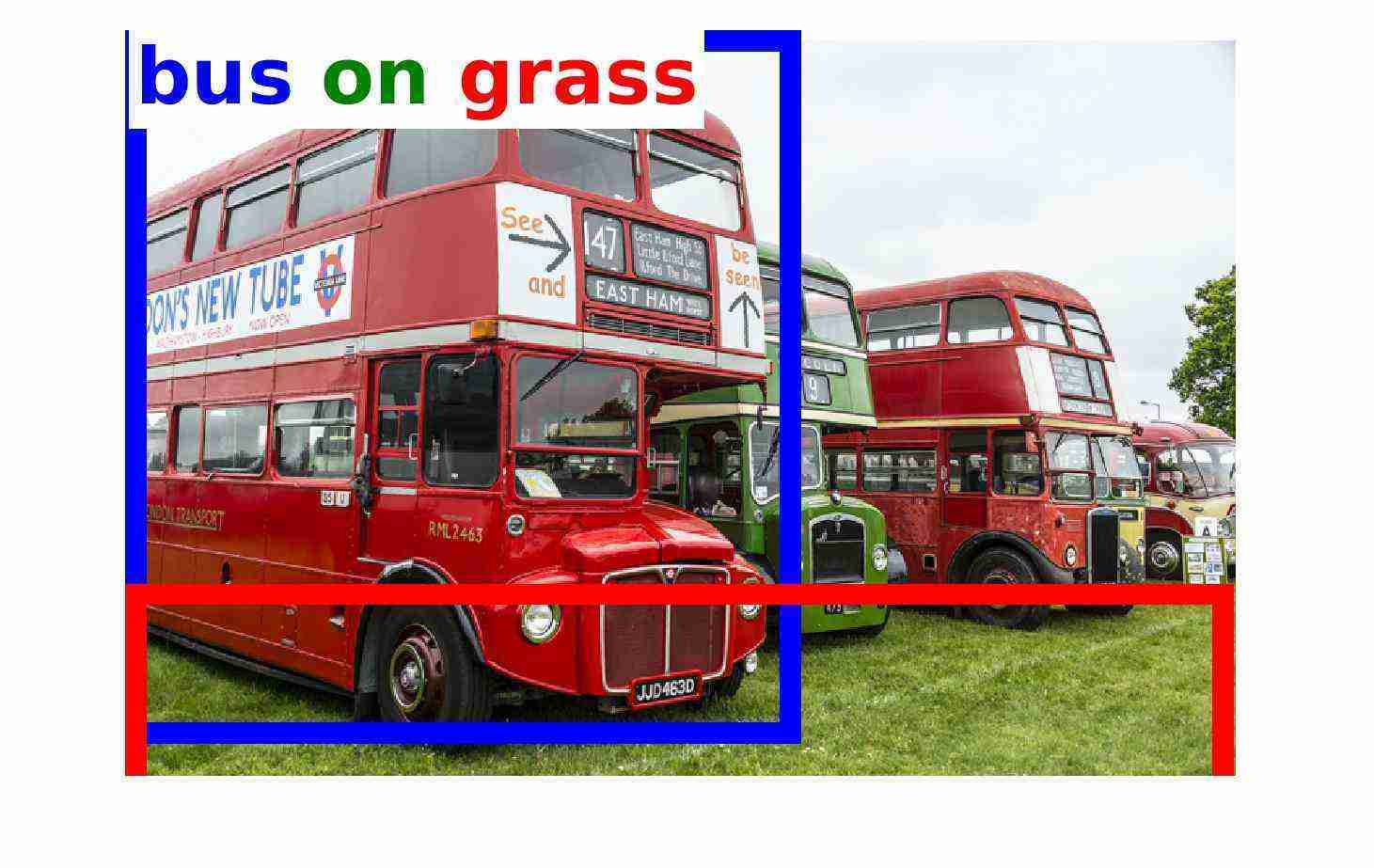}\\
       	\vspace{0.2ex}
       	\small{
       	GT: on\\
       	\cite{Lu16}: above
       	}
       	\vspace{0.3ex}
    \end{minipage}  
    \hspace{0.005\textwidth}
    \begin{minipage}[b]{0.18\textwidth}
    	\centering
       	\includegraphics[trim={0cm 3.5cm 0cm 1.5cm},clip,width=0.95\linewidth,cfbox={green 2pt 2pt}]{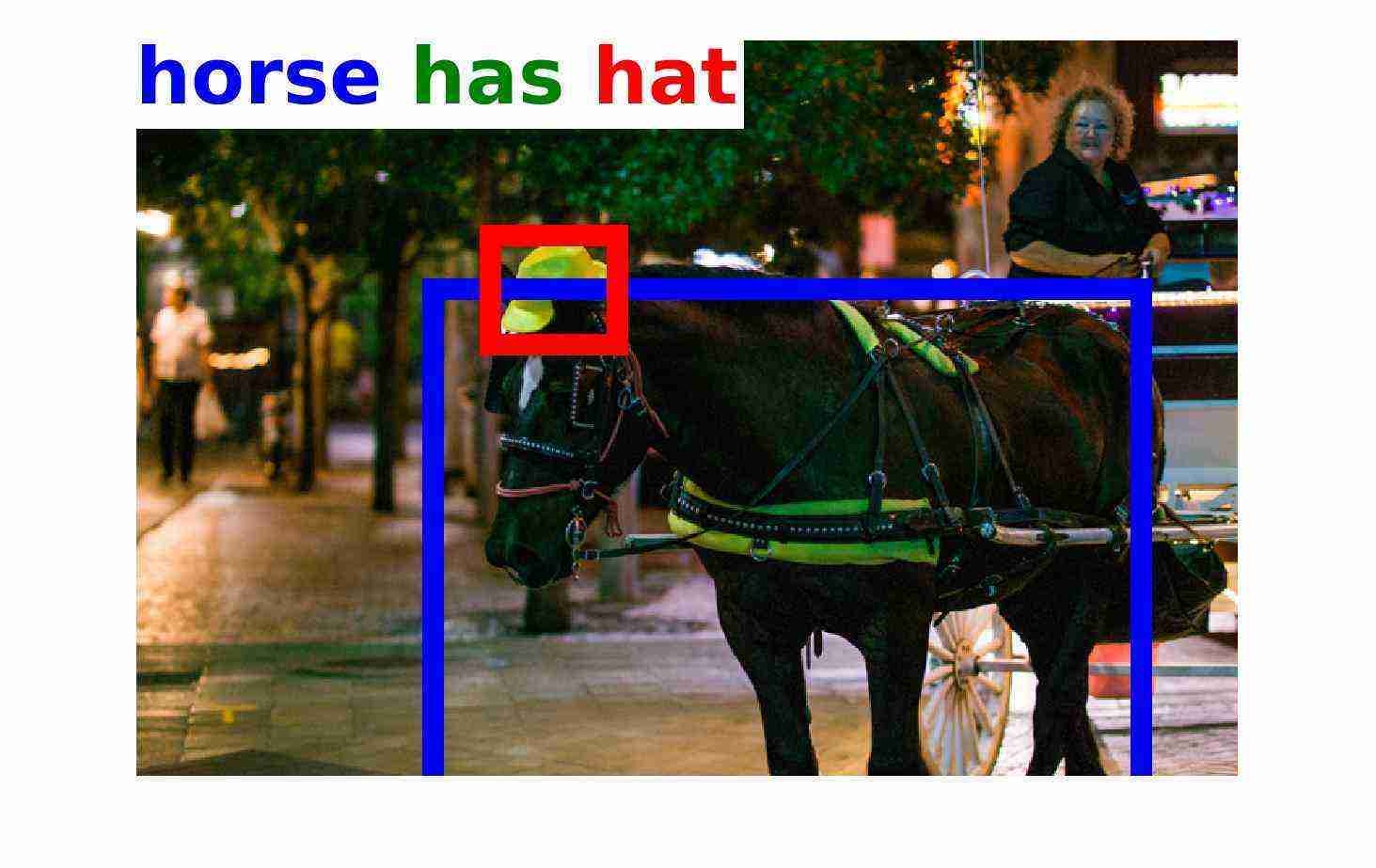}\\
       	\vspace{0.2ex}
       	\small{
       	GT: has, wear\\
       	\cite{Lu16}: wear
       	}
       	\vspace{0.3ex}
    \end{minipage} 
    \hspace{0.005\textwidth}
    \begin{minipage}[b]{0.18\textwidth}
       	\centering
    	\includegraphics[trim={2cm 7cm 2cm 2.5cm},clip,width=0.95\linewidth,cfbox={yellow 2pt 2pt}]{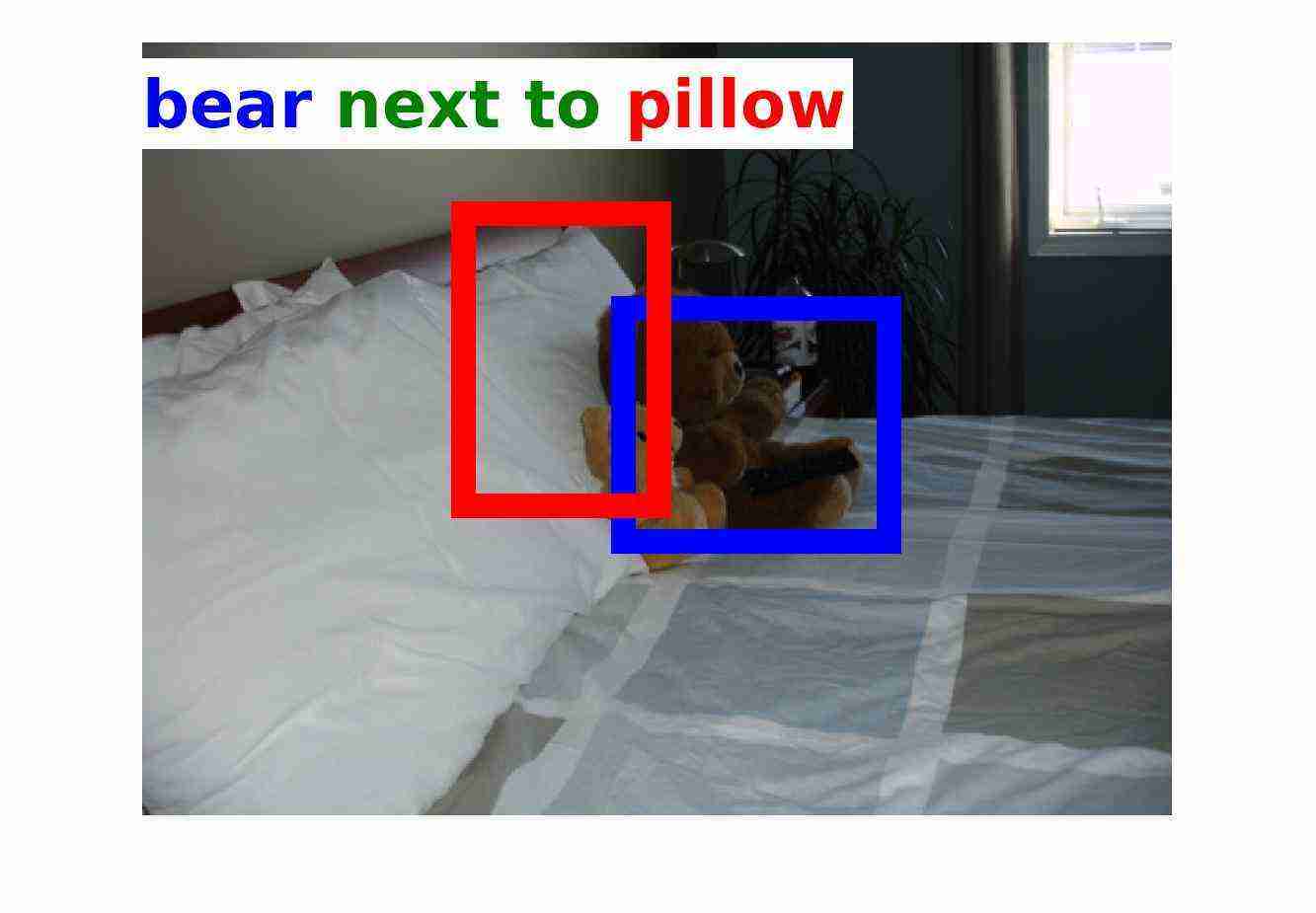}\\
    		\vspace{0.2ex}
    		\small{
       	GT: against\\
       	\cite{Lu16}: next to
       	}
       	\vspace{0.3ex}
    \end{minipage}
    \hspace{0.005\textwidth}
    \begin{minipage}[b]{0.18\textwidth}
    	\centering
       	\includegraphics[trim={0.5cm 4cm 0.5cm 1.5cm},clip,width=0.95\linewidth,cfbox={red 2pt 2pt}]{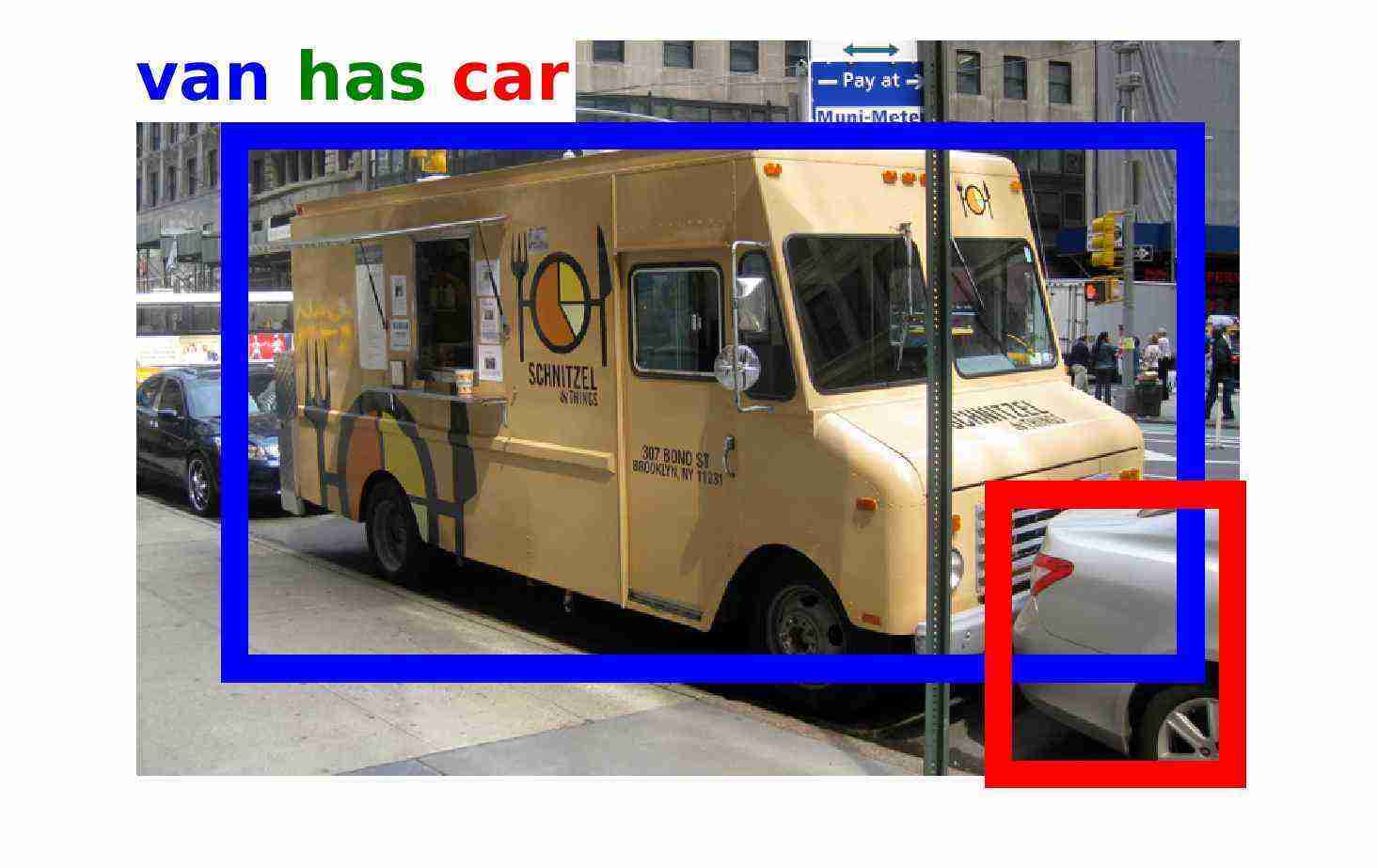}\\
       	\vspace{0.2ex}
       	\small{
       	GT: behind, left of\\
       	\cite{Lu16}: behind
       	}
       	\vspace{0.3ex}
    \end{minipage}   
     
    \setlength\abovecaptionskip{5pt}
    \caption{Relationship detections on the test set of \cite{Lu16}. We show examples among the top scored triplets detected for each relation by our weakly-supervised model described in \ref{model}. The triplet is correctly recognized if both the object detections and the relation match ground truth (in green), else the triplet is incorrect (in red). We also show examples of correctly predicted relations where the ground truth is erroneous : either missing or incomplete (in yellow). The last row shows zero-shot triplets that are not in the training set. See the appendix for additional qualitative results.}
    \vspace{-.4cm}
    \label{fig:relationship_detection_all}
\end{figure*}

\spaceparagraph
\paragraph{Unseen triplets.} Following~\cite{Lu16} we report results on the ``zero-shot split" of the test set containing only the test triplets not seen in training. Results for both of our fully-supervised and weakly-supervised methods are shown in Table \ref{tab:results_vrd} (column Unseen). Interestingly, our fully supervised model almost triples the performance on the unseen triplets compared to the Visual+Language model of \cite{Lu16}. Even using weak supervision, our recall of $19.0\%$ is significantly better than their fully supervised method. We believe that this improvement is due to the strength of our visual features that generalize well to unseen triplets. 

\gotoline
Figure \ref{fig:relationship_detection_all} shows examples of predictions  of both seen and unseen triplets (last row) by our model [S+A] trained with weak-supervision. We note that many of the misclassified relations are in fact due to missing annotations in the dataset (yellow column). First, not all pairs of objects in the image are labeled; second, the pairs that are labeled are not labelled exhaustively, i.e.\ ``person riding horse" can be labelled as ``person on horse" and predicting ``riding" for this pair of objects is considered as an error. Not having exhaustive annotation per object pair is therefore an issue as predicates are not necessary mutually exclusive. We tackle this problem in the next section by introducing a new exhaustively labeled dataset that enables retrieval evaluation. Our real errors (red column) are mostly due to two reasons: either the spatial configuration is challenging (e.g.``person on table"), or the spatial configuration is roughly correct  but the output predicate is incorrect (e.g. ``van has car" has similar configuration to ''person has bag").

\subsection{Retrieval of rare relations on UnRel Dataset}

\paragraph{Dataset.} To address the problem of missing annotations, we introduce a new challenging dataset of unusual relations, UnRel, that contains images collected from the web with unusual language triplet queries such as ``person ride giraffe". We exhaustively annotate these images at box-level for the given triplet queries. 
UnRel dataset has three main advantages. First, it is now possible to evaluate retrieval and localization of triplet queries in a clean setup without problems posed by missing annotations. Second, as the triplet queries of UnRel are rare (and thus likely not seen at training), it enables evaluating the generalization performance of the algorithm. Third, other datasets can be easily added to act as confusers to further increase the difficulty of the retrieval set-up. 
Currently, UnRel dataset contains more than 1000 images queried with 76 triplet queries.

\spaceparagraph
\paragraph{Setup.} 
We use our UnRel dataset as a set of positive pairs to be retrieved among all the test pairs of the Visual Relationship Dataset. We evaluate retrieval and localization with mean average precision (mAP) over triplet queries $t=(s,r,o)$ of UnRel in two different setups. In the first setup (with GT) we rank the manually provided ground truth pairs of boxes $(\bm{o}_s,\bm{o}_o)$ according to their predicate scores $v_{rel}((\bm{o}_s,\bm{o}_o) ~|~ r)$ to evaluate relation prediction without the difficulty of object detection. In the second setup (with candidates) we rank candidate pairs of boxes $(\bm{o}_s,\bm{o}_o)$ provided by the object detector according to predicate scores $v_{rel}((\bm{o}_s,\bm{o}_o) ~|~ r)$. For this second setup we also evaluate the accuracy of localization : a candidate pair of boxes is positive if its IoU with one ground truth pair is above 0.3. We compute different localization metrics : $mAP \textendash subj$ computes the overlap of the predicted subject box with the ground truth subject box, $mAP \textendash union$ computes the overlap of the predicted union of subject and object box with the union of ground truth boxes and $mAP \textendash subj/obj$ computes the overlap of both the subject and object boxes with their respective ground truth. Like in the previous section, we form candidate pairs of boxes by taking the top-scored object detections given by \cite{girshick15fastrcnn}. We keep at most 100 candidate objects per image, and retain at most 500 candidate pairs per image. For this retrieval task where it is important to discriminate the positive from negative pairs, we found it is important to learn an additional ``no relation" class by adding an extra column to $W$ in (\ref{DIFFRAC}). The negative pairs are sampled at random among the candidates that do not match the image-level annotations.

\begin{figure*}[t]
\centering
	\begin{minipage}[t]{0.005\textwidth}
    	\centering
    	\vspace{0.6ex}
    \end{minipage}
    \hspace{0.01\textwidth}
	\begin{minipage}[t]{0.18\textwidth}
    	\centering
    	\emph{{\color{blue}bike} above {\color{red}person}}\\
    	\vspace{0.2ex}
	\end{minipage}
	\hspace{0.005\textwidth}
	\begin{minipage}[t]{0.18\textwidth}
    	\centering
    \emph{{\color{blue}building} has {\color{red}wheel}}\\
    	\vspace{0.2ex}
	\end{minipage}
	\hspace{0.005\textwidth}
	\begin{minipage}[t]{0.18\textwidth}
    \centering
    	\emph{{\color{blue}cat} wear {\color{red}tie}}\\
    	\vspace{0.2ex}
	\end{minipage}
	\hspace{0.005\textwidth}
	\begin{minipage}[t]{0.18\textwidth}
    \centering
    	\emph{{\color{blue}person} above {\color{red}bed}}\\
    	\vspace{0.2ex}
	\end{minipage}
	\hspace{0.005\textwidth}
	\begin{minipage}[t]{0.18\textwidth}
    \centering
    	\emph{{\color{blue}cone} on top of {\color{red}person}}\\
    	\vspace{0.2ex}
	\end{minipage}	
	
	\begin{minipage}[b]{0.005\textwidth}
    	\centering
    	\begin{turn}{90}
    	top 1
    	\end{turn}	
    \vspace{3.5ex}
    \end{minipage}
    \hspace{0.01\textwidth}
    \begin{minipage}[t]{0.18\textwidth}
    	\centering
       	\includegraphics[trim={0cm 5cm 0cm 3cm},clip,width=0.95\linewidth,cfbox={green 2pt 2pt}]{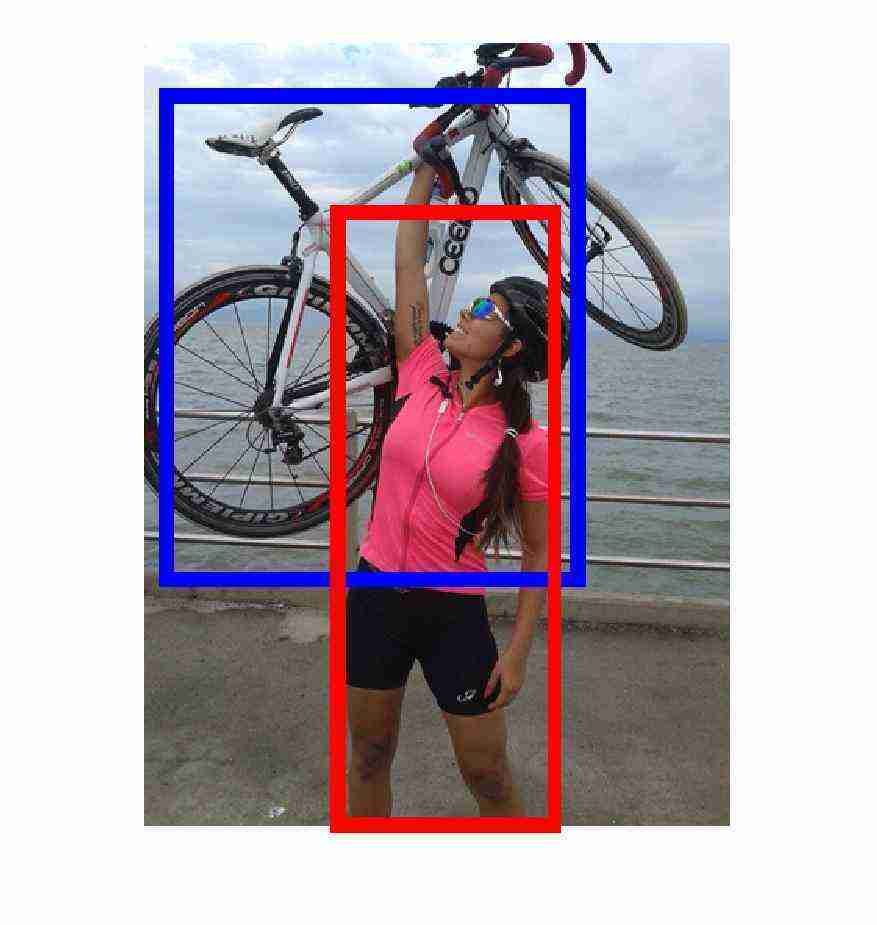}\\
       	\vspace{0.2ex}
    \end{minipage}
    \hspace{0.005\textwidth}
    \begin{minipage}[t]{0.18\textwidth}
       \centering
       \includegraphics[trim={8.5cm 0cm 7cm 0cm},clip,width=0.95\linewidth,cfbox={green 2pt 2pt}]{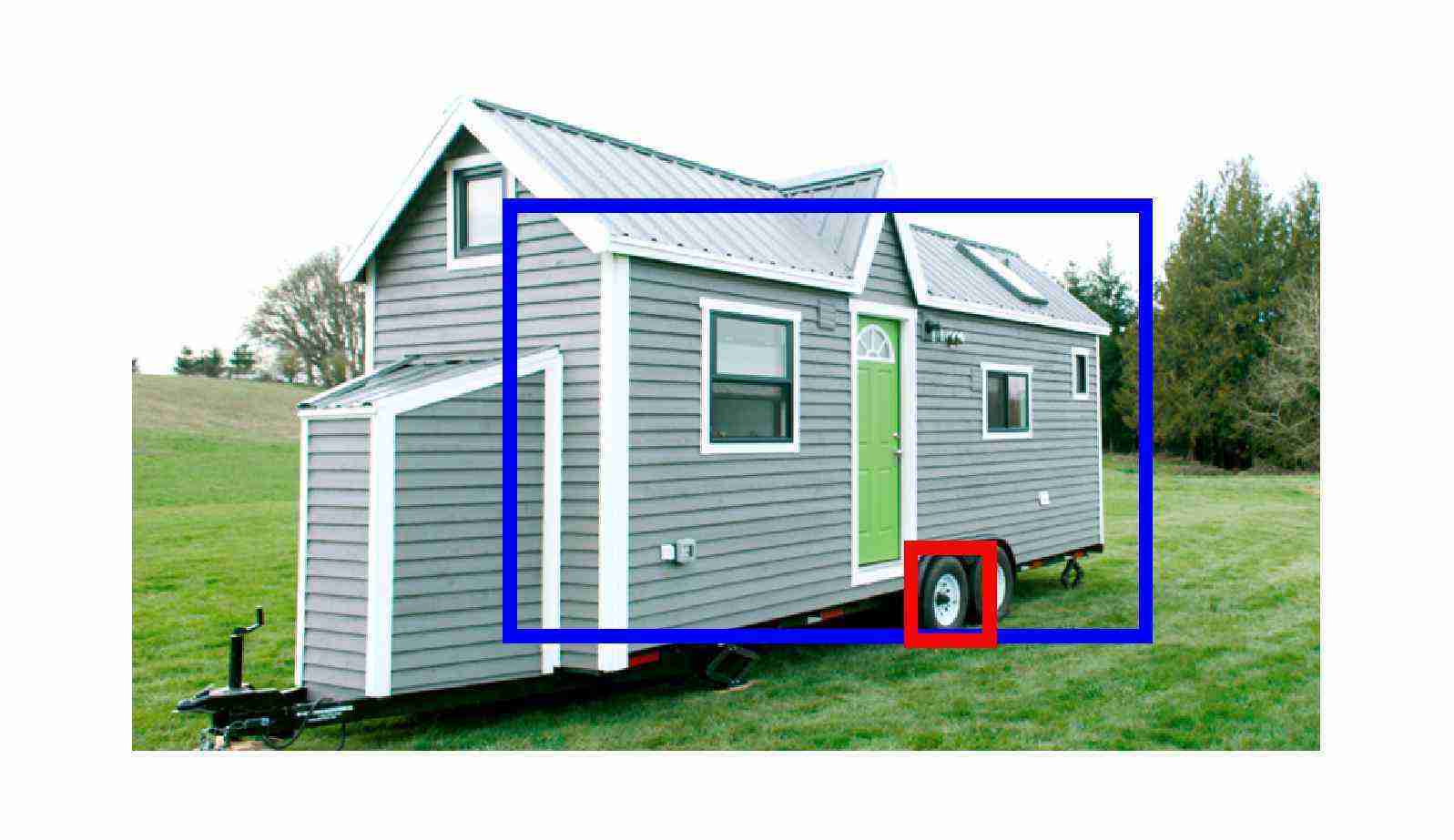}\\
       \vspace{0.2ex}
    \end{minipage}
    \hspace{0.005\textwidth}
    \begin{minipage}[t]{0.18\textwidth}
    	\centering
       	\includegraphics[trim={0cm 9cm 0cm 2cm},clip,width=0.95\linewidth,cfbox={green 2pt 2pt}]{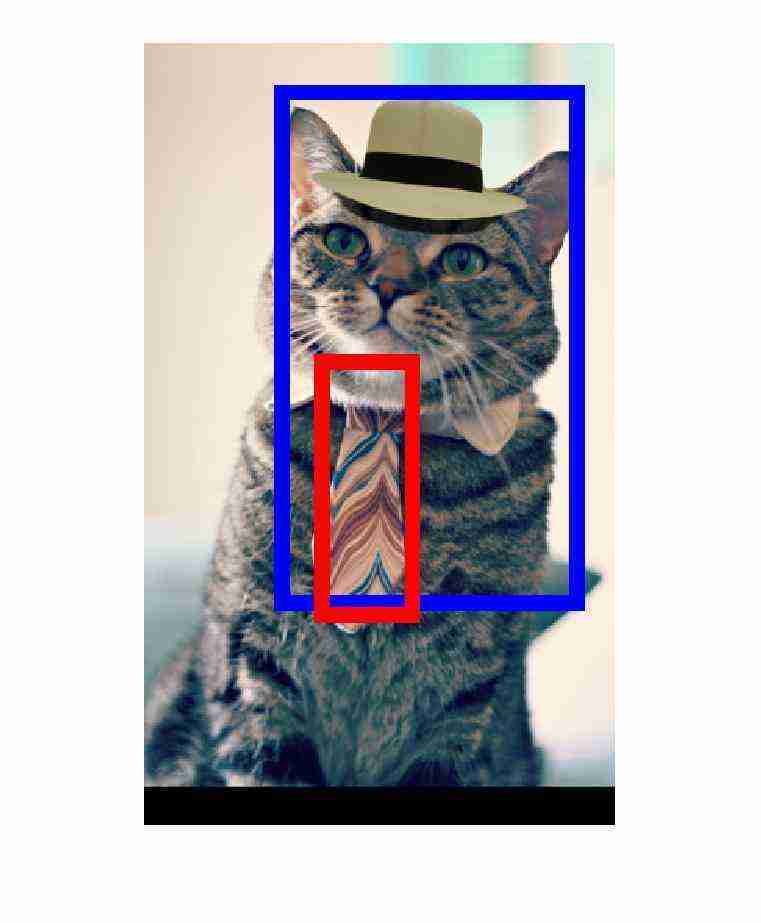}\\
       	\vspace{0.2ex}
    \end{minipage}
    \hspace{0.005\textwidth}
    \begin{minipage}[t]{0.18\textwidth}
    	\centering
       	\includegraphics[trim={4cm 2.3cm 4cm 0.5cm},clip,width=0.95\linewidth,cfbox={green 2pt 2pt}]{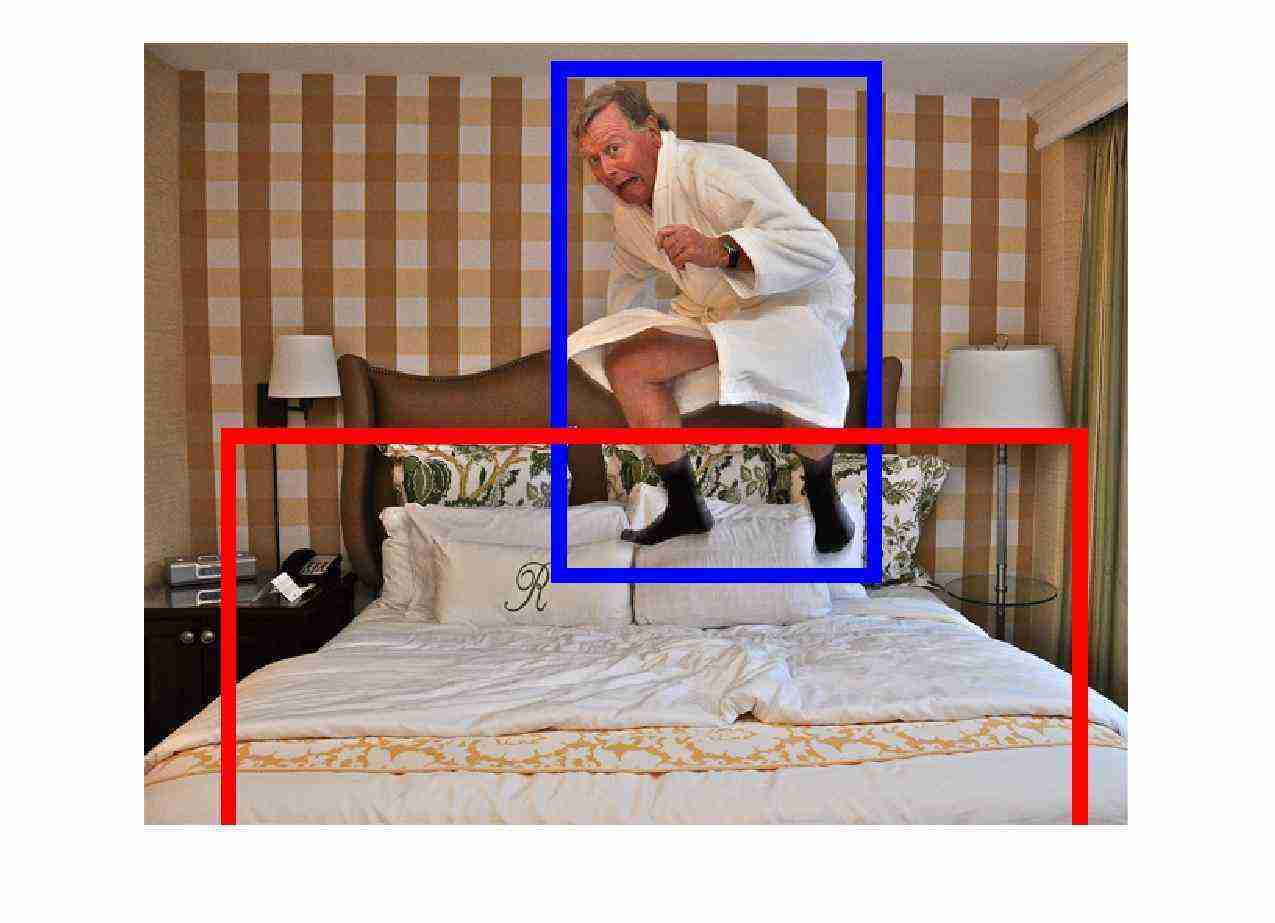}\\
       	\vspace{0.2ex}
    \end{minipage}
    \hspace{0.005\textwidth}  
    \begin{minipage}[t]{0.18\textwidth}
    	\centering
       	\includegraphics[trim={0cm 5cm 0cm 2.7cm},clip,width=0.95\linewidth,cfbox={green 2pt 2pt}]{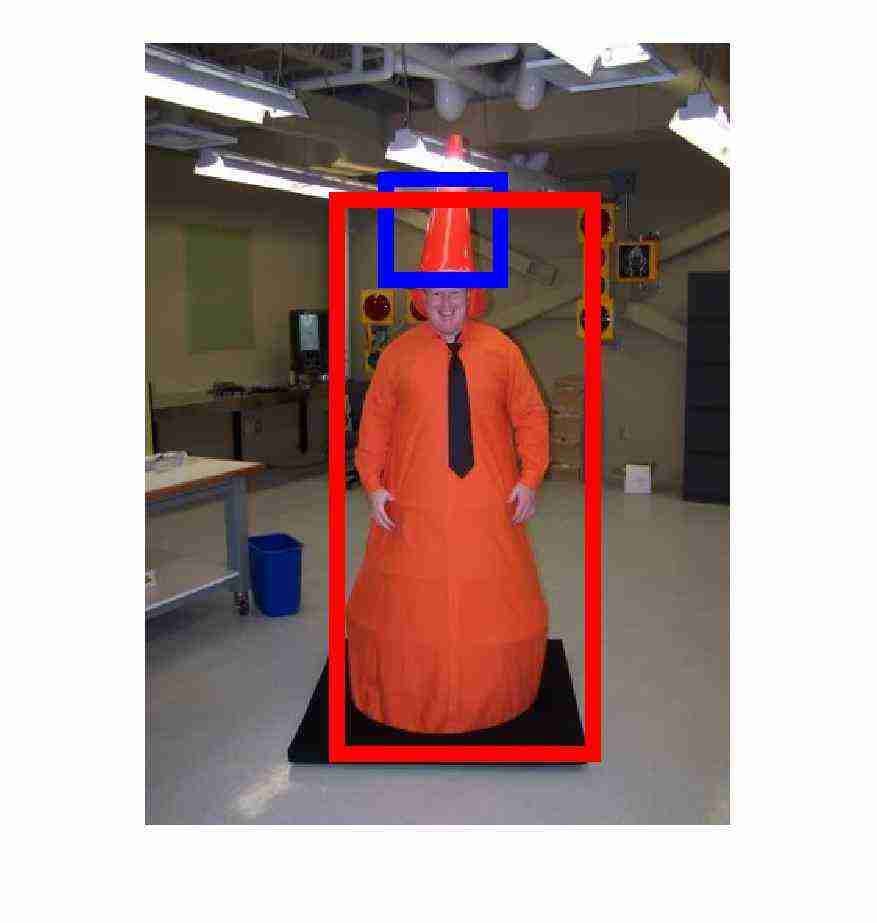}\\
      	\vspace{0.2ex}
    \end{minipage} 
    
	\begin{minipage}[b]{0.005\textwidth}
    	\centering
    	\begin{turn}{90}
    	top 2
    	\end{turn}
    	\vspace{3ex}
    \end{minipage}
    \hspace{0.01\textwidth}
    \begin{minipage}[t]{0.18\textwidth}
    	\centering
       	\includegraphics[trim={0cm 7cm 0cm 2.5cm},clip,width=0.95\linewidth,cfbox={green 2pt 2pt}]{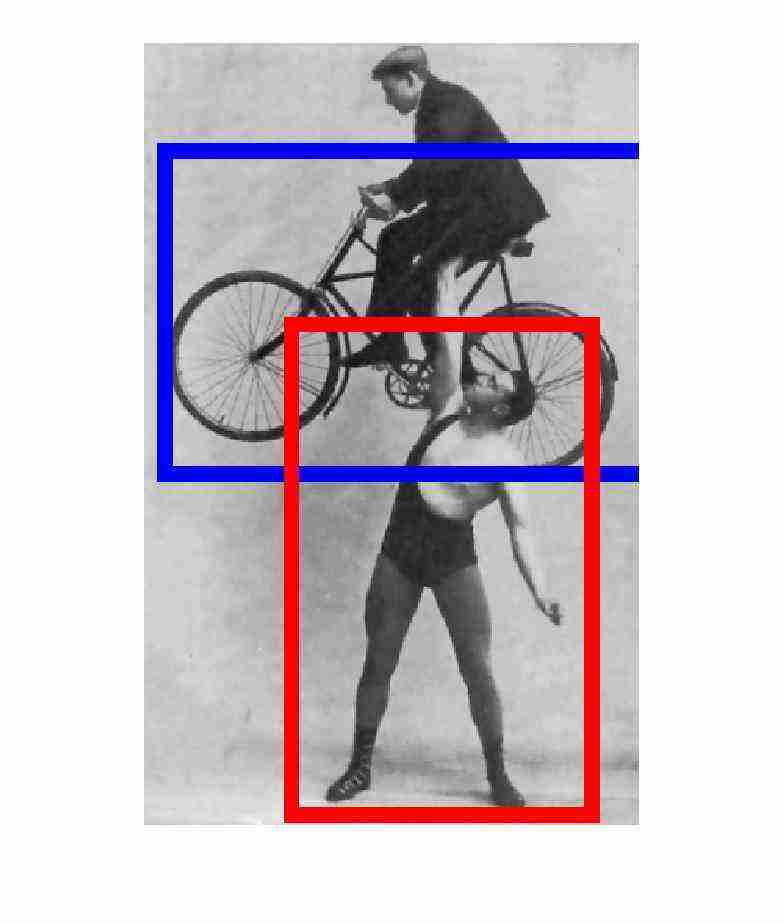}\\
       	\vspace{0.2ex}
    \end{minipage}
    \hspace{0.005\textwidth}
    \begin{minipage}[t]{0.18\textwidth}
       \centering
       \includegraphics[trim={9cm 3cm 5cm 2cm},clip,width=0.95\linewidth,cfbox={red 2pt 2pt}]{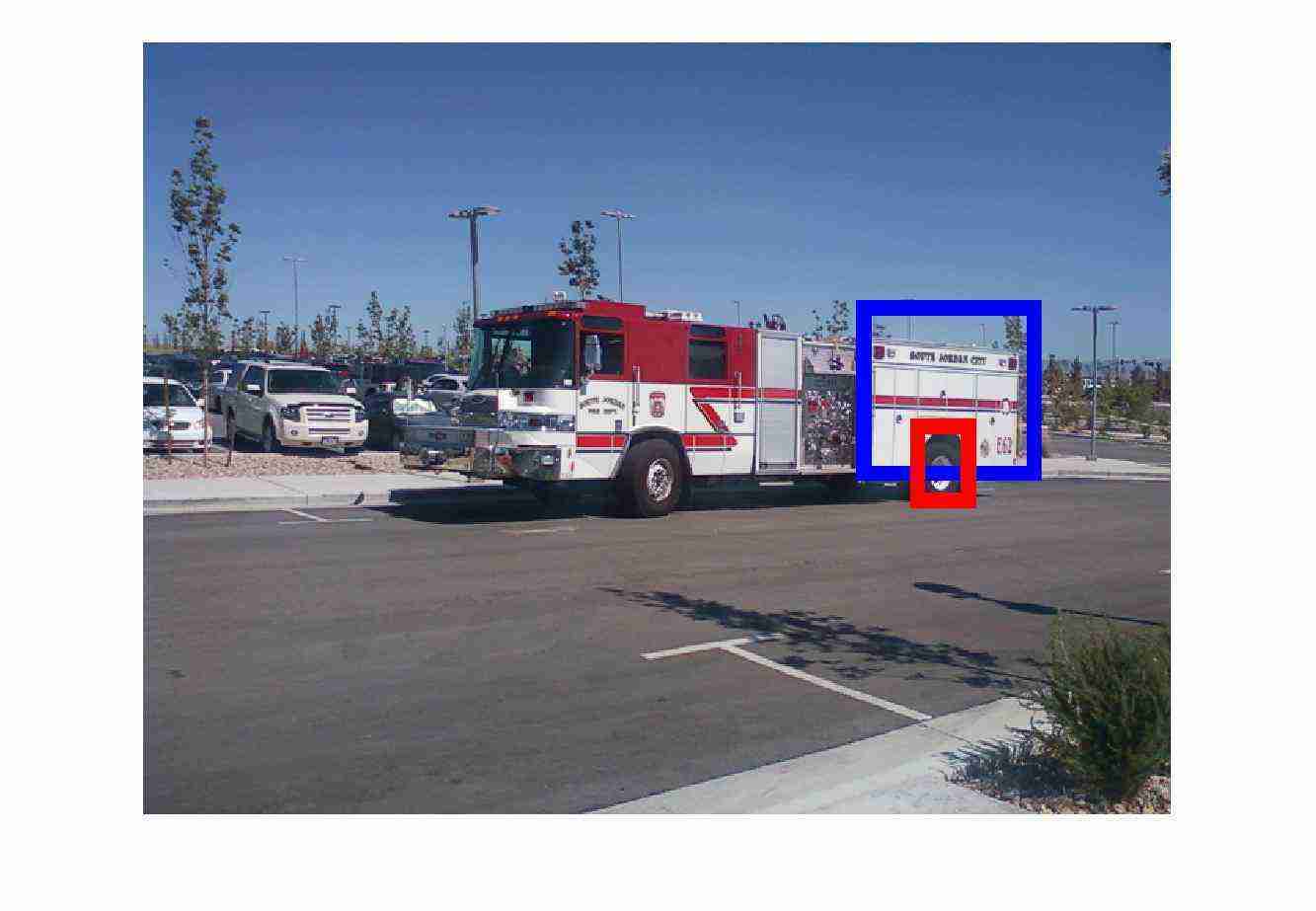}\\
       \vspace{0.2ex}
    \end{minipage}
    \hspace{0.005\textwidth}
    \begin{minipage}[t]{0.18\textwidth}
    	\centering
       	\includegraphics[trim={0cm 5.3cm 0cm 1.5cm},clip,width=0.95\linewidth,cfbox={green 2pt 2pt}]{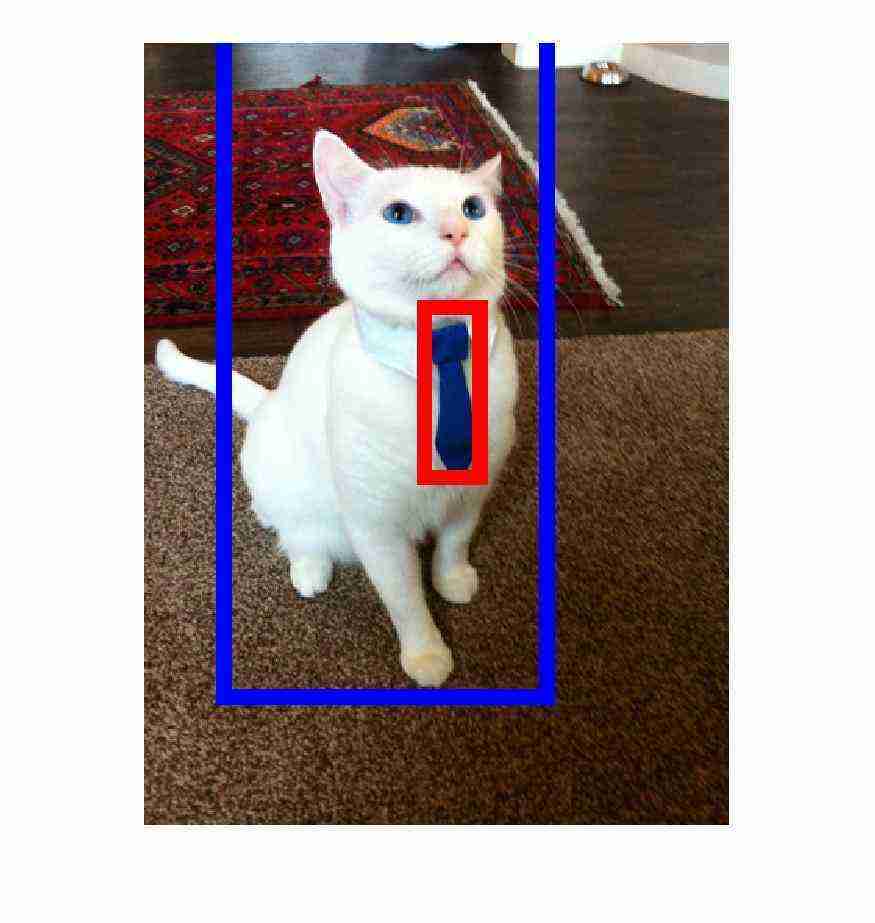}\\
       	\vspace{0.2ex}
    \end{minipage}
    \hspace{0.005\textwidth}
    \begin{minipage}[t]{0.18\textwidth}
    	\centering
       	\includegraphics[trim={7cm 0cm 5cm 0cm},clip,width=0.95\linewidth,cfbox={green 2pt 2pt}]{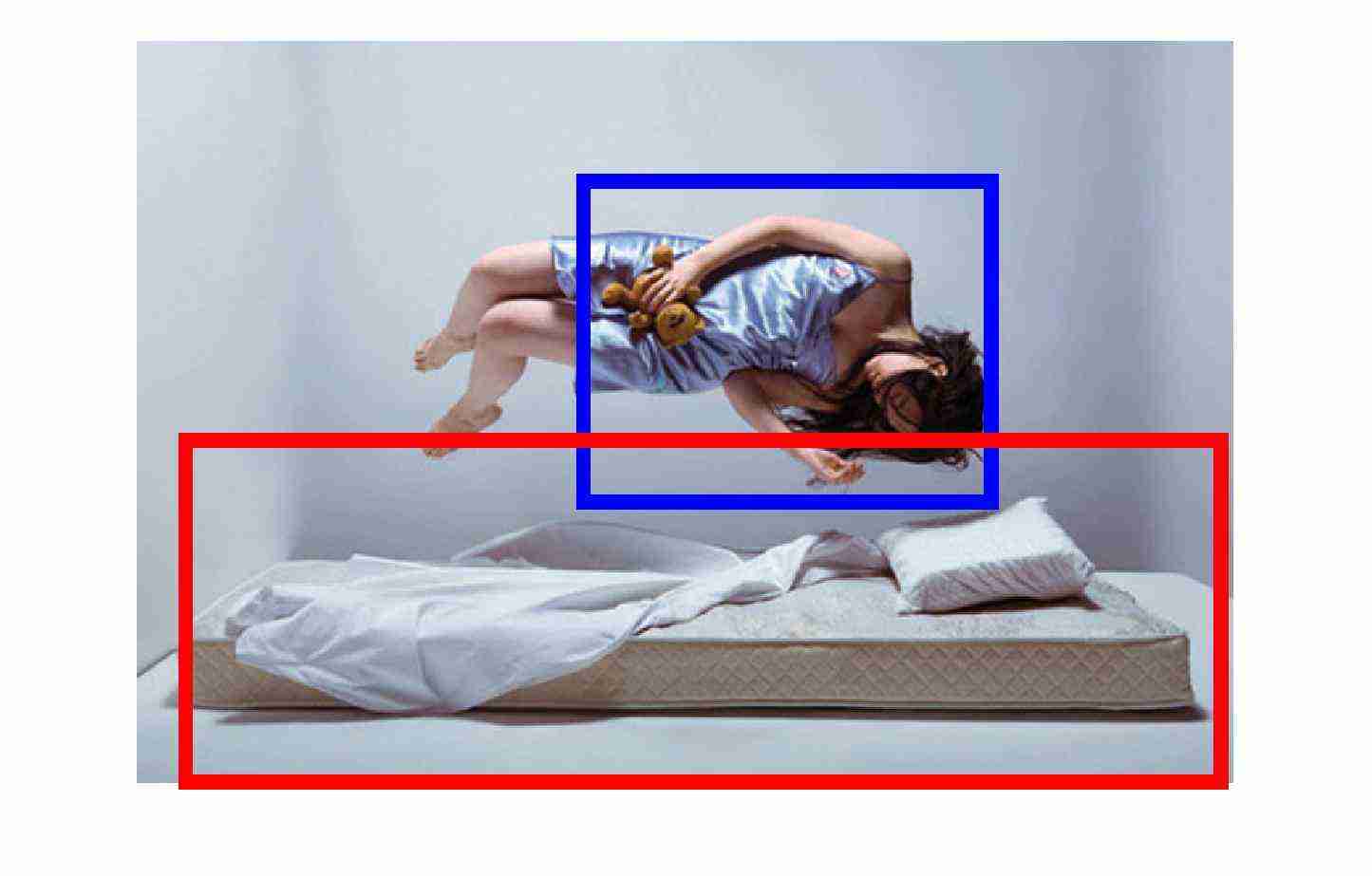}\\
       	\vspace{0.2ex}
    \end{minipage}
    \hspace{0.005\textwidth}  
    \begin{minipage}[t]{0.18\textwidth}
    	\centering
       	\includegraphics[trim={7cm 3cm 6cm 1cm},clip,width=0.95\linewidth,cfbox={green 2pt 2pt}]{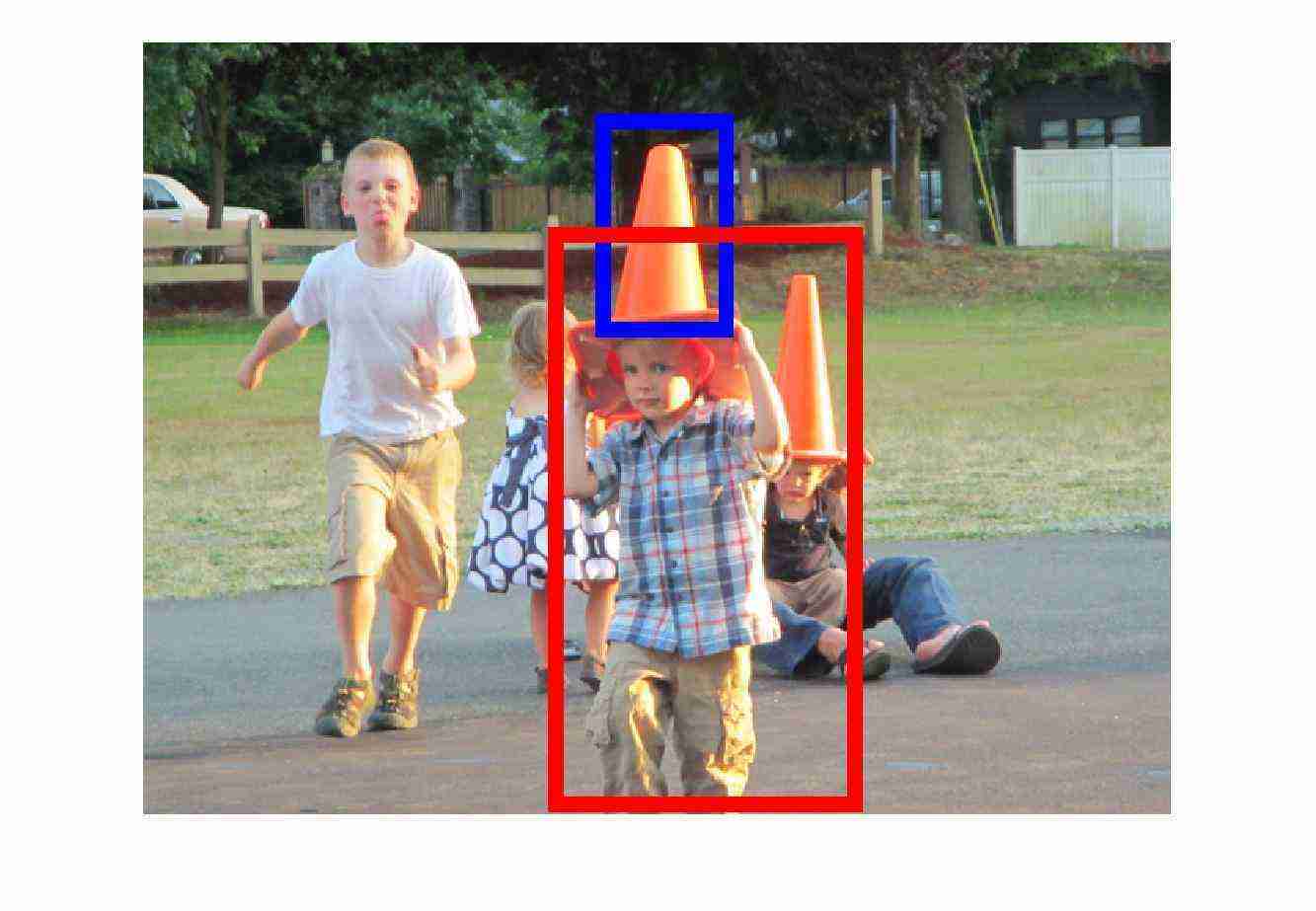}\\
      	\vspace{0.2ex}
    \end{minipage} 
    
	\begin{minipage}[b]{0.005\textwidth}
    	\centering
    	\begin{turn}{90}
    	top 3
    	\end{turn}
    \vspace{2.2ex}
    \end{minipage}
    \hspace{0.01\textwidth}
    \begin{minipage}[t]{0.18\textwidth}
    	\centering
       	\includegraphics[trim={0cm 5cm 0cm 2cm},clip,width=0.95\linewidth,cfbox={green 2pt 2pt}]{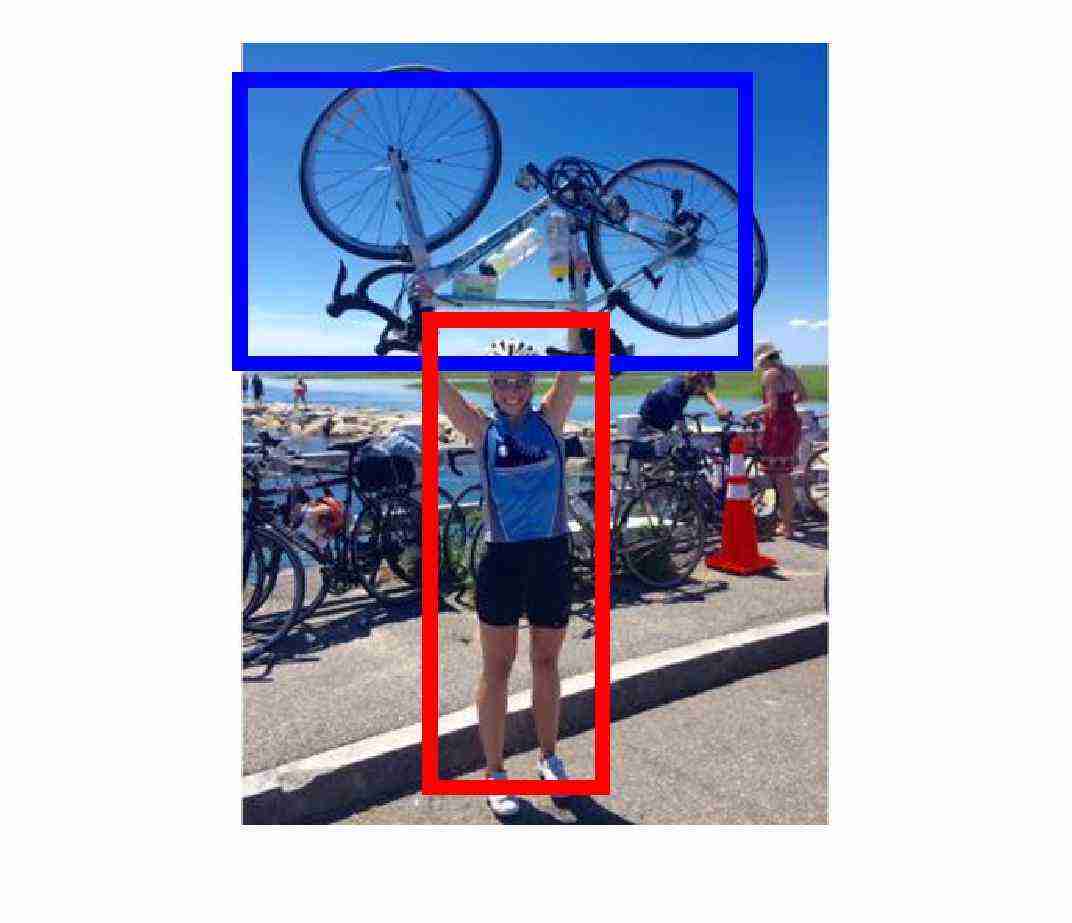}\\
       	\vspace{0.2ex}
    \end{minipage}
    \hspace{0.005\textwidth}
    \begin{minipage}[t]{0.18\textwidth}
       \centering
       \includegraphics[trim={4.5cm 4.2cm 4.5cm 2.5cm},clip,width=0.95\linewidth,cfbox={green 2pt 2pt}]{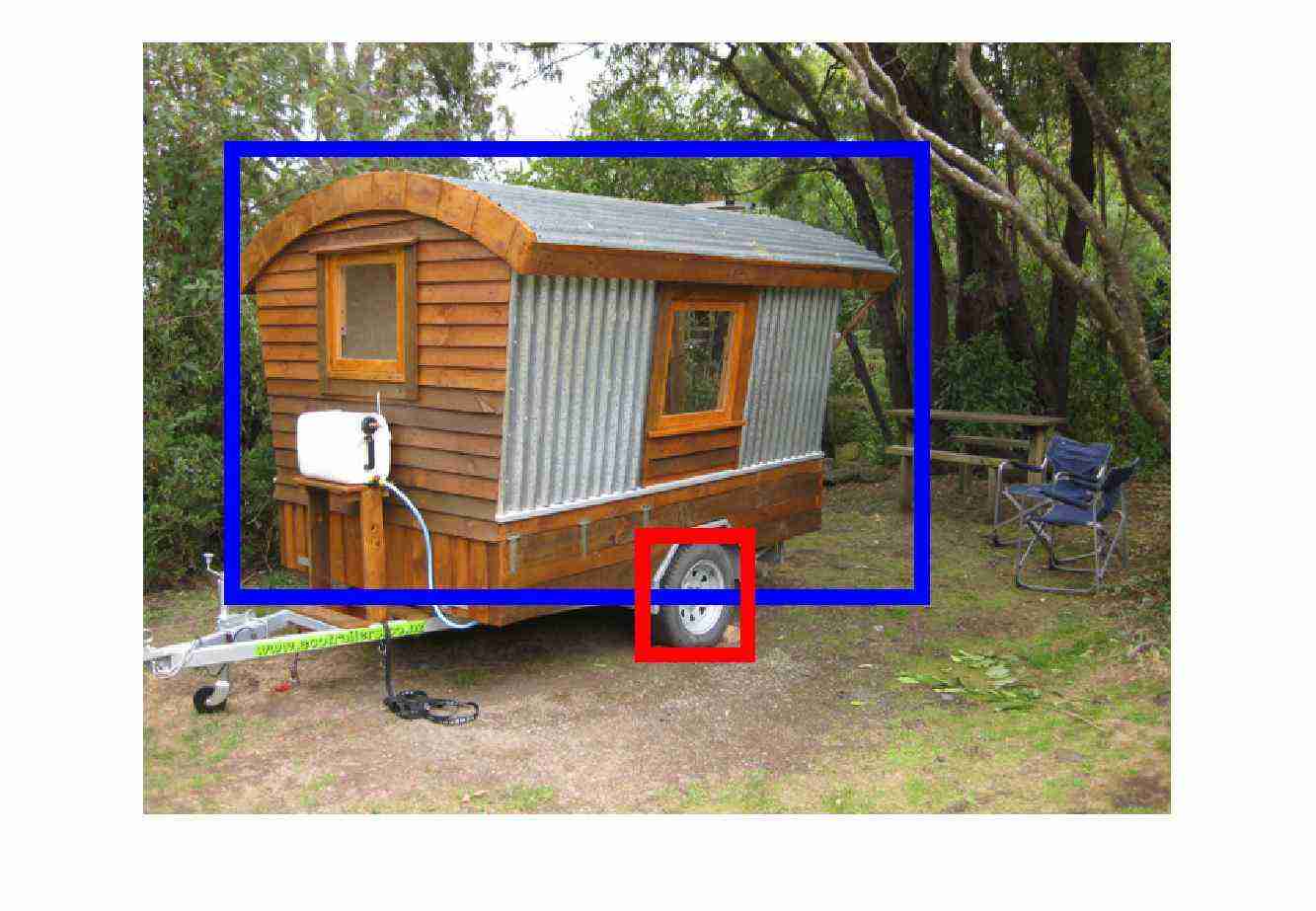}\\
       \vspace{0.2ex}
    \end{minipage}
    \hspace{0.005\textwidth}
    \begin{minipage}[t]{0.18\textwidth}
    	\centering
       	\includegraphics[trim={0cm 3.9cm 0cm 3cm},clip,width=0.95\linewidth,cfbox={green 2pt 2pt}]{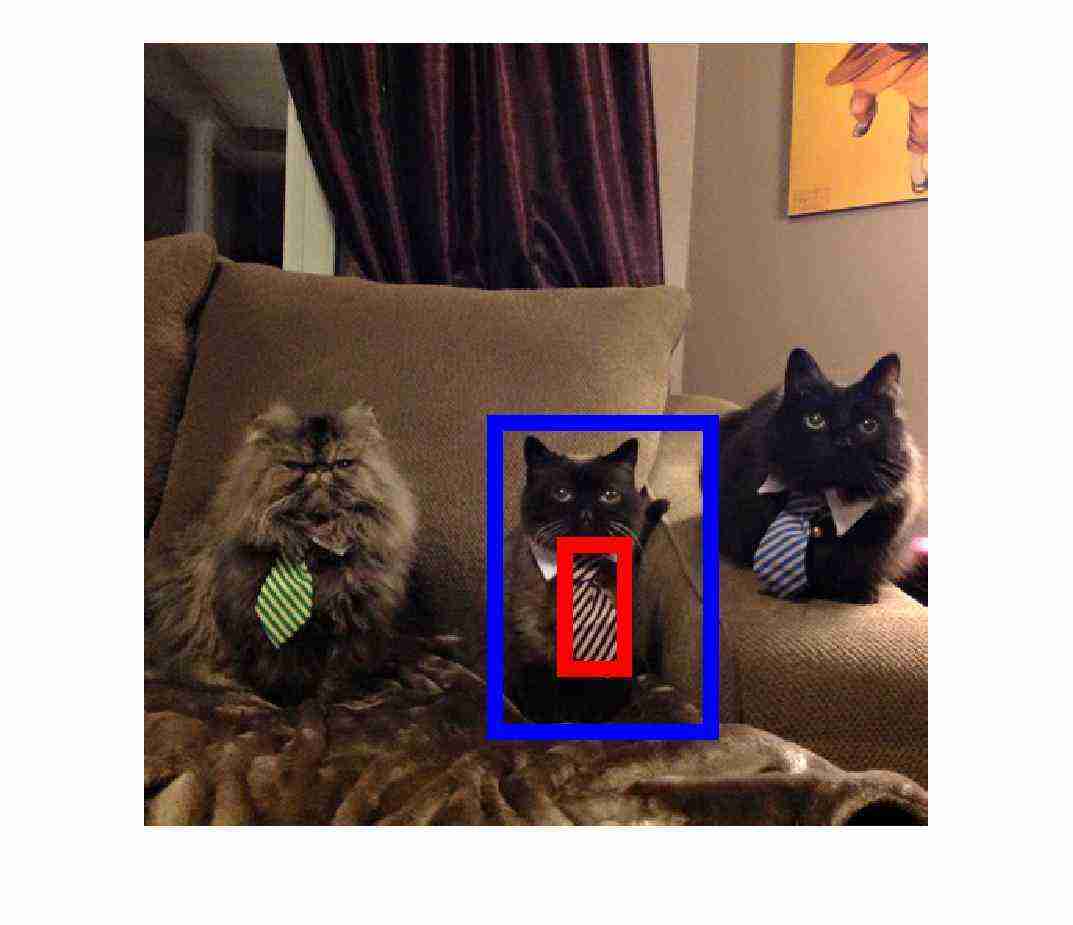}\\
       	\vspace{0.2ex}
    \end{minipage}
    \hspace{0.005\textwidth}
    \begin{minipage}[t]{0.18\textwidth}
    	\centering
       	\includegraphics[trim={5cm 2.5cm 5.3cm 2cm},clip,width=0.95\linewidth,cfbox={green 2pt 2pt}]{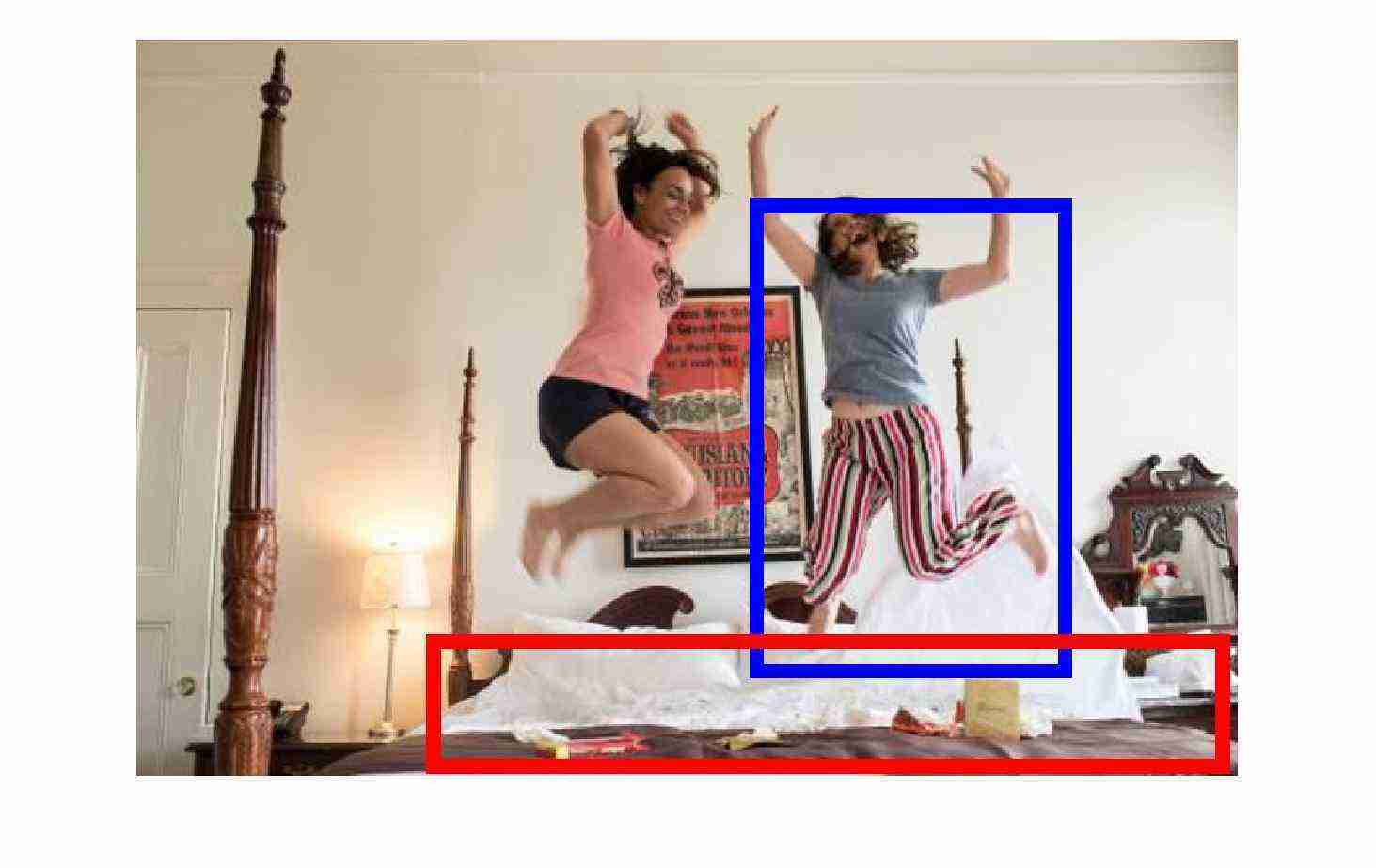}\\
       	\vspace{0.2ex}
    \end{minipage}
    \hspace{0.005\textwidth}  
    \begin{minipage}[t]{0.18\textwidth}
    	\centering
       	\includegraphics[trim={4cm 3cm 4cm 3cm},clip,width=0.95\linewidth,cfbox={red 2pt 2pt}]{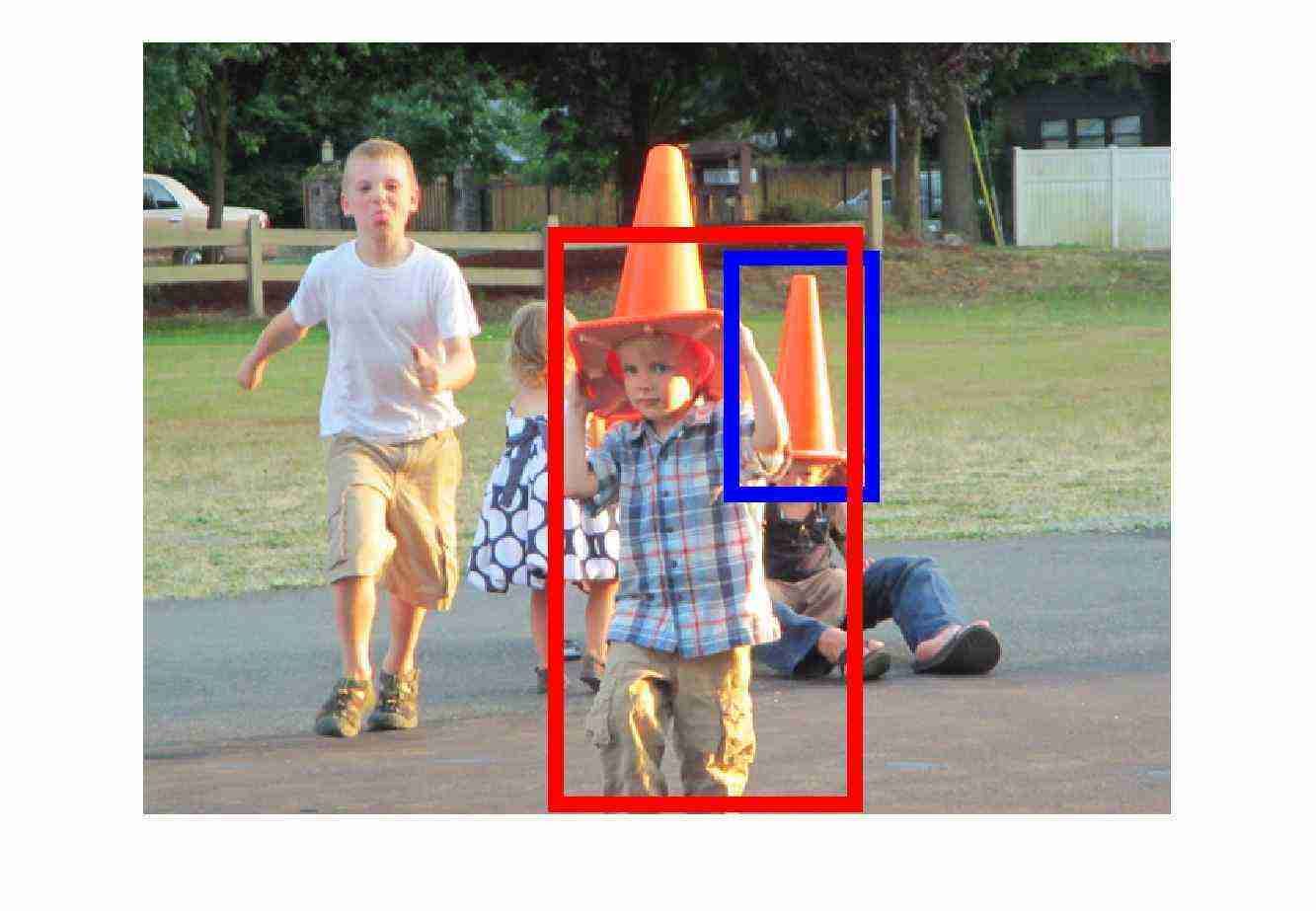}\\
      	\vspace{0.2ex}
    \end{minipage} 
    
    \setlength\abovecaptionskip{5pt}
    \caption{Top 3 retrieved pairs of boxes for a set of UnRel triplet queries (first line is best) with our weakly-supervised model. The pair is marked as positive (green) if the candidate subject and object boxes coincide with a ground truth subject and object boxes with $IoU \ge 0.3$. We provide more qualitative results in appendix.}
    \label{fig:unrel}
        \vspace{-.4cm}
\end{figure*}

\spaceparagraph
\paragraph{Results.}
Retrieval results are shown in Table~\ref{tab:results_rarerel}. Our classifiers are trained on the training subset of the Visual Relationship Dataset. 
We compare with two strong baselines. The first baseline is our implementation of~\cite{Lu16} (their trained models are not available online). For this, we trained a classifier~\cite{ren15} to output predicates given visual features extracted from the union of subject and object bounding boxes. We do not use the language model as its score does not affect the retrieval results (only adding a constant offset to all retrieved images). We verified our implementation on the Visual Relationship Dataset where results of~\cite{Lu16} are available. As the second baseline, we use the DenseCap \cite{Johnson2015} model to generate region candidates for each image and sort them according to the score of the given triplet query. Note that this is a strong baseline as we use the pre-trained model released by the authors which has been trained on 77K images of \cite{Krishna2016} in a fully supervised manner using localized language descriptions, compared to our model trained on only 4K training images of \cite{Lu16}. DenseCap outputs only a single bounding box (not a pair of boxes) but we interpret its output as either a subject box or a union of boxes. We cannot compare with the Visual Phrases~\cite{Sadeghi2011} approach as it requires training data for each triplet, which is not available for these rare queries. We report the chance as the performance given by random ordering of the proposals. Results in Table~\ref{tab:results_rarerel} show significant improvements of our method over the baselines. Note that our weakly-supervised method outperforms these strong baselines that are fully supervised. This confirms our results from the previous section that (i) our visual features are well suited to model relations, (ii) they generalize well to unseen triplets, and (iii)  training from weak image-level supervision is possible and results only in a small loss of accuracy compared to training from fully supervised data. Examples of retrieved unusual relations are shown in Figure \ref{fig:unrel}.

\begin{table}\centering
\small{
\ra{1}
\begin{tabular}{@{}rccccc@{}}\toprule
& \multicolumn{1}{c}{With GT} & \multicolumn{3}{c}{With candidates} \\
& - & union & subj & subj/obj
\\\midrule
\rule{0pt}{2ex}  
Chance 					& 38.4	& 8.6 	& 6.6 	& 4.2 \\
\rule{0pt}{2ex}  
\textbf{Full sup.}\\
\rule{0pt}{1ex} 
DenseCap \cite{Johnson2015} 		& - 		& 6.2 	& 6.8 	& - \\
Reproduce \cite{Lu16} 			& 50.6 	& 12.0 	& 10.0 	& 7.2  \\
Ours [S+A] 				&  \textbf{62.6} & \textbf{14.1} & \textbf{12.1} & \textbf{9.9} \\
\rule{0pt}{3ex}  
\textbf{Weak sup.}\\
\rule{0pt}{1ex} 
Ours [S+A] 					&  58.5 		& 13.4 	& 11.0 	& 8.7 \\
Ours [S+A] - Noisy			&  55.0 		& 13.0 	& 10.6	& 8.5 \\
\bottomrule
\end{tabular}
\setlength\abovecaptionskip{5pt}
\caption{Retrieval on UnRel (mAP) with IoU=0.3} 
\label{tab:results_rarerel}
}
\end{table}

\section{Conclusion}
We have developed a new powerful visual descriptor for representing object relations in images achieving state-of-the-art performance on the Visual Relationship Detection dataset \cite{Lu16}, and in particular significantly improving the current results on unseen object relations. We have also developed a weakly-supervised model for learning object relations and have demonstrated that, given pre-trained object detectors, object relations can be learnt from weak image-level annotations without a significant loss of recognition performance. Finally, we introduced, UnRel, a new evaluation dataset for visual relation detection that enables to evaluate retrieval without missing annotations and assess generalization to unseen triplets. Our work opens-up the possibility of learning a large vocabulary of visual relations directly from large-scale Internet collections annotated with image-level natural language captions.    

\paragraph{Acknowledgements.}
\small{This work was partly supported by ERC grants Activia (no. 307574),
LEAP (no. 336845) and Allegro (no. 320559), CIFAR Learning in Machines \& Brains program and ESIF, OP Research, development and education Project IMPACT
No.\ CZ$.02.1.01/0.0/0.0/15\_003/0000468$.}

{\small
\bibliographystyle{ieee}
\bibliography{ICCV17}

\begin{thebibliography}{10}\itemsep=-1pt

\bibitem{Andreas2016}
J.~Andreas, M.~Rohrbach, T.~Darrell, and D.~Klein.
\newblock Neural module networks.
\newblock In {\em CVPR}, 2016.

\bibitem{bach2008diffrac}
F.~R. Bach and Z.~Harchaoui.
\newblock Diffrac: a discriminative and flexible framework for clustering.
\newblock In {\em NIPS}, 2008.

\bibitem{Bilen16}
H.~Bilen and A.~Vedaldi.
\newblock Weakly supervised deep detection networks.
\newblock In {\em CVPR}, 2016.

\bibitem{Bojanowski2014}
P.~Bojanowski, R.~Lajugie, F.~Bach, I.~Laptev, J.~Ponce, C.~Schmid, and
  J.~Sivic.
\newblock Weakly supervised action labeling in videos under ordering
  constraints.
\newblock In {\em ECCV}, 2014.

\bibitem{chang2015text}
A.~Chang, W.~Monroe, M.~Savva, C.~Potts, and C.~D. Manning.
\newblock Text to 3d scene generation with rich lexical grounding.
\newblock {\em ACL}, 2015.

\bibitem{chen2013neil}
X.~Chen, A.~Shrivastava, and A.~Gupta.
\newblock Neil: Extracting visual knowledge from web data.
\newblock In {\em ICCV}, 2013.

\bibitem{Choi2012}
M.~J. Choi, A.~Torralba, and A.~S. Willsky.
\newblock Context models and out-of-context objects.
\newblock {\em Pattern Recognition Letters}, 2012.

\bibitem{Delaitre11}
V.~Delaitre, J.~Sivic, and I.~Laptev.
\newblock Learning person-object interactions for action recognition in still
  images.
\newblock In {\em NIPS}, 2011.

\bibitem{Desai2010}
C.~Desai, D.~Ramanan, and C.~Fowlkes.
\newblock Discriminative models for static human-object interactions.
\newblock In {\em CVPR Workshops}, 2010.

\bibitem{elhoseiny2015sherlock}
M.~Elhoseiny, S.~Cohen, W.~Chang, B.~Price, and A.~Elgammal.
\newblock Sherlock: Scalable fact learning in images.
\newblock {\em AAAI}, 2016.

\bibitem{fangCVPR15}
H.~Fang, S.~Gupta, F.~N. Iandola, R.~Srivastava, L.~Deng, P.~Doll{\'{a}}r,
  J.~Gao, X.~He, M.~Mitchell, J.~C. Platt, C.~L. Zitnick, and G.~Zweig.
\newblock From captions to visual concepts and back.
\newblock In {\em CVPR}, 2015.

\bibitem{Frome2013}
A.~Frome, G.~S. Corrado, J.~Shlens, S.~Bengio, J.~Dean, M.~A. Ranzato, and
  T.~Mikolov.
\newblock Devise: A deep visual-semantic embedding model.
\newblock In {\em NIPS}. 2013.

\bibitem{galleguillos2008object}
C.~Galleguillos, A.~Rabinovich, and S.~Belongie.
\newblock Object categorization using co-occurrence, location and appearance.
\newblock In {\em CVPR}, 2008.

\bibitem{girshick15fastrcnn}
R.~Girshick.
\newblock Fast {R-CNN}.
\newblock In {\em ICCV}, 2015.

\bibitem{girshick2014rcnn}
R.~Girshick, J.~Donahue, T.~Darrell, and J.~Malik.
\newblock Rich feature hierarchies for accurate object detection and semantic
  segmentation.
\newblock In {\em CVPR}, 2014.

\bibitem{Gupta08}
A.~Gupta and L.~S. Davis.
\newblock Beyond nouns: Exploiting prepositions and comparative adjectives for
  learning visual classifiers.
\newblock In {\em ECCV}, 2008.

\bibitem{Gupta2009}
A.~Gupta, A.~Kembhavi, and L.~S. Davis.
\newblock Observing human-object interactions: Using spatial and functional
  compatibility for recognition.
\newblock {\em PAMI}, 2009.

\bibitem{Hendricks2015}
L.~A. Hendricks, S.~Venugopalan, M.~Rohrbach, R.~Mooney, K.~Saenko, and
  T.~Darrell.
\newblock Deep compositional captioning: Describing novel object categories
  without paired training data.
\newblock In {\em CVPR}, 2016.

\bibitem{Hu2015}
R.~Hu, H.~Xu, M.~Rohrbach, J.~Feng, K.~Saenko, and T.~Darrell.
\newblock Natural language object retrieval.
\newblock {\em CVPR}, 2016.

\bibitem{jenatton2012}
R.~Jenatton, N.~L. Roux, A.~Bordes, and G.~R. Obozinski.
\newblock A latent factor model for highly multi-relational data.
\newblock In {\em NIPS}, 2012.

\bibitem{Johnson2015}
J.~Johnson, A.~Karpathy, and L.~Fei-Fei.
\newblock Densecap: Fully convolutional localization networks for dense
  captioning.
\newblock In {\em CVPR}, 2016.

\bibitem{Johnson15a}
J.~Johnson, R.~Krishna, M.~Stark, L.-J. Li, D.~A. Shamma, M.~S. Bernstein, and
  L.~Fei-Fei.
\newblock Image retrieval using scene graphs.
\newblock In {\em CVPR}, 2015.

\bibitem{joulin2014}
A.~Joulin, K.~Tang, and L.~Fei-Fei.
\newblock Efficient image and video co-localization with frank-wolfe algorithm.
\newblock In {\em ECCV}, 2014.

\bibitem{Karpathy2014}
A.~Karpathy and L.~Fei-Fei.
\newblock Deep visual-semantic alignments for generating image descriptions.
\newblock In {\em CVPR}, 2015.

\bibitem{Karpathy2014a}
A.~Karpathy, A.~Joulin, and L.~Fei-Fei.
\newblock Deep fragment embeddings for bidirectional image sentence mapping.
\newblock In {\em NIPS}, 2014.

\bibitem{Kazemzadeh2014}
S.~Kazemzadeh, V.~Ordonez, M.~Matten, and T.~L. Berg.
\newblock Referitgame: Referring to objects in photographs of natural scenes.
\newblock In {\em EMNLP}, 2014.

\bibitem{Krishna2016}
R.~Krishna, Y.~Zhu, O.~Groth, J.~Johnson, K.~Hata, J.~Kravitz, S.~Chen,
  Y.~Kalantidis, L.-J. Li, D.~A. Shamma, M.~Bernstein, and L.~Fei-Fei.
\newblock Visual genome: Connecting language and vision using crowdsourced
  dense image annotations.
\newblock In {\em IJCV}, 2016.

\bibitem{Lazaridou2014a}
A.~Lazaridou, E.~Bruni, and M.~Baroni.
\newblock Is this a wampimuk? cross-modal mapping between distributional
  semantics and the visual world.
\newblock In {\em ACL}, 2014.

\bibitem{li2012automatic}
C.~Li, D.~Parikh, and T.~Chen.
\newblock Automatic discovery of groups of objects for scene understanding.
\newblock In {\em CVPR}, 2012.

\bibitem{Lin2014a}
T.-Y. Lin, M.~Maire, S.~Belongie, J.~Hays, P.~Perona, D.~Ramanan,
  P.~Doll{\'a}r, and C.~L. Zitnick.
\newblock Microsoft {COCO}: Common objects in context.
\newblock In {\em ECCV}, 2014.

\bibitem{Lu16}
C.~Lu, R.~Krishna, M.~Bernstein, and L.~Fei-Fei.
\newblock Visual relationship detection with language priors.
\newblock In {\em ECCV}, 2016.

\bibitem{Mao2016}
J.~Mao, J.~Huang, A.~Toshev, O.~Camburu, A.~Yuille, and K.~Murphy.
\newblock Generation and comprehension of unambiguous object descriptions.
\newblock {\em CVPR}, 2016.

\bibitem{miech17}
A.~Miech, J.-B. Alayrac, P.~Bojanowski, I.~Laptev, and J.~Sivic.
\newblock Learning from video and text via large-scale discriminative
  clustering.
\newblock {\em ICCV}, 2017.

\bibitem{Movshovitz-attias}
D.~Movshovitz-Attias and W.~W. Cohen.
\newblock Kb-lda: Jointly learning a knowledge base of hierarchy, relations,
  and facts.
\newblock In {\em ACL}, 2015.

\bibitem{Oquab15}
M.~Oquab, L.~Bottou, I.~Laptev, and J.~Sivic.
\newblock Is object localization for free? weakly-supervised learning with
  convolutional neural networks.
\newblock In {\em CVPR}, 2015.

\bibitem{Osokin16}
A.~Osokin, J.-B. Alayrac, I.~Lukasewitz, P.~K. Dokania, and S.~Lacoste-Julien.
\newblock Minding the gaps for block {F}rank-{W}olfe optimization of structured
  {SVM}s.
\newblock In {\em ICML}, 2016.

\bibitem{Plummer2015}
B.~A. Plummer, L.~Wang, C.~M. Cervantes, J.~C. Caicedo, J.~Hockenmaier, and
  S.~Lazebnik.
\newblock Flickr30k entities: Collecting region-to-phrase correspondences for
  richer image-to-sentence models.
\newblock In {\em ICCV}, 2015.

\bibitem{Prest12}
A.~Prest, C.~Schmid, and V.~Ferrari.
\newblock Weakly supervised learning of interactions between humans and
  objects.
\newblock {\em PAMI}, 2011.

\bibitem{ramanathan15}
V.~Ramanathan, C.~Li, J.~Deng, W.~Han, Z.~Li, K.~Gu, Y.~Song, S.~Bengio,
  C.~Rossenberg, and L.~Fei-Fei.
\newblock Learning semantic relationships for better action retrieval in
  images.
\newblock In {\em CVPR}, 2015.

\bibitem{ren15}
S.~Ren, K.~He, R.~Girshick, and J.~Sun.
\newblock Faster {R-CNN}: Towards real-time object detection with region
  proposal networks.
\newblock In {\em NIPS}. 2015.

\bibitem{Rohrbach2015}
A.~Rohrbach, M.~Rohrbach, R.~Hu, T.~Darrell, and B.~Schiele.
\newblock Grounding of textual phrases in images by reconstruction.
\newblock {\em ECCV}, 2016.

\bibitem{sadeghi2015viske}
F.~Sadeghi, S.~K. Divvala, and A.~Farhadi.
\newblock Viske: Visual knowledge extraction and question answering by visual
  verification of relation phrases.
\newblock In {\em CVPR}, 2015.

\bibitem{Sadeghi2011}
M.~A. Sadeghi and A.~Farhadi.
\newblock Recognition using visual phrases.
\newblock In {\em CVPR}, 2011.

\bibitem{Socher2013b}
R.~Socher, D.~Chen, C.~D. Manning, and A.~Ng.
\newblock Reasoning with neural tensor networks for knowledge base completion.
\newblock In {\em NIPS}, 2013.

\bibitem{Socher2013a}
R.~Socher, M.~Ganjoo, C.~D. Manning, and A.~Ng.
\newblock Zero-shot learning through cross-modal transfer.
\newblock In {\em NIPS}, 2013.

\bibitem{Venugopalan2016}
S.~Venugopalan, L.~A. Hendricks, M.~Rohrbach, R.~Mooney, T.~Darrell, and
  K.~Saenko.
\newblock Captioning images with diverse objects.
\newblock {\em arXiv preprint arXiv:1606.07770}, 2016.

\bibitem{Xian16}
Y.~Xian, Z.~Akata, G.~Sharma, Q.~Nguyen, M.~Hein, and B.~Schiele.
\newblock Latent embeddings for zero-shot classification.
\newblock {\em CVPR}, 2016.

\bibitem{Yao2010}
B.~Yao and L.~Fei-Fei.
\newblock Grouplet: A structured image representation for recognizing human and
  object interactions.
\newblock In {\em CVPR}, 2010.

\bibitem{Yao2010a}
B.~Yao and L.~Fei-Fei.
\newblock Modeling mutual context of object and human pose in human-object
  interaction activities.
\newblock In {\em CVPR}, 2010.

\bibitem{Yao11}
B.~Yao, X.~Jiang, A.~Khosla, A.~L. Lin, L.~Guibas, and L.~Fei-Fei.
\newblock Human action recognition by learning bases of action attributes and
  parts.
\newblock In {\em ICCV}, 2011.

\bibitem{yatskar2016stating}
M.~Yatskar, V.~Ordonez, and A.~Farhadi.
\newblock Stating the obvious: Extracting visual common sense knowledge.
\newblock In {\em NAACL}, 2016.

\bibitem{zagoruyko2016multipath}
S.~Zagoruyko, A.~Lerer, T.-Y. Lin, P.~O. Pinheiro, S.~Gross, S.~Chintala, and
  P.~Doll{\'a}r.
\newblock A multipath network for object detection.
\newblock {\em BMVC}, 2016.

\bibitem{Zhu2014}
Y.~Zhu, A.~Fathi, and L.~Fei-Fei.
\newblock Reasoning about object affordances in a knowledge base
  representation.
\newblock In {\em ECCV}, 2014.

\end{thebibliography}
}

\clearpage

\appendix

\section*{Overview of supplementary material}
In this supplementary material, we provide additional qualitative and quantitative results for our weakly-supervised approach described in Section~\ref{model} of the main paper. First, in Section~\ref{part:multimodal} we provide visualization of learned components of our spatial model to illustrate that our spatial features can handle ambiguous multimodal relations. We then show in Section~\ref{part:unrel} additional qualitative results for retrieval on the UnRel dataset that complement Figure \ref{fig:unrel} of the main paper. In Section \ref{part:vrd} we provide additional qualitative results on the Visual Relationship Dataset. In particular, we show additional examples of retrieved triplets for several action predicates, such as ``ride" and ``hold" that appear less frequently. We also provide additional visualizations for the zero-shot retrieval task, comparing our predictions to the Visual+Language model of~\cite{Lu16}. In Section~\ref{part:vg}, we provide results on the Visual Genome dataset. In Section~\ref{part:quantitative}, we provide additional quantitative results using alternative evaluation criteria  on UnRel and Visual Relationship Detection datasets. 
Finally, in section~\ref{part:lu-baseline}, we report a complete set of results we obtained by running the baseline \cite{Lu16} on the Visual Relationship Detection task. We report these results as they are slightly different than those reported in \cite{Lu16}.

\section{Handling multimodal relations}
\label{part:multimodal}
\begin{figure*}[t]
\centering
    \begin{minipage}[b]{0.005\textwidth}
    	\centering
    	\begin{turn}{90}
    GMM 1
    	\end{turn}
    	\vspace{2ex}
    \end{minipage}
    \hspace{0.01\textwidth}
    \begin{minipage}[b]{0.18\textwidth}
    	\centering
       	\includegraphics[trim={4cm 2cm 4cm 3cm},clip,width=\linewidth]{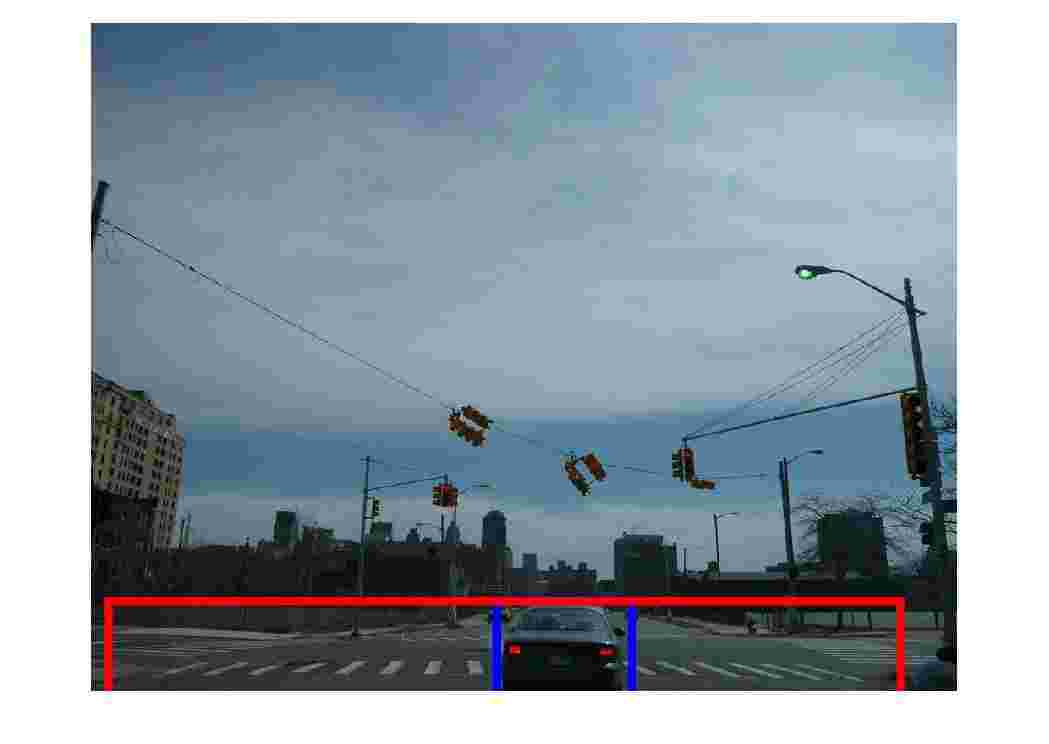}
    \end{minipage}
    \hspace{0.005\textwidth}
    \begin{minipage}[b]{0.18\textwidth}
    	\centering
       	\includegraphics[trim={5cm 2cm 4cm 1.2cm},clip,width=\linewidth]{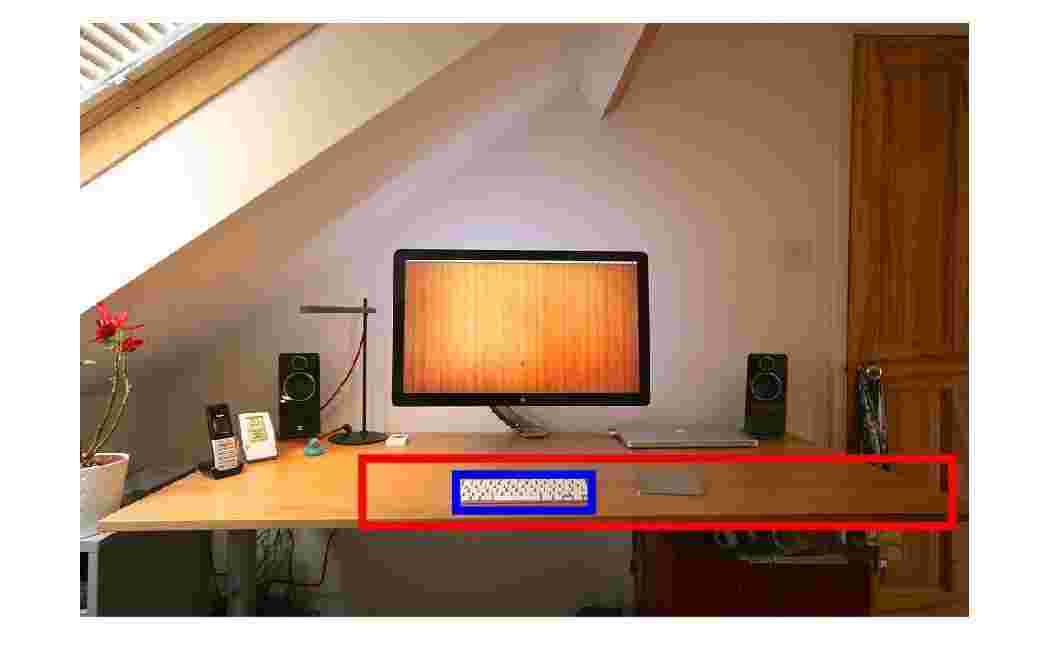}
    \end{minipage}
    \hspace{0.005\textwidth}
    \begin{minipage}[b]{0.18\textwidth}
    	\centering
       	\includegraphics[trim={3cm 3.4cm 4cm 1cm},clip,width=\linewidth]{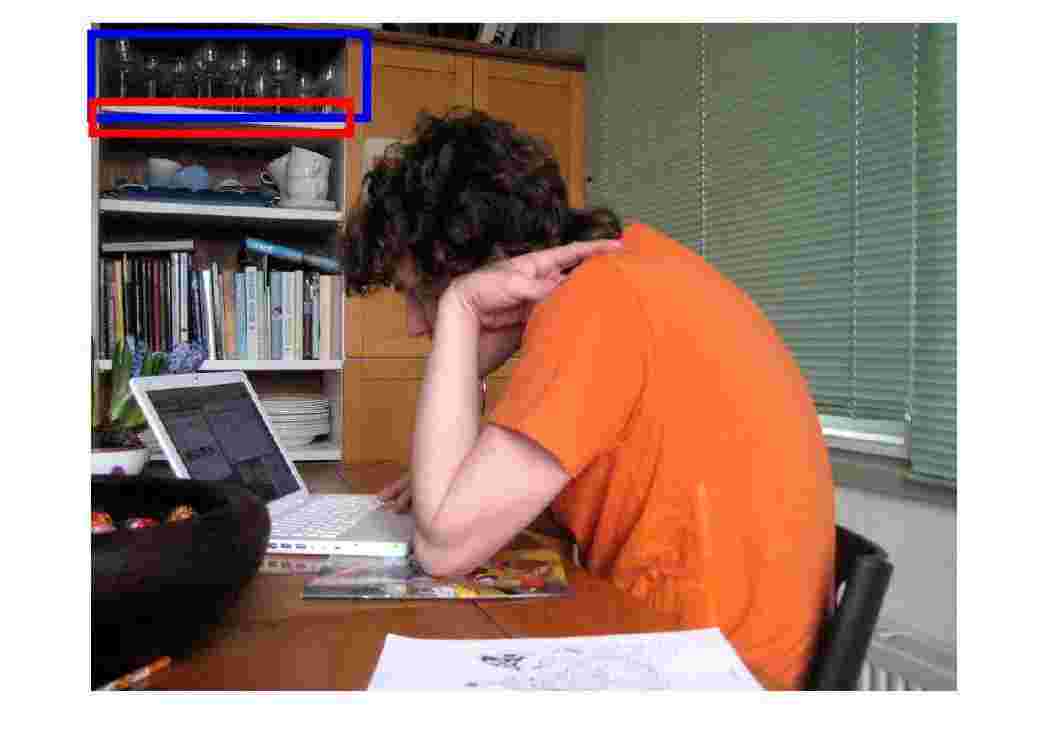}
    \end{minipage}
    \hspace{0.005\textwidth}
	\begin{minipage}[b]{0.18\textwidth}
    	\centering
       	\includegraphics[trim={3cm 2cm 4cm 12.85cm},clip,width=\linewidth]{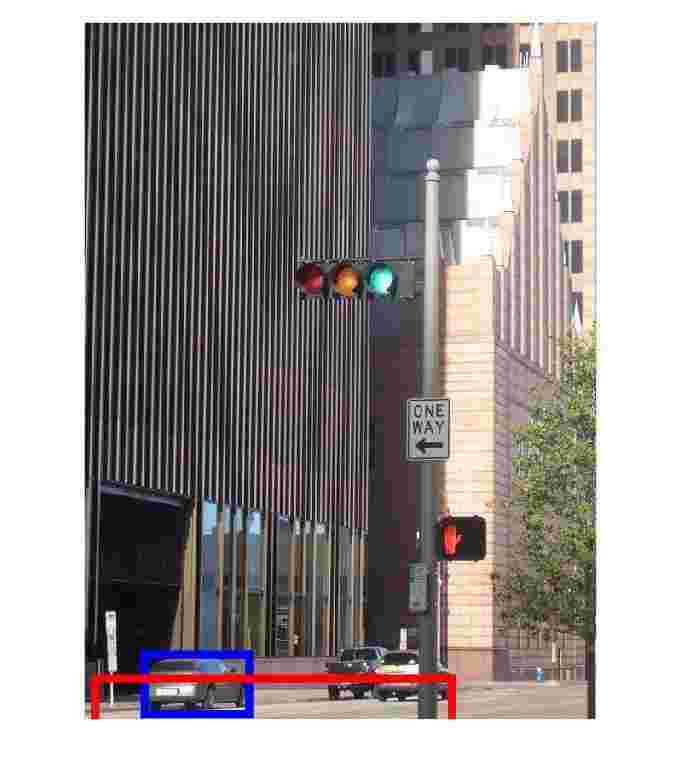}
    \end{minipage}
    \hspace{0.005\textwidth}
    \begin{minipage}[b]{0.18\textwidth}
    	\centering
       	\includegraphics[trim={3.5cm 2cm 3cm 2cm},clip,width=\linewidth]{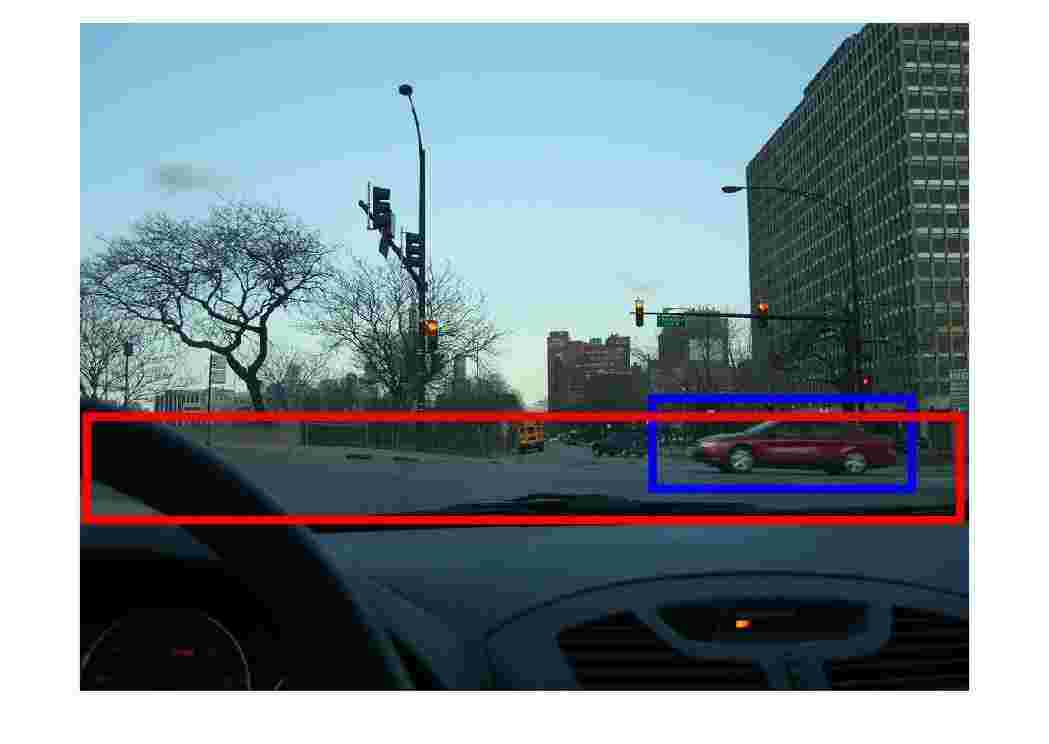}
    \end{minipage} 
	   
    \begin{minipage}[b]{0.005\textwidth}
    	\centering
    	\begin{turn}{90}
    GMM 2
    	\end{turn}
    	\vspace{2.5ex}
    \end{minipage}
    \hspace{0.01\textwidth}
    \begin{minipage}[b]{0.18\textwidth}
    	\centering
       	\includegraphics[trim={3cm 2.1cm 4cm 2.8cm},clip,width=\linewidth]{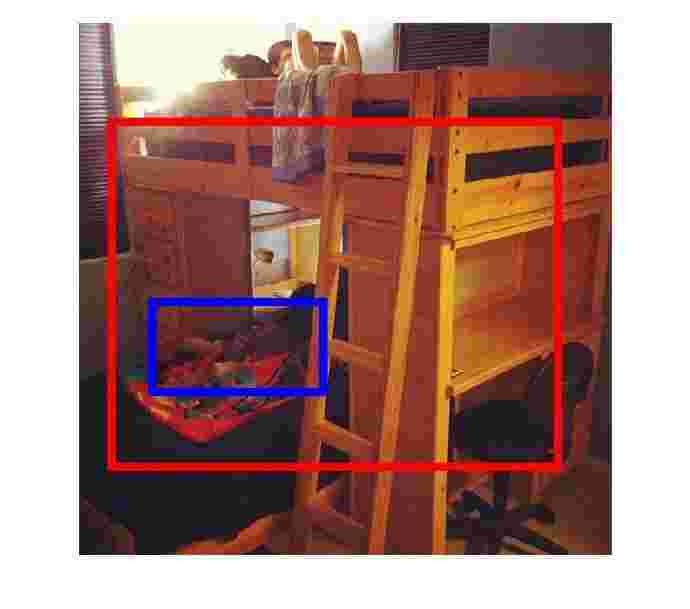}
    \end{minipage}
    \hspace{0.005\textwidth}
    \begin{minipage}[b]{0.18\textwidth}
    	\centering
       	\includegraphics[trim={8cm 2cm 4cm 1cm},clip,width=\linewidth]{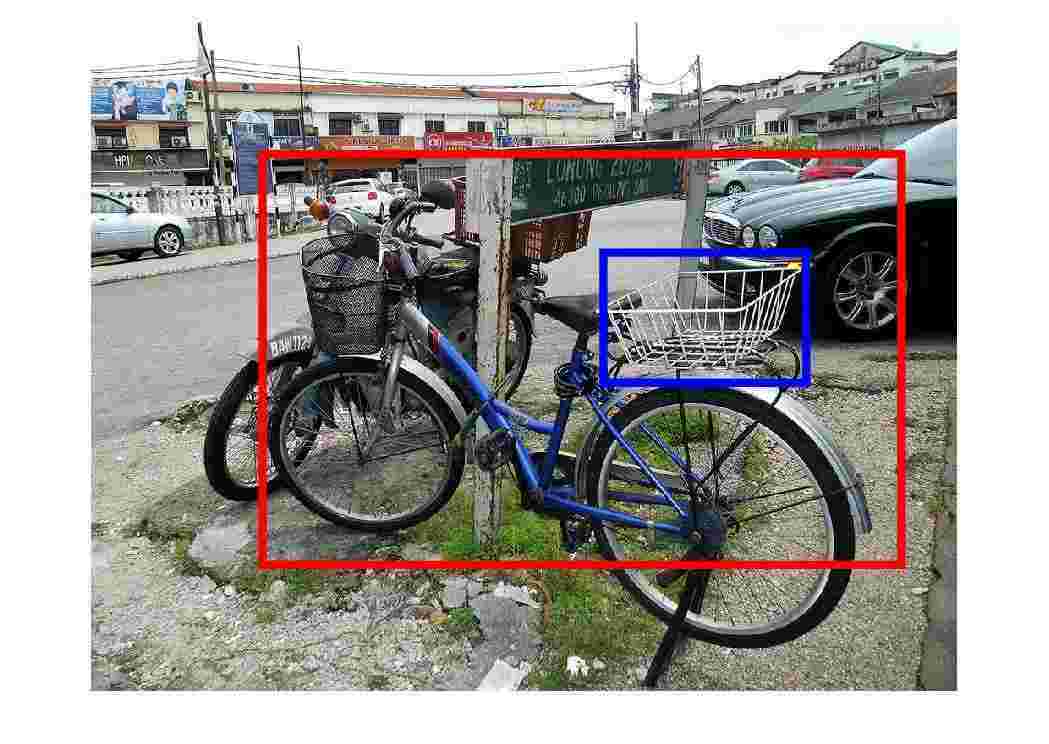}
    \end{minipage}
    \hspace{0.005\textwidth}
    \begin{minipage}[b]{0.18\textwidth}
    	\centering
       	\includegraphics[trim={11.3cm 2cm 4cm 1cm},clip,width=\linewidth]{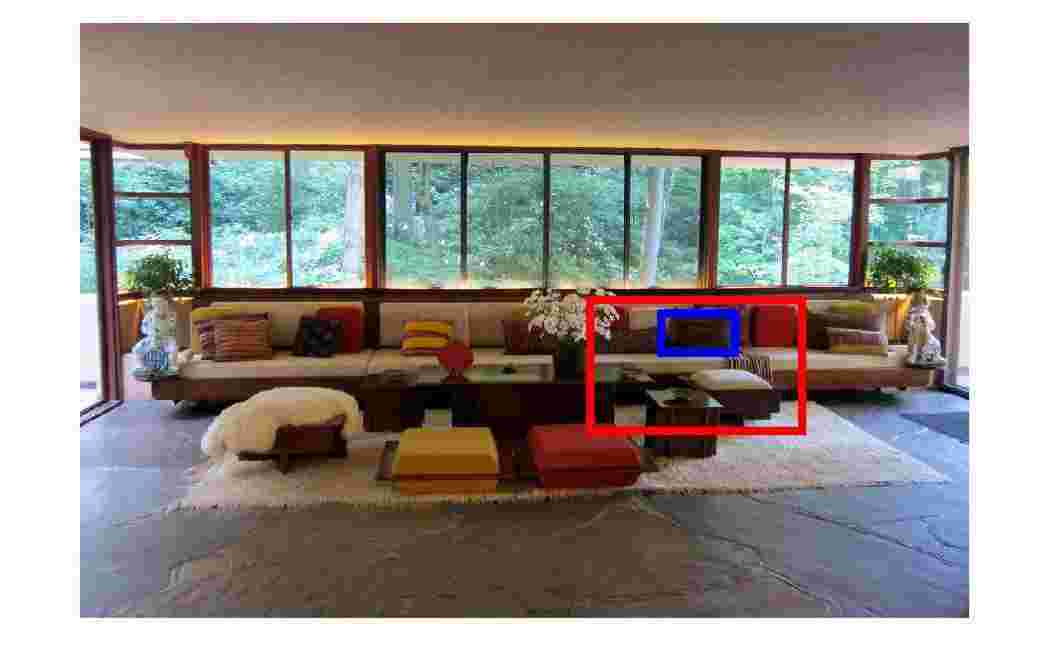}
    \end{minipage}
    \hspace{0.005\textwidth}
	\begin{minipage}[b]{0.18\textwidth}
    	\centering
       	\includegraphics[trim={4cm 2cm 8.6cm 1cm},clip,width=\linewidth]{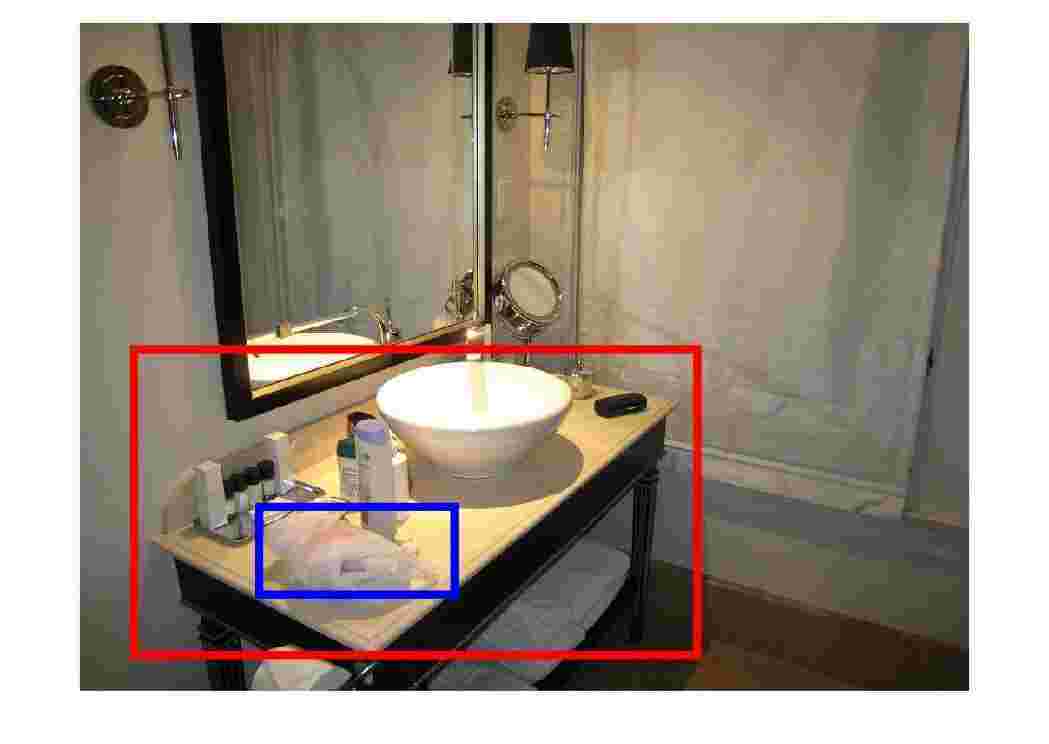}
    \end{minipage}
    \hspace{0.005\textwidth}
    \begin{minipage}[b]{0.18\textwidth}
    	\centering
       	\includegraphics[trim={5cm 2cm 8.6cm 2cm},clip,width=\linewidth]{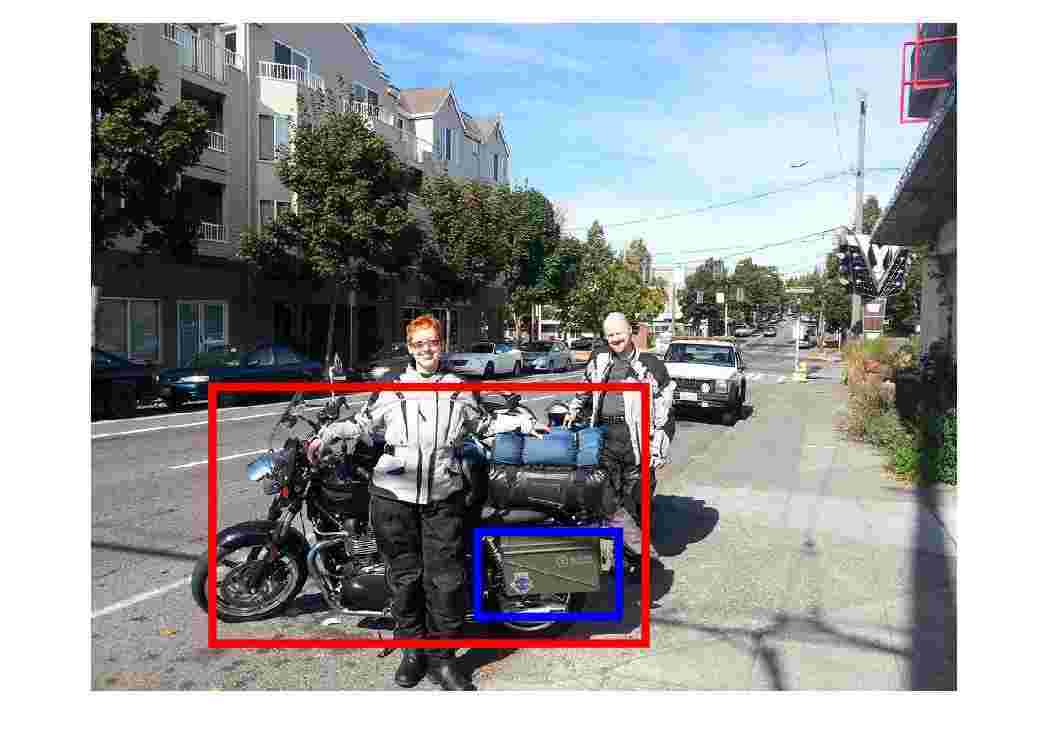}
    \end{minipage}

    \begin{minipage}[b]{0.005\textwidth}
    	\centering
    	\begin{turn}{90}
    GMM 3
    	\end{turn}
    	\vspace{2ex}
    \end{minipage}
    \hspace{0.01\textwidth}
    \begin{minipage}[b]{0.18\textwidth}
    	\centering
       	\includegraphics[trim={4.1cm 1.8cm 6.3cm 1cm},clip,width=\linewidth]{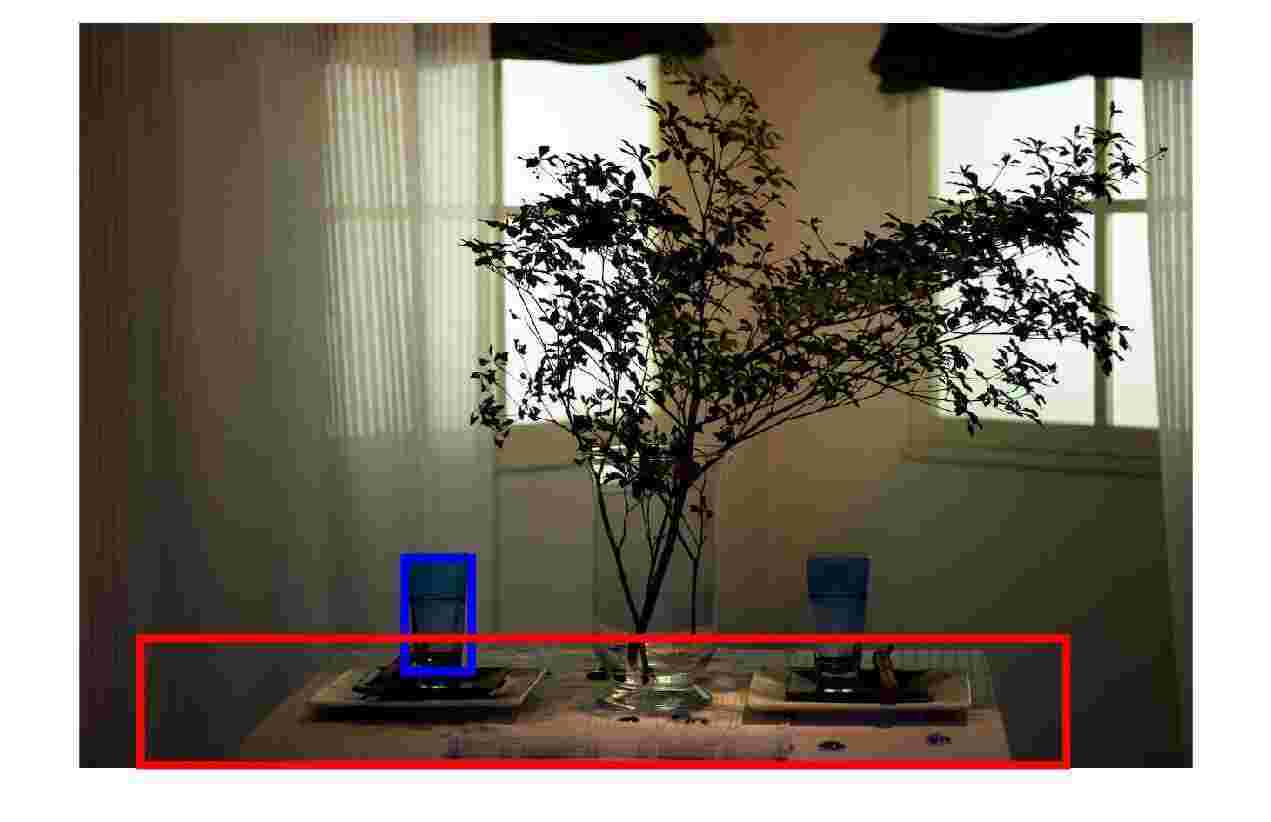}
    \end{minipage}
    \hspace{0.005\textwidth}
    \begin{minipage}[b]{0.18\textwidth}
    	\centering
       	\includegraphics[trim={3.2cm 1.9cm 6.4cm 1cm},clip,width=\linewidth]{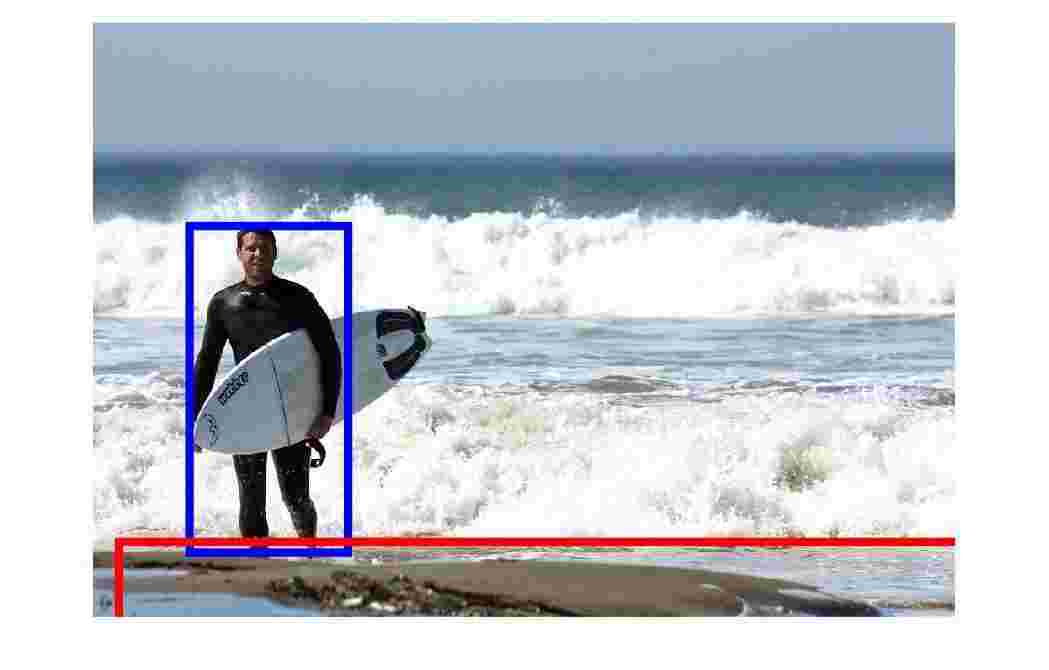}
    \end{minipage}
    \hspace{0.005\textwidth}
    \begin{minipage}[b]{0.18\textwidth}
    	\centering
       	\includegraphics[trim={3.3cm 1.8cm 9.4cm 1cm},clip,width=\linewidth]{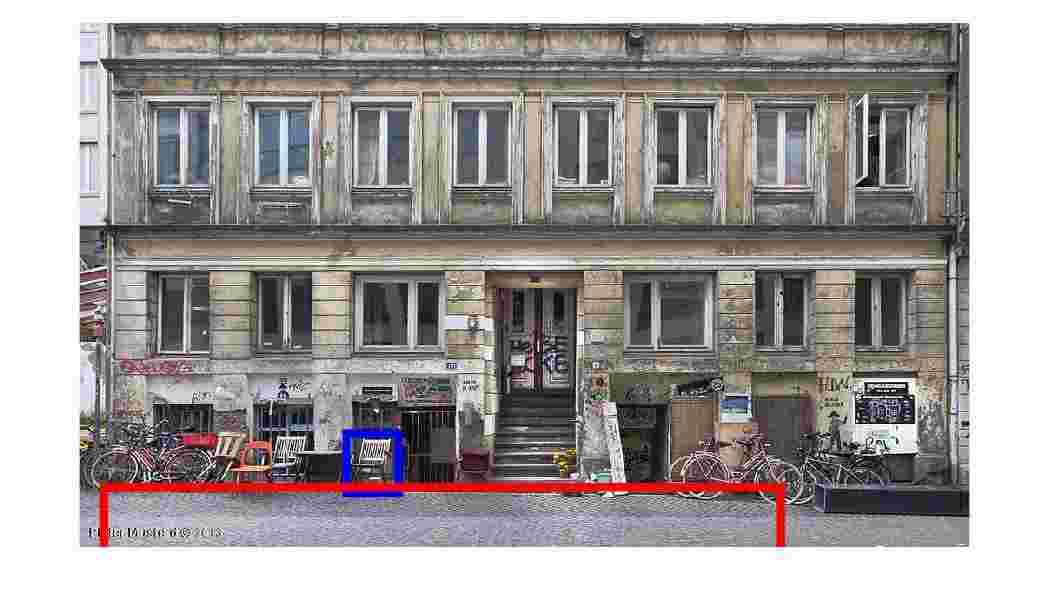}
    \end{minipage}
    \hspace{0.005\textwidth}
	\begin{minipage}[b]{0.18\textwidth}
    	\centering
       	\includegraphics[trim={3.4cm 2.1cm 3.2cm 1.2cm},clip,width=\linewidth]{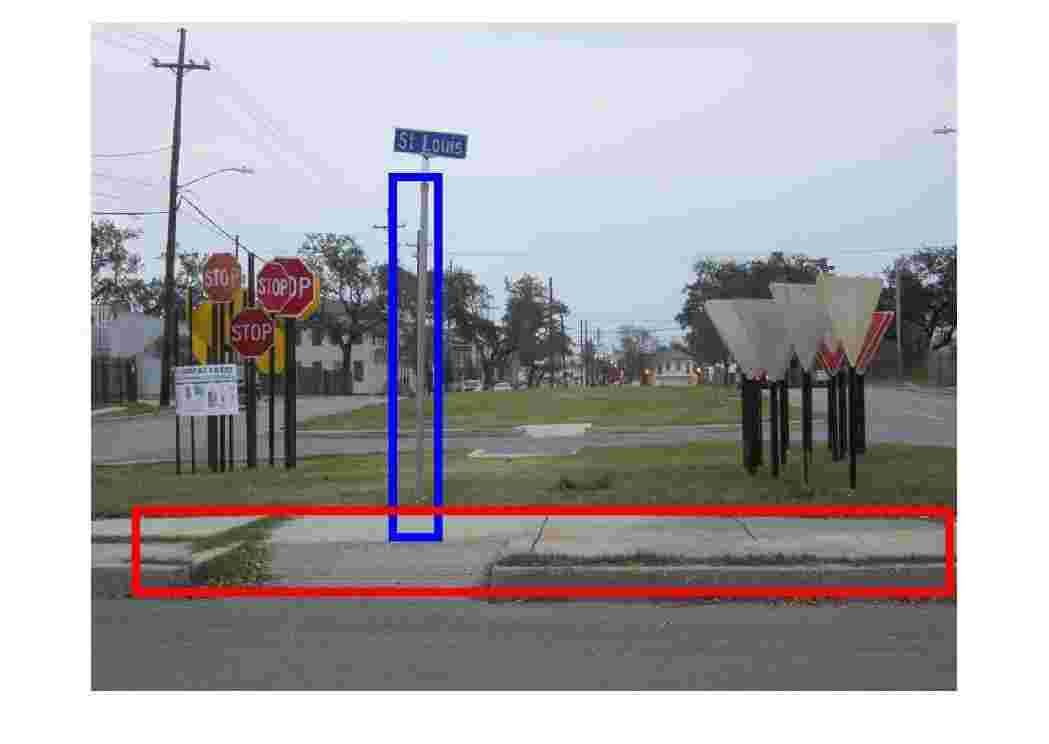}
    \end{minipage}
    \hspace{0.005\textwidth}
    \begin{minipage}[b]{0.18\textwidth}
    	\centering
       	\includegraphics[trim={3.2cm 2cm 3.2cm 1cm},clip,width=\linewidth]{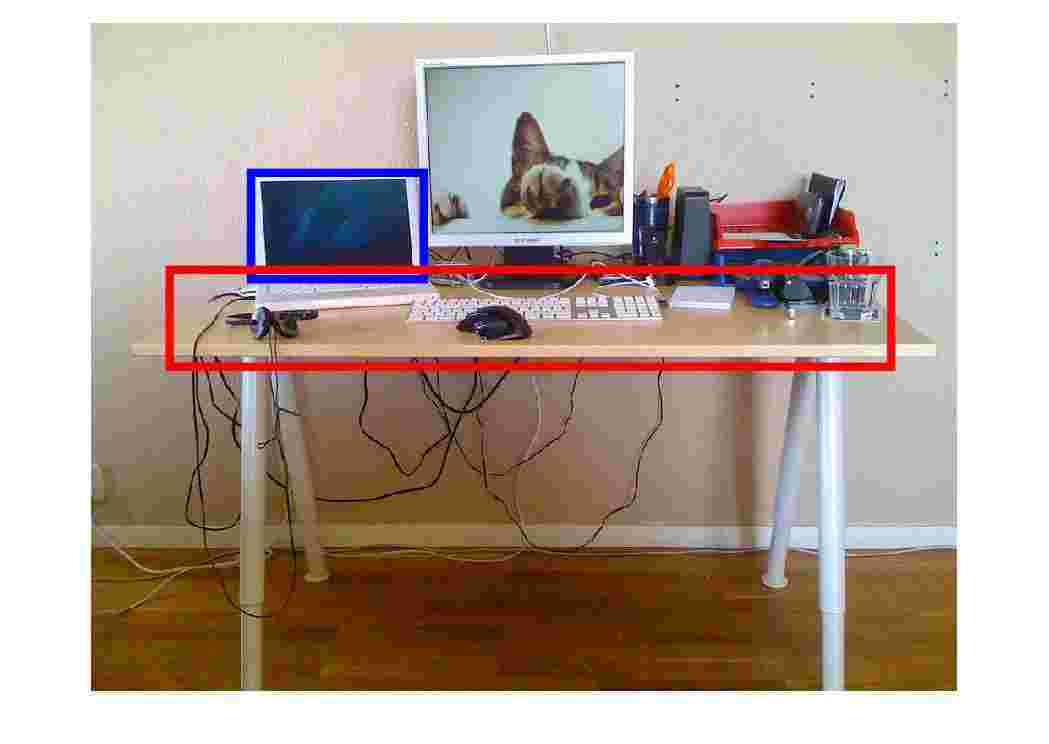}
    \end{minipage}

    \begin{minipage}[b]{0.005\textwidth}
    	\centering
    	\begin{turn}{90}
    GMM 4
    	\end{turn}
    	\vspace{1.5ex}
    \end{minipage}
    \hspace{0.01\textwidth}
    \begin{minipage}[b]{0.18\textwidth}
    	\centering
       	\includegraphics[trim={3.1cm 2cm 7cm 3cm},clip,width=\linewidth]{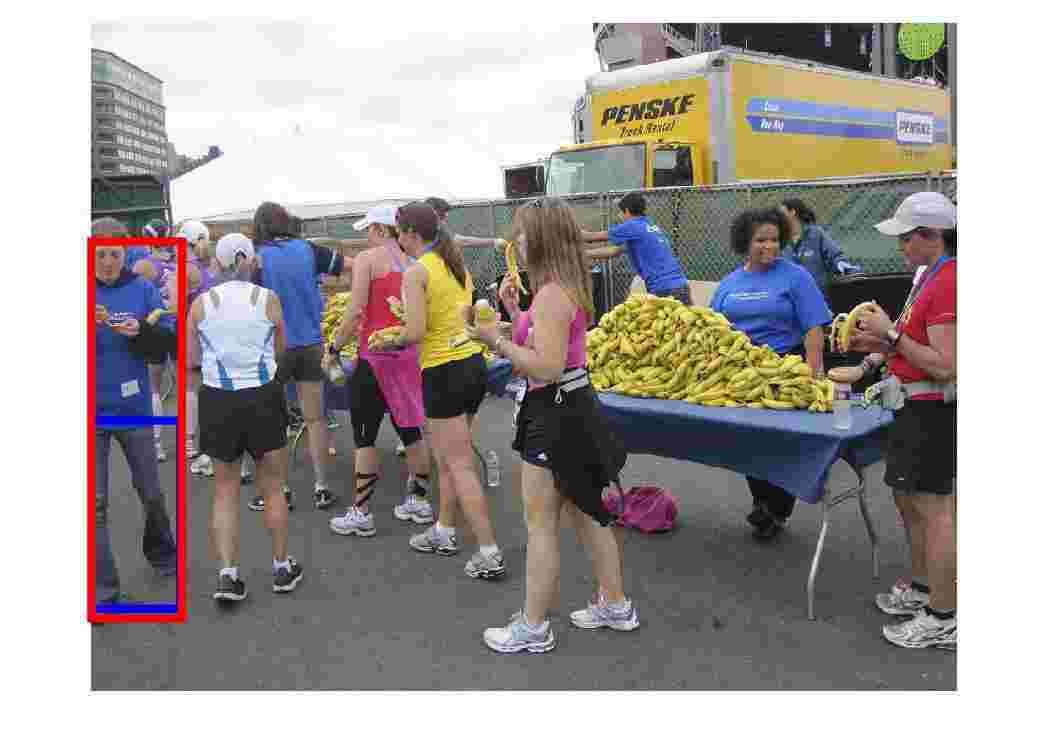}
    \end{minipage}
    \hspace{0.005\textwidth}
    \begin{minipage}[b]{0.18\textwidth}
    	\centering
       	\includegraphics[trim={9.8cm 2cm 3cm 2.5cm},clip,width=\linewidth]{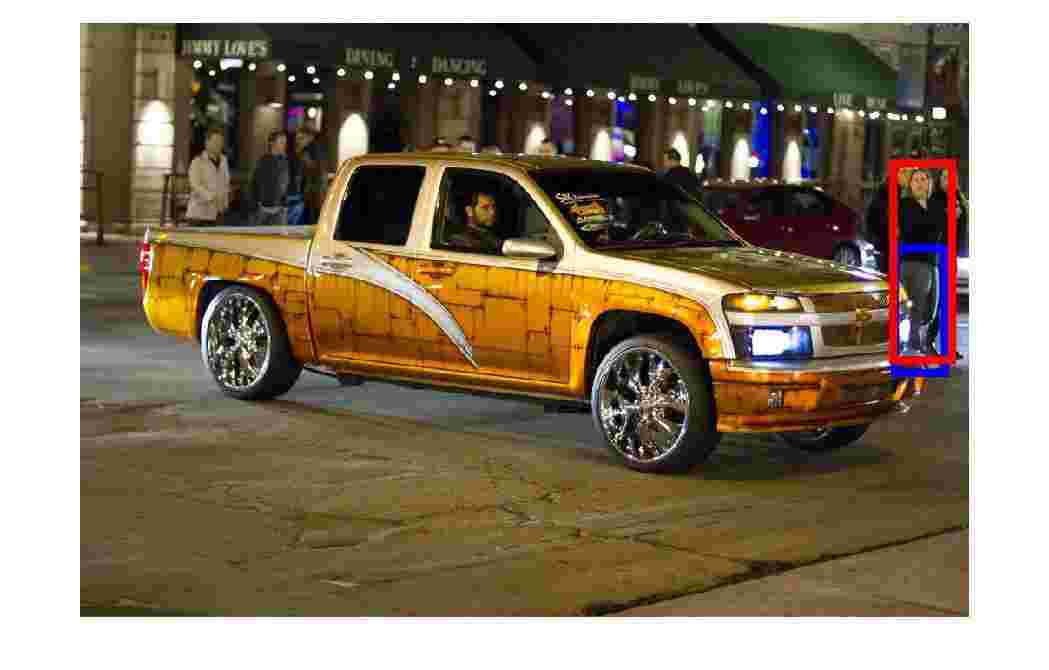}
    \end{minipage}
    \hspace{0.005\textwidth}
    \begin{minipage}[b]{0.18\textwidth}
    	\centering
       	\includegraphics[trim={8.7cm 2cm 3.5cm 2cm},clip,width=\linewidth]{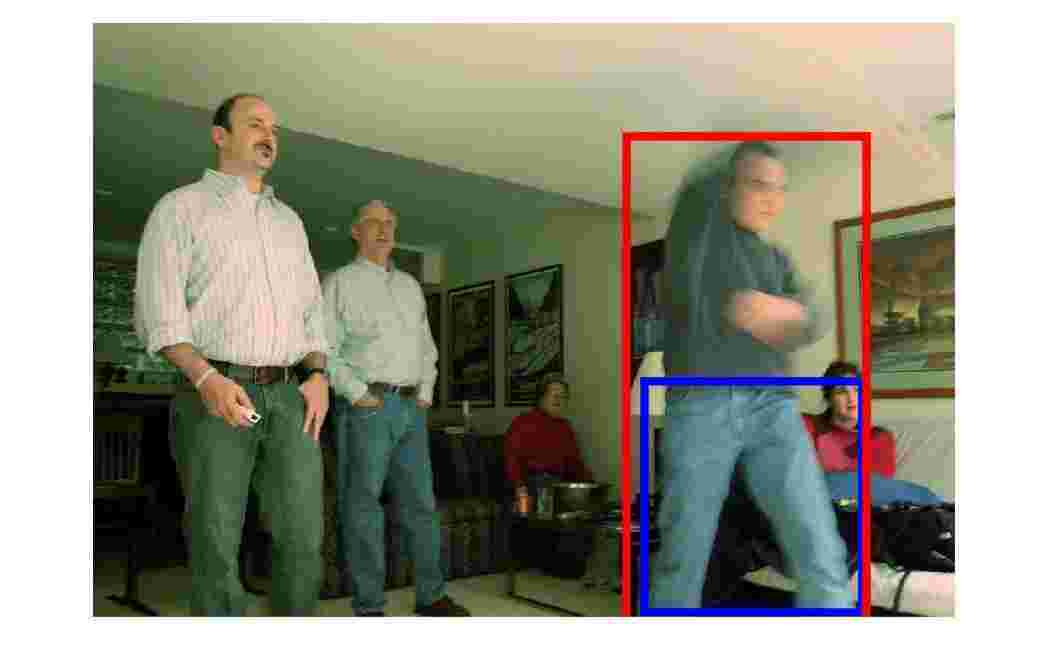}
    \end{minipage}
    \hspace{0.005\textwidth}
	\begin{minipage}[b]{0.18\textwidth}
    	\centering
       	\includegraphics[trim={8.1cm 2cm 3cm 4cm},clip,width=\linewidth]{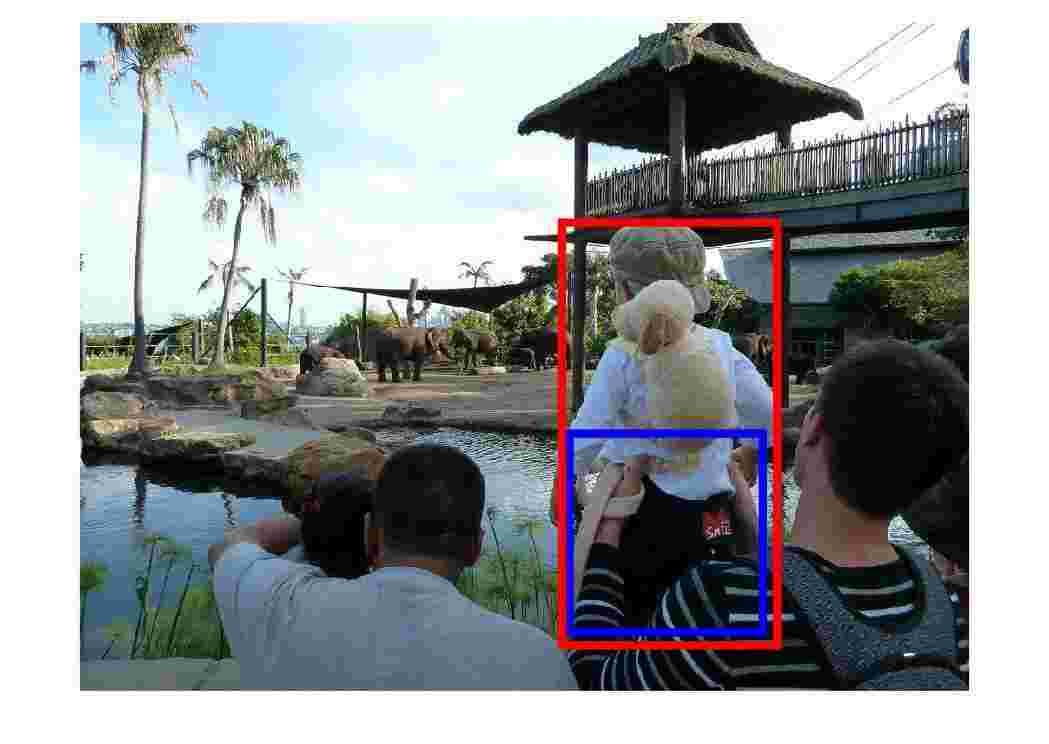}
    \end{minipage}
    \hspace{0.005\textwidth}
    \begin{minipage}[b]{0.18\textwidth}
    	\centering
       	\includegraphics[trim={3cm 2cm 3.5cm 2cm},clip,width=\linewidth]{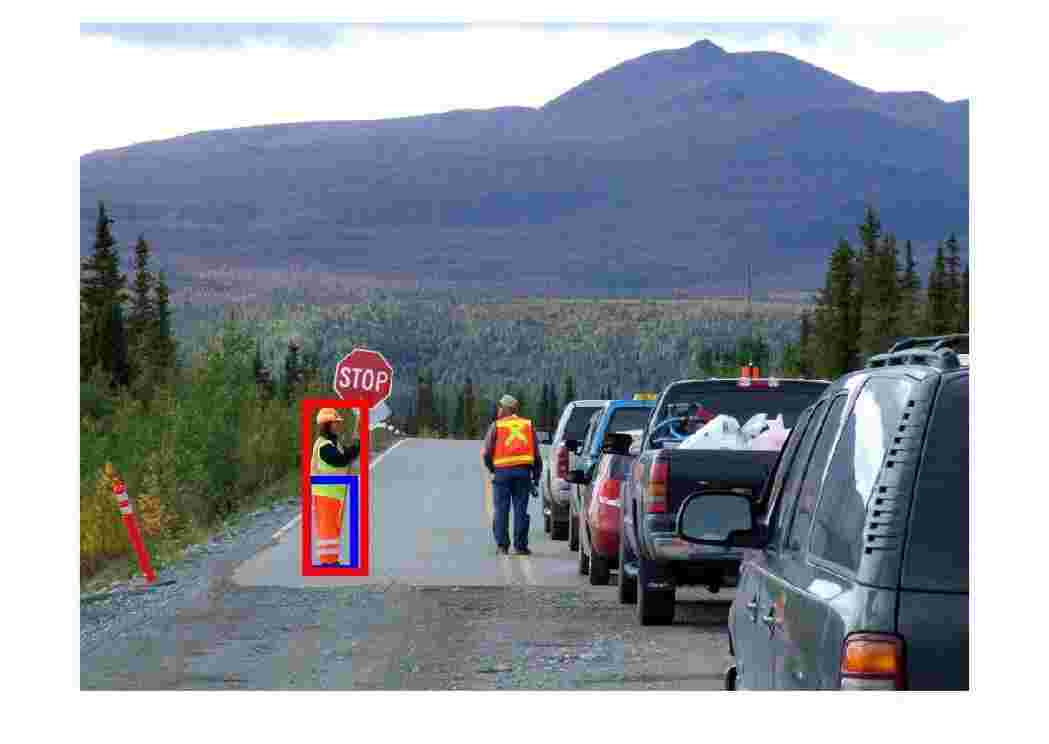}
    \end{minipage}             
     
   \caption{
      High-scoring GMM components for the relation ``on'' learned by our weakly-supervised model. Each row shows examples of pairs of boxes ({\color{blue}blue} on {\color{red}red}) for one GMM component. Note that our GMM-based spatial model can capture different configurations for highly multimodal relations such as ``on".} 
      \vspace{-.2cm}
    \label{fig:multimodal}
\end{figure*}

In Figure \ref{fig:multimodal} we show examples of pairs of boxes belonging to the top-4 scoring GMM components for the ambiguous relation ``on". Each row corresponds to a different learned mode of ``on". In particular, components 1 and 3 represent an object being on top of another object, component 4 corresponds to a garment worn by a person, which is often described by an ``on'' relation, such as ``pants on person''. Component 2 often corresponds to the ``on top'' configuration, where the two objects are captured from an elevated viewpoint.

\section{Qualitative results on UnRel dataset}
\label{part:unrel}

\begin{figure*}[t]
\vspace{-0.7cm}
\centering
	\begin{minipage}[t]{0.005\textwidth}
    	\centering
    	\vspace{0.6ex}
    \end{minipage}
    \hspace{0.01\textwidth}
	\begin{minipage}[t]{0.58\textwidth}
    	\centering
    \textit{	Top true positives}\\
    	\vspace{0.2ex}
	\end{minipage}
	\hspace{0.005\textwidth}
	\begin{minipage}[t]{0.18\textwidth}
    \centering
    \textit{	Top false positives}\\
    	\vspace{0.2ex}
	\end{minipage}
	\hspace{0.005\textwidth}
	\begin{minipage}[t]{0.18\textwidth}
    \centering
    	\textit{Missed detections}\\
    	\vspace{0.2ex}
	\end{minipage}	
	
	\begin{minipage}[b]{0.005\textwidth}
    	\centering
    	\begin{turn}{90}
    \small{{\color{blue}dog} ride {\color{red}bike}}
    	\end{turn}	
    \vspace{-0.5ex}
    \end{minipage}
    \hspace{0.01\textwidth}
    \begin{minipage}[t]{0.18\textwidth}
    	\centering
       	\includegraphics[trim={1cm 3cm 1cm 1.5cm},clip,width=0.95\linewidth,cfbox={green 2pt 2pt}]{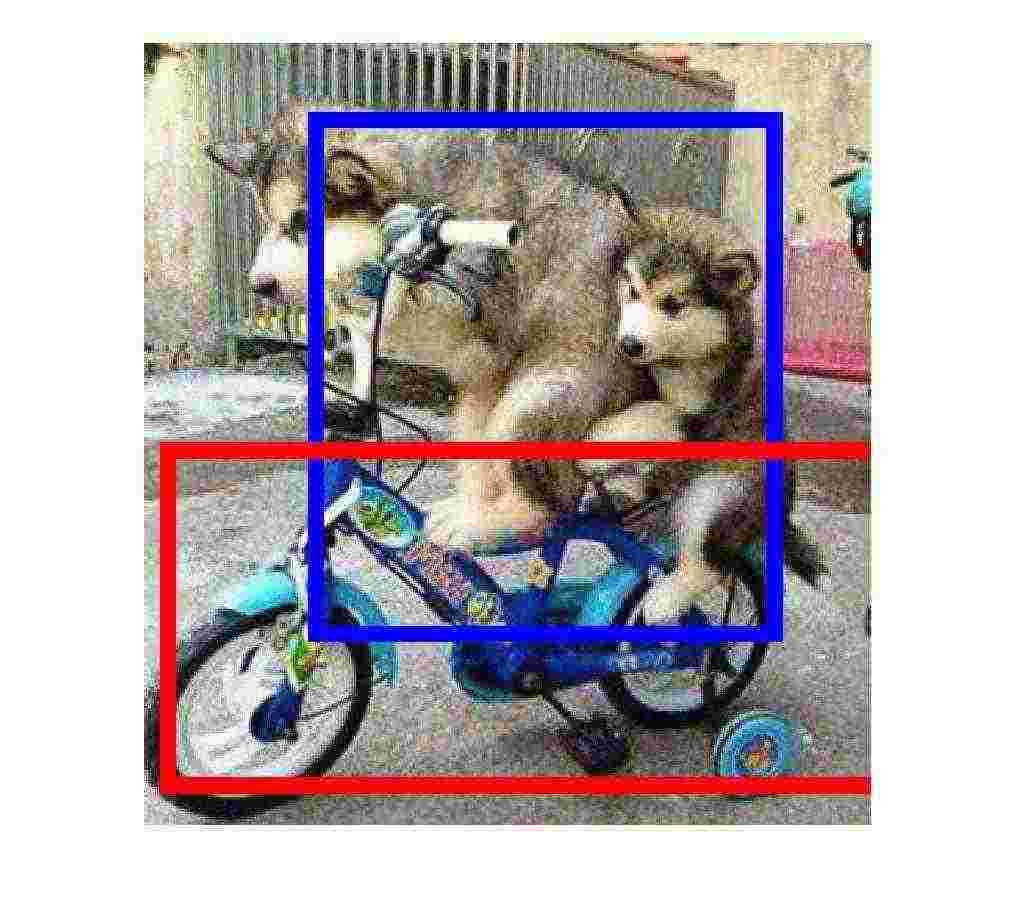}\\
       	\vspace{1.5ex}
    \end{minipage}
    \hspace{0.005\textwidth}
    \begin{minipage}[t]{0.18\textwidth}
    	\centering
       	\includegraphics[trim={3cm 2cm 3.2cm 1cm},clip,width=0.95\linewidth,cfbox={green 2pt 2pt}]{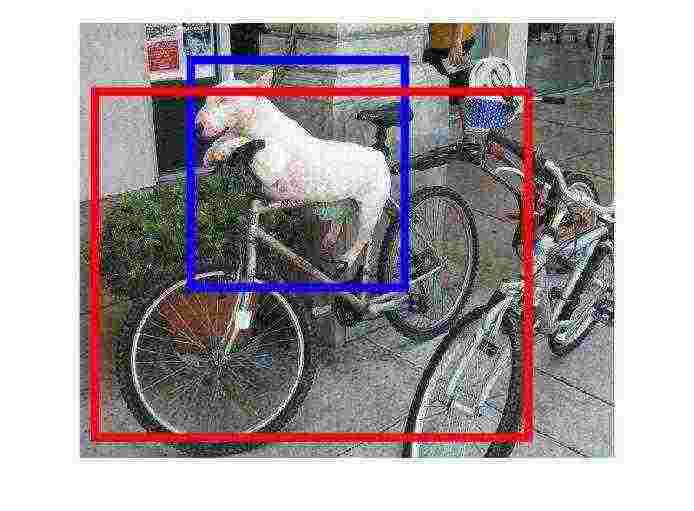}\\
       	\vspace{1.5ex}
    \end{minipage}
    \hspace{0.005\textwidth}
    \begin{minipage}[t]{0.18\textwidth}
       \centering
       \includegraphics[trim={0cm 3cm 0cm 2cm},clip,width=0.95\linewidth,cfbox={green 2pt 2pt}]{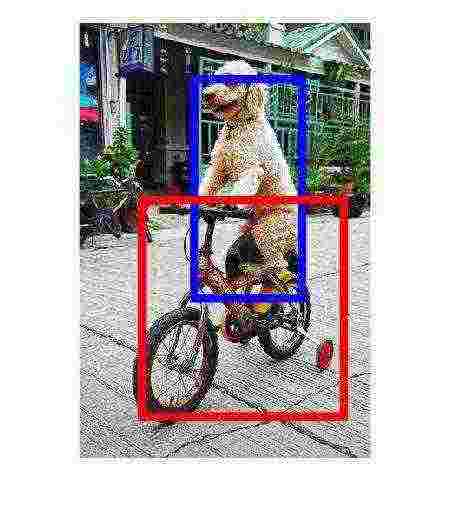}\\
       \vspace{1.5ex}
    \end{minipage}
    \hspace{0.005\textwidth}
    \begin{minipage}[t]{0.18\textwidth}
    	\centering
       	\includegraphics[trim={3.1cm 1.95cm 7cm 1cm},clip,width=0.95\linewidth,cfbox={red 2pt 2pt}]{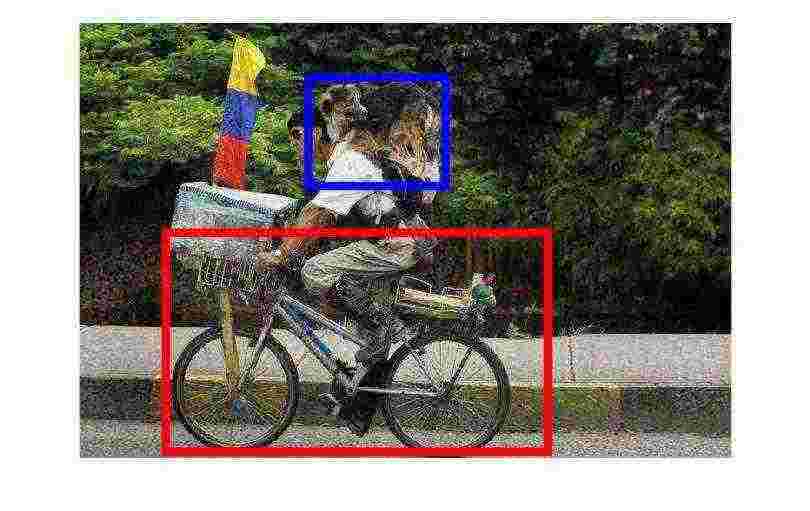}\\
      	\vspace{1.5ex}
    \end{minipage} 
    \hspace{0.005\textwidth}  
    \begin{minipage}[t]{0.18\textwidth}
    	\centering
       	\includegraphics[trim={5.5cm 1.7cm 5cm 1cm},clip,width=0.95\linewidth,cfbox={blue 2pt 2pt}]{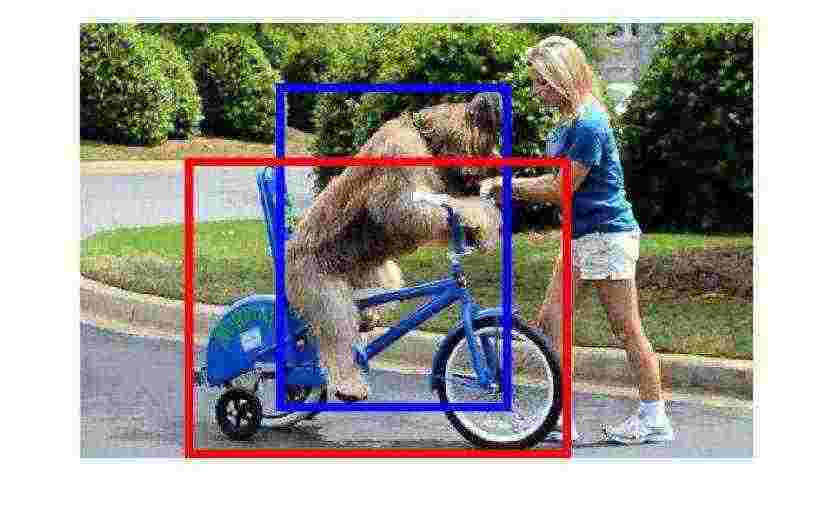}\\
       	\vspace{1.5ex}
    \end{minipage}

	\begin{minipage}[b]{0.005\textwidth}
    	\centering
    	\begin{turn}{90}
    	\small{{\color{blue}umbrella} cover {\color{red}dog}}
    	\end{turn}	
    \vspace{-3.5ex}
    \end{minipage}
    \hspace{0.01\textwidth}
    \begin{minipage}[t]{0.18\textwidth}
    	\centering
       	\includegraphics[trim={8.5cm 3cm 5cm 2cm},clip,width=0.95\linewidth,cfbox={green 2pt 2pt}]{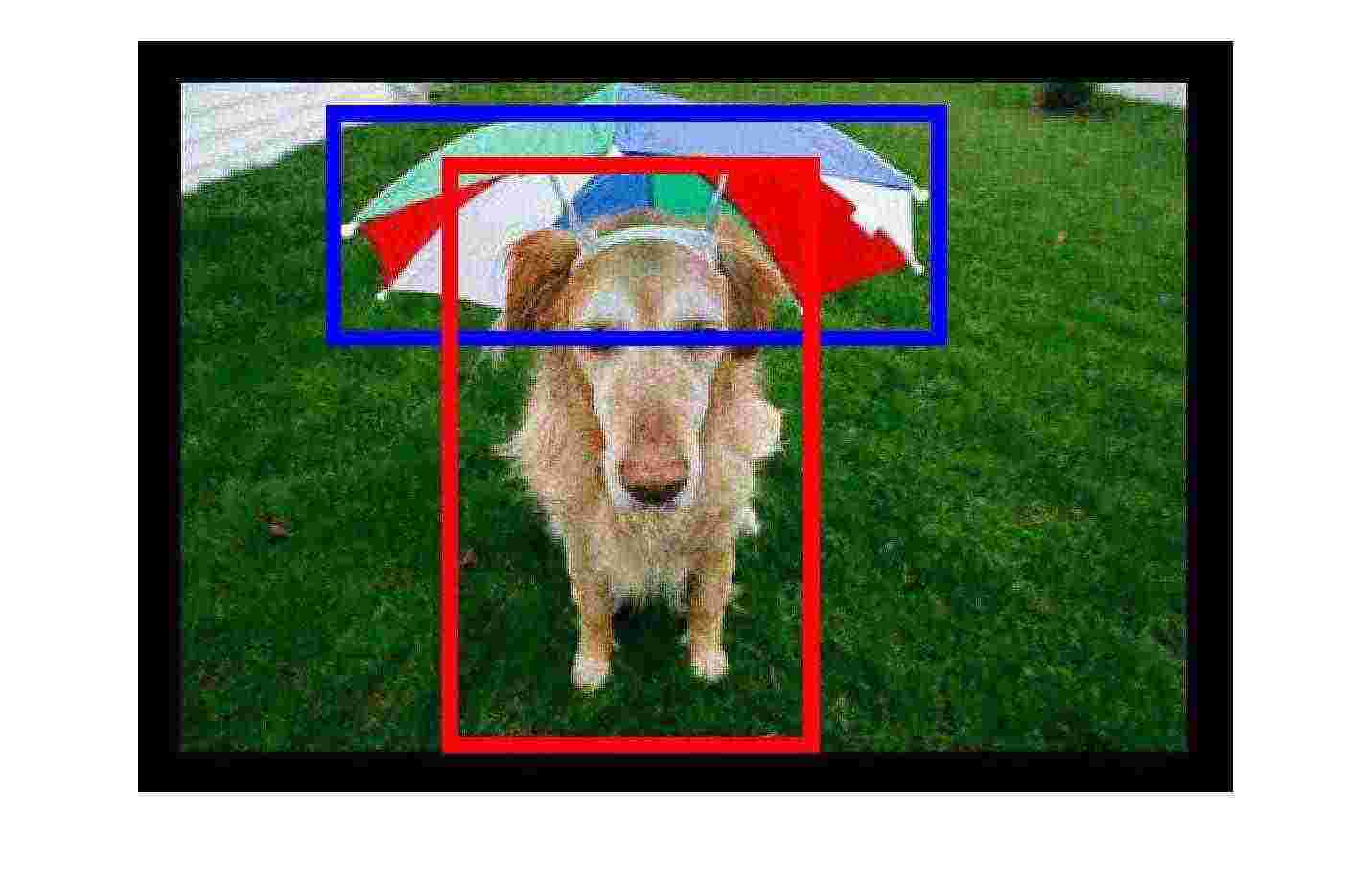}\\
       	\vspace{1.5ex}
    \end{minipage}
    \hspace{0.005\textwidth}
    \begin{minipage}[t]{0.18\textwidth}
       \centering
       \includegraphics[trim={0cm 5.5cm 0cm 4.4cm},clip,width=0.95\linewidth,cfbox={green 2pt 2pt}]{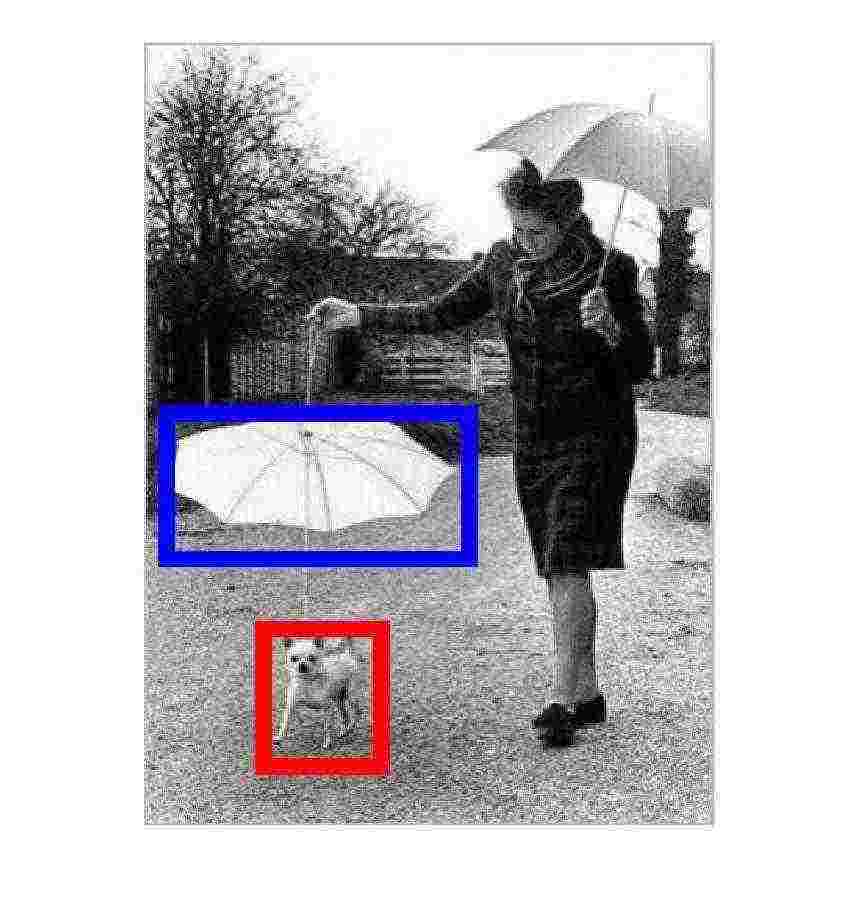}\\
       \vspace{1.5ex}
    \end{minipage}
    \hspace{0.005\textwidth}
    \begin{minipage}[t]{0.18\textwidth}
    	\centering
       	\includegraphics[trim={6cm 2cm 5cm 0cm},clip,width=0.95\linewidth,cfbox={green 2pt 2pt}]{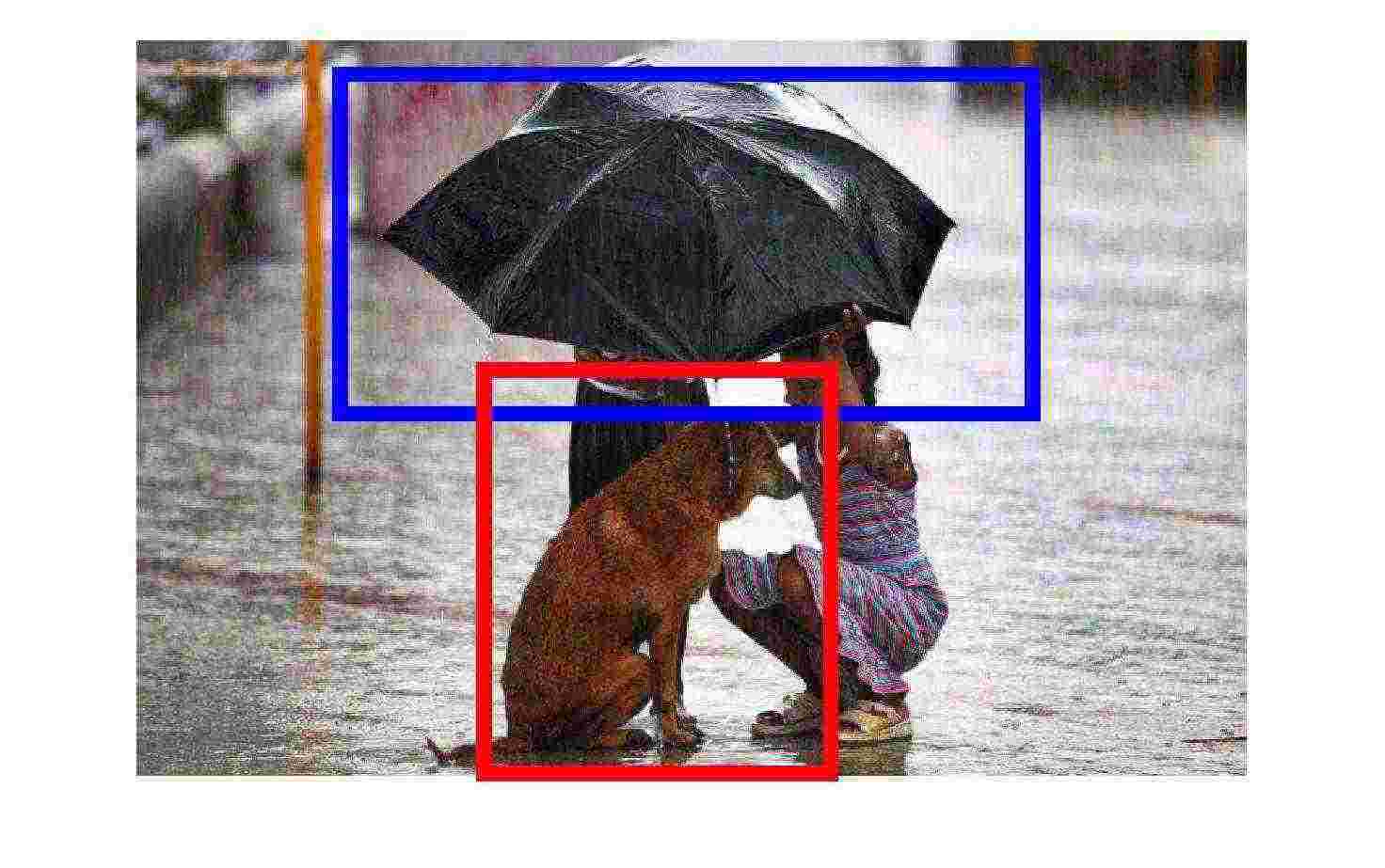}\\
       	\vspace{1.5ex}
    \end{minipage}
    \hspace{0.005\textwidth}  
    \begin{minipage}[t]{0.18\textwidth}
    	\centering
       	\includegraphics[trim={1.6cm 4.5cm 1.6cm 2cm},clip,width=0.95\linewidth,cfbox={red 2pt 2pt}]{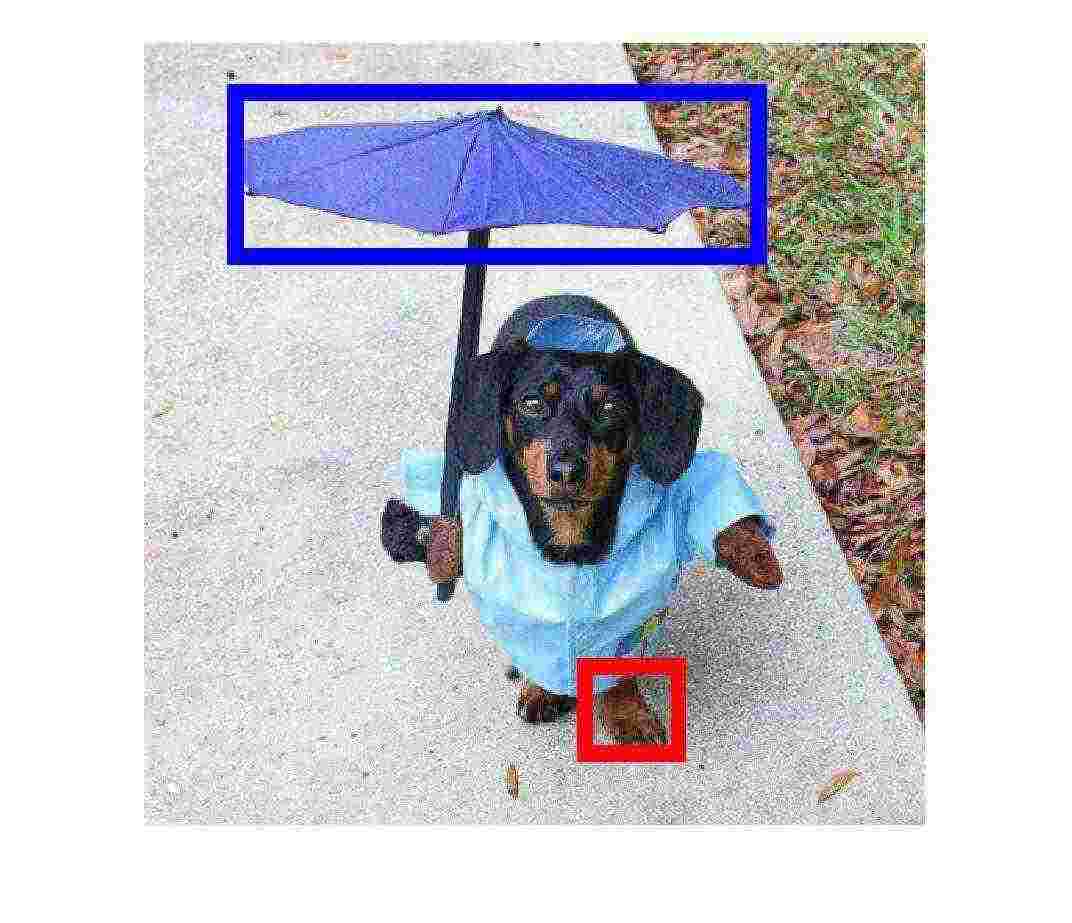}\\
      	\vspace{1.5ex}
    \end{minipage} 
    \hspace{0.005\textwidth}
    \begin{minipage}[t]{0.18\textwidth}
    	\centering
       	\includegraphics[trim={6.5cm 3.5cm 6.5cm 0cm},clip,width=0.95\linewidth,cfbox={blue 2pt 2pt}]{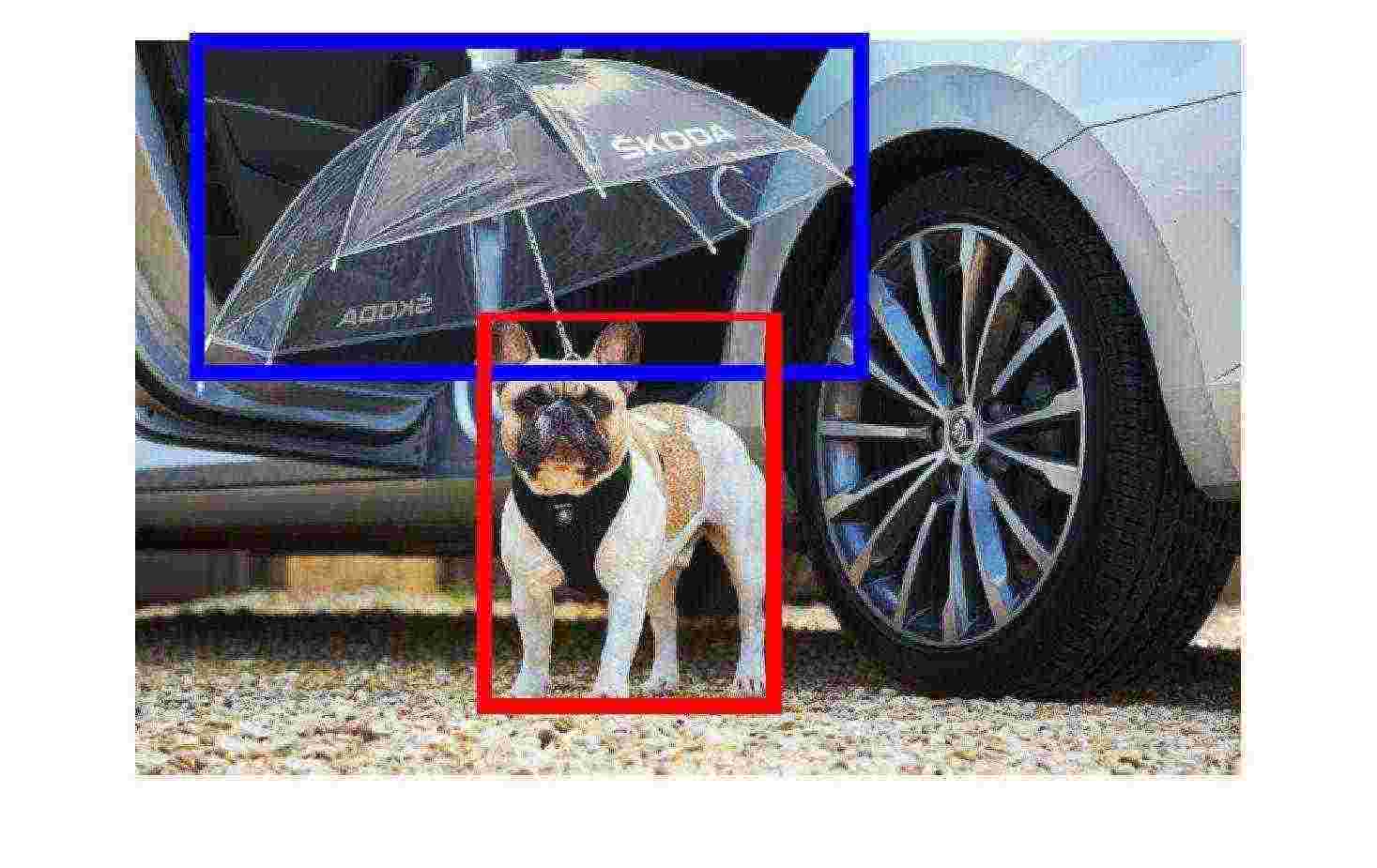}\\
       	\vspace{1.5ex}
    \end{minipage}

	\begin{minipage}[b]{0.005\textwidth}
    	\centering
    	\begin{turn}{90}
    \small{{\color{blue}person} hold {\color{red}car}}
    	\end{turn}	
    \vspace{-2.5ex}
    \end{minipage}
    \hspace{0.01\textwidth}
    \begin{minipage}[t]{0.18\textwidth}
    	\centering
       	\includegraphics[trim={0.9cm 3cm 0.9cm 1cm},clip,width=0.95\linewidth,cfbox={green 2pt 2pt}]{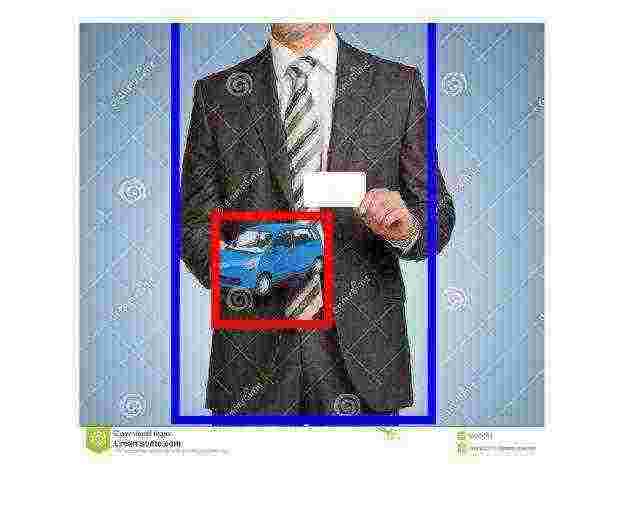}\\
       	\vspace{1.5ex}
    \end{minipage}
    \hspace{0.005\textwidth}
    \begin{minipage}[t]{0.18\textwidth}
    	\centering
       	\includegraphics[trim={4cm 1.8cm 2.5cm 1cm},clip,width=0.95\linewidth,cfbox={green 2pt 2pt}]{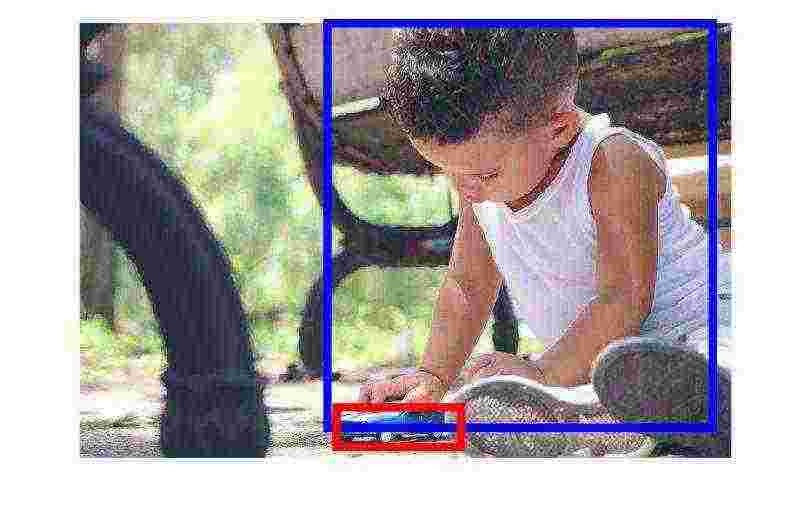}\\
       	\vspace{1.5ex}
    \end{minipage}
    \hspace{0.005\textwidth}  
    \begin{minipage}[t]{0.18\textwidth}
       \centering
       \includegraphics[trim={3cm 2cm 4.6cm 1.5cm},clip,width=0.95\linewidth,cfbox={green 2pt 2pt}]{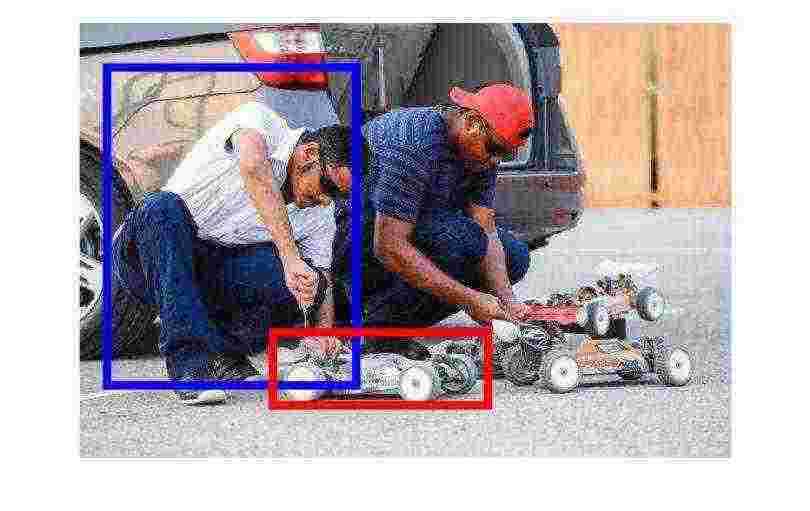}\\
       \vspace{1.5ex}
    \end{minipage}
    \hspace{0.005\textwidth}
    \begin{minipage}[t]{0.18\textwidth}
    	\centering
       	\includegraphics[trim={3cm 2cm 3.7cm 0.9cm},clip,width=0.95\linewidth,cfbox={red 2pt 2pt}]{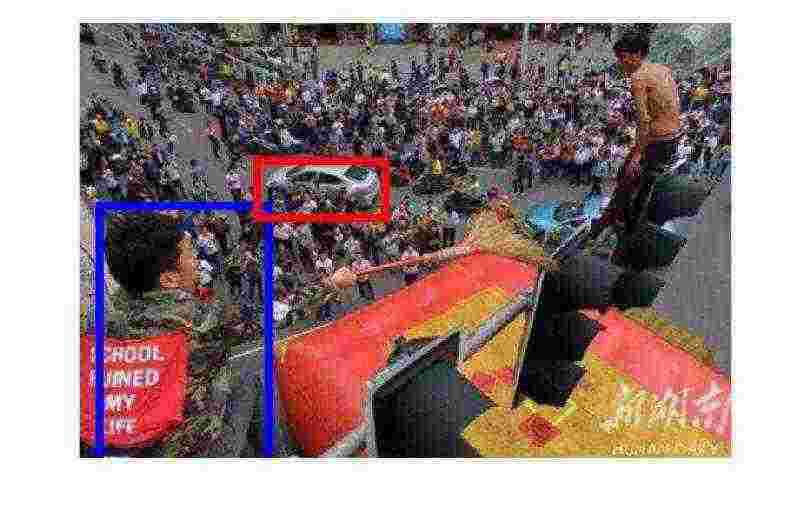}\\
      	\vspace{1.5ex}
    \end{minipage} 
    \hspace{0.005\textwidth}
    \begin{minipage}[t]{0.18\textwidth}
    	\centering
       	\includegraphics[trim={3cm 2cm 3cm 1cm},clip,width=0.95\linewidth,cfbox={blue 2pt 2pt}]{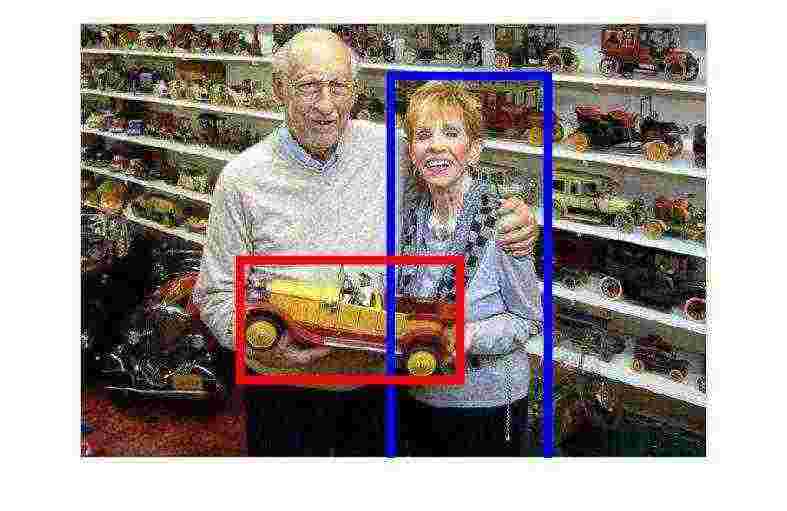}\\
       	\vspace{1.5ex}
    \end{minipage}

	\begin{minipage}[b]{0.005\textwidth}
    	\centering
    	\begin{turn}{90}
    \small{{\color{blue}person} ride {\color{red}giraffe}}
    	\end{turn}	
    \vspace{-1.2ex}
    \end{minipage}
    \hspace{0.01\textwidth}
    \begin{minipage}[t]{0.18\textwidth}
    	\centering
       	\includegraphics[trim={4.6cm 0cm 4cm 0cm},clip,width=0.95\linewidth,cfbox={green 2pt 2pt}]{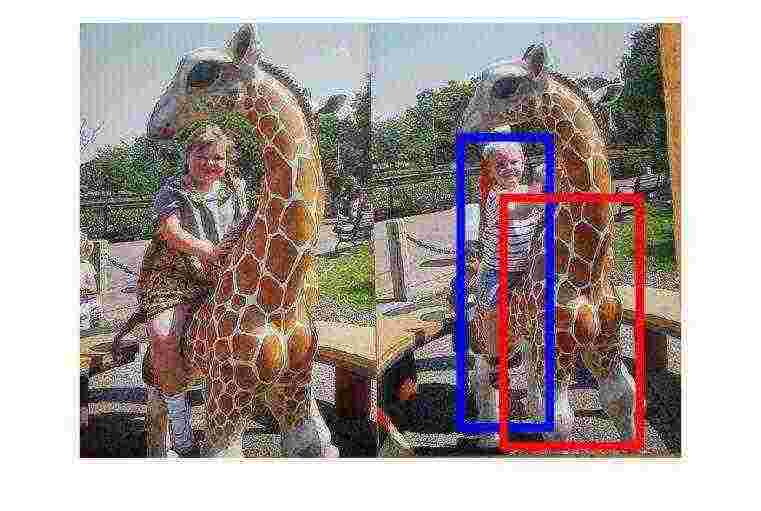}\\
       	\vspace{1.5ex}
    \end{minipage}
    \hspace{0.005\textwidth}
    \begin{minipage}[t]{0.18\textwidth}
    	\centering
       	\includegraphics[trim={3cm 0cm 4.2cm 0cm},clip,width=0.95\linewidth,cfbox={green 2pt 2pt}]{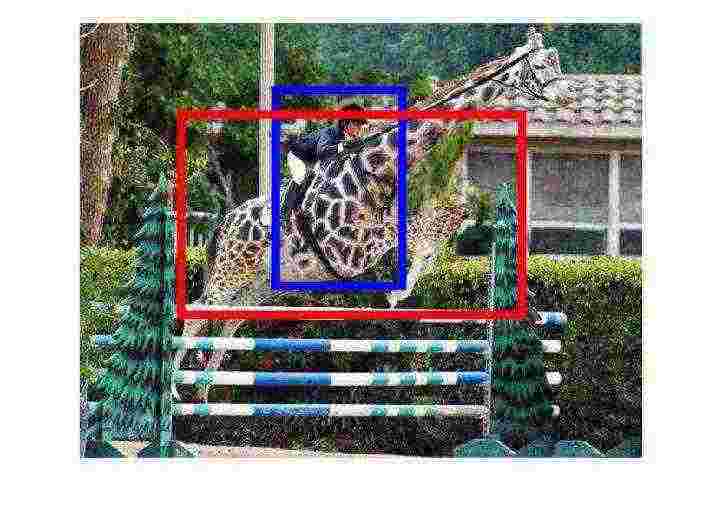}\\
       	\vspace{1.5ex}
    \end{minipage}
    \hspace{0.005\textwidth}
    \begin{minipage}[t]{0.18\textwidth}
       \centering
       \includegraphics[trim={4cm 0cm 4.6cm 0cm},clip,width=0.95\linewidth,cfbox={green 2pt 2pt}]{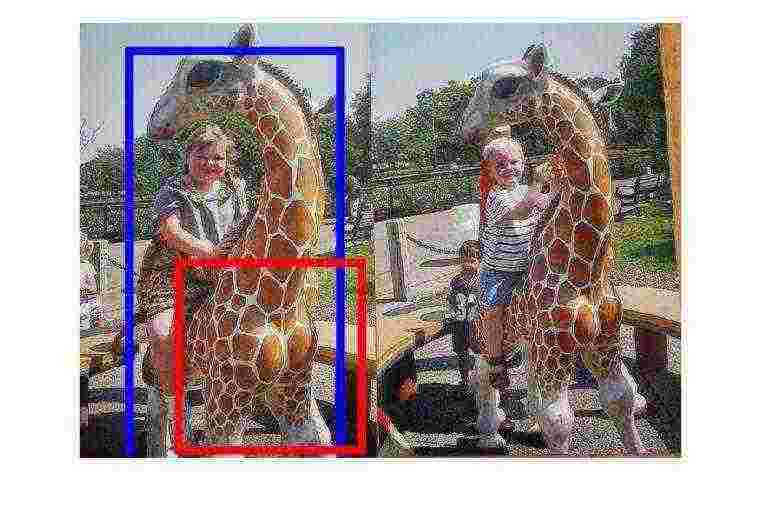}\\
       \vspace{1.5ex}
    \end{minipage}
    \hspace{0.005\textwidth}  
    \begin{minipage}[t]{0.18\textwidth}
    	\centering
       	\includegraphics[trim={0.7cm 2cm 0.7cm 0.6cm},clip,width=0.95\linewidth,cfbox={red 2pt 2pt}]{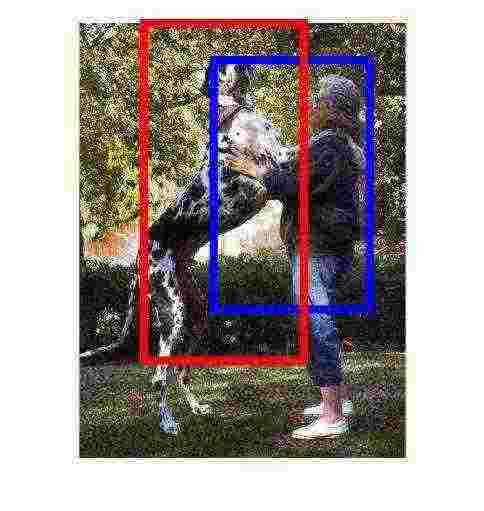}\\
      	\vspace{1.5ex}
    \end{minipage}  
    \hspace{0.005\textwidth}
    \begin{minipage}[t]{0.18\textwidth}
    	\centering
       	\includegraphics[trim={0cm 1.8cm 0cm 0.7cm},clip,width=0.95\linewidth,cfbox={blue 2pt 2pt}]{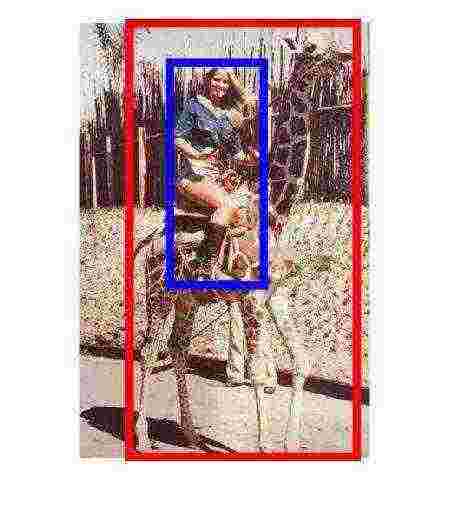}\\
       	\vspace{1.5ex}
    \end{minipage}

	\begin{minipage}[b]{0.005\textwidth}
    	\centering
    	\begin{turn}{90}
    	\small{{\color{blue}person} stand on {\color{red}bench}}
    	\end{turn}	
    \vspace{-3.5ex}
    \end{minipage}
    \hspace{0.01\textwidth}
    \begin{minipage}[t]{0.18\textwidth}
    	\centering
       	\includegraphics[trim={3cm 0.2cm 3cm 0cm},clip,width=0.95\linewidth,cfbox={green 2pt 2pt}]{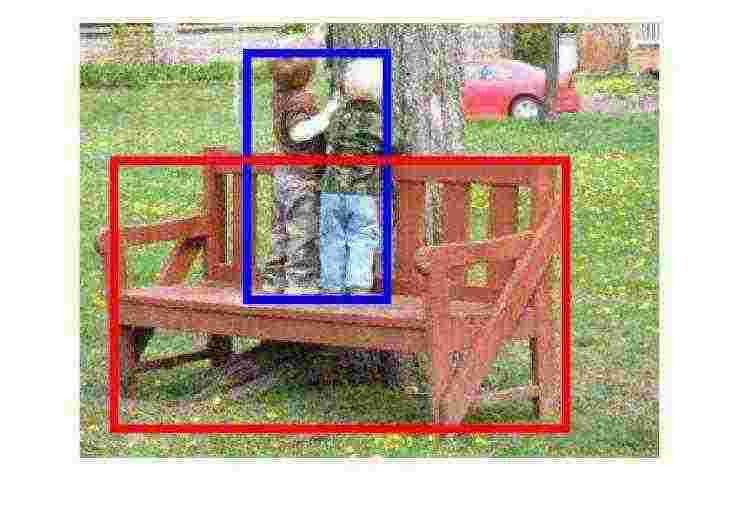}\\
       	\vspace{1.5ex}
    \end{minipage}
    \hspace{0.005\textwidth}  
    \begin{minipage}[t]{0.18\textwidth}
    	\centering
       	\includegraphics[trim={1.5cm 3.8cm 1.5cm 4cm},clip,width=0.95\linewidth,cfbox={green 2pt 2pt}]{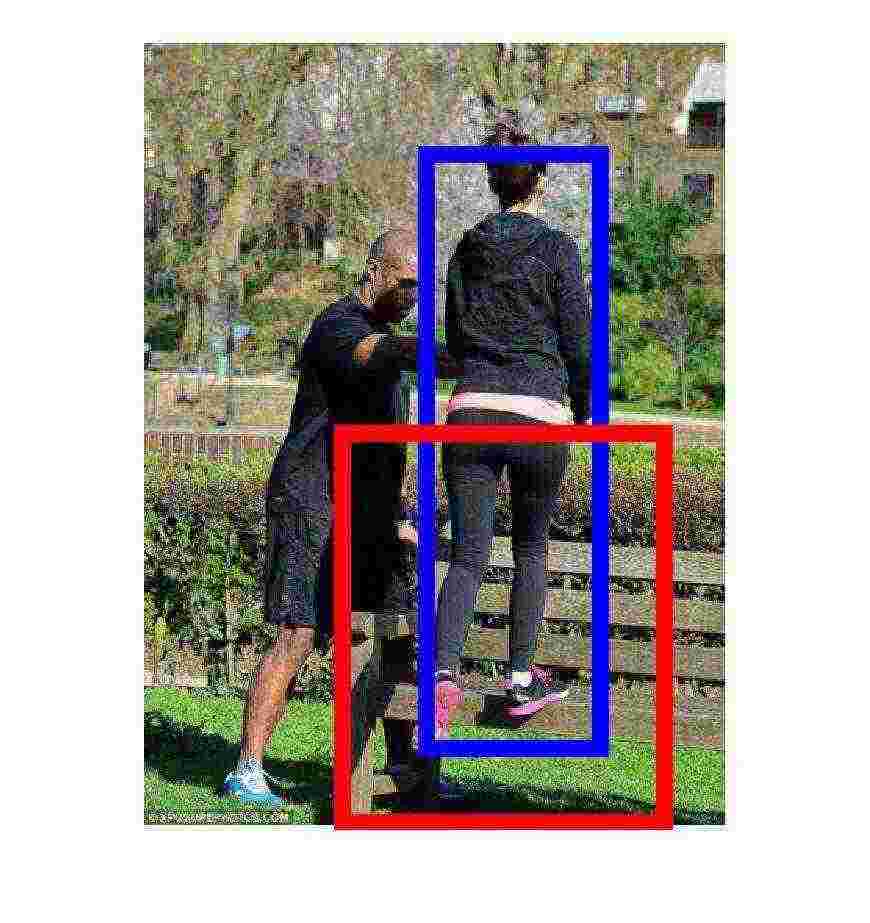}\\
       	\vspace{1.5ex}
    \end{minipage}
    \hspace{0.005\textwidth}
    \begin{minipage}[t]{0.18\textwidth}
       \centering
       \includegraphics[trim={0cm 1.8cm 0cm 2cm},clip,width=0.95\linewidth,cfbox={green 2pt 2pt}]{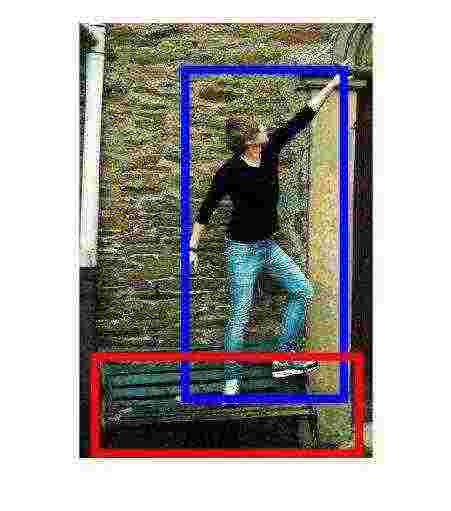}\\
       \vspace{1.5ex}
    \end{minipage}
    \hspace{0.005\textwidth}
    \begin{minipage}[t]{0.18\textwidth}
    	\centering
       	\includegraphics[trim={3cm 0cm 5.3cm 0cm},clip,width=0.95\linewidth,cfbox={red 2pt 2pt}]{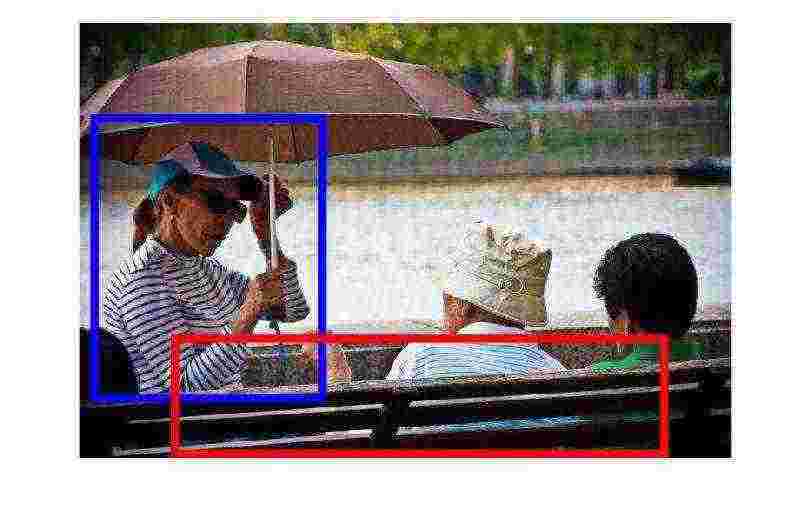}\\
      	\vspace{1.5ex}
    \end{minipage} 
    \hspace{0.005\textwidth}
    \begin{minipage}[t]{0.18\textwidth}
    	\centering
       	\includegraphics[trim={0.6cm 1.8cm 0.5cm 2cm},clip,width=0.95\linewidth,cfbox={blue 2pt 2pt}]{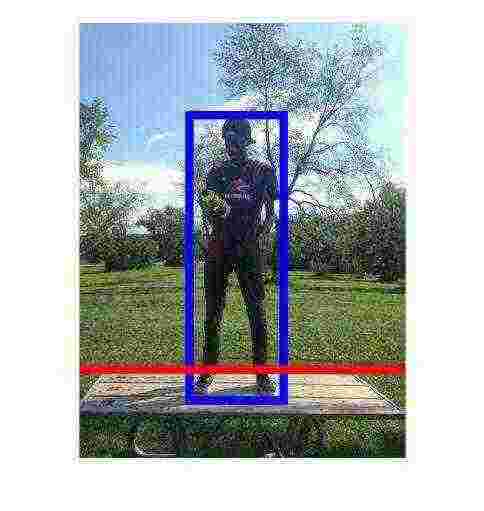}\\
       	\vspace{1.5ex}
    \end{minipage}

	\begin{minipage}[b]{0.005\textwidth}
    	\centering
    	\begin{turn}{90}
    \small{{\color{blue}person} inside {\color{red}tree}}
    	\end{turn}	
    \vspace{-1.5ex}
    \end{minipage}
    \hspace{0.01\textwidth}
    \begin{minipage}[t]{0.18\textwidth}
    	\centering
    		\includegraphics[trim={5cm 0.25cm 7cm 0cm},clip,width=0.95\linewidth,cfbox={green 2pt 2pt}]{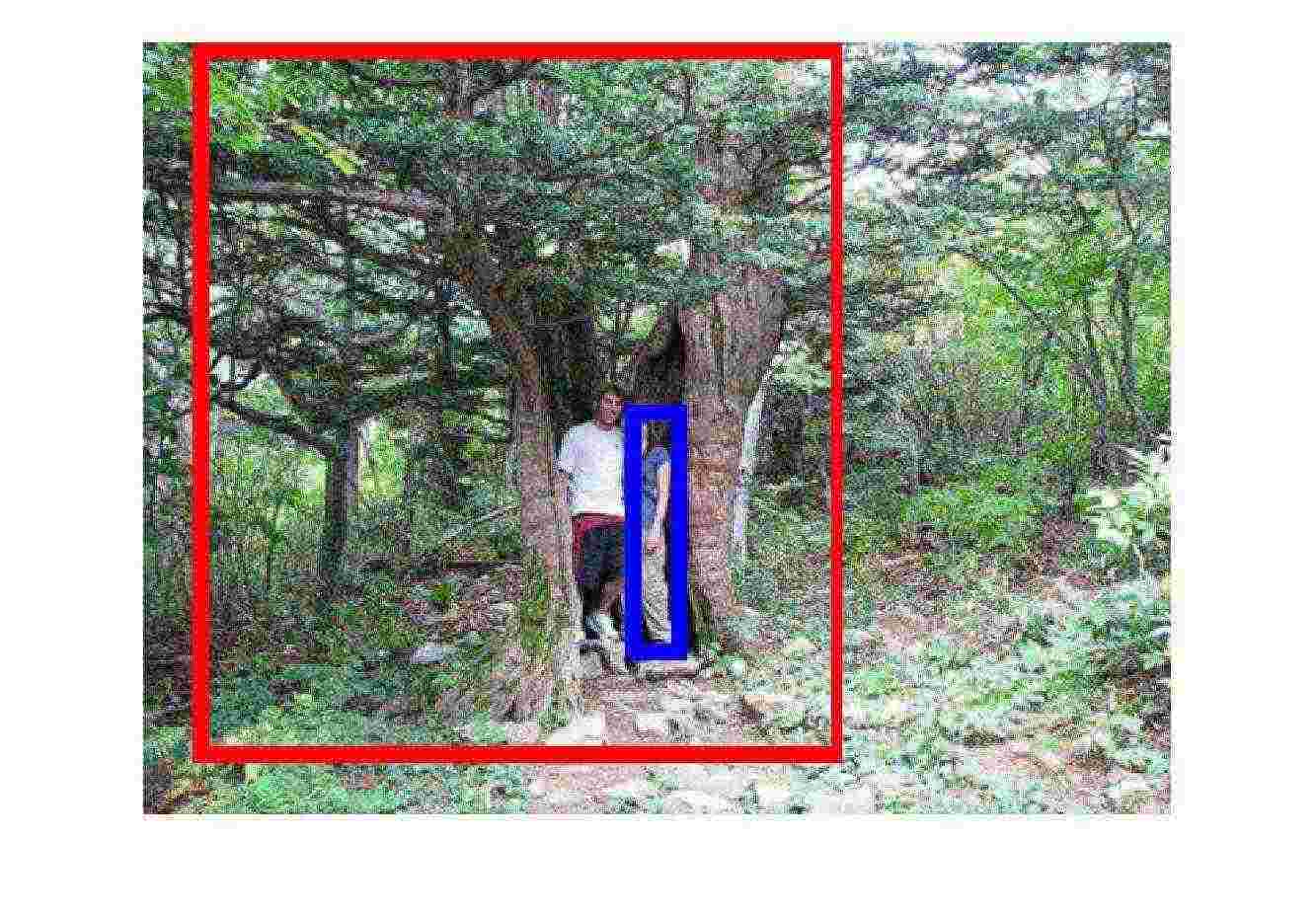}\\
       	\vspace{1.5ex}
    \end{minipage}
    \hspace{0.005\textwidth}  
    \begin{minipage}[t]{0.18\textwidth}
    	\centering
       	\includegraphics[trim={3.5cm 0cm 3cm 0cm},clip,width=0.95\linewidth,cfbox={green 2pt 2pt}]{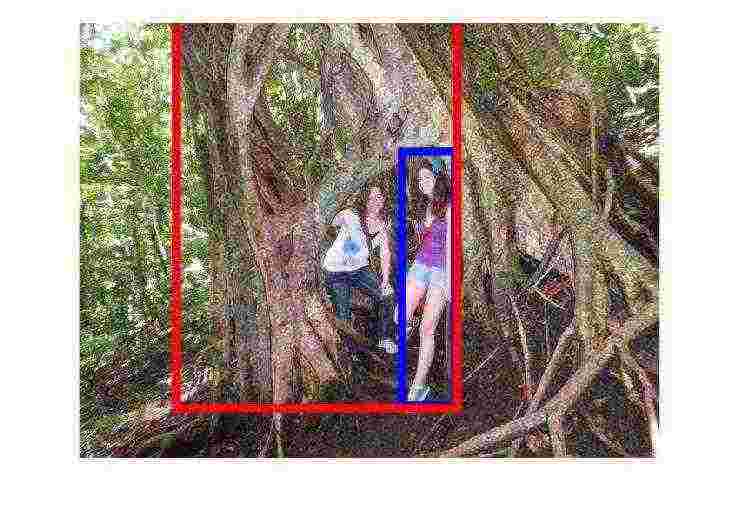}\\
       	\vspace{1.5ex}
    \end{minipage}
    \hspace{0.005\textwidth}
    \begin{minipage}[t]{0.18\textwidth}
       \centering
       \includegraphics[trim={3.5cm 0cm 3cm 0cm},clip,width=0.95\linewidth,cfbox={green 2pt 2pt}]{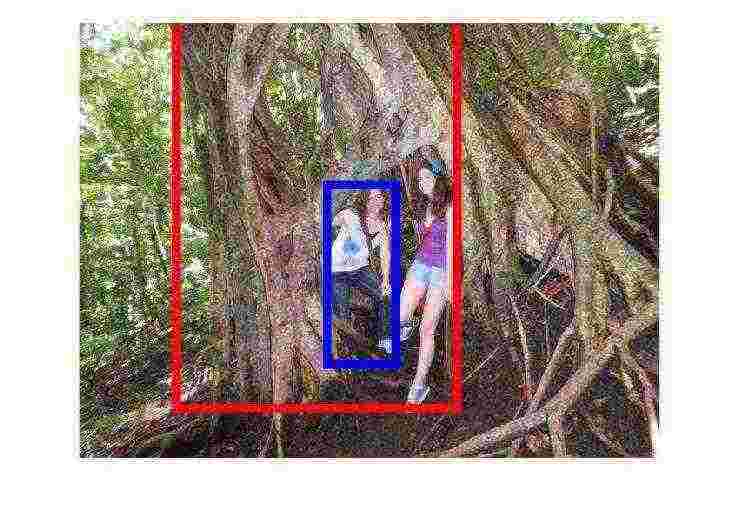}\\
       \vspace{1.5ex}
    \end{minipage}
    \hspace{0.005\textwidth}
    \begin{minipage}[t]{0.18\textwidth}
    	\centering
       	\includegraphics[trim={5cm 2cm 11.2cm 2cm},clip,width=0.95\linewidth,cfbox={red 2pt 2pt}]{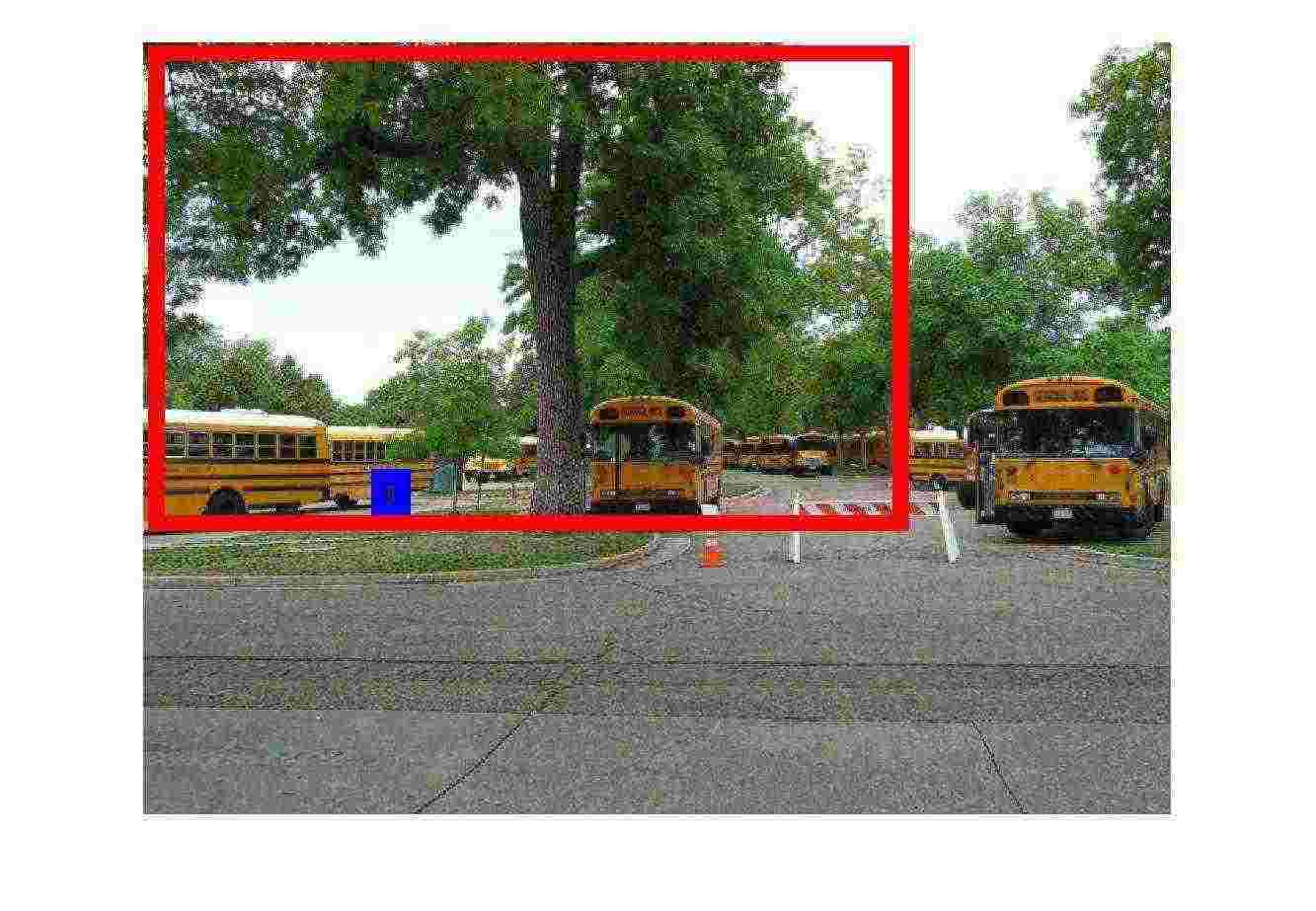}\\
      	\vspace{1.5ex}
    \end{minipage} 
    \hspace{0.005\textwidth}
    \begin{minipage}[t]{0.18\textwidth}
    	\centering
       	\includegraphics[trim={5.3cm 0cm 4cm 0cm},clip,width=0.95\linewidth,cfbox={blue 2pt 2pt}]{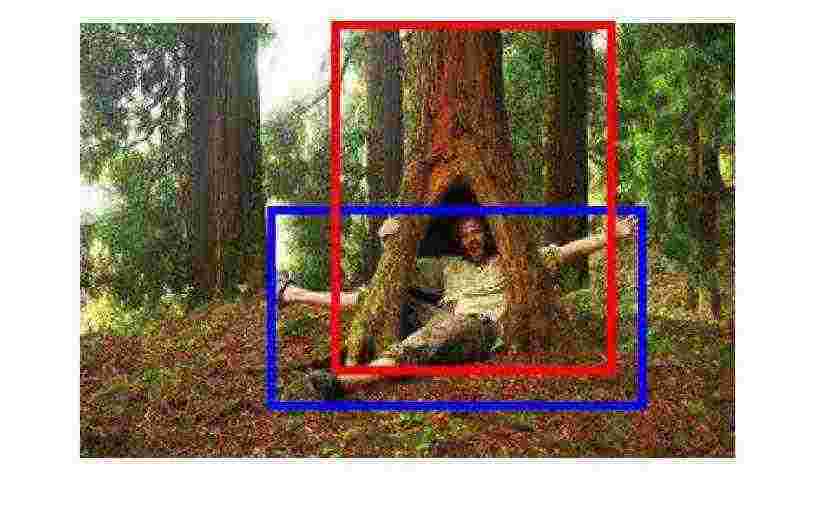}\\
       	\vspace{1.5ex}
    \end{minipage}
       
	\begin{minipage}[b]{0.005\textwidth}
    	\centering
    	\begin{turn}{90}
    \small{{\color{blue}elephant} sleep on {\color{red}person}}
    	\end{turn}	
    \vspace{-5ex}
    \end{minipage}
    \hspace{0.01\textwidth}
    \begin{minipage}[t]{0.18\textwidth}
    	\centering
       	\includegraphics[trim={4cm 2.6cm 4.4cm 0.7cm},clip,width=0.95\linewidth,cfbox={green 2pt 2pt}]{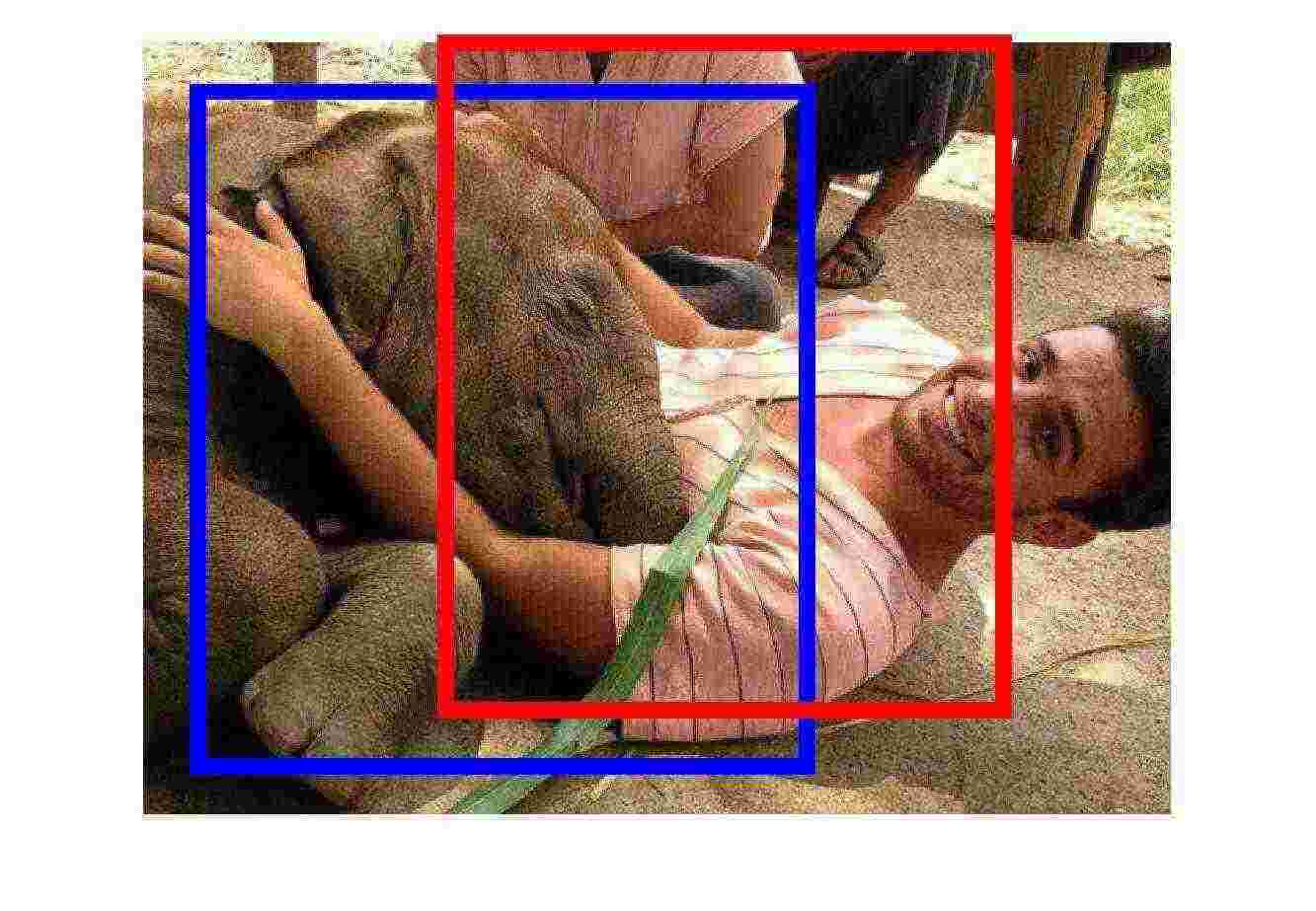}\\
       	\vspace{1.5ex}
    \end{minipage}
    \hspace{0.005\textwidth}
    \begin{minipage}[t]{0.18\textwidth}
       \centering
       \includegraphics[trim={3cm 2.6cm 3cm 1cm},clip,width=0.95\linewidth,cfbox={green 2pt 2pt}]{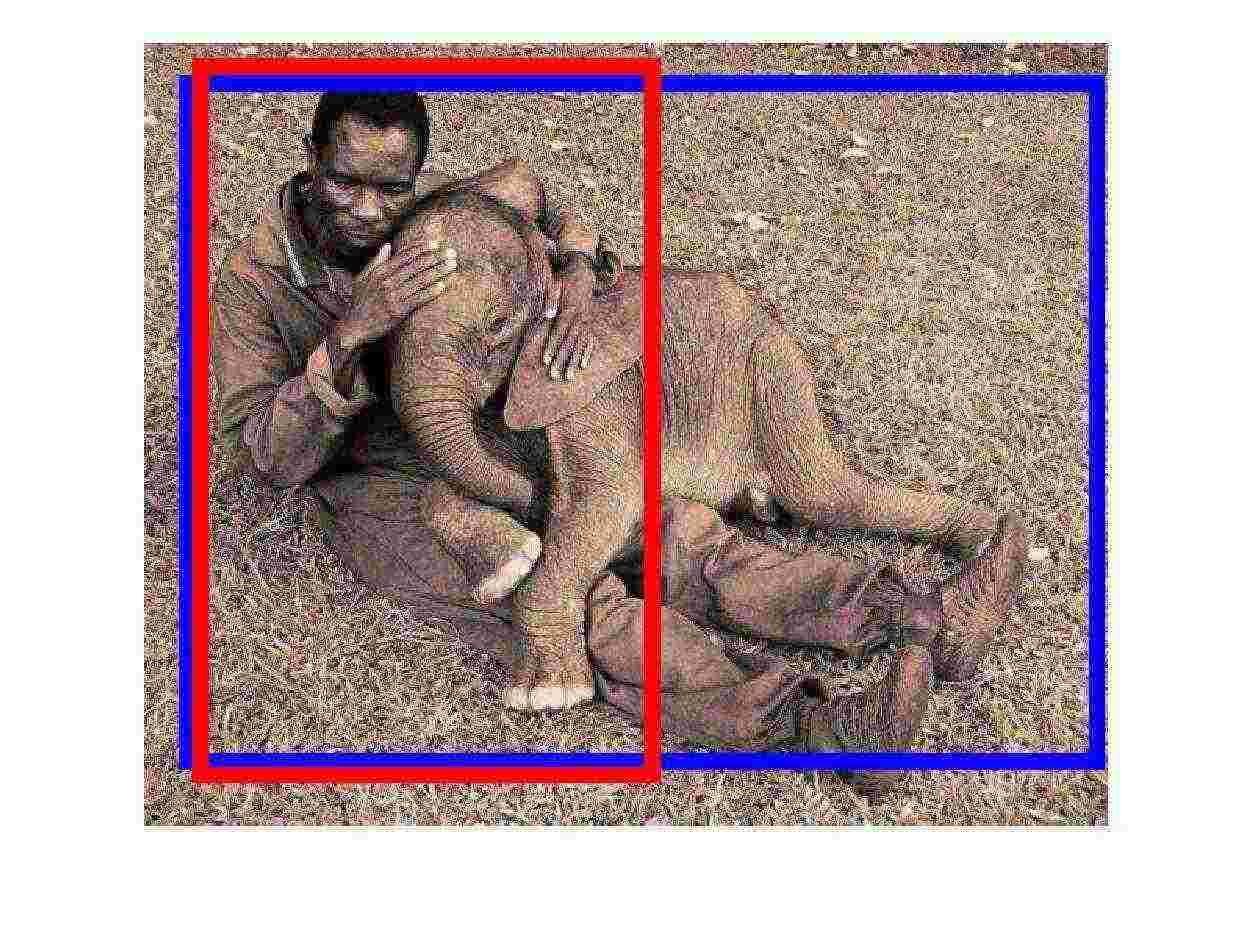}\\
       \vspace{1.5ex}
    \end{minipage}
    \hspace{0.005\textwidth}
    \begin{minipage}[t]{0.18\textwidth}
    	\centering
       	\includegraphics[trim={0.9cm 4cm 0.9cm 1.2cm},clip,width=0.95\linewidth,cfbox={green 2pt 2pt}]{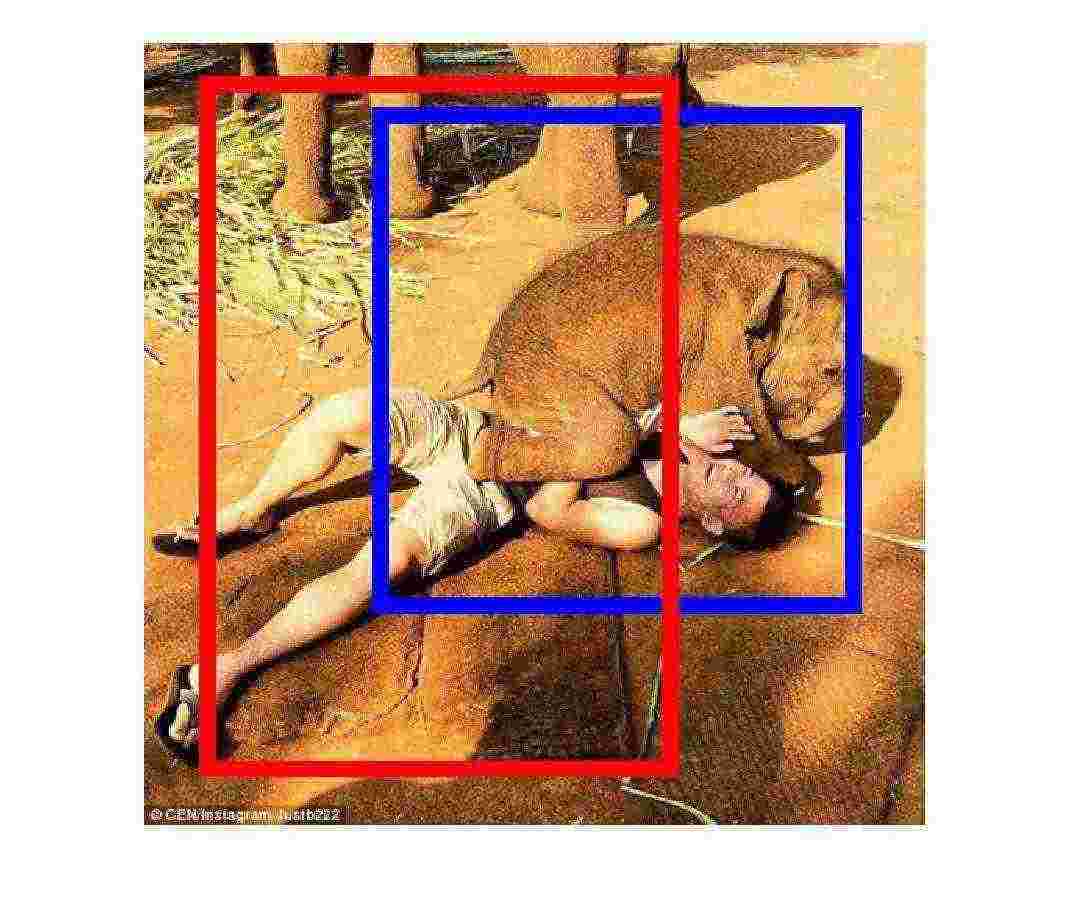}\\
       	\vspace{1.5ex}
    \end{minipage}
   	\hspace{0.005\textwidth}  
    \begin{minipage}[t]{0.18\textwidth}
    	\centering
       	\includegraphics[trim={4cm 2.6cm 4.3cm 0.7cm},clip,width=0.95\linewidth,cfbox={red 2pt 2pt}]{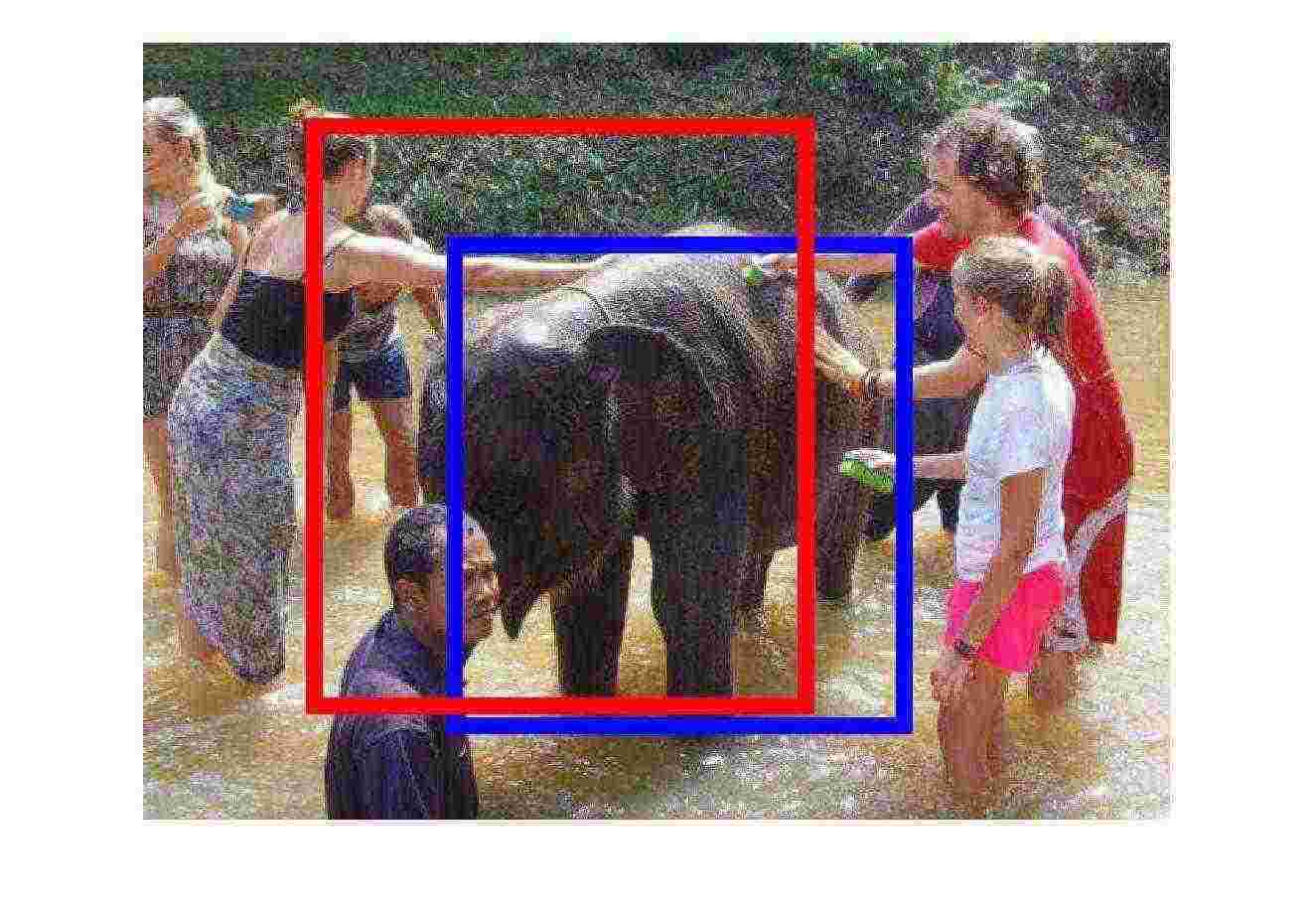}\\
      	\vspace{1.5ex}
    \end{minipage} 
    \hspace{0.005\textwidth}
    \begin{minipage}[t]{0.18\textwidth}
    	\centering
       	\includegraphics[trim={5cm 0cm 15cm 0cm},clip,width=0.95\linewidth,cfbox={blue 2pt 2pt}]{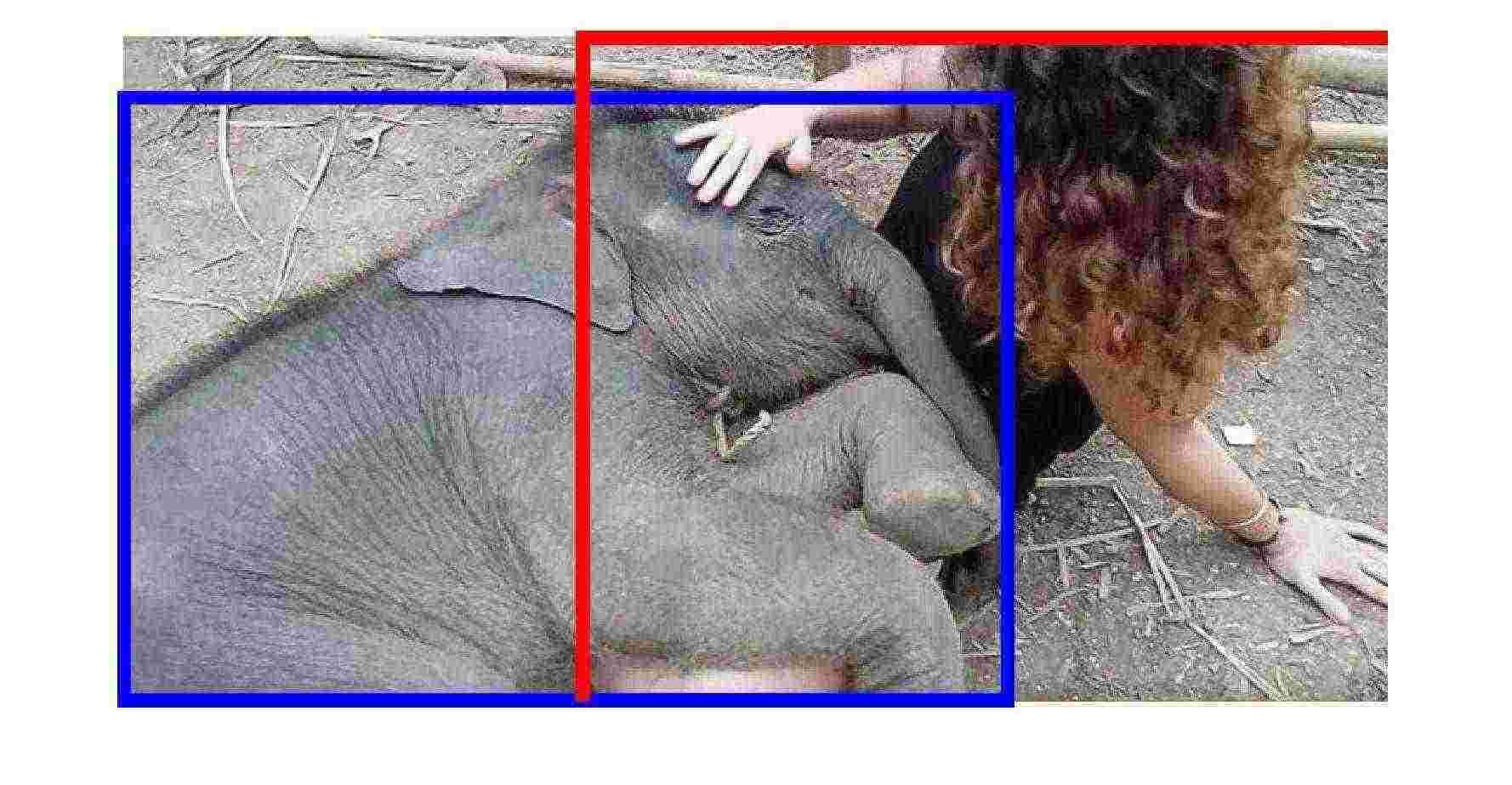}\\
       	\vspace{1.5ex}
    \end{minipage}

    \setlength\abovecaptionskip{1pt}
    \caption{Examples of retrieved results for triplet queries on UnRel. Each row corresponds to one triplet query and displays true positives among the top-retrieved pairs (green), top-scored incorrect pairs (red) and missed detections (blue). These retrieval results were computed using our weakly-supervised method with candidate proposals. A pair of boxes is considered as positive if both subject and object candidates overlap with the corresponding subject and object ground truth with IoU $\ge 0.3$.}
    \label{fig:unrel_appendix}
\end{figure*}

In Figure~\ref{fig:unrel_appendix} we show additional qualitative results for triplet retrieval on the UnRel dataset using our weakly-supervised model. Each line corresponds to one unusual triplet query and we plot examples of top-scoring retrieved pairs of boxes (green), top-scoring incorrect pairs (red) and missed detections (blue). The false positives (red) are either due to incorrect object detection/localization (e.g.~the dog in ``person ride giraffe" is confused with a giraffe) or failures of the relation classifier in challenging configurations (e.g.~``person hold car'', ``person stand on bench'').
The missed detections are often due to the failure of the object detector, which is by itself challenging, as the UnRel dataset contains images of objects in unusual contexts (e.g.~``dog ride bike'').

\section{Qualitative results for Visual Relationship Detection}
\label{part:vrd}

\begin{figure*}[t]
\centering
	\begin{minipage}[b]{0.6\textwidth}
    \centering
    	\textit{correctly recognized relations}\\
    	\vspace{0.4ex}
	\end{minipage}
	\hspace{0.005\textwidth}
	\begin{minipage}[b]{0.18\textwidth}
    \centering
    	\textit{missing / ambiguous}\\
    	\vspace{0.4ex}
	\end{minipage}
	\hspace{0.005\textwidth}
	\begin{minipage}[b]{0.19\textwidth}
    \centering
    	\textit{incorrectly recognized}\\
    	\vspace{0.4ex}
	\end{minipage}
	
	\begin{minipage}[t]{0.005\textwidth}
    	\centering
    	\vspace{-8.5ex}
    	\begin{turn}{90}
    	ride
    	\end{turn}
    	\vspace{2ex}
    \end{minipage}
    \hspace{0.01\textwidth}
    \begin{minipage}[t]{0.18\textwidth}
    	\centering
       	\includegraphics[trim={1cm 1cm 0.5cm 0cm},clip,width=0.95\linewidth,cfbox={green 2pt 2pt}]{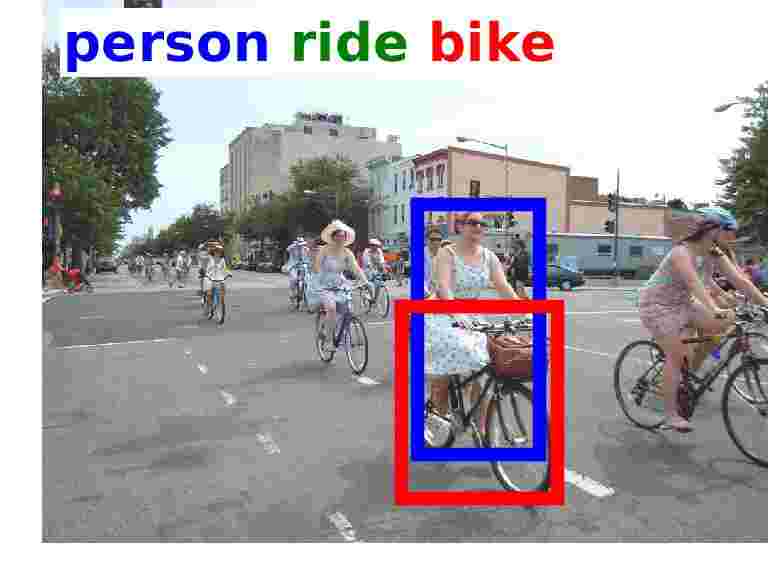}\\
       	\vspace{0.3ex}
       	GT: on, ride, above,\\
       	with, has
       	\vspace{2ex}
    \end{minipage}
    \hspace{0.005\textwidth}
    \begin{minipage}[t]{0.18\textwidth}
    	\centering
       	\includegraphics[trim={1.5cm 1cm 3cm 0cm},clip,width=0.95\linewidth,cfbox={green 2pt 2pt}]{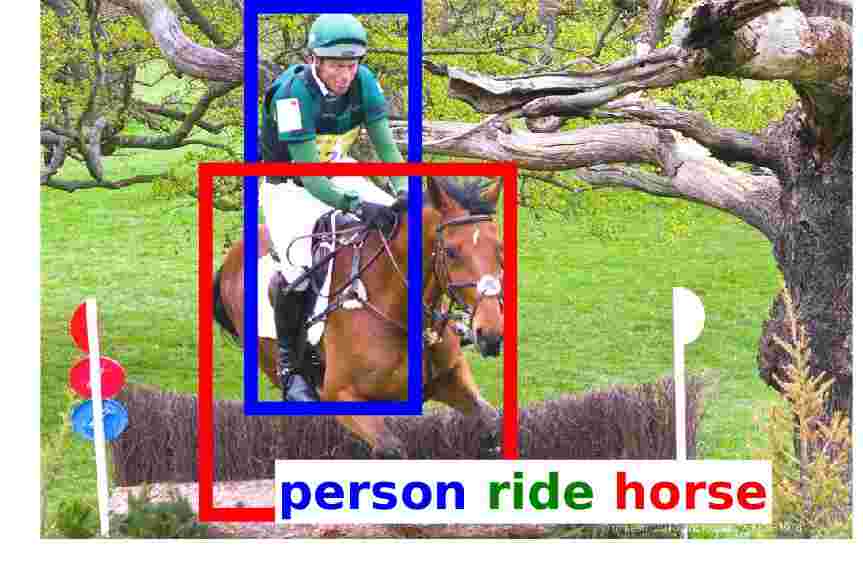}\\
       	\vspace{0.3ex}
       	GT: ride, above
       	\vspace{0.2ex}
    \end{minipage}
    \hspace{0.005\textwidth}
    \begin{minipage}[t]{0.18\textwidth}
    	\centering
       	\includegraphics[trim={9.5cm 1cm 0cm 0cm},clip,width=0.95\linewidth,cfbox={green 2pt 2pt}]{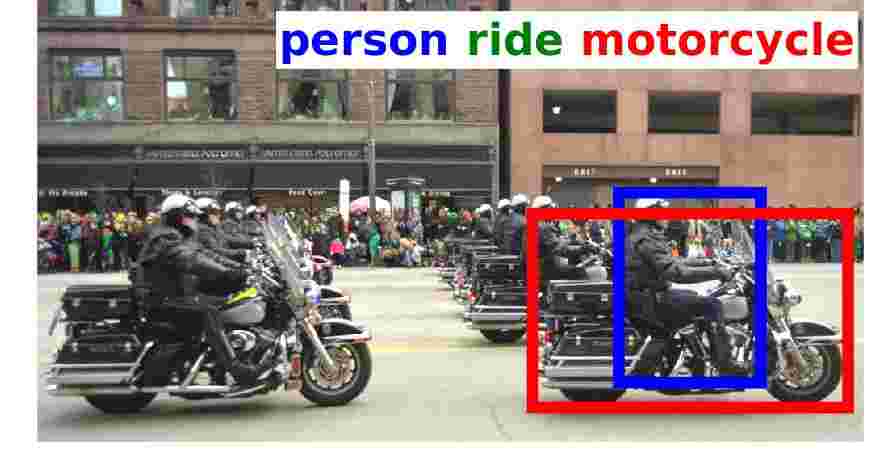}\\
       	\vspace{0.3ex}
       	GT: on, ride
       	\vspace{0.2ex}
    \end{minipage}
    \hspace{0.005\textwidth}  
    \begin{minipage}[t]{0.18\textwidth}
    	\centering
       	\includegraphics[trim={2cm 1cm 6.5cm 0cm},clip,width=0.95\linewidth,cfbox={yellow 2pt 2pt}]{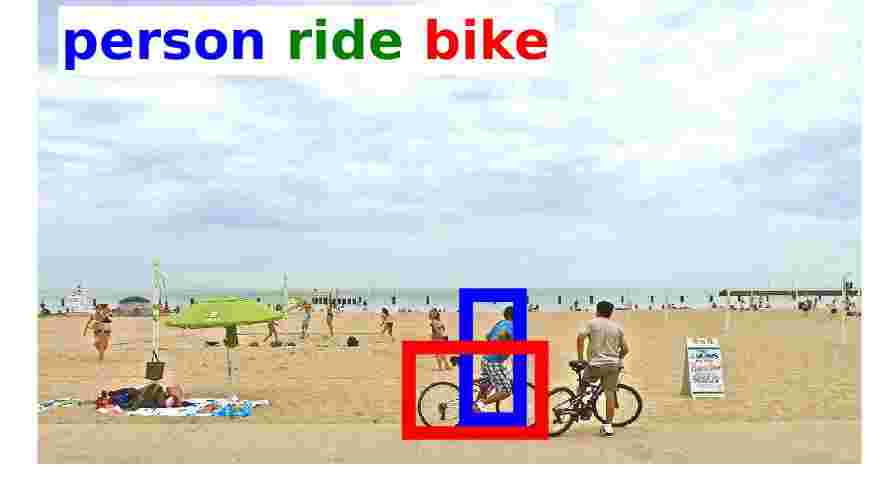}\\
       	\vspace{0.3ex}
       	GT: on, sit on
       	\vspace{0.2ex}
    \end{minipage}
    \hspace{0.005\textwidth}  
    \begin{minipage}[t]{0.18\textwidth}
    	\centering
       	\includegraphics[trim={1cm 1cm 6.5cm 0cm},clip,width=0.95\linewidth,cfbox={red 2pt 2pt}]{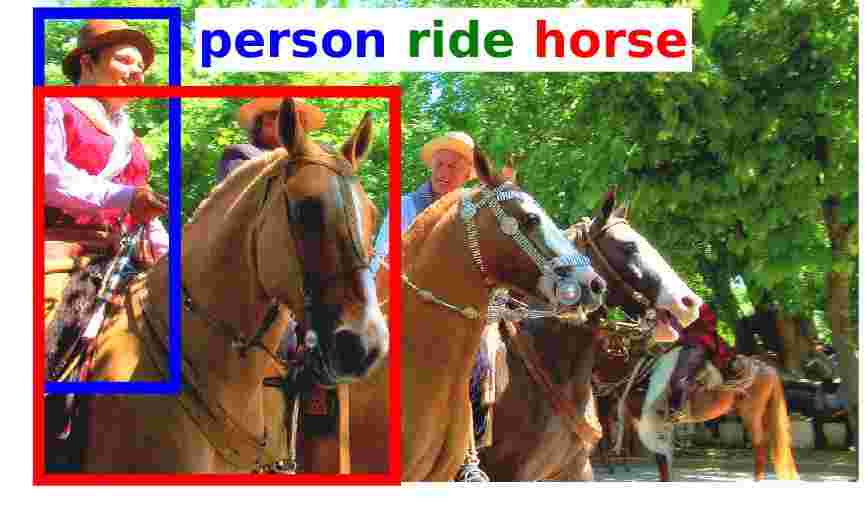}\\
       	\vspace{0.3ex}
       	GT: on
      	\vspace{0.2ex}
    \end{minipage} 
    
	\begin{minipage}[t]{0.005\textwidth}
    	\centering
    	\vspace{-9ex}
    	\begin{turn}{90}
    	carry
    	\end{turn}
    	\vspace{2ex}
    \end{minipage}
    \hspace{0.01\textwidth}
	\begin{minipage}[t]{0.18\textwidth}
    	\centering
       	\includegraphics[trim={0.8cm 0.8cm 1cm 0cm},clip,width=0.95\linewidth,cfbox={green 2pt 2pt}]{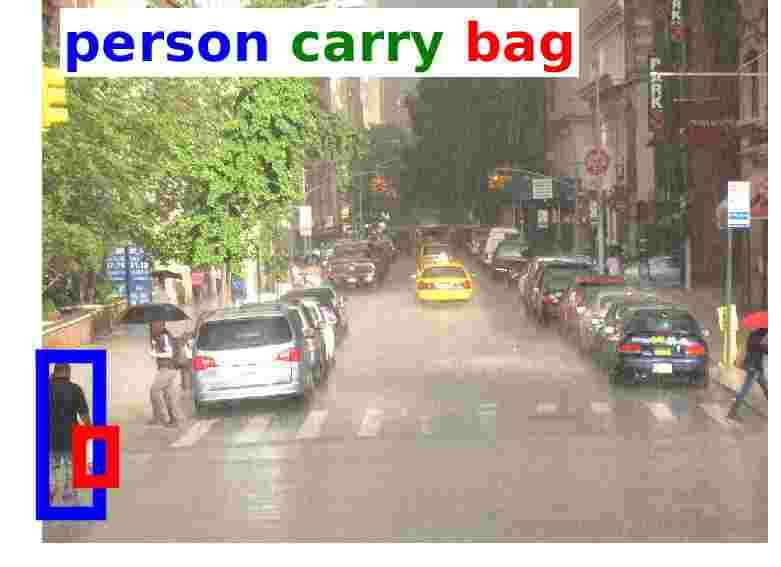}\\
       	\vspace{0.3ex}
       	GT: hold, carry
       	\vspace{2ex}
    \end{minipage}
    \hspace{0.005\textwidth}
    \begin{minipage}[t]{0.18\textwidth}
    	\centering
       	\includegraphics[trim={1cm 0.5cm 0.5cm 0cm},clip,width=0.95\linewidth,cfbox={green 2pt 2pt}]{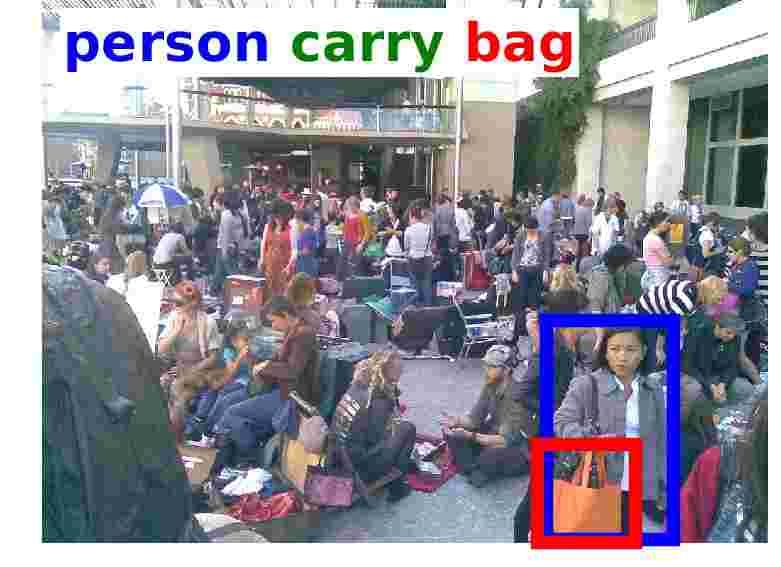}\\
       	\vspace{0.3ex}
       	GT: has, carry, hold
       	\vspace{0.2ex}
    \end{minipage}
    \hspace{0.005\textwidth}
	\begin{minipage}[t]{0.18\textwidth}
    	\centering
       	\includegraphics[trim={1.3cm 0.5cm 2.8cm 0cm},clip,width=0.95\linewidth,cfbox={green 2pt 2pt}]{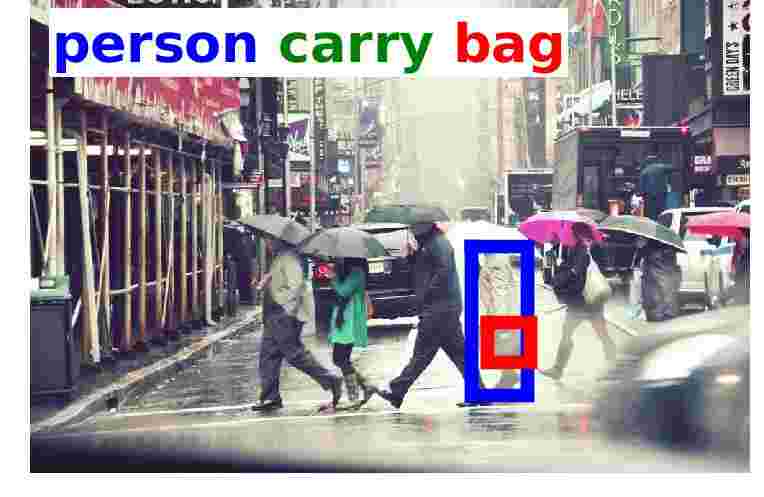}\\
       	\vspace{0.3ex}
       	GT: carry
       	\vspace{0.2ex}
    \end{minipage}
    \hspace{0.005\textwidth}
    \begin{minipage}[t]{0.18\textwidth}
    	\centering
       	\includegraphics[trim={1.2cm 2cm 5.4cm 0cm},clip,width=0.95\linewidth,cfbox={yellow 2pt 2pt}]{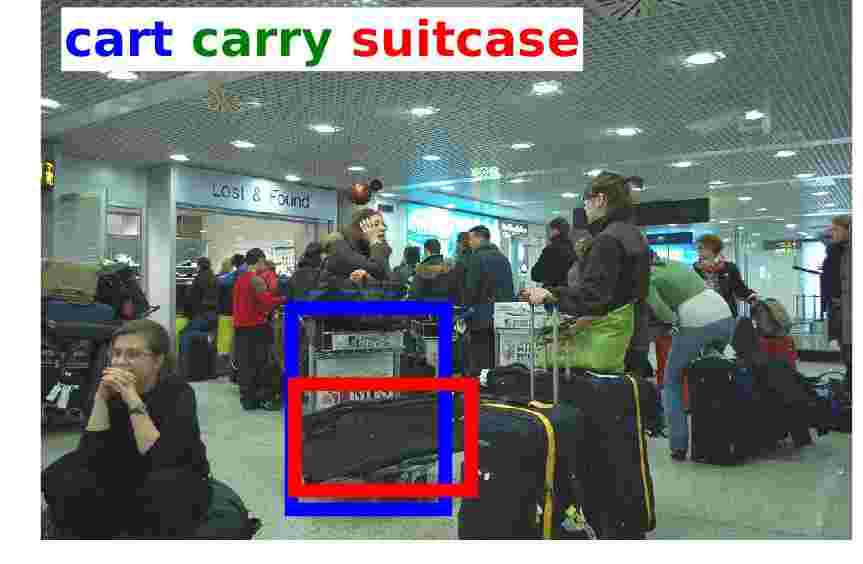}\\
       	\vspace{0.3ex}
       	GT: hold
       	\vspace{0.2ex}
   	\end{minipage}
    \hspace{0.005\textwidth}
    \begin{minipage}[t]{0.18\textwidth}
    	\centering
       	\includegraphics[trim={4.5cm 1.9cm 2cm 0.2cm},clip,width=0.95\linewidth,cfbox={red 2pt 2pt}]{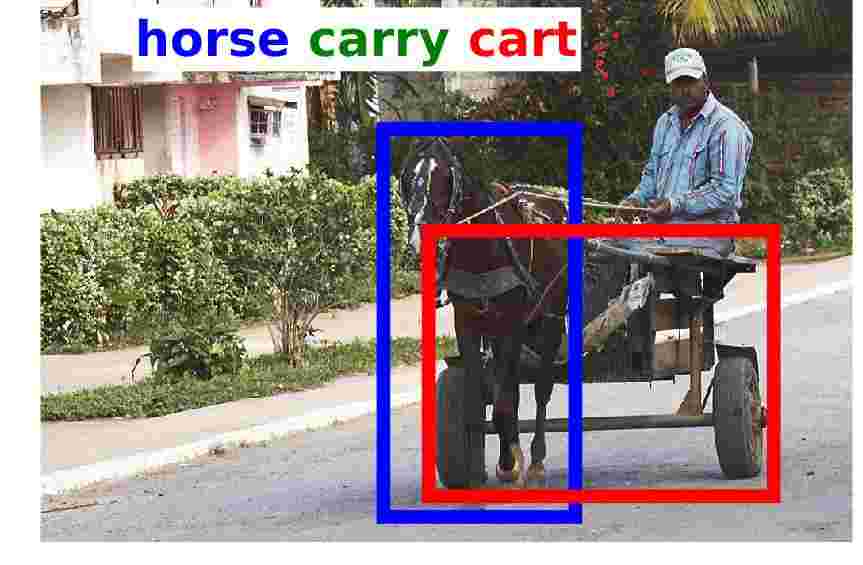}\\
       	\vspace{0.3ex}
       	GT: pull
       	\vspace{0.2ex}
    \end{minipage}   

	\begin{minipage}[t]{0.005\textwidth}
    	\centering
    	\vspace{-9.5ex}
    	\begin{turn}{90}
    	hold
    	\end{turn}
    	\vspace{2ex}
    \end{minipage}
    \hspace{0.01\textwidth}
	\begin{minipage}[t]{0.18\textwidth}
    	\centering
       	\includegraphics[trim={2cm 6.4cm 0.5cm 1.5cm},clip,width=0.95\linewidth,cfbox={green 2pt 2pt}]{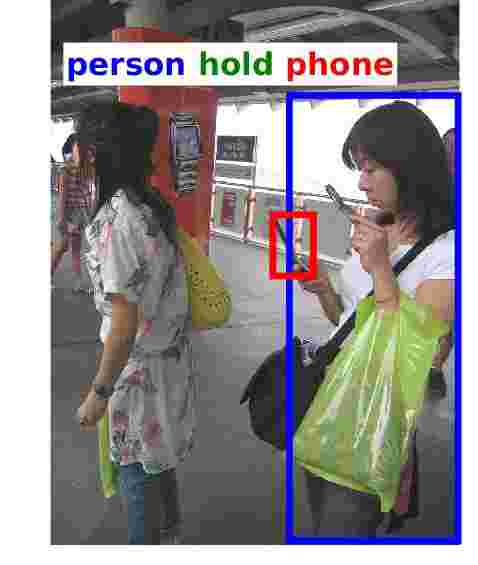}\\
       	\vspace{0.3ex}
       	GT: hold
       	\vspace{2ex}
    \end{minipage}
    \hspace{0.005\textwidth}
	\begin{minipage}[t]{0.18\textwidth}
    	\centering
       	\includegraphics[trim={2.5cm 0.5cm 2cm 0cm},clip,width=0.95\linewidth,cfbox={green 2pt 2pt}]{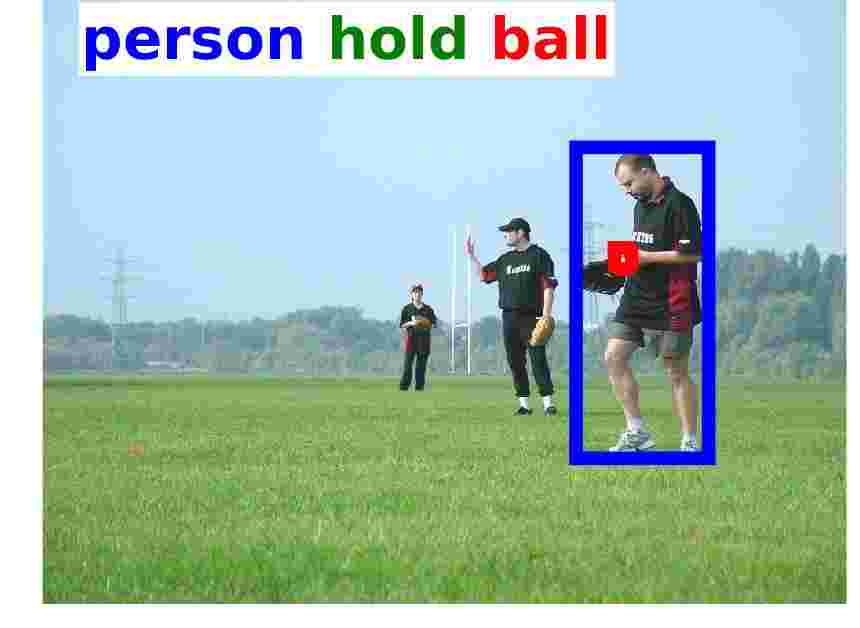}\\
       	\vspace{0.3ex}
       	GT: hold
       	\vspace{0.2ex}
    \end{minipage}
    \hspace{0.005\textwidth}
    \begin{minipage}[t]{0.18\textwidth}
    	\centering
      	\includegraphics[trim={2cm 1.1cm 1cm 0cm},clip,width=0.95\linewidth,cfbox={green 2pt 2pt}]{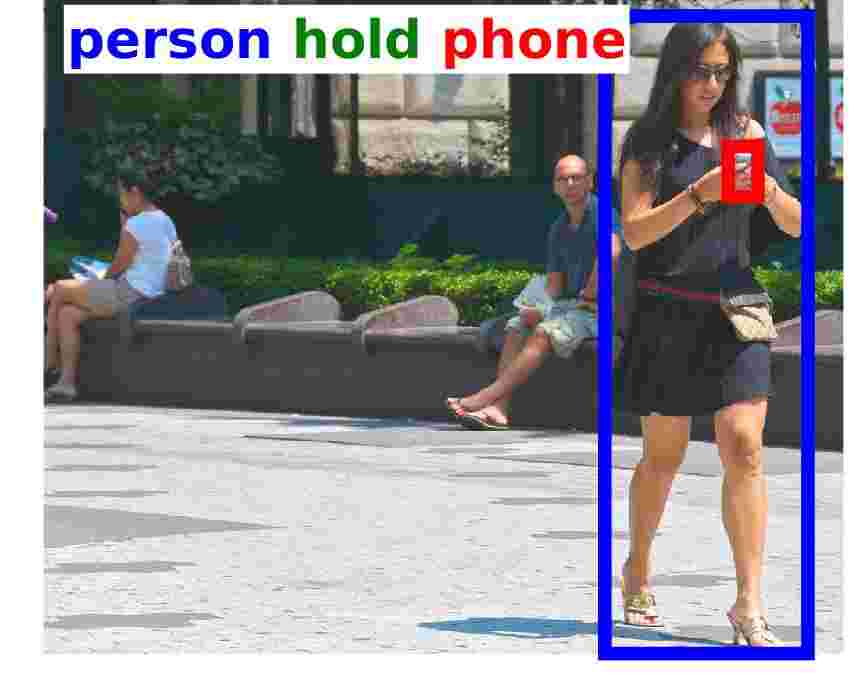}\\
      	\vspace{0.3ex}
       	GT: has, hold
       	\vspace{0.2ex}
    \end{minipage}
    \hspace{0.005\textwidth}
	\begin{minipage}[t]{0.18\textwidth}
    	\centering
       	\includegraphics[trim={2cm 0.7cm 6cm 0cm},clip,width=0.95\linewidth,cfbox={yellow 2pt 2pt}]{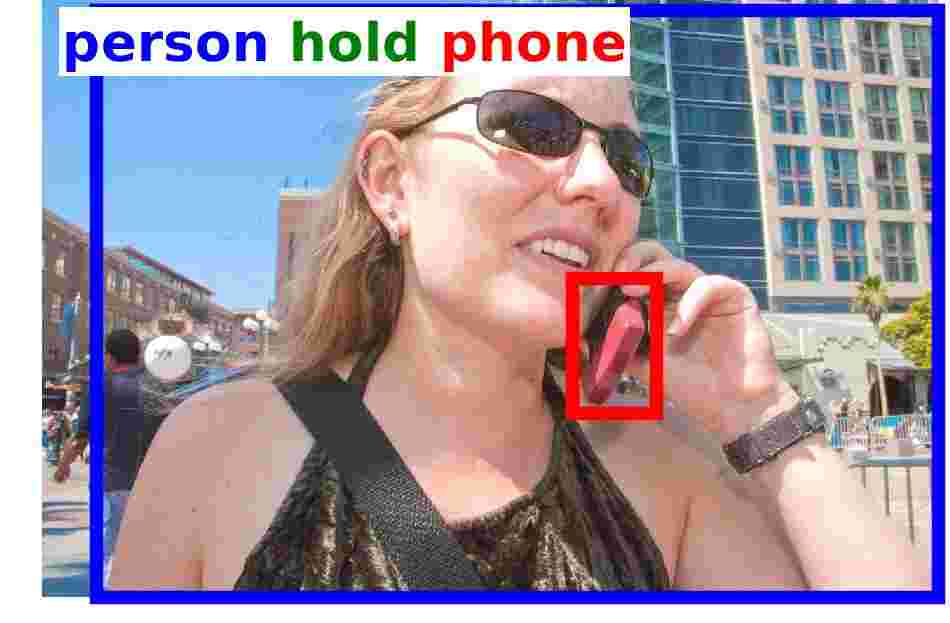}\\
       	\vspace{0.3ex}
       	GT: talk, use
       	\vspace{0.2ex}
    \end{minipage}
    \hspace{0.005\textwidth}
    \begin{minipage}[t]{0.18\textwidth}
    	\centering
       	\includegraphics[trim={1.5cm 1cm 2.6cm 0cm},clip,width=0.95\linewidth,cfbox={red 2pt 2pt}]{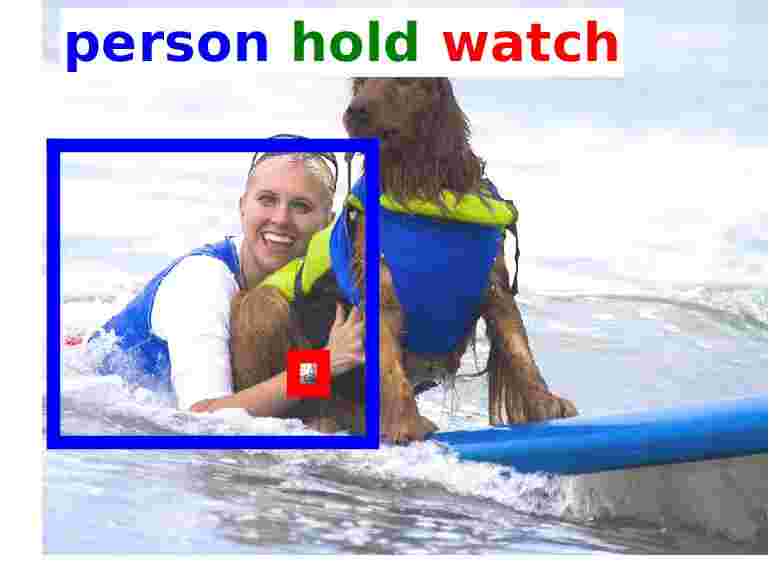} \\
       	\vspace{0.3ex}
       	GT: wear
       	\vspace{0.2ex}
    \end{minipage}
     	
	\begin{minipage}[t]{0.005\textwidth}
    	\centering
	\vspace{-8.5ex}
    	\begin{turn}{90}
    	drive
    	\end{turn}
    	\vspace{2ex}
   	\end{minipage}
    \hspace{0.01\textwidth}
    \begin{minipage}[t]{0.18\textwidth}
    	\centering
       	\includegraphics[trim={1cm 0.6cm 2cm 0cm},clip,width=0.95\linewidth,cfbox={green 2pt 2pt}]{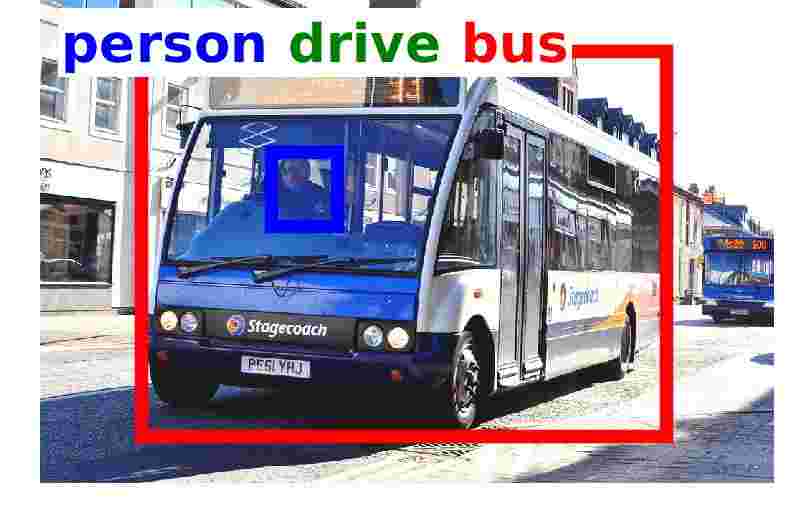}\\
       	\vspace{0.3ex}
       	GT: inside, drive, \\
       	sit on
       	\vspace{2ex}
    \end{minipage}
    \hspace{0.005\textwidth}
    \begin{minipage}[t]{0.18\textwidth}
    	\centering
       	\includegraphics[trim={1cm 1.3cm 0cm 0cm},clip,width=0.95\linewidth,cfbox={green 2pt 2pt}]{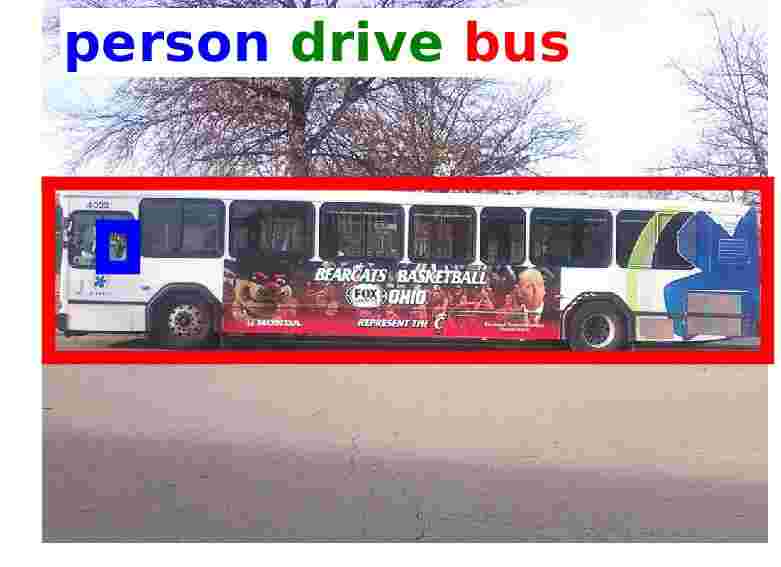}\\
       	\vspace{0.3ex}
       	GT: in, drive
       	\vspace{0.2ex}
    \end{minipage}  
    \hspace{0.005\textwidth}
    \begin{minipage}[t]{0.18\textwidth}
    	\centering
       	\includegraphics[trim={1.5cm 1cm 2cm 0cm},clip,width=0.95\linewidth,cfbox={green 2pt 2pt}]{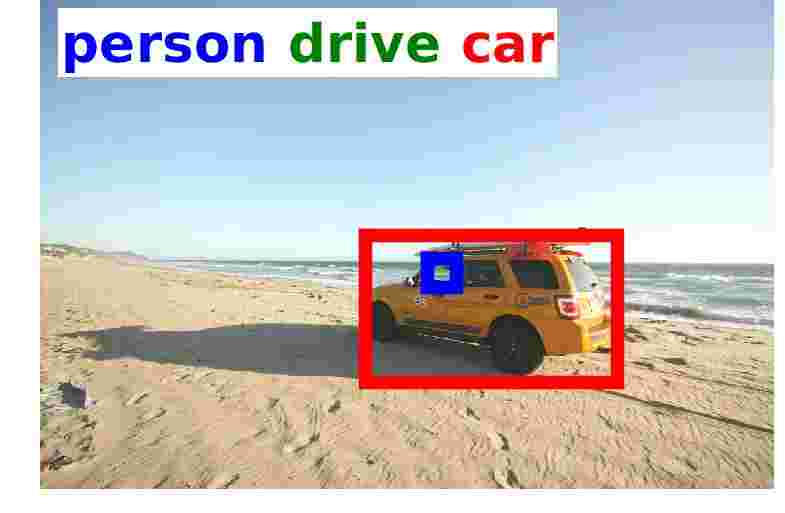}\\
       	\vspace{0.3ex}
       	GT: in, drive
       	\vspace{0.2ex}
    \end{minipage} 
    \hspace{0.005\textwidth}
    \begin{minipage}[t]{0.18\textwidth}
       	\centering
    	\includegraphics[trim={2cm 2cm 2cm 0cm},clip,width=0.95\linewidth,cfbox={yellow 2pt 2pt}]{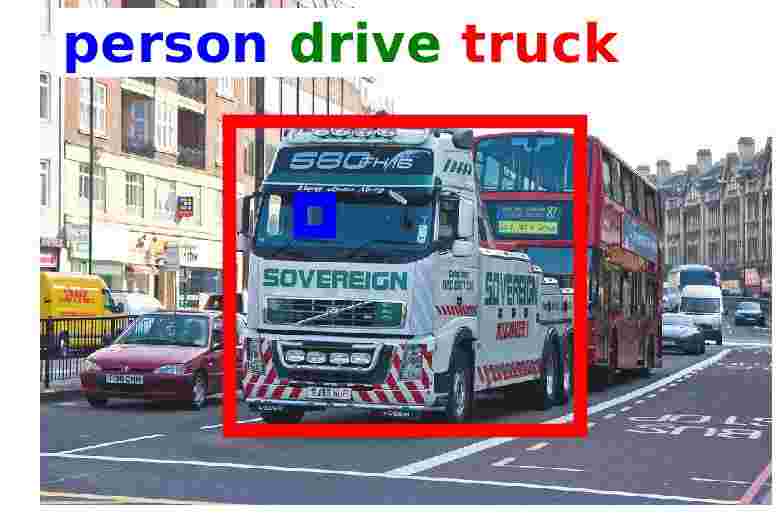}\\
    	\vspace{0.3ex}
       	GT: in
       	\vspace{0.2ex}
    \end{minipage}
    \hspace{0.005\textwidth}
    \begin{minipage}[t]{0.18\textwidth}
    	\centering
       	\includegraphics[trim={4cm 4cm 4cm 1cm},clip,width=0.95\linewidth,cfbox={red 2pt 2pt}]{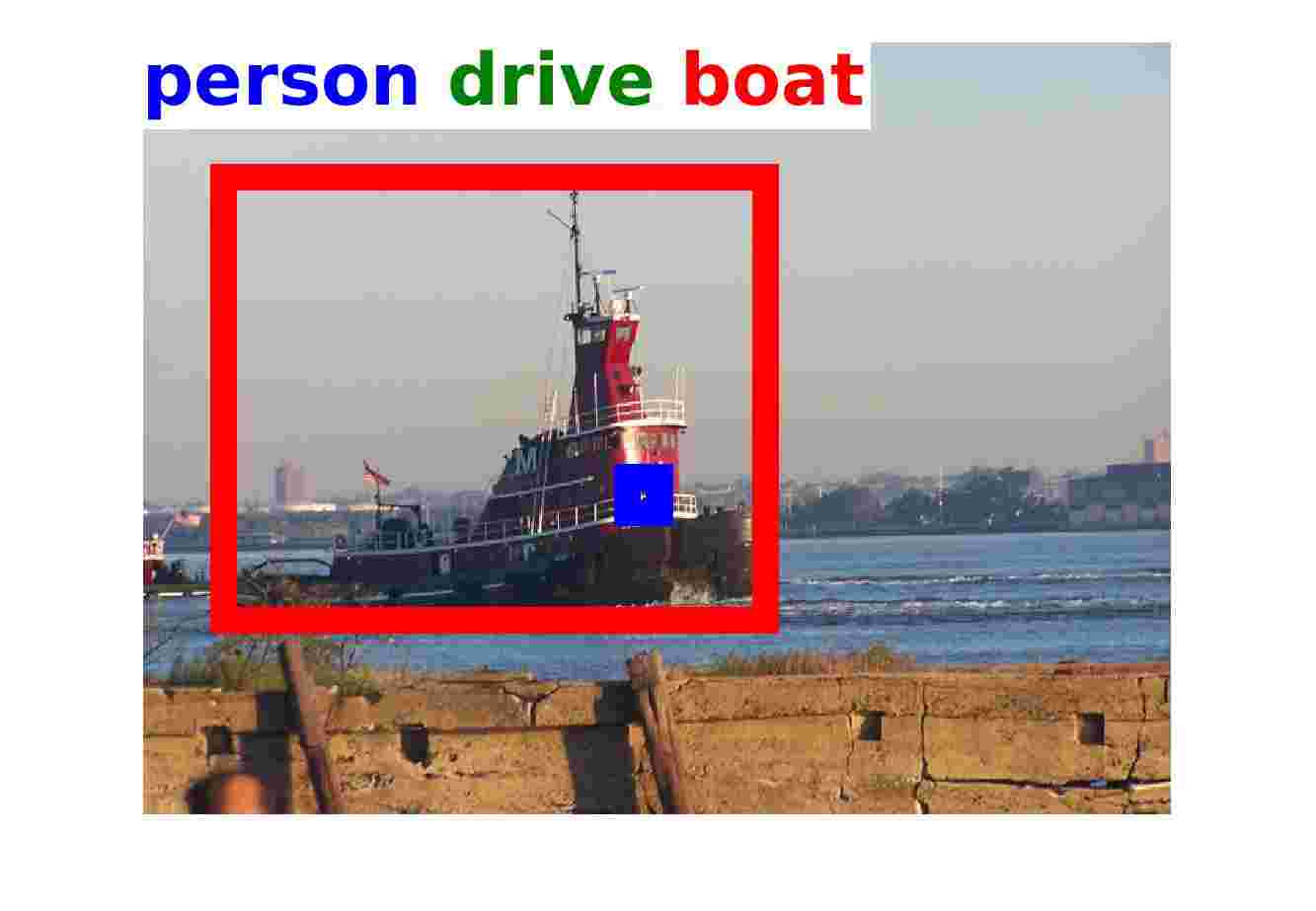}\\
       	\vspace{0.3ex}
       	GT: ride
       	\vspace{0.2ex}
    \end{minipage}

	\begin{minipage}[t]{0.005\textwidth}
    	\centering
    	\vspace{-9.5ex}
    	\begin{turn}{90}
    	stand on
    	\end{turn}
    	\vspace{5ex}
   	\end{minipage}
    \hspace{0.01\textwidth}
    \begin{minipage}[t]{0.18\textwidth}
    	\centering
       	\includegraphics[trim={1cm 0.6cm 0cm 0cm},clip,width=0.95\linewidth,cfbox={green 2pt 2pt}]{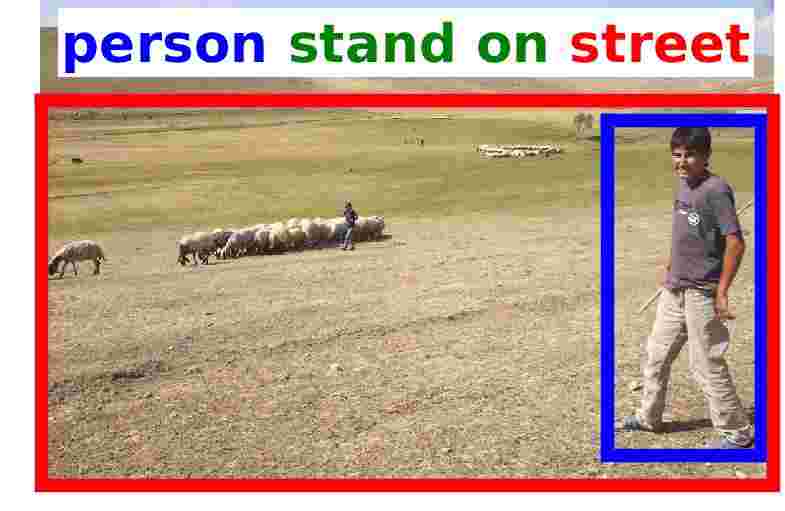}\\
       	\vspace{0.3ex}
       	GT: stand on
       	\vspace{0.2ex}
    \end{minipage}
    \hspace{0.005\textwidth}
    \begin{minipage}[t]{0.18\textwidth}
    	\centering
       	\includegraphics[trim={0cm 6.5cm 0cm 6.1cm},clip,width=0.95\linewidth,cfbox={green 2pt 2pt}]{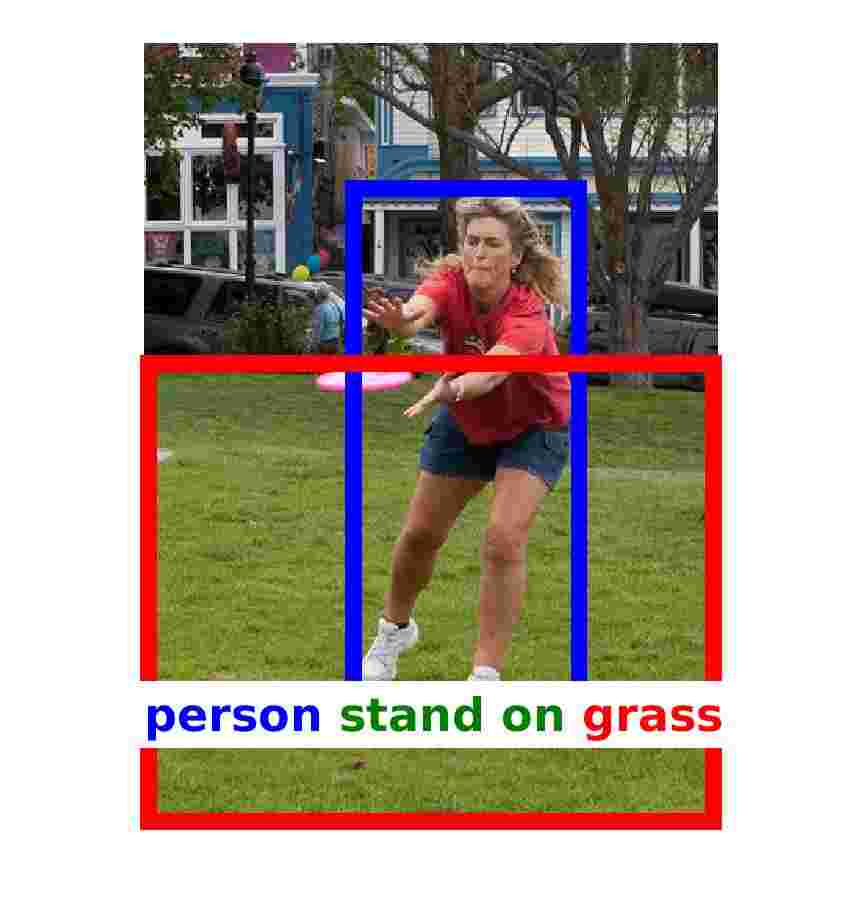}\\
       	\vspace{0.3ex}
       	GT: on, stand on
       	\vspace{0.2ex}
    \end{minipage} 
    \hspace{0.005\textwidth}
    \begin{minipage}[t]{0.18\textwidth}
    	\centering
       	\includegraphics[trim={1cm 0.6cm 1cm 0.8cm},clip,width=0.95\linewidth,cfbox={green 2pt 2pt}]{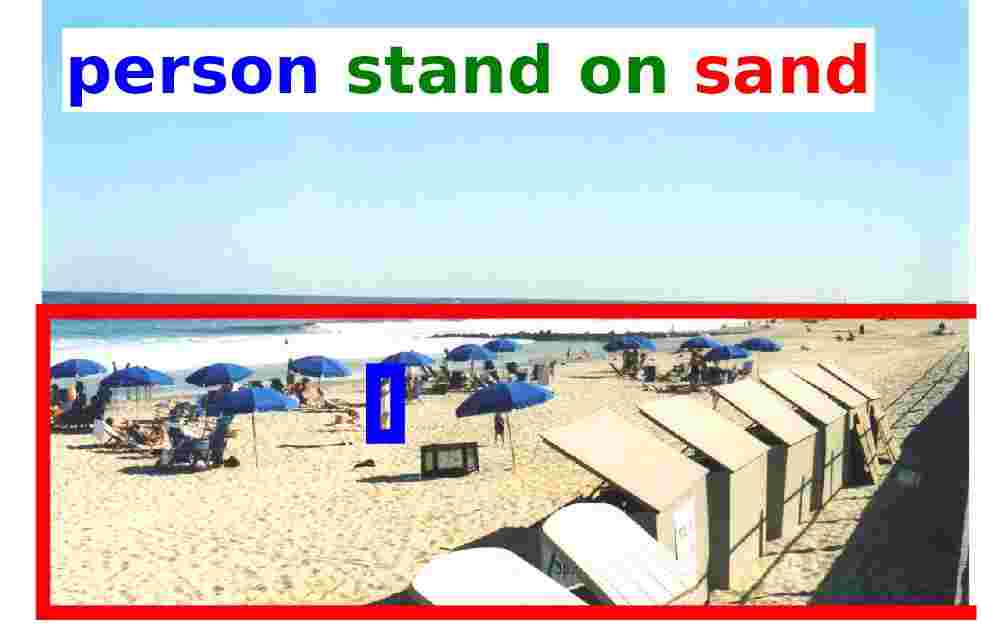}\\
       	\vspace{0.3ex}
       	GT: stand on
       	\vspace{0.2ex}
    \end{minipage}  
    \hspace{0.005\textwidth}
    \begin{minipage}[t]{0.18\textwidth}
       	\centering
    	\includegraphics[trim={2cm 2cm 0.2cm 1.7cm},clip,width=0.95\linewidth,cfbox={yellow 2pt 2pt}]{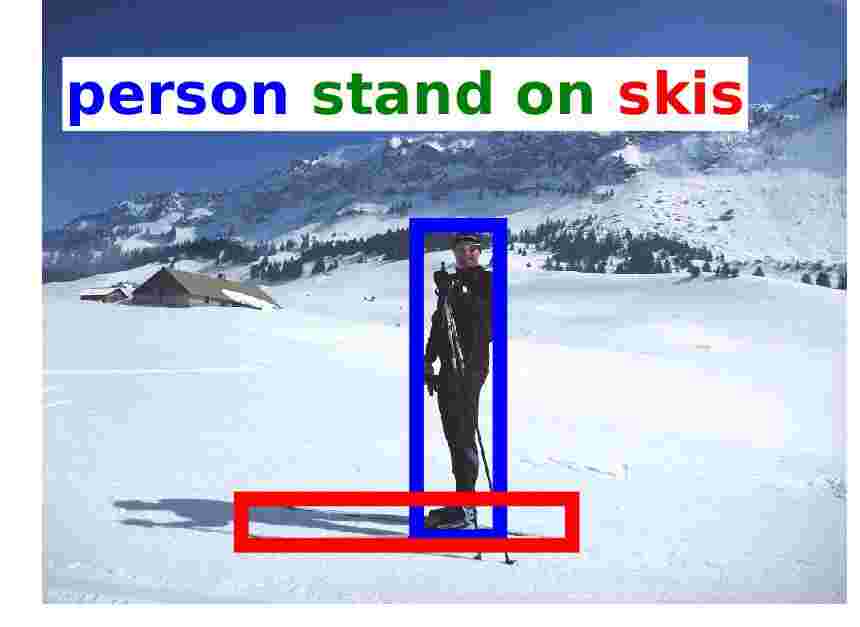}\\
    	\vspace{0.3ex}
       	GT: on
       	\vspace{0.2ex}
    \end{minipage}
    \hspace{0.005\textwidth}
    \begin{minipage}[t]{0.18\textwidth}
    	\centering
       	\includegraphics[trim={4cm 3cm 4cm 1cm},clip,width=0.95\linewidth,cfbox={red 2pt 2pt}]{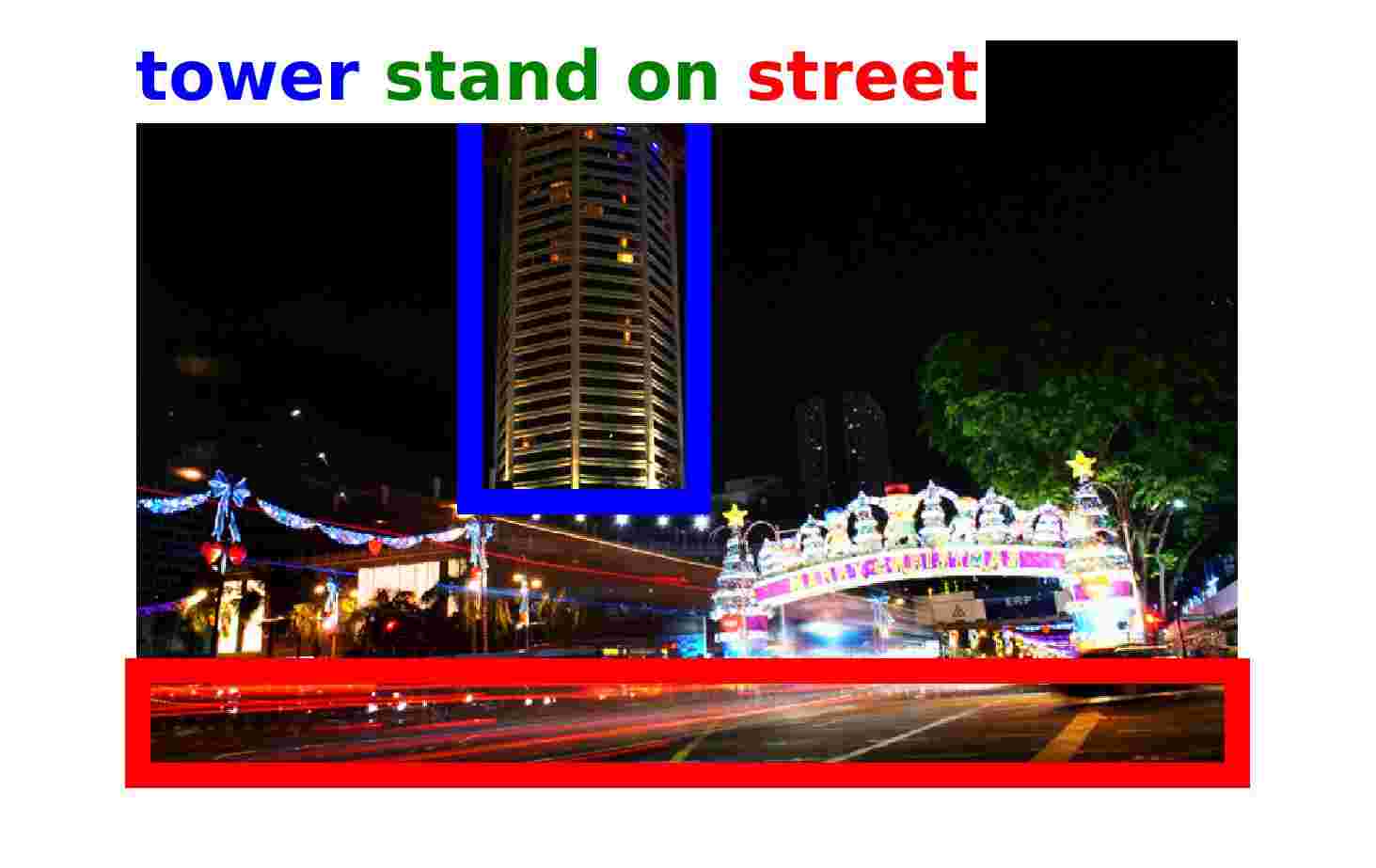}\\
       	\vspace{0.3ex}
       	GT: above
       	\vspace{0.2ex}
    \end{minipage}      

	\begin{minipage}[t]{0.005\textwidth}
    	\centering
    	\vspace{-8.5ex}
    	\begin{turn}{90}
    	look
    	\end{turn}
    	\vspace{3ex}
   	\end{minipage}
    \hspace{0.01\textwidth}
    \begin{minipage}[t]{0.18\textwidth}
    	\centering
       	\includegraphics[trim={1cm 0.6cm 1cm 3.3cm},clip,width=0.95\linewidth,cfbox={green 2pt 2pt}]{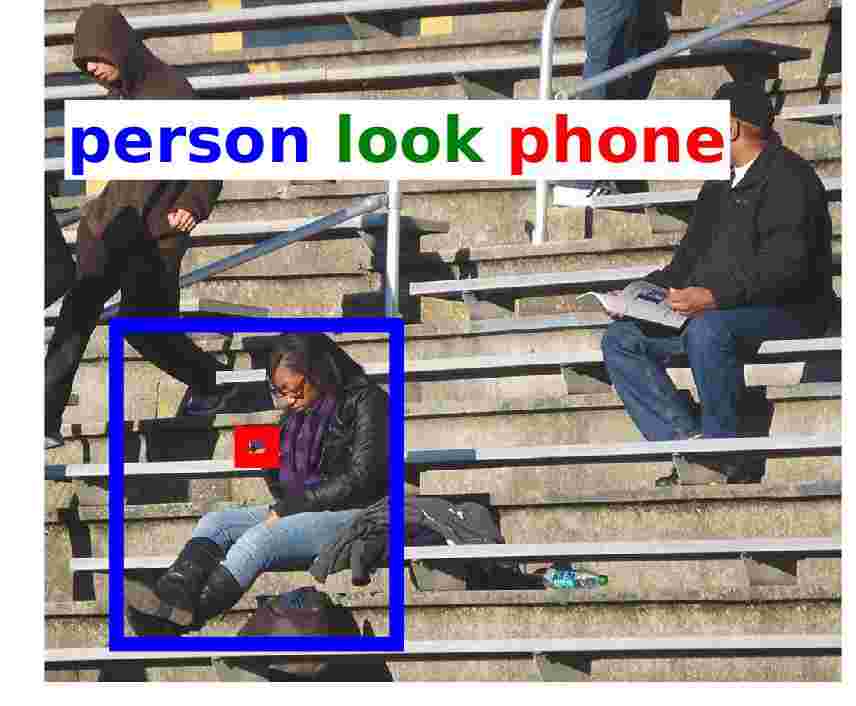}\\
       	\vspace{0.3ex}
       	GT: look
       	\vspace{0.2ex}
    \end{minipage}  
    \hspace{0.005\textwidth}
    \begin{minipage}[t]{0.18\textwidth}
    	\centering
       	\includegraphics[trim={1.3cm 0.5cm 0cm 0cm},clip,width=0.95\linewidth,cfbox={green 2pt 2pt}]{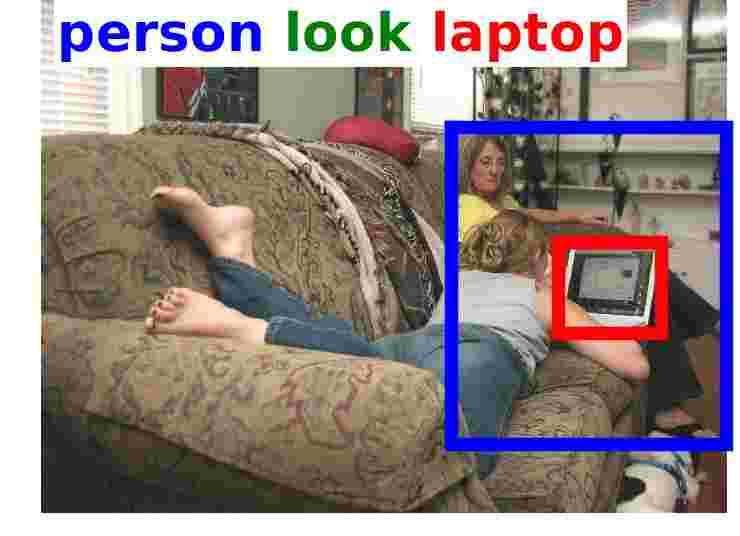}\\
       	\vspace{0.3ex}
       	GT: look
       	\vspace{0.2ex}
    \end{minipage} 
    \hspace{0.005\textwidth}
    \begin{minipage}[t]{0.18\textwidth}
    	\centering
       	\includegraphics[trim={6cm 1.6cm 6cm 0.4cm},clip,width=0.95\linewidth,cfbox={green 2pt 2pt}]{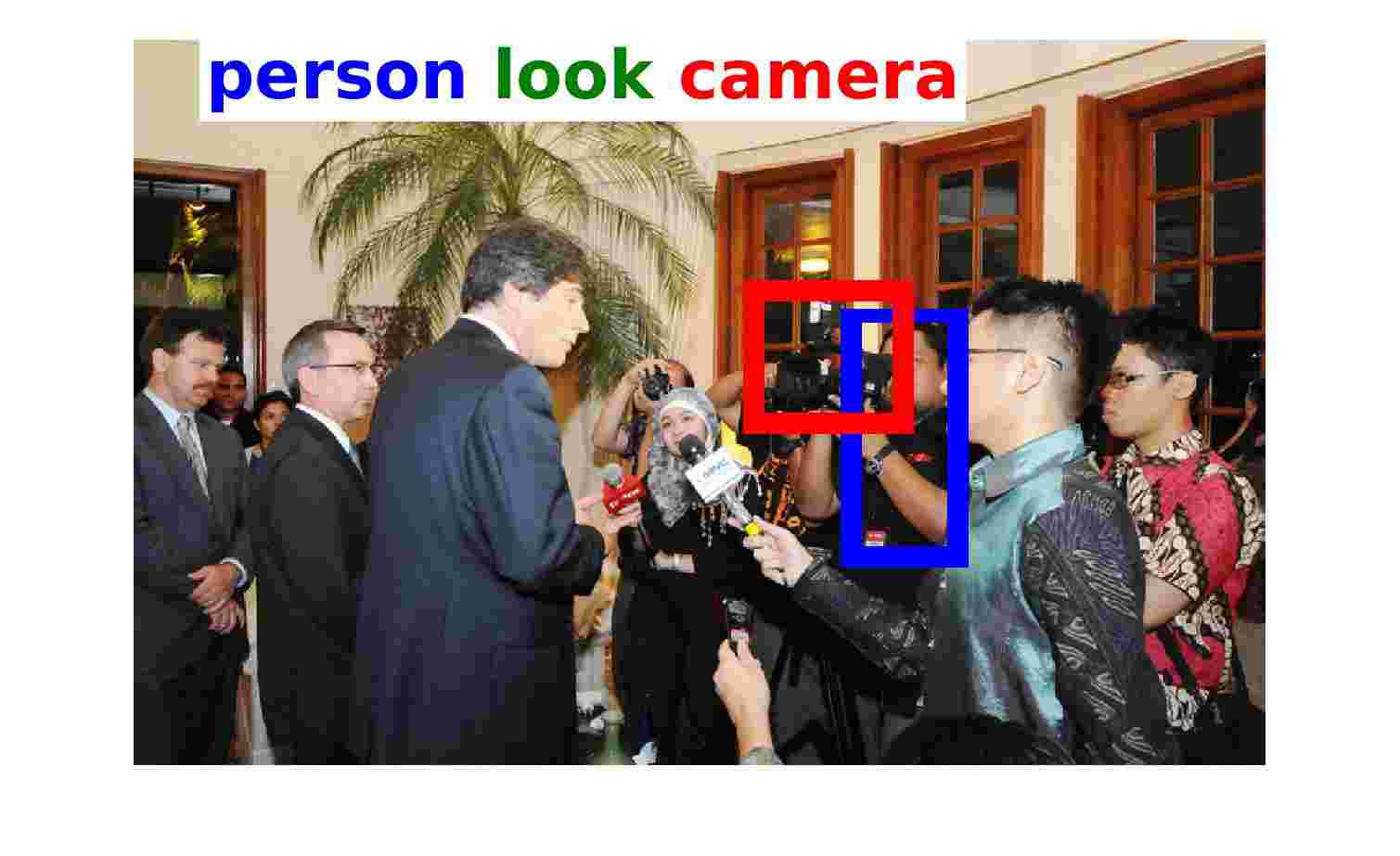}\\
       	\vspace{0.3ex}
       	GT: look, hold, with
       	\vspace{0.2ex}
    \end{minipage}
    \hspace{0.005\textwidth}
    \begin{minipage}[t]{0.18\textwidth}
       	\centering
    	\includegraphics[trim={4.3cm 1.2cm 3cm 0cm},clip,width=0.95\linewidth,cfbox={yellow 2pt 2pt}]{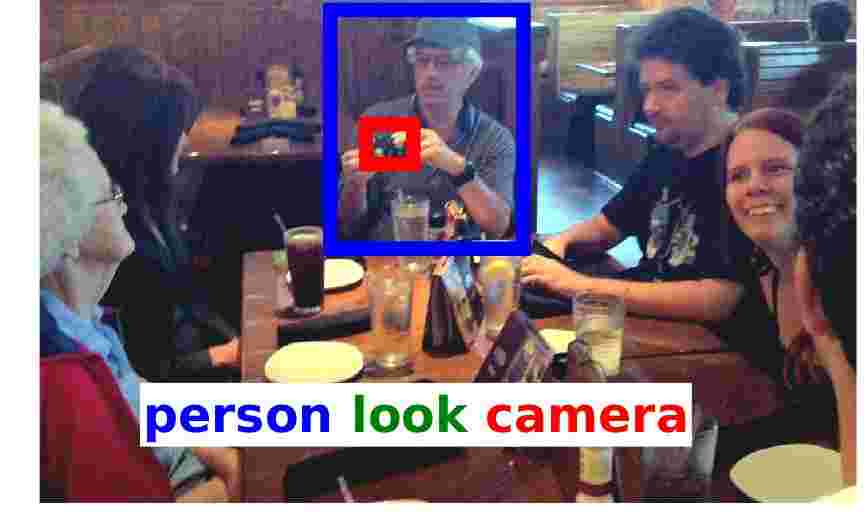}\\
    	\vspace{0.3ex}
       	GT: hold
       	\vspace{0.2ex}
    \end{minipage}
    \hspace{0.005\textwidth}
    \begin{minipage}[t]{0.18\textwidth}
    	\centering
       	\includegraphics[trim={1.7cm 8cm 0.5cm 1.1cm},clip,width=0.95\linewidth,cfbox={red 2pt 2pt}]{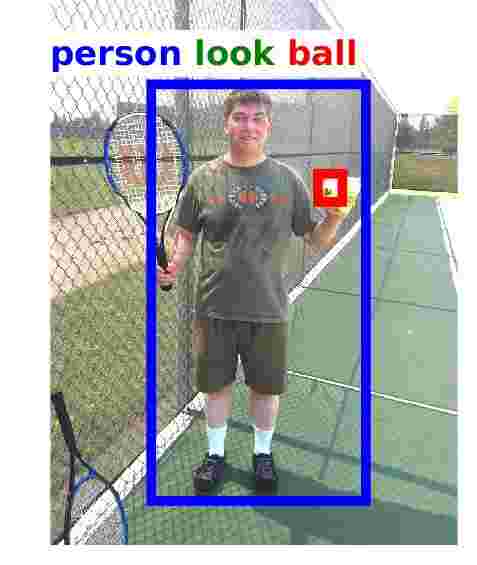}\\
       	\vspace{0.3ex}
       	GT: has
       	\vspace{0.2ex}
    \end{minipage}    
     
    \setlength\abovecaptionskip{5pt}
    \caption{Predicate detections on the test set of \cite{Lu16}. We show examples among the top 100 scored triplets for some action relations retrieved by our weakly-supervised model described in Section~\ref{model}. In this task, the candidate object boxes are the ground truth boxes. The triplet is correctly recognized if the relation matches ground truth (in green), else the triplet is incorrect (in red). We also show examples of correctly predicted relations where the ground truth is erroneous, either missing or incomplete (in yellow). Below each image, we indicate the ground truth predicates ('GT') for the pair of boxes shown.}
    \label{fig:predicate_detection_action}
\end{figure*}

\paragraph{What is learnt by action predicates?}
In Figure~\ref{fig:predicate_detection_action}, we show examples of predictions with our weakly-supervised model (Section~\ref{model} in the main paper) for the task of predicate detection. In this task, candidate object boxes are fixed to ground truth boxes and the goal is to predict the relation between a pair of objects. We perform retrieval per class, i.e. for each predicate (one row in Figure~\ref{fig:predicate_detection_action}) we show examples of top-scoring object pairs for this relation. This allows us to visualize what our model has learnt for less frequent predicates such as ``ride", ``carry" or ``drive", which are less frequently predicted by our model, as biases in the dataset favor predicates such as ``on", ``has" or ``wear". 
Similar to prepositions, we see that the spatial configuration of
object boxes plays a key role in the prediction of verbs. Indeed, the
top-ranked pairs in each row share similar spatial patterns. The
top-ranked negatives (in red) demonstrate that it is still challenging
to disambiguate subtle differences between relations (e.g. ``person
ride horse" versus ``person on horse", or ``person hold watch" versus
``person wear watch"). 
Ground truth can also be incomplete or ambiguous (in
yellow), i.e. ``person ride bike'' is predicted correctly, whereas the ground
truth ``sit on'' is less relevant for this example. 

\begin{table}\centering
\ra{1}
\begin{tabular}{@{}rrrcrr@{}}\toprule
\textbf{Relationship} & \multicolumn{1}{c}{Top-1} & \multicolumn{1}{c}{Top-5}
\\\midrule
\cite{Krishna2016} full 		& 8.7 	& 26.6 \\
Ours [S+A] full 			& \textbf{46.5} & \textbf{76.4}\\
Ours [S+A] weak 			& 35.5 & 70.1 \\
\bottomrule
\end{tabular}
{\vskip-0.5ex}
\caption{Results on the Visual Genome dataset \cite{Krishna2016} for the Relationship recognition task.} 
\label{tab:results_visualgenome}
\vspace{-0.2cm}
\end{table}

\begin{figure*}[t]
\centering
	\begin{minipage}[b]{0.6\textwidth}
    \centering
    	\textit{correctly recognized relations}\\
    	\vspace{0.4ex}
	\end{minipage}
	\hspace{0.005\textwidth}
	\begin{minipage}[b]{0.18\textwidth}
    \centering
    	\textit{missing / ambiguous}\\
    	\vspace{0.4ex}
	\end{minipage}
	\hspace{0.005\textwidth}
	\begin{minipage}[b]{0.19\textwidth}
    \centering
    	\textit{incorrectly recognized}\\
    	\vspace{0.4ex}
	\end{minipage}
	
    \begin{minipage}[t]{0.185\textwidth}
    	\centering
       	\includegraphics[trim={1cm 0.5cm 0.5cm 0cm},clip,width=0.95\linewidth,cfbox={green 2pt 2pt}]{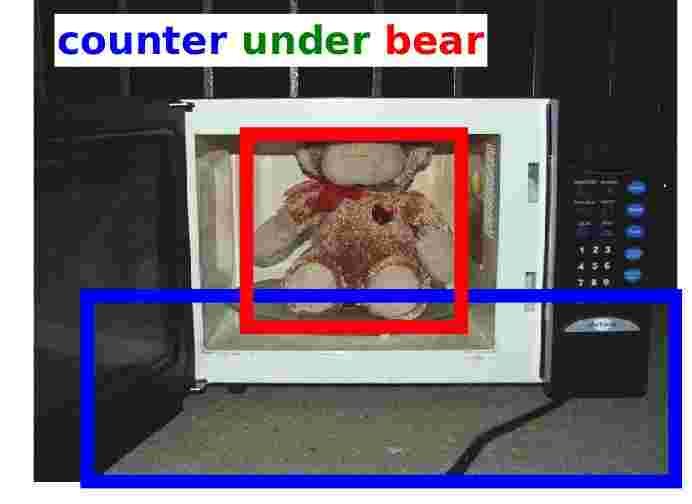}\\
       	\vspace{0.3ex}
       	\cite{Lu16}: contain \\
       	GT: under
       	\vspace{2ex}
    \end{minipage}
    \hspace{0.005\textwidth}
    \begin{minipage}[t]{0.185\textwidth}
    	\centering
       	\includegraphics[trim={0.5cm 0.5cm 1cm 0cm},clip,width=0.95\linewidth,cfbox={green 2pt 2pt}]{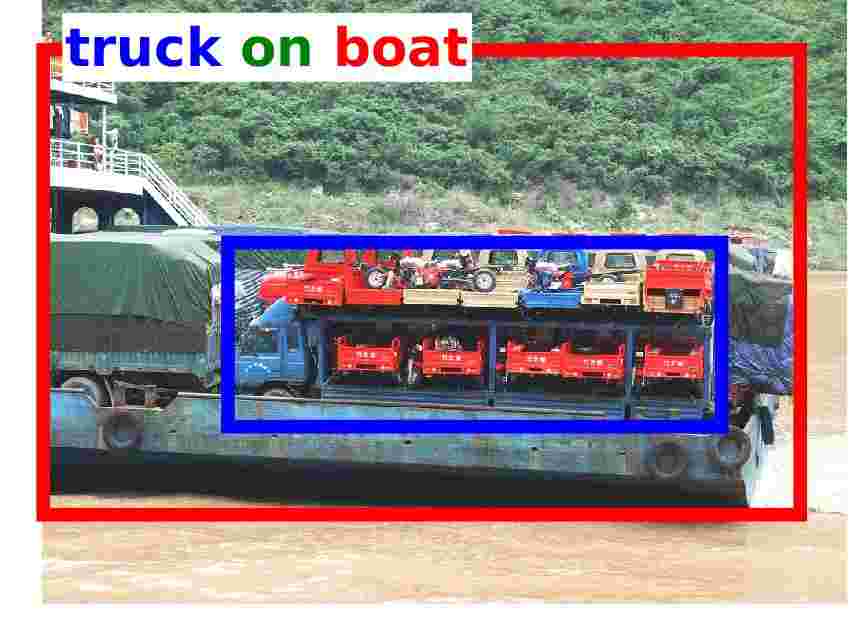}\\
       	\vspace{0.3ex}
       	\cite{Lu16}: behind \\
       	GT: on
       	\vspace{0.2ex}
    \end{minipage}
    \hspace{0.005\textwidth}
    \begin{minipage}[t]{0.185\textwidth}
    	\centering
       	\includegraphics[trim={1cm 0.5cm 0.5cm 0cm},clip,width=0.95\linewidth,cfbox={green 2pt 2pt}]{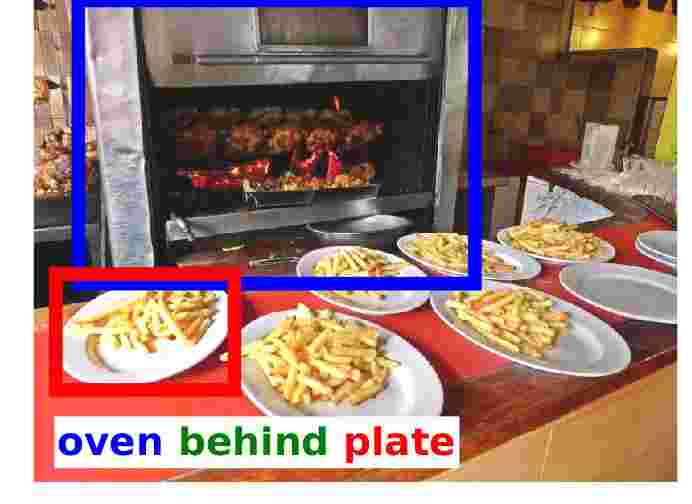}\\
		\vspace{0.3ex}       	
       	\cite{Lu16}: under \\
       	GT: behind
       	\vspace{0.2ex}
    \end{minipage}
    \hspace{0.005\textwidth}
    \begin{minipage}[t]{0.185\textwidth}
    	\centering
       	\includegraphics[trim={1cm 0.3cm 0.5cm 0cm},clip,width=0.95\linewidth,cfbox={yellow 2pt 2pt}]{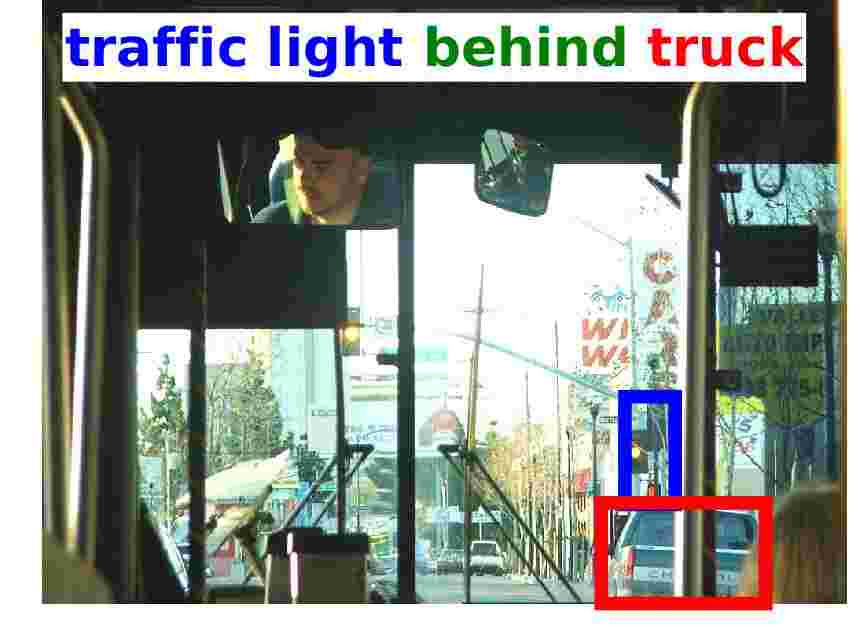}\\
		\vspace{0.3ex}       	
       	\cite{Lu16}: above \\
       	GT: in the front of
       	\vspace{0.2ex}
    \end{minipage}
    \hspace{0.005\textwidth}  
    \begin{minipage}[t]{0.185\textwidth}
    	\centering
       	\includegraphics[trim={2cm 5cm 2cm 2cm},clip,width=0.95\linewidth,cfbox={red 2pt 2pt}]{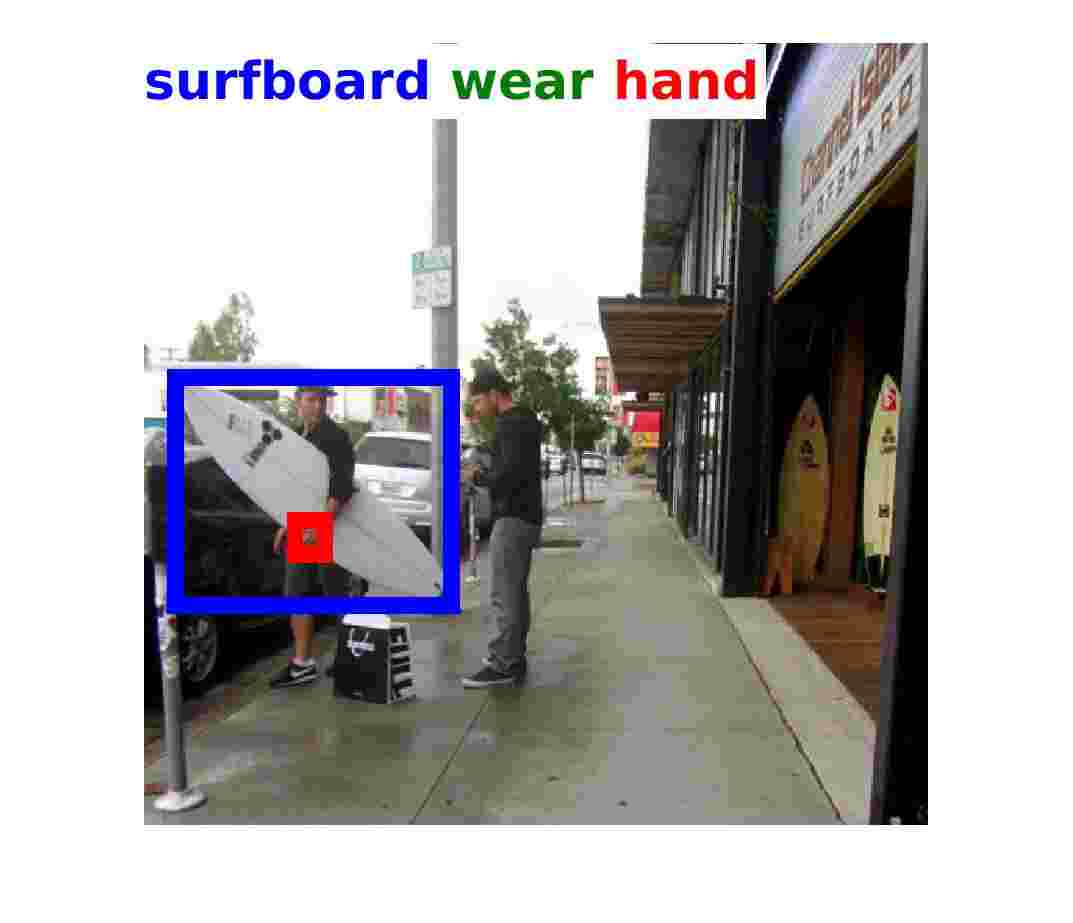}\\
		\vspace{0.3ex}       	
       	\cite{Lu16}: in \\
       	GT: in
      	\vspace{0.2ex}
    \end{minipage} 

    \begin{minipage}[t]{0.185\textwidth}
    	\centering
       	\includegraphics[trim={1.5cm 0.5cm 1.3cm 0cm},clip,width=0.95\linewidth,cfbox={green 2pt 2pt}]{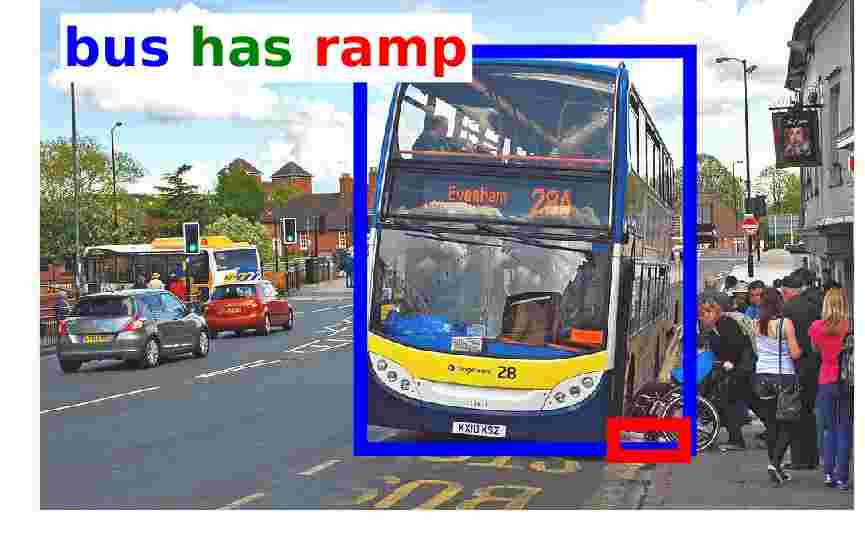}\\
       	\vspace{0.3ex}
       	\cite{Lu16}: on \\
       	GT: has
       	\vspace{2ex}
    \end{minipage}
    \hspace{0.005\textwidth}
    \begin{minipage}[t]{0.185\textwidth}
       \centering
       \includegraphics[trim={1cm 0.5cm 0cm 0cm},clip,width=0.95\linewidth,cfbox={green 2pt 2pt}]{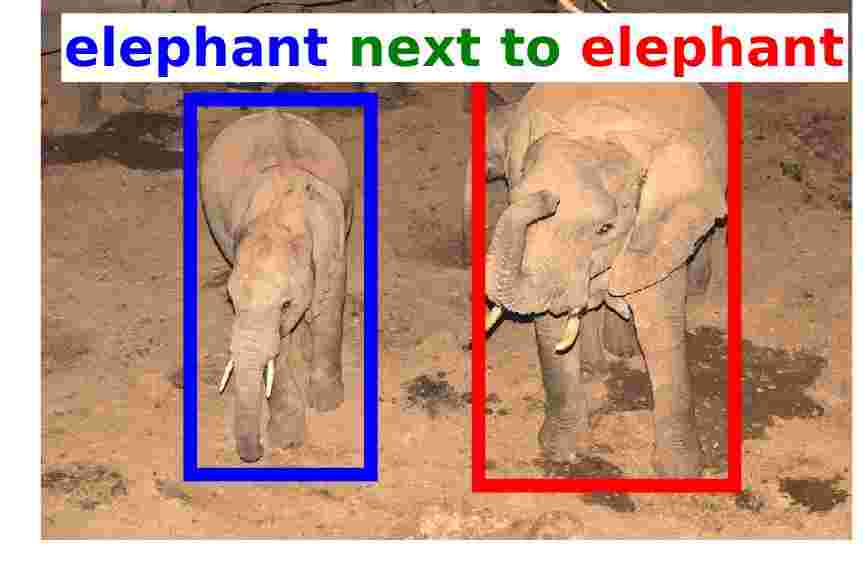}\\
       \vspace{0.3ex}
       \cite{Lu16}: feed \\
       GT: next to
       \vspace{0.2ex}
    \end{minipage}
    \hspace{0.005\textwidth}
    \begin{minipage}[t]{0.185\textwidth}
       \centering
       \includegraphics[trim={1cm 0.5cm 0cm 0cm},clip,width=0.95\linewidth,cfbox={green 2pt 2pt}]{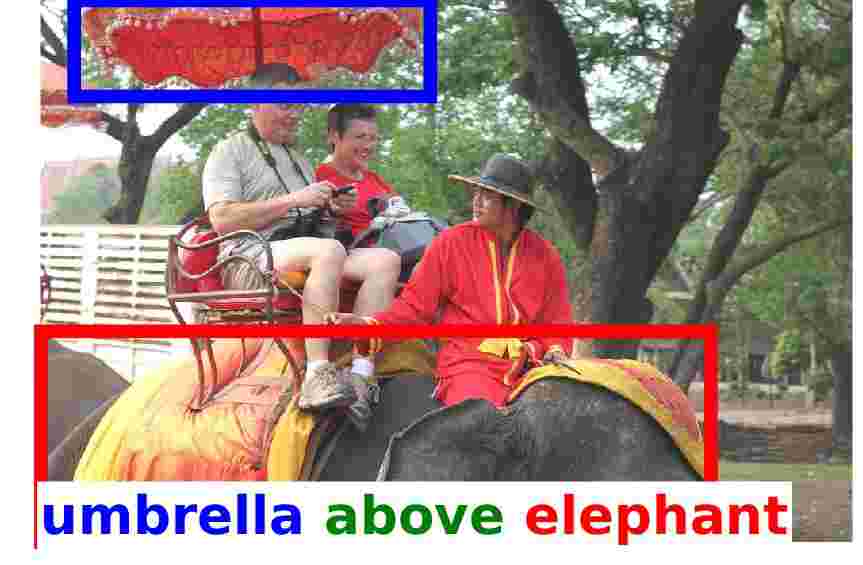}\\
       \vspace{0.3ex}
       \cite{Lu16}: behind \\
       GT: above
       \vspace{0.2ex}
    \end{minipage}
    \hspace{0.005\textwidth}
    \begin{minipage}[t]{0.185\textwidth}
    	\centering
       	\includegraphics[trim={4cm 3cm 5.5cm 0.2cm},clip,width=0.95\linewidth,cfbox={yellow 2pt 2pt}]{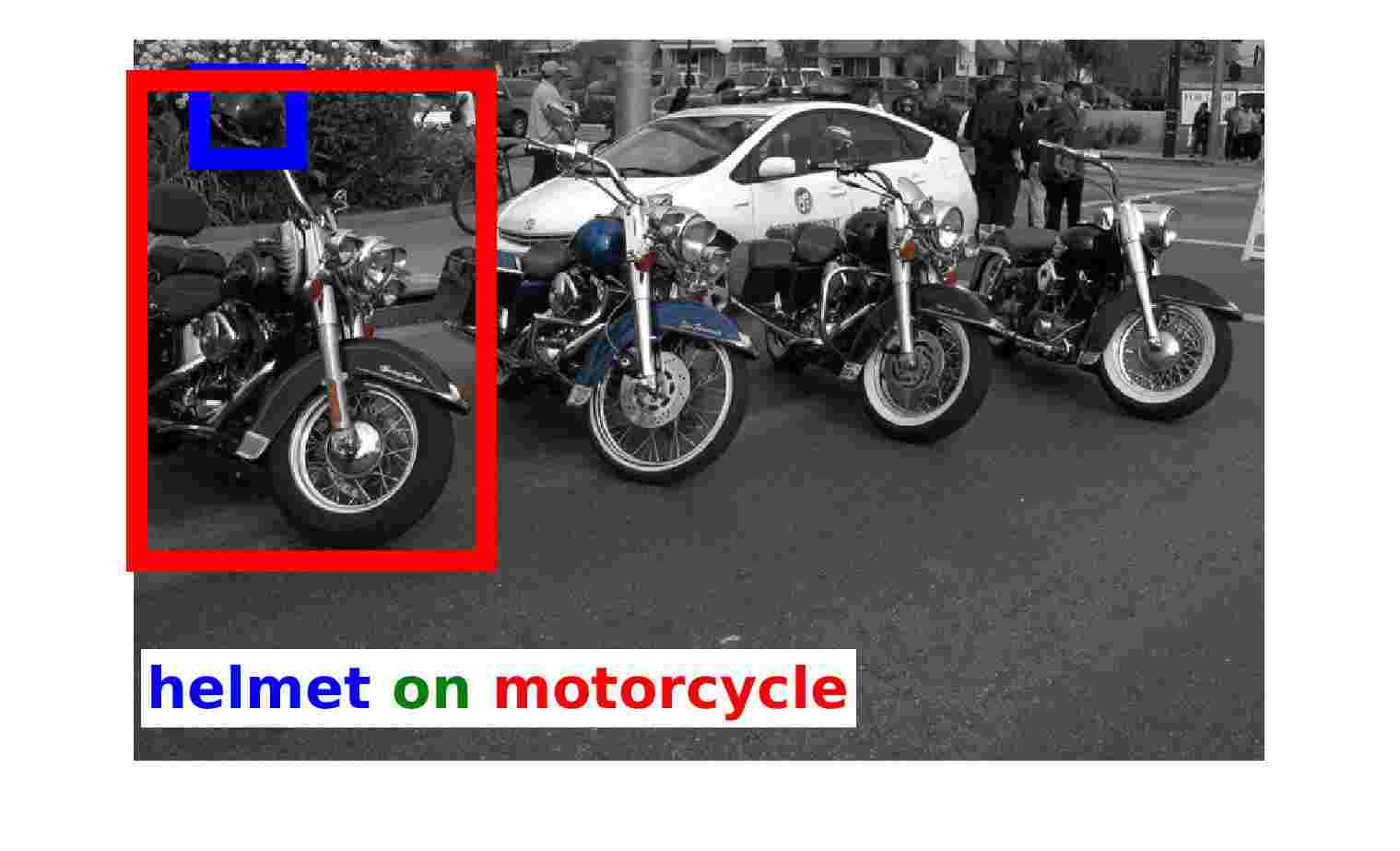}\\
       	\vspace{0.3ex}
       	\cite{Lu16}: on \\
       	GT: on the top of
       	\vspace{0.2ex}
    \end{minipage}
    \hspace{0.005\textwidth} 
    \begin{minipage}[t]{0.185\textwidth}
    	\centering
       	\includegraphics[trim={1cm 1cm 0cm 0.1cm},clip,width=0.95\linewidth,cfbox={red 2pt 2pt}]{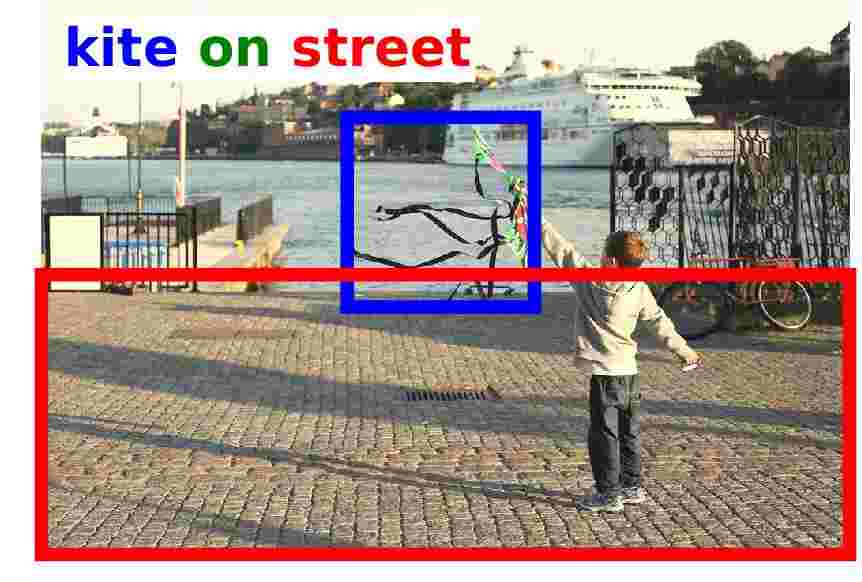}\\
       	\vspace{0.3ex}
       	\cite{Lu16}: above \\
       	GT: above
      	\vspace{0.2ex}
    \end{minipage}  

    \begin{minipage}[t]{0.185\textwidth}
    	\centering
       	\includegraphics[trim={2cm 0.5cm 0.5cm 0cm},clip,width=0.95\linewidth,cfbox={green 2pt 2pt}]{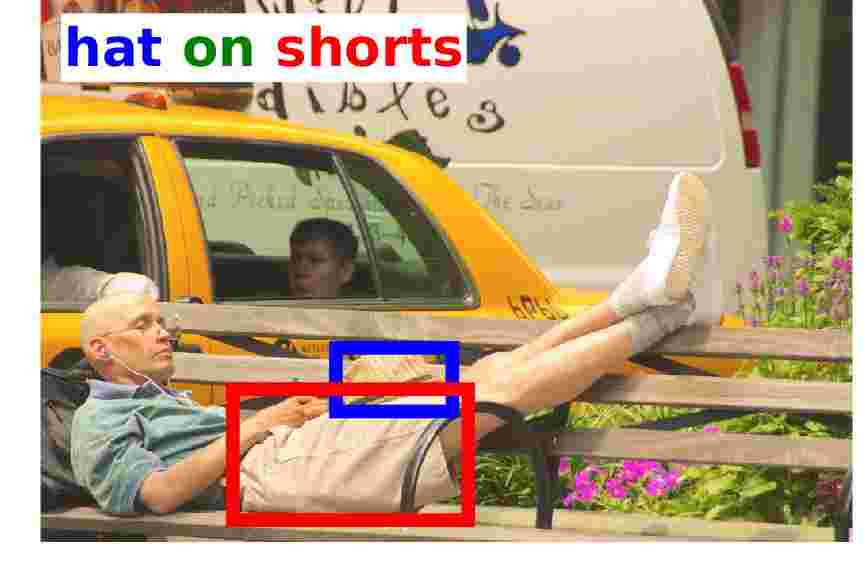}\\
       	\vspace{0.3ex}
       	\cite{Lu16}: above \\
       	GT: on
       	\vspace{2ex}
    \end{minipage}
    \hspace{0.005\textwidth}
    \begin{minipage}[t]{0.185\textwidth}
    	\centering
       	\includegraphics[trim={3cm 3cm 3cm 1cm},clip,width=0.95\linewidth,cfbox={green 2pt 2pt}]{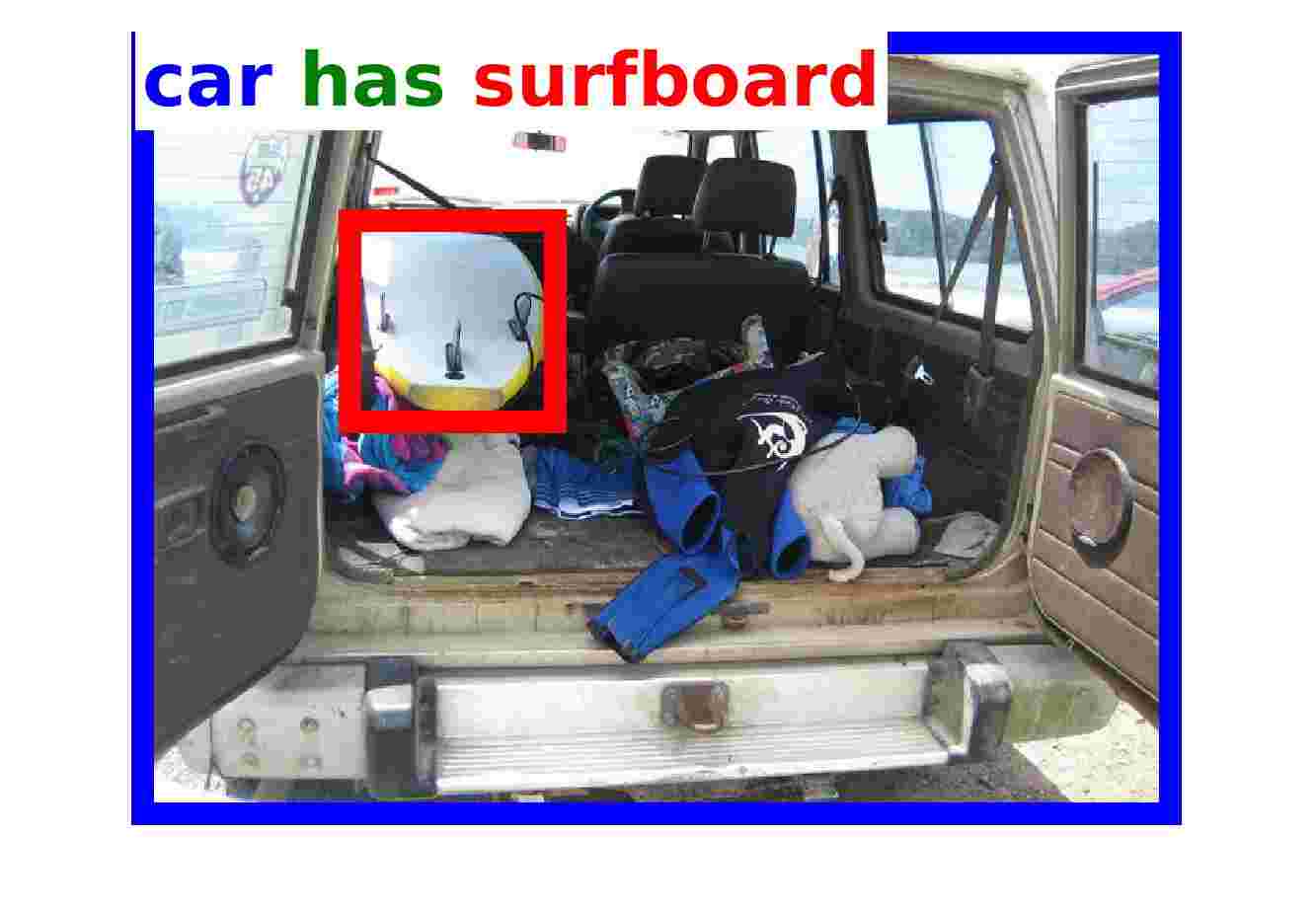}\\
       	\vspace{0.3ex}
       	\cite{Lu16}: on \\
       	GT: has
       	\vspace{0.2ex}
    \end{minipage}
    \hspace{0.005\textwidth}
    \begin{minipage}[t]{0.185\textwidth}
       \centering
       \includegraphics[trim={1cm 0.5cm 5.2cm 0cm},clip,width=0.95\linewidth,cfbox={green 2pt 2pt}]{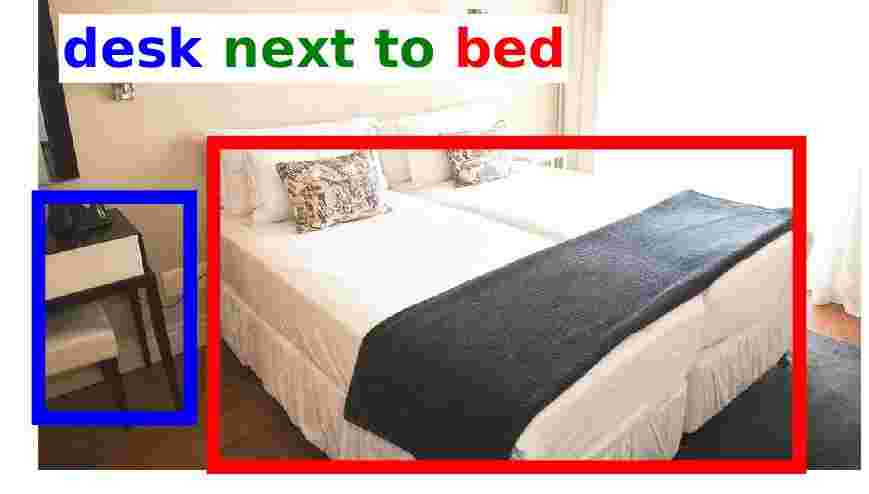}\\
       \vspace{0.3ex}
       \cite{Lu16}: beneath \\
       GT.: next to
       \vspace{0.2ex}
    \end{minipage}
    \hspace{0.005\textwidth}
    \begin{minipage}[t]{0.185\textwidth}
    	\centering
       	\includegraphics[trim={1cm 0.1cm 1cm 0cm},clip,width=0.95\linewidth,cfbox={yellow 2pt 2pt}]{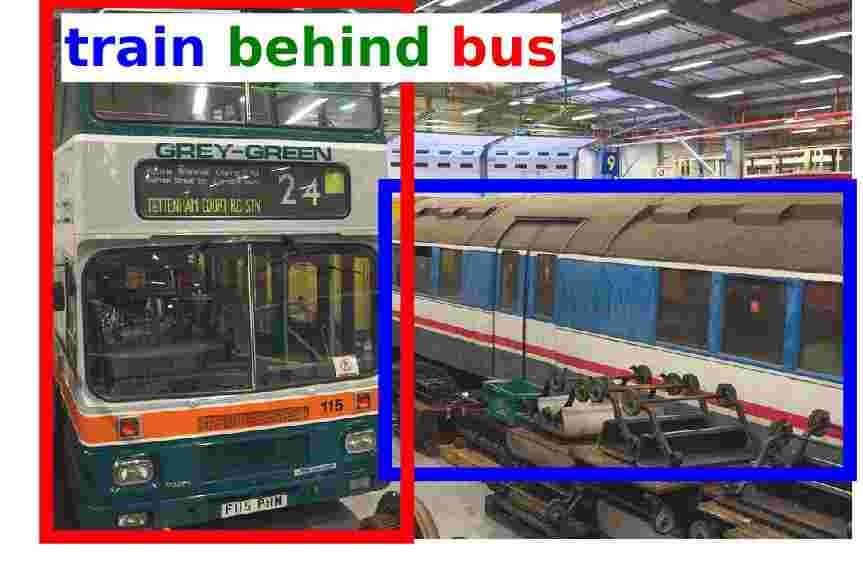}\\
       	\vspace{0.3ex}
       	\cite{Lu16}: next to \\
       	GT: next to
       	\vspace{0.2ex}
    \end{minipage}
    \hspace{0.005\textwidth}  
    \begin{minipage}[t]{0.185\textwidth}
    	\centering
       	\includegraphics[trim={2.2cm 0.5cm 0.2cm 0.1cm},clip,width=0.95\linewidth,cfbox={red 2pt 2pt}]{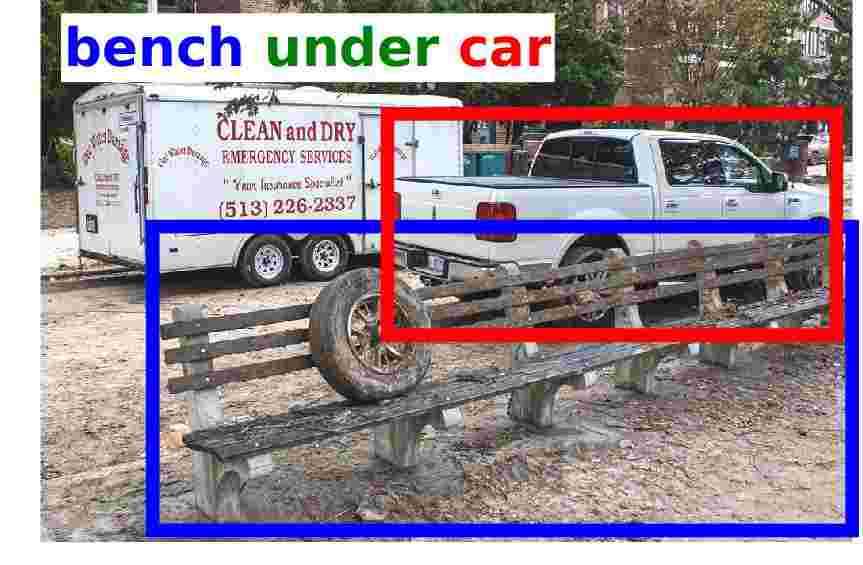}\\
		\vspace{0.3ex}       	
       	\cite{Lu16}: under \\
       	GT: in the front of
      	\vspace{0.2ex}
    \end{minipage} 

    \begin{minipage}[t]{0.185\textwidth}
    	\centering
       	\includegraphics[trim={2cm 0.5cm 0.5cm 0cm},clip,width=0.95\linewidth,cfbox={green 2pt 2pt}]{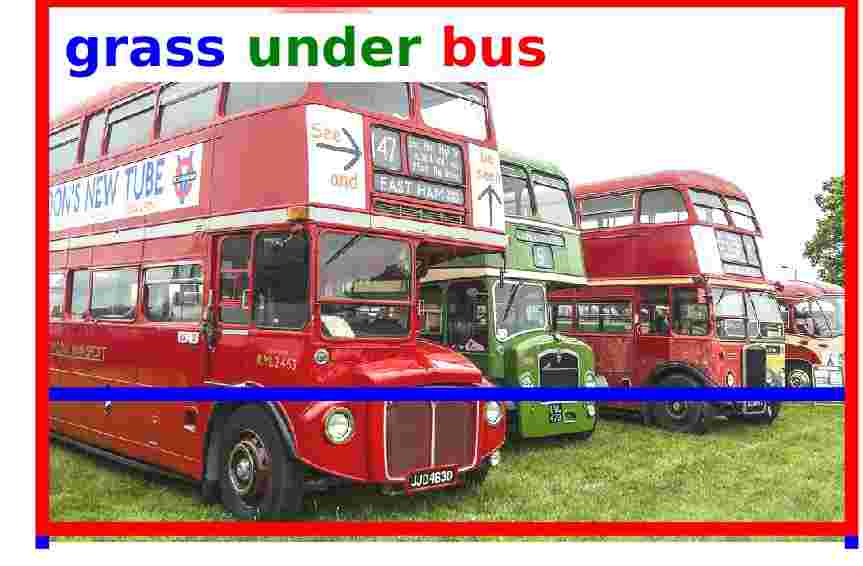}\\
       	\vspace{0.3ex}
       	\cite{Lu16}: behind \\
       	GT: under
       	\vspace{2ex}
    \end{minipage}
    \hspace{0.005\textwidth}
    \begin{minipage}[t]{0.185\textwidth}
    	\centering
       	\includegraphics[trim={1cm 1cm 0.7cm 0cm},clip,width=0.95\linewidth,cfbox={green 2pt 2pt}]{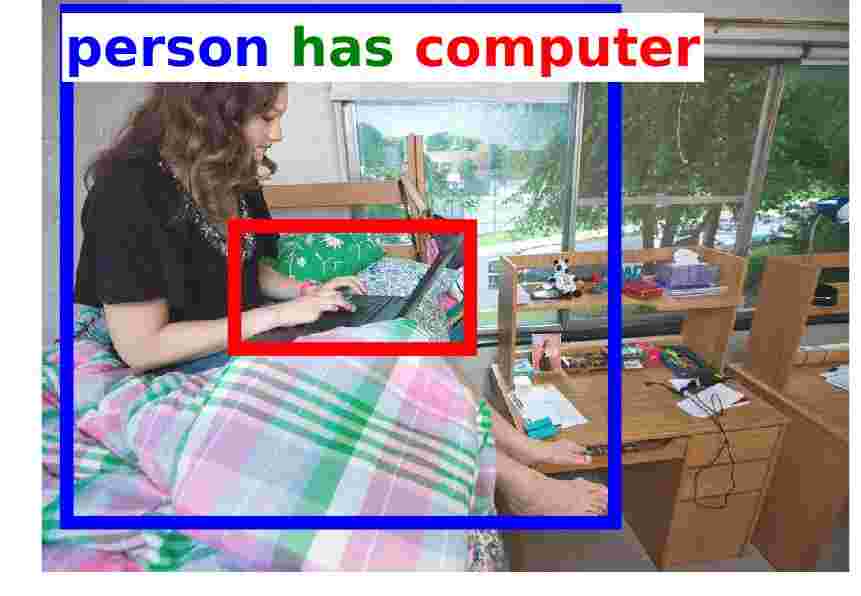}\\
       	\vspace{0.3ex}
       	\cite{Lu16}: wear \\
       	GT: has
       	\vspace{0.2ex}
    \end{minipage}
    \hspace{0.005\textwidth}
    \begin{minipage}[t]{0.185\textwidth}
       \centering
       \includegraphics[trim={1.5cm 0.5cm 1cm 0cm},clip,width=0.95\linewidth,cfbox={green 2pt 2pt}]{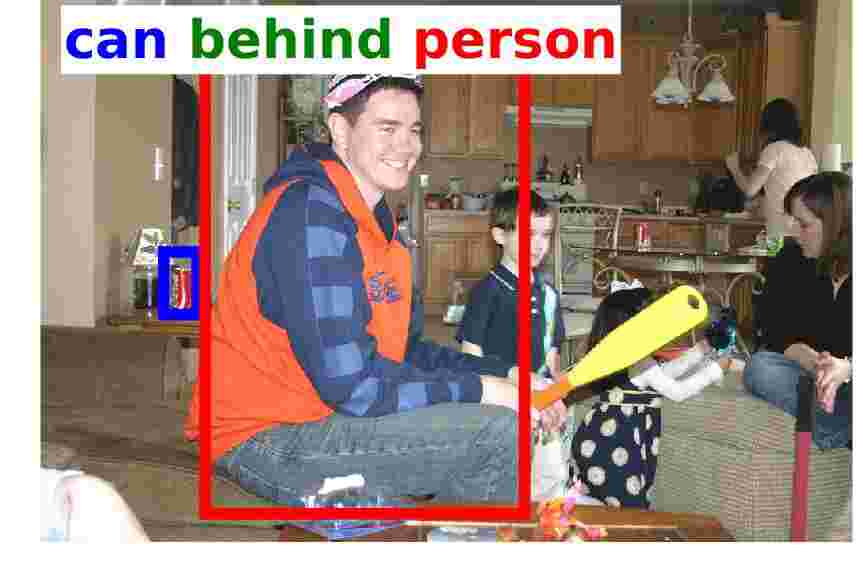}\\
       \vspace{0.3ex}
       \cite{Lu16}: on \\
       GT: behind
       \vspace{0.2ex}
    \end{minipage}
    \hspace{0.005\textwidth}
    \begin{minipage}[t]{0.185\textwidth}
    	\centering
       	\includegraphics[trim={1cm 2cm 1cm 0.3cm},clip,width=0.95\linewidth,cfbox={yellow 2pt 2pt}]{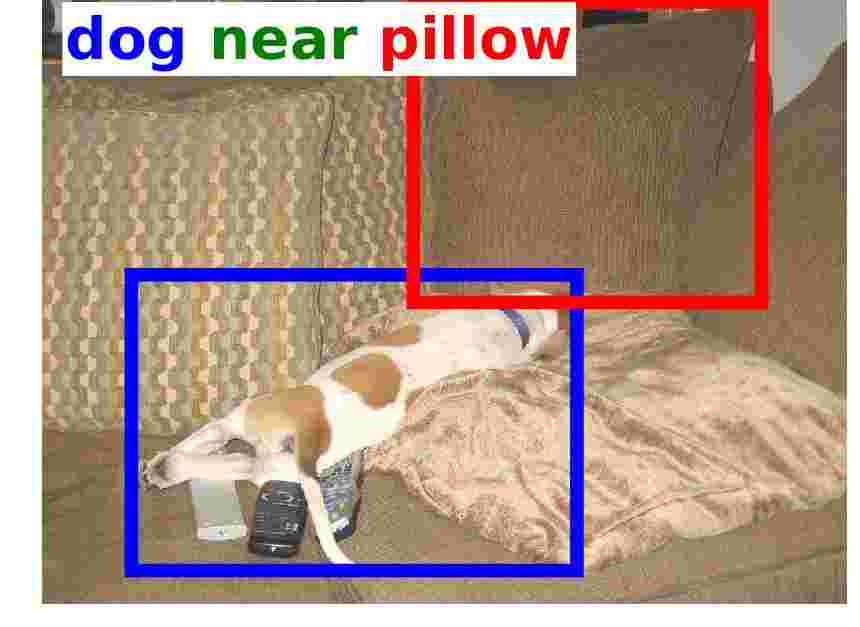}\\
       	\vspace{0.3ex}
       	\cite{Lu16}: next to \\
       	GT: beside, next to
       	\vspace{0.2ex}
    \end{minipage}
    \hspace{0.005\textwidth}  
    \begin{minipage}[t]{0.185\textwidth}
    	\centering
       	\includegraphics[trim={1cm 0.9cm 2cm 0cm},clip,width=0.95\linewidth,cfbox={red 2pt 2pt}]{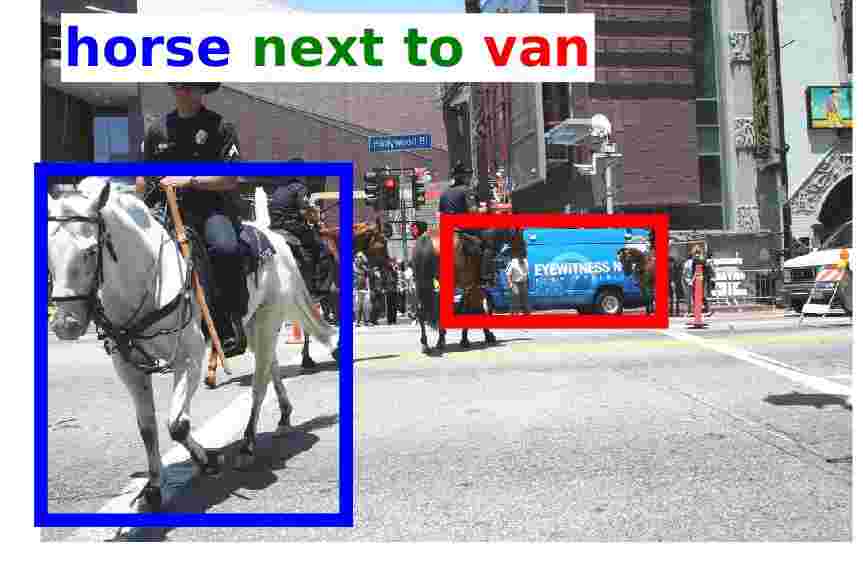}\\
		\vspace{0.3ex}       	
       	\cite{Lu16}: next to \\
       	GT: in the front of
      	\vspace{0.2ex}
    \end{minipage} 

    \begin{minipage}[t]{0.185\textwidth}
    	\centering
       	\includegraphics[trim={1.7cm 0.5cm 0.5cm 0cm},clip,width=0.95\linewidth,cfbox={green 2pt 2pt}]{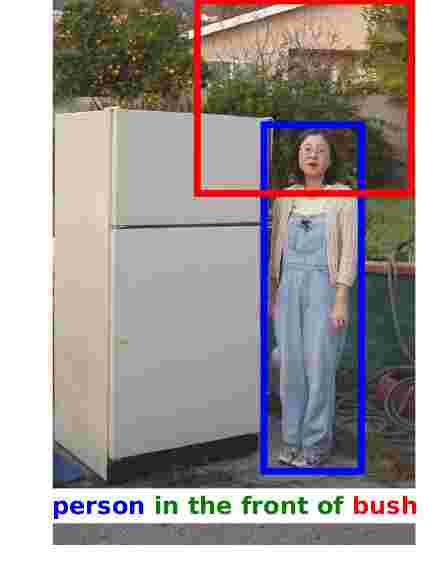}\\
		\vspace{0.3ex}       	
       	\cite{Lu16}: wear \\
       	GT: in the front of
       	\vspace{3ex}
    \end{minipage}
    \hspace{0.005\textwidth}
    \begin{minipage}[t]{0.185\textwidth}
    	\centering
       	\includegraphics[trim={10cm 0cm 15.1cm 0cm},clip,width=0.95\linewidth,cfbox={green 2pt 2pt}]{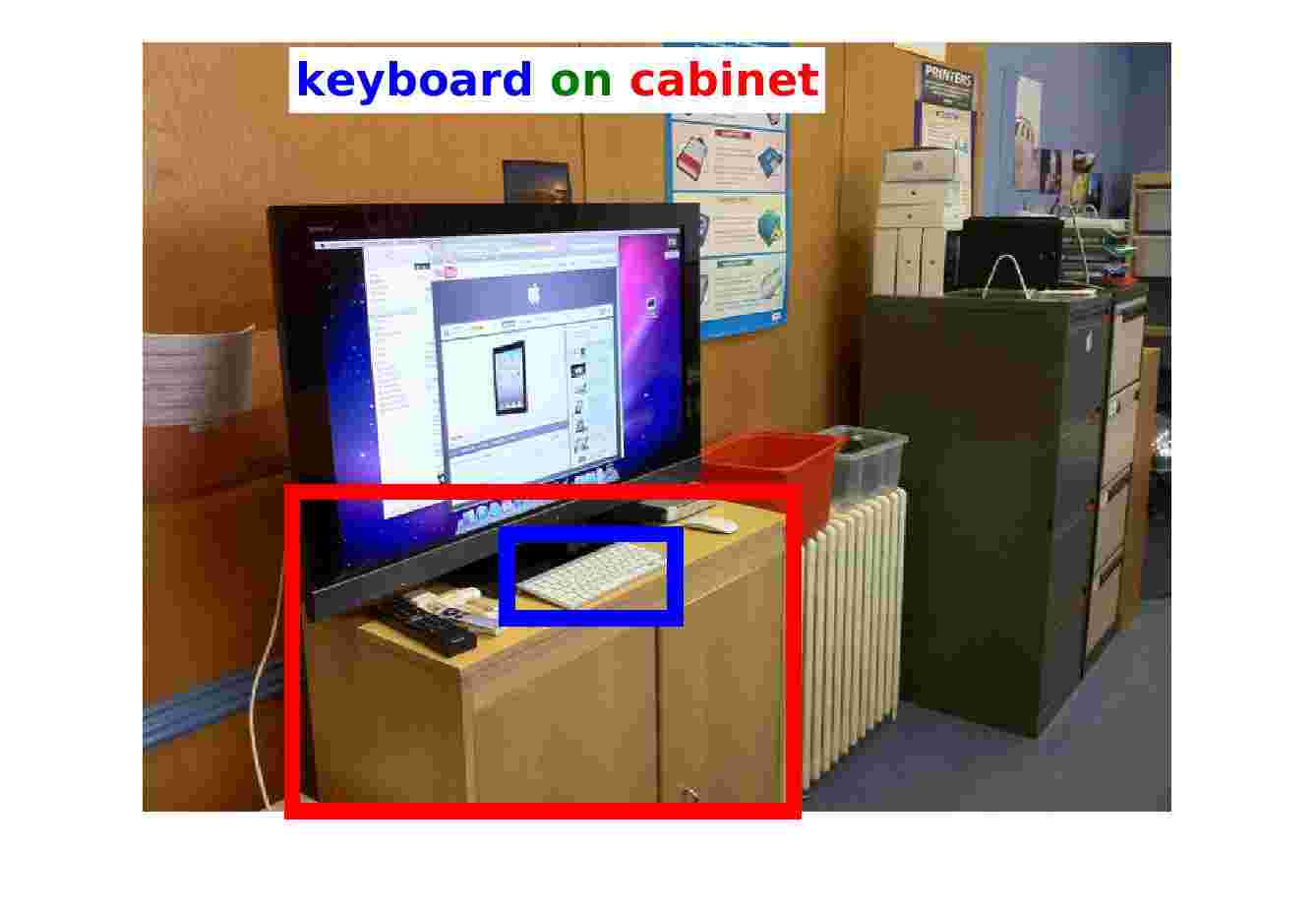}\\
       	\vspace{0.3ex}
       	\cite{Lu16}: in the front of \\
       	GT: on
       	\vspace{0.2ex}
    \end{minipage}
    \hspace{0.005\textwidth}
    \begin{minipage}[t]{0.185\textwidth}
       \centering
       \includegraphics[trim={2.3cm 1.7cm 1.2cm 0.5cm},clip,width=0.95\linewidth,cfbox={green 2pt 2pt}]{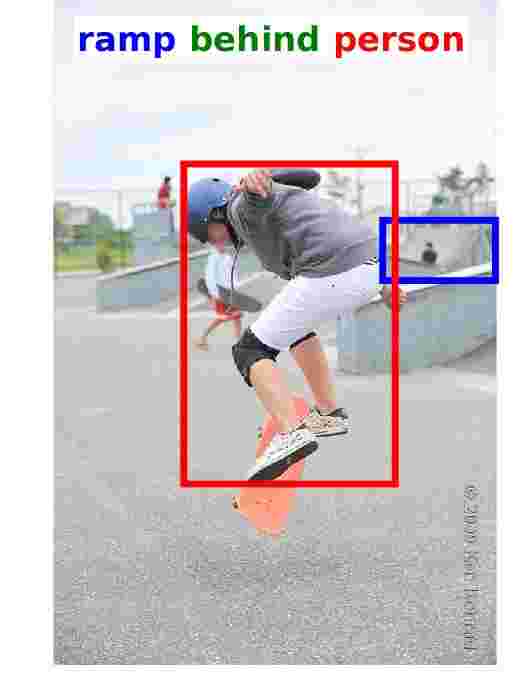}\\
       \vspace{0.3ex}
       \cite{Lu16}: beneath \\
       GT: behind
       \vspace{0.2ex}
    \end{minipage}
    \hspace{0.005\textwidth}
    \begin{minipage}[t]{0.185\textwidth}
    	\centering
       	\includegraphics[trim={2cm 0.9cm 2cm 0cm},clip,width=0.95\linewidth,cfbox={yellow 2pt 2pt}]{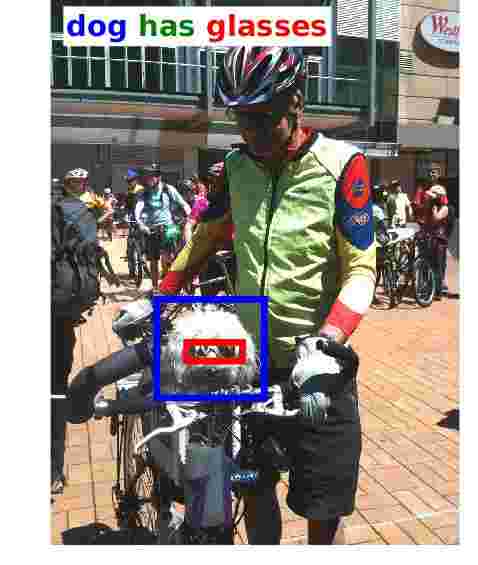}\\
       	\vspace{0.3ex}
       	\cite{Lu16}: wear \\
       	GT: wear
       	\vspace{0.2ex}
    \end{minipage}
    \hspace{0.005\textwidth}  
    \begin{minipage}[t]{0.185\textwidth}
    	\centering
       	\includegraphics[trim={27cm 0cm 5.3cm 0cm},clip,width=0.95\linewidth,cfbox={red 2pt 2pt}]{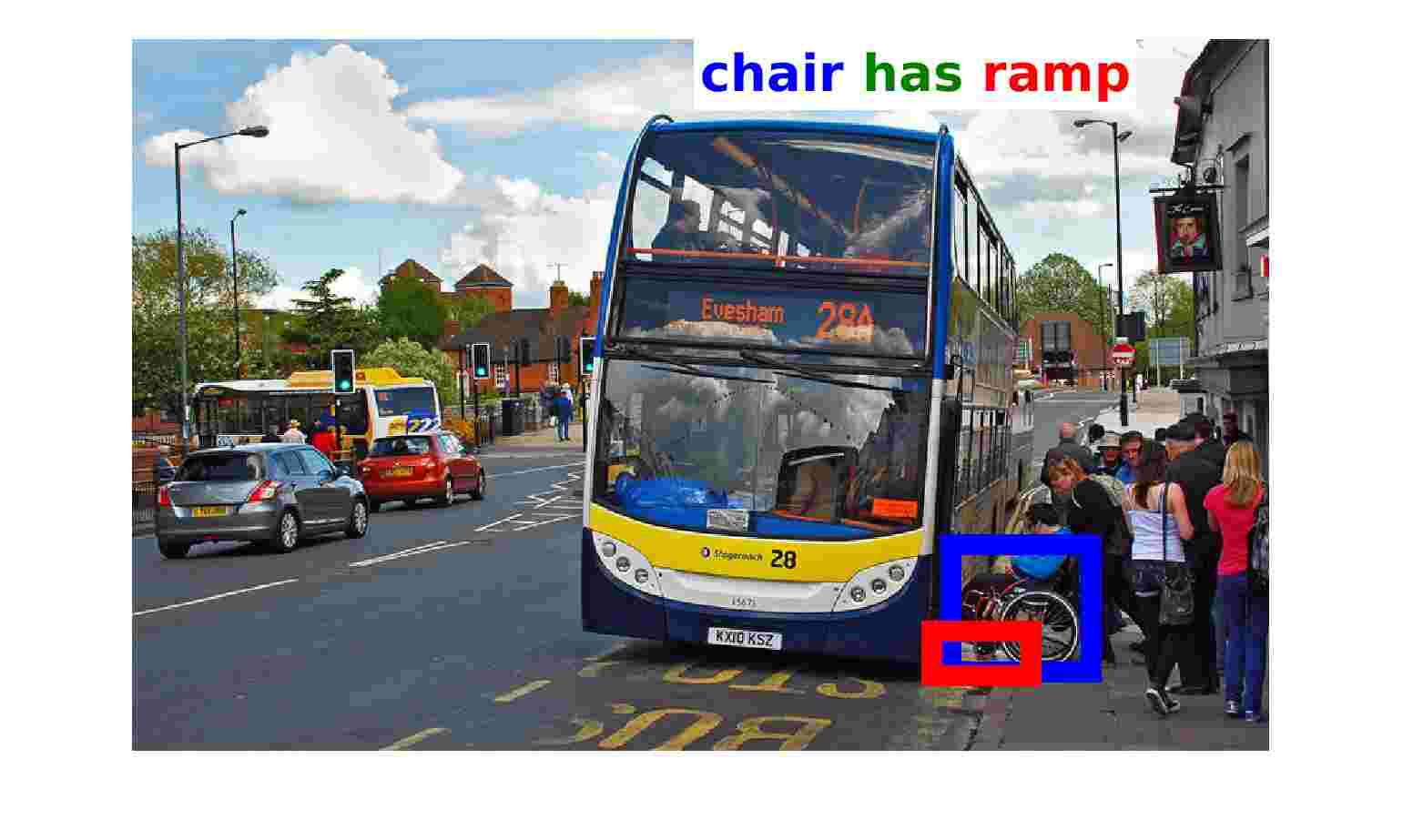}\\
       	\vspace{0.3ex}
       	\cite{Lu16}: behind \\
       	GT: on
      	\vspace{0.2ex}
    \end{minipage}

    \setlength\abovecaptionskip{5pt}
    \caption{Examples of predicate detections on the unseen test triplets of \cite{Lu16} by our weakly-supervised model described in Section~\ref{model} using ground truth object boxes. The triplet is correctly recognized if the relation matches ground truth (in green), else the triplet is incorrect (in red). We also show examples where the ground truth is missing or ambiguous (in yellow). Below each image, we report the prediction of the Visual+Language model of \cite{Lu16}, as well as the correct ground truth predicates ('GT') for the pair of boxes.}
    \label{fig:predicate_detection_zeroshot}
\end{figure*}

\paragraph{Predicting unseen triplets.}
In Figure~\ref{fig:predicate_detection_zeroshot} we provide additional examples for retrieval of zero-shot triplets. Similar to the previous set-up, we assume the ground truth object boxes to be given and focus on predicting relations.  We compare predictions of our weakly-supervised model with the fully supervised Visual+Language model of \cite{Lu16}. 
For each pair of boxes, we indicate below each image the output of~\cite{Lu16}. We also report the ground truth predicates as 'GT'. These examples demonstrate the benefit of our visual features for predicting zero-shot triplets. In some cases, the Visual+Language model of~\cite{Lu16} appears to heavily rely on language (e.g. ``elephant feed elephant'' instead of ``elephant next to elephant'', which is transferred from ``person feed elephant'' via language) where our spatial features  predict the correct relation. In other cases the language model suppresses  incorrect relations such as ``surfboard wear hand'' as well as disambiguates subtly different spatial configurations (``kite on street'' instead of ``kite above street'').

\section{Results on Visual Genome Dataset}

\label{part:vg}
Here we show results for the new challenging dataset
of~\cite{Krishna2016} that depicts complex scenes (21 objects and 17
relationships per image on
average). Table~\ref{tab:results_visualgenome} compares the accuracy
of our method with the method of~\cite{Krishna2016}.
Since not all details are given in the paper, we have reproduced the experimental setting as well as we could based on a communication with the authors. In particular, we
keep a vocabulary corresponding to the 100 most frequent relations and nouns that
occur at least 50 times in one of these relations (we end up with 2618
nouns). We use the training/validation/test splits given
by~\cite{Johnson2015} and retain around 27K unique triplets for
training and validation as in~\cite{Krishna2016} while testing on the
whole test split. We compare with the fully-supervised baseline
experiment \textbf{Relationship} in~\cite{Krishna2016} which trains a
VGG16 network to predict the predicate for a pair of boxes given the
appearance of the union of the boxes. We train our model described in
Section~\ref{model} first with full supervision (Ours [S+A] full) then
with weak supervision (Ours [S+A] weak). Our appearance features are
extracted from a VGG16 network trained on COCO (we do not
fine-tune). For the weak supervision we use ground truth object boxes
to form the candidate pairs of boxes in the image. This would
correspond to the case where a perfect object detector is given and we
only have to disambiguate the correct configuration. The evaluation
measure checks the per-instance accuracy to predict the right
predicate for each pair of boxes.

\begin{table}\centering
{\small
\centering
\ra{1}
\begin{tabular}{@{}rccccc@{}}\toprule
& \multicolumn{1}{c}{With GT} & \multicolumn{3}{c}{With candidates} \\
& - & union & subj & subj/obj
\\\midrule
Chance 							& 38.4	& 6.7 	& 4.9 	& 2.8 \\
\textbf{Full sup.}\\
DenseCap \cite{Johnson2015} 		& - 		& 2.9 	& 2.3 	& - \\
Reproduce \cite{Lu16} 			& 50.6 	& 10.4 	& 7.8 	& 5.1  \\
Ours [S+A] 				&  \textbf{62.6} & \textbf{11.8} 	& \textbf{9.2} & \textbf{6.7} \\
\rule{0pt}{1ex}  
\textbf{Weak sup.}\\
Ours [S+A] 						&  58.5 	& 11.2 	& 8.5 	& 5.9 \\
Ours [S+A] - Noisy				&  55.5 	& 11.0	& 8.2	& 5.7 \\
\bottomrule
\end{tabular}
\caption{Retrieval on UnRel (mAP) with IoU=0.5} 
\label{tab:results_rarerel_appendix}
}
\end{table}

\begin{table*}\centering
\ra{1}
\begin{tabular}{@{}rrrrrcrrcrr@{}}\toprule
&& & \multicolumn{2}{c}{Predicate Det.} & \phantom{abc} & \multicolumn{2}{c}{Phrase Det.} & \phantom{abc}& \multicolumn{2}{c}{Relationship Det.}\\
&&								& All & Unseen && All & Unseen && All & Unseen \\\midrule
&& \textbf{Full sup.}\\
a. && Visual Phrases \cite{Sadeghi2011} 	&  1.9 	& -		&& 0.07 & -		&&  -    & - \\
b. && Visual \cite{Lu16} 					&  7.1	& 3.5	&& 2.6 	& 1.1	&&  1.8	 & 0.7 \\
c. && Language (likelihood) \cite{Lu16} 	&  18.2 	& 5.1 	&& 0.08	& 0.01	&&  0.08 & 0.01	\\
d. && Visual + Language \cite{Lu16} 		&  47.8 	& 8.4	&& 17.0 	& 3.7	&&  14.7 & 3.5 	\\
e. && Language (full) \cite{Lu16} 		&  48.4 	& 12.9 	&& 17.3	& 5.5	&&  15.3 & 5.1	\\
\rule{0pt}{3ex}  
f. && Ours [S] 							&  42.2 & 22.2	&& 15.0	& 8.7	&& 	13.5 & 8.1	\\
g. && Ours [A] 							&  46.3 & 16.1	&& 16.4	& 6.0	&&	14.4 & 5.4	\\
h. && Ours [S+A] 						&  50.4 & \textbf{23.6} && 18.1  &\textbf{8.7}	&& 16.1 &  \textbf{8.2}	\\
i. && Ours [S+A] + Language \cite{Lu16} 
										& \textbf{52.6} & 21.6 && \textbf{19.5} & 7.8 && \textbf{17.1} & 7.4 \\
\rule{0pt}{3ex}  
 && \textbf{Weak sup.}\\
j. && Ours [S+A]							& 46.8 	& 19.0	&& 	17.4 & 7.4	&& 15.3 & 7.1	\\
k. && Ours [S+A] - Noisy 				& 46.4  	& 17.6	&& 	16.9 & 6.7	&& 15.0 & 6.4	\\
\bottomrule
\end{tabular}
\caption{Results for Visual Relationship Detection on the dataset of~\cite{Lu16} for R@100.} 
\vspace{-.1cm}
\label{tab:results_vrd_appendix}
\end{table*}

\section{Results for different evaluation criteria}
\label{part:quantitative}

\paragraph{R@100 on Visual Relationship Dataset~\cite{Lu16}.} Complementary to results with $R@50$ provided in the main paper, we show results with $R@100$ in~Table \ref{tab:results_vrd_appendix}. Similar to previous observations, our method outperforms~\cite{Lu16,Sadeghi2011}, in particular on the zero-shot split. Also, the relatively high performance of appearance features alone (g.), which can incorporate only limited context around objects, and the language model only (e.), which ignores image information, reveals a bias in this dataset: knowing object categories already provides a strong prior on the set of possible relations. This emphasizes the value of our UnRel dataset which is created to remove such bias by considering unusual relations among objects.

\paragraph{Retrieval on UnRel with IoU=0.5.}
In addition to results on the UnRel dataset presented in the main paper for IoU=0.3, Table~\ref{tab:results_rarerel_appendix} shows UnRel results for IoU=0.5. Our results show similar patterns for both IoU thresholds. 

\section{Reproducing results of \cite{Lu16}}
\label{part:lu-baseline}
When reproducing results of \cite{Lu16} for Visual Relationship Detection task using their evaluation code we obtained slightly higher performance than reported in \cite{Lu16}. Hence we report the full set of obtained results in Table \ref{tab:baseline_lu}.

\begin{table*}\centering
\ra{1}
\begin{tabular}{@{}rrrrrcrrcrr@{}}\toprule
&& & \multicolumn{2}{c}{Predicate Det.} & \phantom{abc} & \multicolumn{2}{c}{Phrase Det.} & \phantom{abc}& \multicolumn{2}{c}{Relationship Det.}\\
&&								& All & Unseen && All & Unseen && All & Unseen \\\midrule
&& \textbf{Recall@50}\\
b. && Visual \cite{Lu16} 				&  7.2	& 5.4	&& 2.3 	& 1.5	&&  1.6	 & 1.0 \\
e. && Visual + Language \cite{Lu16} 		&  48.7 	& 12.9	&& 16.5 	& 5.1	&&  14.1 & 4.8 	\\
\rule{0pt}{3ex}  
 && \textbf{Recall@100}\\
b. && Visual \cite{Lu16} 				&  7.2	& 5.4	&& 2.7 	& 1.7	&&  1.9	 & 1.2 \\
e. && Visual + Language \cite{Lu16} 		&  48.7 	& 12.9	&& 17.3 	& 5.7	&&  15.0 & 5.4 	\\
\bottomrule
\end{tabular}
\caption{Results for Visual Relationship Detection on the dataset of~\cite{Lu16} recomputed for the approach of~\cite{Lu16}. Despite using the evaluation code of~\cite{Lu16} we have obtained slightly higher results than those reported in~\cite{Lu16}.} 
\vspace{-.1cm}
\label{tab:baseline_lu}
\end{table*}

\end{document}